\documentclass[journal]{IEEEtran}
\usepackage{amsmath,amsfonts,amssymb}
\usepackage{algorithmic}
\usepackage{algorithm}
\usepackage{array}
\newcolumntype{L}[1]{>{\raggedright\let\newline\\\arraybackslash}p{#1}}
\newcolumntype{C}[1]{>{\centering\let\newline\\\arraybackslash}p{#1}}
\newcolumntype{R}[1]{>{\raggedleft\let\newline\\\arraybackslash}p{#1}}
\usepackage{graphicx}
\usepackage{subcaption}
\usepackage{textcomp}
\usepackage{stfloats}
\usepackage{url}
\usepackage{verbatim}
\usepackage{graphicx}
\usepackage[hidelinks]{hyperref}
\usepackage{cite}
\usepackage{booktabs}
\usepackage[table,x11names]{xcolor}         
\usepackage{mdframed}
\usepackage{array, boldline, rotating}
\usepackage{longtable}
\usepackage{multirow,tabularx}
\usepackage{orcidlink}

\newcommand{\RRR}{\ensuremath{\mathbb{R}}}
\newcommand{\NNN}{\ensuremath{\mathbb{N}}}
\newcommand{\ZZZ}{\ensuremath{\mathbb{Z}}}

\def\eg{\textit{e.\,g.}}
\def\ie{\textit{i.\,e.}}
\def\cf{\textit{cf.}}

\def\t4c{\textit{Traffic4cast}\relax}
\def\mcswts{\textit{MeTS-10}\relax}
\def\mcswtslong{\textit{Metropolitan Segment Traffic Speeds from Massive Floating Car Data in 10 Cities}\relax}

\usepackage{pifont}
\def\cir#1{%
        \raisebox{.9pt}{\textcircled{\raisebox{-.9pt}{#1}}}%
}

\newcommand{\showoutlines}{show them}  
        {
            \ifthenelse{\isundefined{\showoutlines}}%
                    {\expandafter\comment}%
                    {\begin{mdframed} \textcolor{blue}{{\bf Outline}}}%
                    }%
         {%
            \ifthenelse{\isundefined{\showoutlines}}%
                    {\expandafter\endcomment}%
                    {\end{mdframed}}%
          }

\begin{document}

\title{\mcswtslong}

\author{
    \IEEEauthorblockN{Moritz~Neun 
    \orcidlink{0000-0003-2076-8714},
    Christian~Eichenberger 
    \orcidlink{0000-0003-4460-5194}, 
    Yanan~Xin 
    \orcidlink{0000-0003-3866-821X}, 
    Cheng~Fu 
    \orcidlink{0000-0003-1562-7941}, 
    Nina~Wiedemann 
    \orcidlink{0000-0002-8160-7634},
    Henry~Martin 
    \orcidlink{0000-0002-0456-8539}, 
    Martin~Tomko 
    \orcidlink{0000-0002-5736-4679}, 
    Lukas~Ambühl 
    \orcidlink{0000-0001-8835-0950}, 
    Luca~Hermes 
    \orcidlink{0000-0002-7568-7981}, 
    Michael~Kopp
    \orcidlink{0000-0002-1385-1109}}
    \thanks{
    Final manuscript of accepted paper submitted  23 June 2023 to the IEEE
Transactions on Intelligent Transportation Systems (T-ITS).
    {\it (Moritz Neun  and Christian Eichenberger contributed equally to this work.) (Corresponding authors: Moritz Neun; Christian Eichenberger.)}}
    \thanks{Moritz Neun, Christian Eichenberger, Henry Martin, and Michael Kopp are with the Institute of Advanced Research in Artificial Intelligence (IARAI), Vienna, Austria (e-mail:\{first.last\}@iarai.ac.at).}
    \thanks{Yanan Xin, Nina Wiedemann and Henry Martin are with the Institute of Cartography and Geoinformation, ETH Zurich, Switzerland.}
    \thanks{Cheng Fu is with the Department of Geography, University of Zurich, Switzerland.}
    \thanks{Martin Tomko is with the Department of Infrastructure Engineering, The University of Melbourne, Australia.}
    \thanks{Lukas Ambühl is with the Dept. of Civil, Environmental and Geomatic Engineering, Traffic Engineering Group, ETH Zurich, Switzerland.}
    \thanks{Luca Hermes is with the Machine Learning Group, Bielefeld University, Germany}
}

\markboth{IEEE Transactions on Intelligent Transportation Systems \href{https://doi.org/10.1109/TITS.2023.3291737}{10.1109/TITS.2023.3291737}}
{Neun \MakeLowercase{\textit{et al.}}: \mcswtslong{}}

\IEEEpubid{
\begin{minipage}{\textwidth}\centering
\copyright~2023 IEEE. \href{https://doi.org/10.1109/TITS.2023.3291737}{Digital Object Identifier 10.1109/TITS.2023.3291737}.
\end{minipage}} 

\maketitle



\begin{abstract}
Traffic analysis is crucial for urban operations and planning,
while
the availability of dense urban traffic data beyond loop detectors is still scarce.
We present a large-scale floating vehicle dataset of per-street segment traffic information, \mcswtslong~(\mcswts),
available for 10 global cities with a 15-minute resolution for collection periods ranging between 108 and 361 days in 2019--2021 and covering
more than 1500 square kilometers
per metropolitan area.
\mcswts{} features traffic speed information at all street levels from main arterials to local streets for \textit{Antwerp, Bangkok, Barcelona, Berlin, Chicago, Istanbul, London, Madrid, Melbourne}, and \textit{Moscow}.
The dataset leverages the industrial-scale floating vehicle \t4c data with speeds and vehicle counts provided in a privacy-preserving spatio-temporal aggregation.
We detail the efficient matching approach mapping the data to the OpenStreetMap (OSM) road graph.
We evaluate the dataset by comparing it with publicly available stationary vehicle detector data (for Berlin, London, and Madrid) and the Uber traffic speed dataset (for Barcelona, Berlin, and London). The comparison highlights the differences across datasets in spatio-temporal coverage and variations in the reported traffic caused by the binning method.
\mcswts{}  enables novel, city-wide analysis of mobility and traffic patterns for ten major world cities, overcoming current limitations of spatially sparse vehicle detector data. The large spatial and temporal coverage offers an opportunity for joining the \mcswts{} with other datasets, such as traffic surveys in traffic planning studies or vehicle detector data in traffic control settings.

\end{abstract}



\section{Introduction}

\begin{table}[!t]    
    \caption{\mcswts{} dataset overview.
    }
    \label{tab:dataset}

    \begin{tabular}
    {p{2.1cm}p{0.3cm}p{1.6cm}p{1.5cm}p{1.2cm}}
\toprule
City (Traffic4cast competition year) & Days & Date ranges & 8:00--18:00 coverage (\#edges w. speed / total \#edges) & Mapped ratio (\%~GPS probes mapped) \\
\midrule
 Antwerp (2021) & 361  & 2019-01---06, 2020-01---06  & 0.17 & 0.89\\
 Bangkok (2021) & 361  & 2019-01---06, 2020-01---06  & 0.05 & 0.91\\
 Barcelona (2021) & 361  & 2019-01---06, 2020-01---06  & 0.09 & 0.81\\
 Berlin (2021) & 180  & 2019-01---06  & 0.36 & 0.94\\
 Chicago (2021) & 180  & 2019-01---06  & 0.11 & 0.93\\
 Istanbul (2021) & 180  & 2019-01---06  & 0.63 & 0.96\\
 Melbourne (2021) & 180  & 2019-01---06  & 0.08 & 0.93\\
 Moscow (2021) & 361  & 2019-01---06, 2020-01---06  & 0.71 & 0.81\\
 London (2022) & 110  & 2019-07---12, 2020-01  & 0.24 & 0.95\\
 Madrid (2022) & 109  & 2021-06---12  & 0.36 & 0.93\\
 Melbourne (2022) & 108  & 2020-06---12  & 0.08 & 0.90\\
\bottomrule
    \end{tabular}
\end{table}

Urban traffic analysis and prediction are highly complex as traffic arises from crowd behavior and human interactions that are difficult to model. It also shows a strong spatial variation depending on the layout, regulations, and operations of the street network. 
Meanwhile, in light of the impact of urban traffic on sustainability, health, and the environment, policy leaders have a strong interest in understanding traffic patterns and in the deployment of data-driven methods for efficient and innovative mobility, called smart cities~\cite{angelidou2017role}.

\IEEEpubidadjcol

Accordingly, the identification, analysis, modeling, simulation, and forecasting of traffic patterns have received much attention in research for years~\cite{vaughan1987urban, paveri1975boltzmann, greenberg1959analysis}.
Traffic analysis has become a high-tech business
aiming at
accurately capturing and predicting traffic patterns online~\cite{batty2013new} to ultimately create a ``digital twin'' of city-wide traffic that could
support operational decisions, navigation, and infrastructure planning~\cite{deng2021systematic}.

However, the lack of accurate and fine-grained open-source traffic data hinders methodological progress in traffic analysis. Currently, most analyses are based on data from stationary vehicle detectors installed at fixed locations in the urban environment, such as loop counters~\cite{bui2021uvds, cui2018deep, loder2019understanding} or cameras~\cite{snyder2019streets, bharadwaj2016large}. Such vehicle detector datasets have three main shortcomings: 1) The installation of stationary detectors is costly and laborious. Thus, there are only a few cities globally with sufficiently dense sensor coverage to gain a comprehensive picture of traffic behavior. 2) These data are captured with varying temporal and spatial granularity. 3) Even with high investments, the sensors are integrated only at selected fixed locations, providing incomplete reflection of the traffic flow in urban streets. In particular, the available datasets are usually guided by the requirements of traffic control, therefore are biased towards highways or urban streets of high throughput. 

In summary, datasets based on stationary detectors are usually prone to biases and insufficient spatial coverage for fine-grained traffic analysis. An alternative to data based on stationary detectors with high spatial coverage is provided by vehicle tracking datasets. Such datasets are currently provided by owners (or administrators) of vehicle fleets, such as taxi fleets~\cite{illinoisdatabankIDB-9610843}, private vehicle tracks~\cite{jiang2019vluc}, or from ride-hailing services such as the Uber dataset~\cite{uber-movement-speeds-calculation-methodology,uber-movement-faqs}. However, access to these datasets is usually expensive and has additional issues. For example, the Uber dataset provides a high spatial resolution with many street segments, but it comes at the cost of a lower temporal resolution with only hourly traffic speed data available. An important obstacle to publishing such datasets is usually privacy concerns~\cite{boeckelt2005probe}, which prevents the sharing of data from sensors installed in private vehicles or smartphones. As exceptions, there are a few publicly available vehicle-probe datasets, but these are limited to single cities (\eg{} the Didi dataset~\cite{didi}), short time periods, a low temporal~\cite{jiang2019vluc} or spatial resolution~\cite{liao2018deep}, or a single type of vehicle. Another difficulty is the representation of such datasets: For simplicity and privacy, data are usually aggregated as origin-destination matrices or in a raster format (\eg{}~\cite{jiang2019vluc}), making the mapping of traffic speed and flow to the street network impossible.

Here, we introduce \mcswtslong{} (\mcswts{}), a multi-city traffic speed dataset that provides high-coverage probe-vehicle data mapped onto street segments of the city network. We present a pipeline to derive the dataset from the \t4c data \cite{here:sample-data}. The \t4c data was recently published by HERE Technologies, a company providing a platform for the visualization and analysis of location data. To ensure data privacy, the \t4c dataset was published as rasterized and aggregated cell-based data, nevertheless providing a high spatial and temporal resolution. 
We use data from the OpenStreetMap street network to yield the \mcswts{} with the following properties:

\begin{itemize} 
    \item \emph{Multi-city coverage}: the dataset includes 10 large metropolitan regions in geographically and culturally diverse locations across the globe; 
    \item \emph{High city coverage}: spanning all roads segments covered by the vehicle fleet, in contrast to the sparsity of vehicle detectors and biases of other datasets towards roads of high throughput;
    \item \emph{Long-term coverage}: between 108 and 361 days of continuously sampled data;
    \item \emph{Fleet coverage}: in contrast to other vehicle-probe datasets, the dataset is not restricted to \eg{} taxis only;
    \item \emph{Graph representation}: in contrast to raster-datasets, we provide data mapped to street segments, enabling the ascription of traffic properties to local regulations or infrastructure;
    \item \emph{High temporal resolution}:
    \mcswts{} provides traffic speed data at 15-min bins as default to balance the trade-off between quality and data size. If needed, data can be readily generated in 5-min bins using the provided pipeline.
\end{itemize}

With the large spatio-temporal coverage and high resolution, our dataset enables multi-city analysis and studies on spatial transfer learning.
Segment-wise representation allows enriching of the data with urban properties and spatial context data at an unprecedented coverage and level of detail, see \autoref{tab:dataset}. Therefore, we believe that the \mcswts{} data can facilitate the development of new methodologies for future smart cities.

The contributions of this paper are as follows:
\begin{itemize}
     \item the introduction of the \mcswts{} dataset of traffic speeds disaggregated and matched on the segments of a road graph from massive floating car data aggregated in a privacy-preserving format (\emph{spot binning});
     \item the open-source implementation of the data pipeline, enabling re-processing with refined methods and different road graphs;
     \item the comparison with stationary vehicle detector speeds (ground-truth-like but spatially sparse measurements) and Uber Movement speeds (also from GPS probe data using trajectory-based aggregation and a different vehicle fleet);
     \item the analysis and discussion of the effects of spot binning and trajectory-based speed aggregation.
\end{itemize}

The remainder of this paper is organized as follows: We first describe related work in \autoref{sec:related_work}, and then describe our novel pipeline to derive the dataset from the HERE raster-data with OSM data in \autoref{sec:dataset}. \autoref{sec:methods_and_data} gives a description of the technicalities of the dataset and usage instructions. In \autoref{sec:validation}, we compare and validate our data with stationary vehicle detector data as well as probe vehicle data from the Uber dataset, exploiting partial temporal and spatial overlaps of our dataset with others. Finally, we discuss the opportunities and limitations of \mcswts{} in \autoref{sec:discussion}.

\section{Related Work}\label{sec:related_work}
We compare existing traffic datasets in \autoref{tab:datasets} in terms of type, size, temporal, and spatial resolution. While many cities worldwide publish (parts of) their sensor data, we focus on a few datasets that were selected based on one of the following properties: 1) size (pooling of multiple sensor-based datasets), 2) their relevance for research (mentioned in methodological work on traffic analysis and prediction), or 3) their similarity to our dataset (probe-vehicle data).

\subsection{Vehicle Detector Datasets}

Most available data come from stationary vehicle detectors installed on highways and main roads. The reason for the availability of such data is their low privacy sensitivity, in contrast to information about individual vehicle tracks. Although many studies have been conducted in collaboration with local authorities on proprietary vehicle detector data, such data was systematically collected, pre-processed, and published for specific regions, for example, in the PEMS~\cite{pems} and METR-LA~\cite{li2018dcrnn_traffic} datasets. Smaller public datasets exist, for example, for Seattle~\cite{cui2018deep}, Guangzhou~\cite{chen2018spatial}, or Portland~\cite{portlanddata}.  The UTD-19~\cite{loder2019understanding} dataset is an effort to combine data from 40 cities into a unified representation but has been used primarily for the analysis and simulation of traffic data~\cite{ambuhl2021disentangling, bramich2022fitting}. Others have introduced image-based datasets from cameras installed on streets~\cite{snyder2019streets, bharadwaj2016large, xia2022dutraffic}. Since we focus on traffic speed and since the use cases of such datasets are very different from ours, we refer to other work for more details.

\begin{table*}[htb]
    \centering
    \caption{Overview of related public traffic datasets.}
    \label{tab:datasets}
    \resizebox{\textwidth}{!}{
    \begin{tabular}{l|llllll}
    \toprule
    Dataset & Type & Number of records & Temporal resolution & Spatial resolution & Collection Area & Biases (vehicle / road types)\\ 
    \midrule
PEMS~\cite{pems} & Loop detectors & & 5 min & 39k detectors  & California & highway network \\
METR-LA~\cite{li2018dcrnn_traffic} & Speed (loop detectors) & 	34,272 & 	5 min & 207 nodes & Los Angeles County & highways 
\\
UVDS~\cite{bui2021uvds} & Flow (loop detectors)	& 	25,632 	& 	5 min & 104 nodes &Daejeon (South Korea) & urban main roads 
\\
Seattle Loop \cite{cui2018deep} & loop detectors &  & 5 min & 323 detectors &  Seattle & freeways \\
NYC Taxi & pick-up / drop-offs & 1.5 B.  & timestamp & full city & New York& taxis \\
UTD-19~\cite{loder2019understanding} & loop detectors & & 3-5 min & 23541 sensors  & 40 cities & main roads \\
VLUC~\cite{jiang2019vluc} & vehicle-probes & 4800 x 2 cities & 30 min & $450\,\textrm{m} \times 450\,\textrm{m}$ grid & 2 cities & all urban streets \\ 
Q-traffic~\cite{liao2018deep} & Speed data (Baidu) & 265 mio & 15 min & 15k road segments&  Beijing, 1x1 km grid & urban areas  \\ 
Uber~\cite{uber-movement-speeds-calculation-methodology} & Speed data &  & 1 hour & OSM segments & 11 cities & Uber drivers\\
Didi (Gaia Initiative) & Speed data &  & 10min & Road segments & Chengdu \& Xi’an & Didi ride \\
Mobile Millenium~\cite{hunter2011scaling} & GPS trajectories & (not open data) & timestamp & GPS trajectories & San Francisco & crowd-sourced collection\\
T-Drive~\cite{yuan2011driving, yuan2010t} & GPS trajectories & 15 million & timestamp & GPS trajectories & Beijing & taxis\\
Kaggle congestion & Vehicle count & 48,120 & 1h & 4 Junctions & Unknown & only 4 junctions \\
Traffic4cast~\cite{here:sample-data} & Speed and probe count & 108--361 days & 5 min & $\sim 100\,\textrm{m} \times 100\,\textrm{m} \times 4$ headings & 10 cities, $\sim 50\,\textrm{km} \times 50\,\textrm{km}$ & vehicle fleet\\
    \bottomrule
    \end{tabular}
    }
\vspace{-2mm}
\end{table*}

\subsection{GPS Probes Datasets}

A similar approach as in the HERE traffic movie dataset is taken by the VLUC dataset~\cite{jiang2019vluc}, pooling several GPS-based datasets in traffic movie representations. In \autoref{tab:datasets}, we only report the statistics for their new dataset from two big cities in Japan (Tokyo and Osaka), although they also include previously published datasets such as NYC Taxi and NYC bike datasets in the VLUC dataset. These new data were collected over 100 days, using a GPS enabled app installed by about 1 million users, ``approximately 1\% of the total population of Japan''\cite{jiang2019vluc}. The data was later aggregated into 30-minute intervals and $450\,m \times 450\,m$ grid cells. In contrast, Microsoft presented work on two datasets of GPS trajectories that were not aggregated in grid cells. While only a small fraction of the Mobile Millenium dataset~\cite{hunter2011scaling} is publicly available, the T-Drive data~\cite{yuan2010t, yuan2011driving} offers a large set of taxi trajectories for trajectory analysis. Such data is rich in information but requires extensive preprocessing before any analysis. Finally,  crowdsourced GPS probes have been used for collecting lane-based road information \cite{CLRIC}. 

Furthermore, the ride-hailing services Didi and Uber published segment-wise data that are closest to ours in terms of representation~\cite{uber-movement-speeds-calculation-methodology,uber-movement-faqs, didi}. While the data are biased toward taxi traffic behavior, they provide high spatial coverage of speed estimates. The main drawback of the Uber dataset is the low temporal resolution of one hour, preventing not only the evaluation of short-term prediction methods, but also in many cases peak hour is around 1h, so 1h bins come at the risk of missing peak hours or averaging them out.

\section{The Dataset: Segment Median Speeds} \label{sec:dataset} 

In this section, we give an overview of \mcswts{}, before going into the technical details in Section~\ref{sec:methods_and_data}.
The \mcswts{} dataset provides segment-wise speeds derived from aggregated GPS probes via a spatial join with a road graph.
We use a road graph derived from OSM.
The GPS data was made publicly available by HERE Technologies as a spatio-temporal aggregation \cite{pmlr-v123-kreil20a,pmlr-v133-kopp21a,pmlr-v176-eichenberger22a,here:sample-data}. 
\autoref{tab:dataset} shows the cities and date ranges of the available aggregated data.  The raw source probes or trajectories are not publicly available.
The dataset comprises 10 cities with data from 108 up to 361 days publicly available for download from HERE Technologies for the corresponding \t4c competition years.

The GPS speeds and the number of probe points (probe volumes) come at a 5-min resolution. Segment-wise speeds can thus be aggregated to 5-min or any coarser resolution (as an integer multiple); here, we default to 15 minutes to balance between quality and storage. The \t4c bounding box covers $\sim50\,\textrm{km}\times50\,\textrm{km}$ in every city.
\autoref{tab:dataset} also shows that the different cities have different coverages (relative number of speed data points at 15-minute resolution), depending on the size and constitution of the contributing vehicle fleets, but also on the road topology (segment lengths). 
The high coverage in Moscow is partially due to the exclusion of the ``fat tail'' as reflected by the lower  ratio of mapped GPS probes.
Here,  most inner-block roads are modeled as service roads, which we do not include by default in the road graph derivation from OSM data.
More details can be found in Supplement~E~\cite{neun2023metropolitansupplementary} and the documentation in the code repository.

\begin{figure}[!t]
\centering
\includegraphics[width=0.47\textwidth]{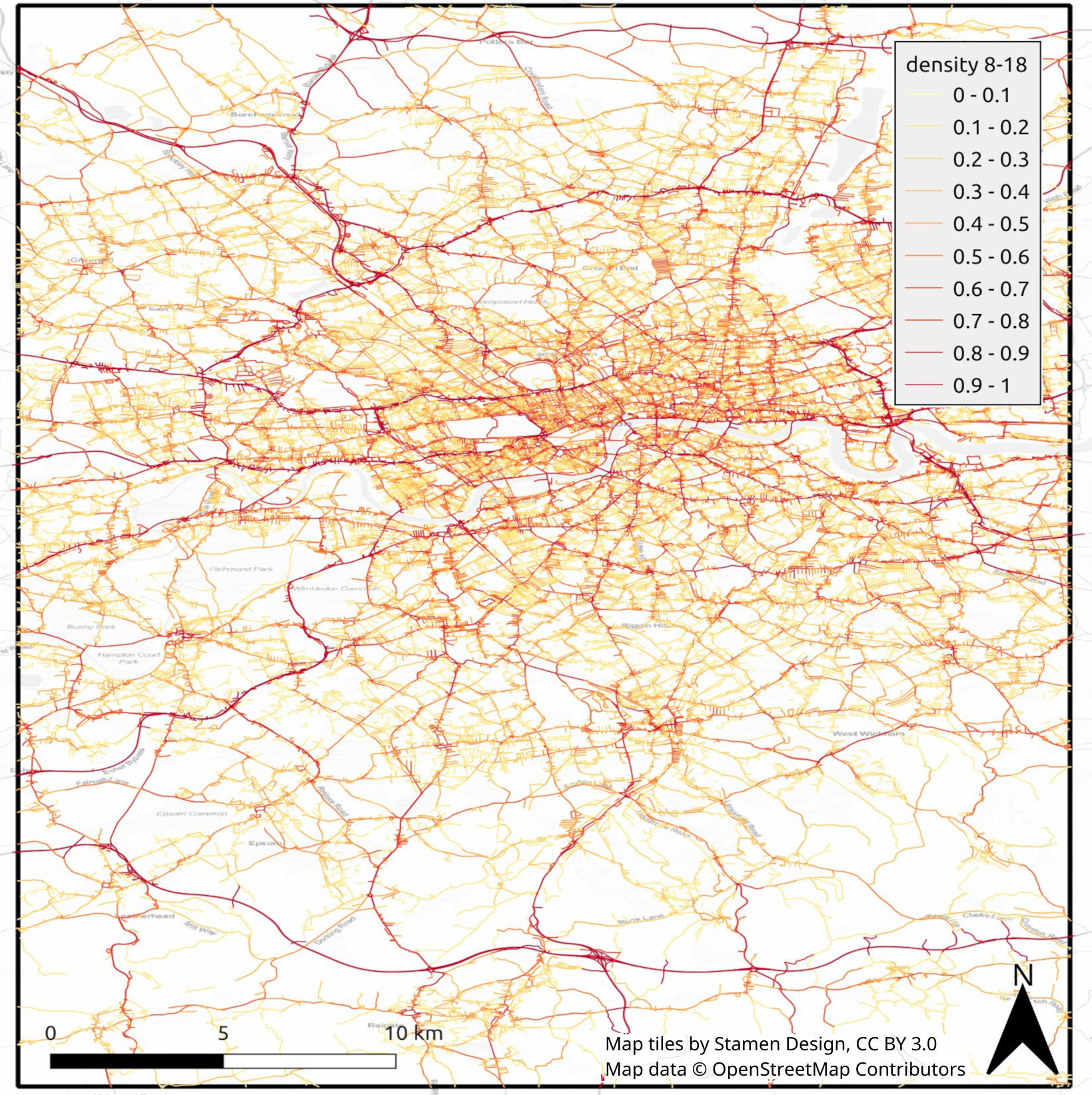}
\vspace{-1mm}
\caption{Segment density during 08:00--18:00 for London.} 
\label{fig:03_dataset/coverage_daytime_london}
\vspace{-1mm}
\end{figure}

For illustration, we pick the city of London, one of the cities for which we have other datasets to compare with in \autoref{sec:validation}.
\autoref{fig:03_dataset/coverage_daytime_london} shows the densities and geographical coverage of the available speed segments across the whole city of London. 
Densities (or temporal coverage) is the ratio of edges with speed values between 8am and 6pm (from 20 sampled days) with respect to the selected OSM road graph
(see also 8:00--18:00 coverage in Table \ref{tab:dataset}).
For motorway highways, the coverage is close to 100\% during day time, and also trunk and primary streets have a speed value during 50\% of the day, see Supplementary Material~\cite{neun2023metropolitansupplementary}.
\autoref{fig:05val02-counter-situations} 
shows the daily speed profiles for two motorway situations,
reflecting the speed limits as well as the daily fluctuation due to increased traffic (dips during the morning and afternoon peak hours). 
London has cross-country motorways cutting through the bounding box, resulting in high speed levels.

\section{Input Data and Data Pipeline}\label{sec:methods_and_data}

\begin{figure}[!t]
\centering
\includegraphics[width=3.5in]{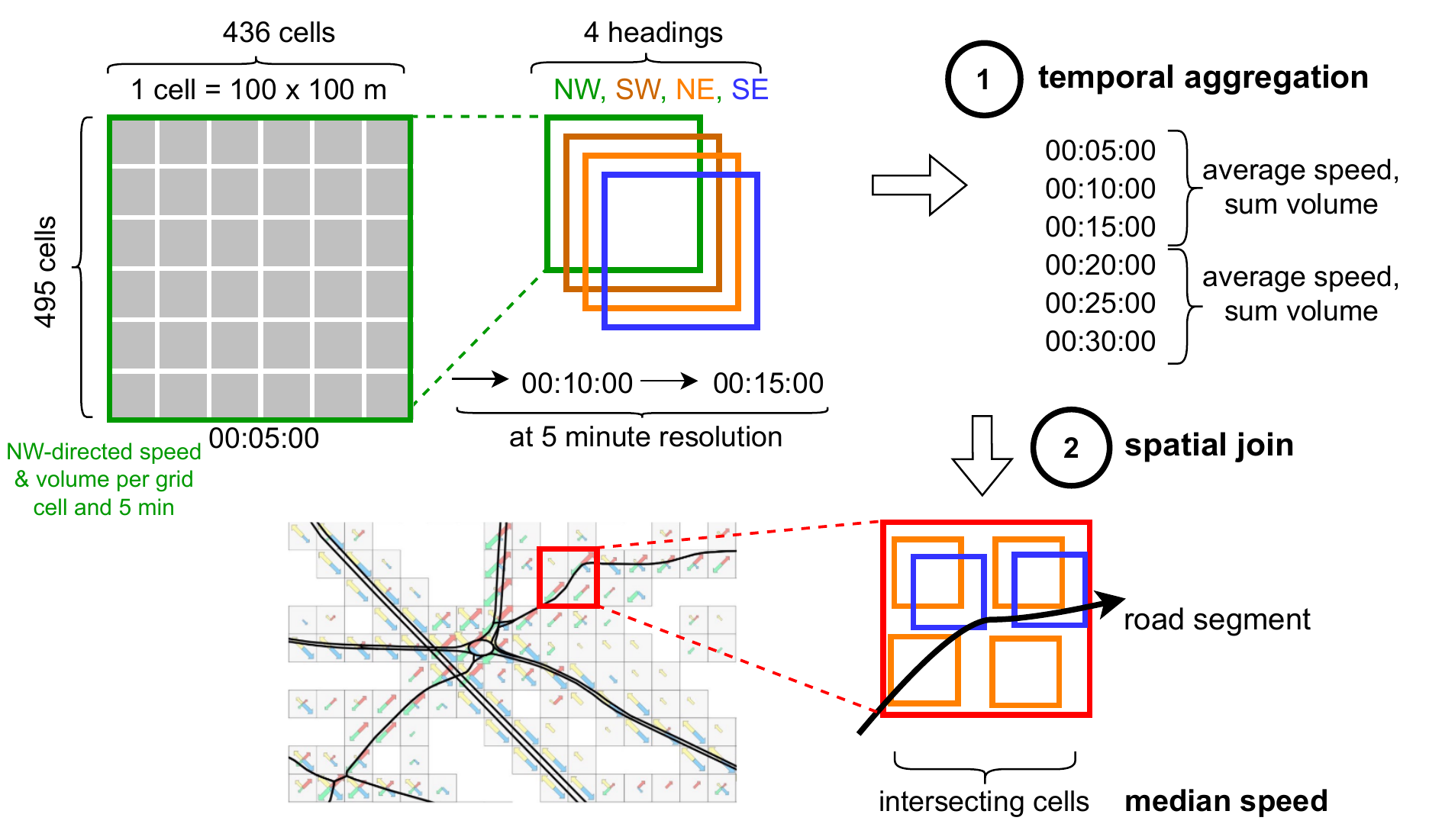}
\caption{\mcswts{} method overview. We leverage data from HERE traffic map movies (GPS probes that are aggregated spatially into $\sim100\,\textrm{m}\times100\,\textrm{m}$, and 4 headings, and temporally in 5-minute bins). We derive segment-wise 15-min speeds by temporal (\cir{1}) and spatial (\cir{2}) aggregation. 
}
\label{fig:04_methods_and_data/visual_abstract}
\vspace{-4mm}
\end{figure}

In this paper, we are demonstrating 
how the aggregated GPS data can be matched with an OSM road graph. The same methodology can also be used on further road graph variants other than OSM.
Due to the license limitations of the two input data sources (see Supplement~A-A~\cite{neun2023metropolitansupplementary}), users need to generate the dataset by running the data pipeline.
As we provide the complete code for generating our multi-city segment-wise traffic speed dataset, users are welcome to improve the methodology we describe in this article and refine it for specific applications.

\autoref{fig:04_methods_and_data/visual_abstract} gives an overview of our method implementing the spatial join:
The road graph comes from OpenStreetMap (OSM) and the spatio-temporally aggregated GPS probes come as Traffic Map Movies from HERE Technologies. After downloading the aggregated GPS probes from HERE Technologies, the pipeline code can be executed and automatically takes care of downloading a suitable navigable road graph from OSM. 
Further technical details on input data and data pipeline can be found in Supplement~A~\cite{neun2023metropolitansupplementary}.

\subsection{Input Data}

\subsubsection{Traffic Map Movies}\label{sec:methods_and_data:traffic_map_movies}

HERE Technologies provides spatially and temporally aggregated GPS probe data from multiple culturally and socially diverse metropolitan areas around the world. The data comprises between 3 months to one year of data per city from 2018, 2019, and 2020 (see \autoref{tab:dataset}). 
In Traffic Map Movies \cite{here:sample-data,pmlr-v123-kreil20a,pmlr-v133-kopp21a,pmlr-v176-eichenberger22a}, each snapshot (or movie frame) covers $\sim50\,\textrm{km}\times50\,\textrm{km}$ of the urban area at a 5-minute time bin, thus providing comprehensive coverage of complex cities. The city bounding boxes were defined in the context of the \t4c competition series with the same height and width for all cities (for some cities, the shape is rotated). The spatial binning divides the data into grid cells of $0.001^\circ$ (\ie{} $\sim100\,\textrm{m}\times100\,\textrm{m}$) and 4 headings (NE,SE,SW,NW), as shown in \autoref{fig:04_methods_and_data:speed-aggregation-overview}. In each bin, probe volume (number of GPS recordings, not vehicles) and mean speed are collected. 
More precisely, data for one city and one day comes in the shape of $(288,495,436,8)$ uint8, representing 288 time slots, 495 rows, 436 columns, and 8 channels (volume and speed for 4 headings).


Intuitively, this \emph{spot binning} favors lower speeds as slower cars stay longer in the same spatio-temporal bin and are counted multiple times.
Under idealized conditions (see Supplement~B~\cite{neun2023metropolitansupplementary}), 
the spatio-temporally aggregated speed 
represents the total distance divided by the total travel time, 
which is the harmonic sum of the speeds.
In particular, this requires controlling the probe rate of vehicles and depends on traffic volume and homogeneity.

\subsubsection{OpenStreetMap}

OpenStreetMap (OSM) is a database of GIS data built by an open community of contributors \cite{OSM:about}.
The OSM data model does not directly describe a road graph, a traversable graph needs to be derived from the OSM elements. 

\subsection{Data Pipeline}

\begin{figure*}[!t]
\centering
\includegraphics[width=0.9\textwidth]{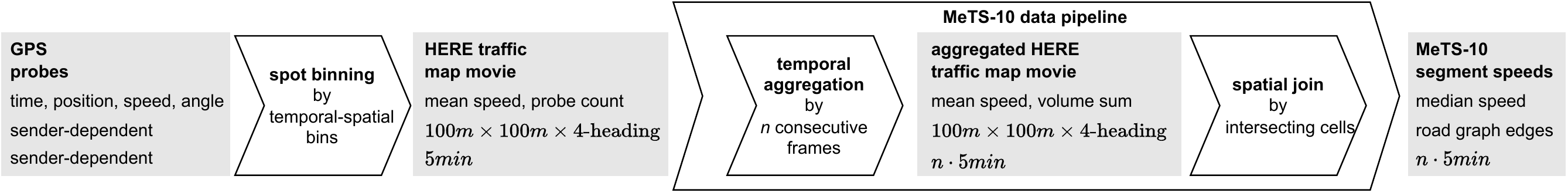}
\caption{GPS probe aggregation overview showing the different steps (arrowheads) and the data (grey areas). For the data, their main attributes and the spatial and temporal resolution are listed.}
\label{fig:04_methods_and_data:speed-aggregation-overview}
\vspace{-3mm}
\end{figure*}

\autoref{fig:04_methods_and_data:speed-aggregation-overview} gives an overview of the GPS probe aggregation.
The HERE traffic map movies are the source of GPS probe data to be leveraged by our method. Optionally, depending on the application, further temporal aggregation helps to smoothen sparsity in the data collection and to align with other data sources like vehicle counters. Finally, the gridded data is mapped to a road graph through the spatial join. We refer to Supplement~B \cite{neun2023metropolitansupplementary} for technical details.

\subsubsection{OpenStreetMap Data Download and Road Graph Construction (dp03)} 

As described above, OSM does not directly come in the form of a road graph.
We generate a basic road graph using source data from OSM data for the city's bounding box using \texttt{OSMnx} \cite{BOEING2017126}.

\subsubsection{Spatial Intersection of Road Graph and City Cells (dp04)}

This step generates lists of intersecting cells for each road segment in the corresponding Traffic Movie grid.
By interpolation over the edge geometries
we get a list of directed cells (row, column, and heading) partially overlapping with this segment.

\subsubsection{Temporal Aggregation of HERE Traffic Map Movies (dp01)} 
HERE Traffic Movies come in 5-minute aggregation, \ie{} the data of one day for each city is represented by 288 movie frames per day.
Further aggregation might be advisable to match with other data sources (such as vehicle counters in the \t4c competition \cite{neun2023traffic4cast}) or to smoothen the sparsity of data. Here, we aggregate the gridded data temporally in 15-minute resolution, see \autoref{fig:04_methods_and_data/visual_abstract} (top right, \cir{1}) 
and \autoref{fig:04_methods_and_data:speed-aggregation-overview}.

\subsubsection{Segment Speeds from Spatial Join (dp06a)}

We derive the segment speeds and volumes by taking the median speed and total probe volume over all intersecting cells from the aggregated \t4c movie, see \autoref{fig:04_methods_and_data/visual_abstract} (bottom, \cir{2}). 
The intersecting cells take into account the heading, and there is a positional and angular margin. In \autoref{fig:04_methods_and_data/visual_abstract}, we show the intersecting cells within the red box for one directed road segment; the orange NE cells are intersecting by the positional margins; the blue SE cells are intersecting because of the angular margins.
We choose the median in order to be robust to distortions coming from erroneous GPS signals and the different intersecting cells going into the edge. In particular, cars stopping for delivery, alighting, or boarding, even in a parking space close to the road, emit many GPS signals at the same location, which can have a large impact on the mean speed from the \t4c movies under low flow conditions. 

\subsubsection{Confidence Filtering of Segment Speeds (dp06b) 
Based on Speed Clustering (dp02) and Free Flow Speeds from Spatial Join and Quantile Selection (dp05)}

As confidence filtering is optional and it does not change the data values, we discuss it last.
Spot binning of GPS probes does neither include any map-matching nor segmentation. Therefore, many of the traditional data cleansing methods of trajectory data mining \cite{zheng2015_trajectory_data_mining_an_overview} do not apply in our setting. 
As a pragmatic approach to trading off quality and coverage, we apply confidence filtering on segment-wise 15~min median speeds. 
The rationale for our filtering is the following: 
First, as just mentioned, there are many situations where small speed values can distort the current traffic situation, which is particularly grave in low-volume situations.
Second, we want to avoid too many false positive low-speed congestion outputs.
Therefore, in order to be confident that a situation is congested, we require more ``proof'' to confidently accept the overall low speed; if we are not confident enough, we reject the median speed computed, \ie{} we do not correct or impute, but filter it out.

Our confidence filtering is based on the concept of a congestion factor \cite{here_mapping_traffic_congestion}, which is the ratio of the current segment speed by the free flow speed. The derivation of free flow speeds is based on k-Means clustering of the speed data.
This is in line with similarly motivated approaches to derive free flow, \eg{} \cite{uber-movement-faqs} uses the 
85th percentile of all speed values observed.   
%

\section{Validation}\label{sec:validation}  

In this section, we validate the \mcswts{} dataset by comparing it with two other datasets: (A) the Uber Movement Speeds dataset, which partially overlaps with the \mcswts{} dataset spatially and temporally for cities 
Barcelona, Berlin, London. (B) ground-truth-like speed readings from stationary vehicle detectors for cities Berlin, London, and Madrid. As additional baseline and sanity check, in (C), we compare the Uber Movement Speeds with the stationary vehicle detector speed readings. 

\subsection{Comparison with Uber Movement Speeds}\label{sec:validation_uber}

\subsubsection{Dataset Comparison}
Uber Movement \cite{uber-movement} provides speeds, travel time, and mobility heatmap across world cities. Uber uses the mean trip speed of a segment as the signature of speed \cite{uber-movement-speeds-calculation-methodology}. Trips as GPS waypoints are map-matched to OSM road segments. Trip speed is then defined as the length of a segment divided by the time of a trip that passes the segment, including the waiting time at exits. For the mean trip speed calculation, trip speeds are further aggregated hourly, excluding trips that have a drop-off or pick-up in the segment. In addition, if a road segment has fewer than 5 valid trips during an hour, the mean trip speed is not provided \cite{uber-movement-speeds-calculation-methodology}. Table~\ref{tab:here-uber} compares HERE Traffic Map Movies \cite{here:sample-data,here-launches-advanced-real-time-traffic-service,here-real-time-traffic, here:terms-and-conditions} and Uber Movement Speeds data sources \cite{uber-movement,uber-movement-speeds-calculation-methodology,uber-movement-faqs}.

\begin{table}[!thb]
\caption{Comparison of HERE Traffic Map Movies and Uber Movement Speeds data sources.\label{tab:here-uber}}
\centering
\begin{tabular}{p{2.4cm} V{2} p{2.6cm}|p{2.6cm}}
\toprule
 & HERE Traffic Map Movies \cite{here:sample-data}& Uber Movement Speeds \cite{uber-movement}\\ 
\midrule
contributing vehicles & connected vehicles, different providers & Uber driver app\\ 
\noalign{\smallskip} temporal resolution & 5 min & 60 min \\ 
\noalign{\smallskip} spatial resolution & $\sim$100m x 100m x 4 headings & segments (map matched)\\ 
\noalign{\smallskip} speed aggregation & mean of instantaneous GPS probe speeds & mean of trajectory distance over time \\ 
\noalign{\smallskip} cities & 10 & 11 \\
\noalign{\smallskip} license & academic and non-commercial, custom HERE T\&Cs & CC BY-NC \\ 
\noalign{\smallskip} probe vol & number of readings & --\\ 
\noalign{\smallskip} continents & AS, EU, NA, OC & AF, AS, EU, NA, OC, SA\\ 
\noalign{\smallskip} coverage & 108--361 days& 3--27 months \\ 
\noalign{\smallskip} collection period & 2019.01--2021.12& 2018.01--2020.01\\ 
\noalign{\smallskip} overlap & \multicolumn{2}{p{5cm}}{\hspace{.2cm}Barcelona: 3 months (2020-01 -- 2020-03)}\\
        & \multicolumn{2}{p{5cm}}{\hspace{.2cm}Berlin: 6 months (2019-01 -- 2019-06)}\\
        & \multicolumn{2}{p{5cm}}{\hspace{.2cm}London: 7 months (2019-07 -- 2020-01)}\\
        & \multicolumn{2}{p{5cm}}{\hspace{.2cm}Madrid: no temporal overlap}\\
\bottomrule
\end{tabular}
\vspace{-3mm}
\end{table}

\subsubsection{Spatial and Temporal Coverage}

\begin{figure}[ht!]
  \centering
  \includegraphics[width=0.45\textwidth]{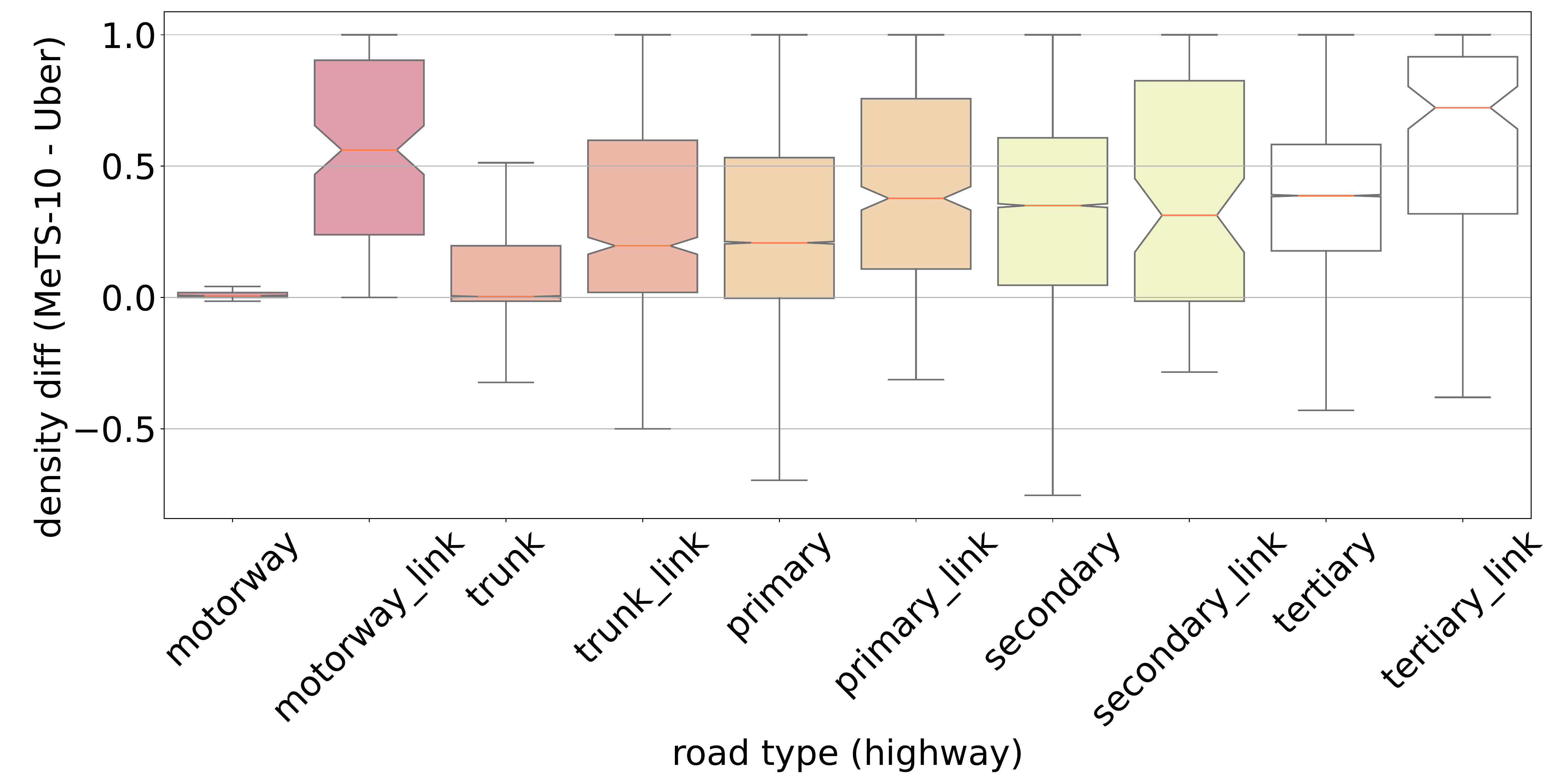}
  \caption{Segment density differences Uber and \mcswts{} London daytime (8am--6pm, \t4c bounding box only) by road type.  Color scheme: OSM Carto.
  }
  \label{fig:London_Uber_density_diff_barplot}
  \vspace{-2mm}
\end{figure}

%
As Uber data is aggregated hourly, we take the 1h-mean of our 15-min \mcswts{} speeds for the comparison.
Figure~\ref{fig:London_Uber_density_diff_barplot} shows the differences by road class for all the segments within the \mcswts{} bounding box for London.
We show the mean density difference by road class (\ie{} OSM highway attribute); positive density difference means higher temporal coverage of \mcswts{} and negative mean lower temporal coverage.
We see that \mcswts{} provides generally higher temporal coverage.

\subsubsection{Speed Differences}

\begin{figure}[th]
  \centering
  \includegraphics[width=0.45\textwidth]{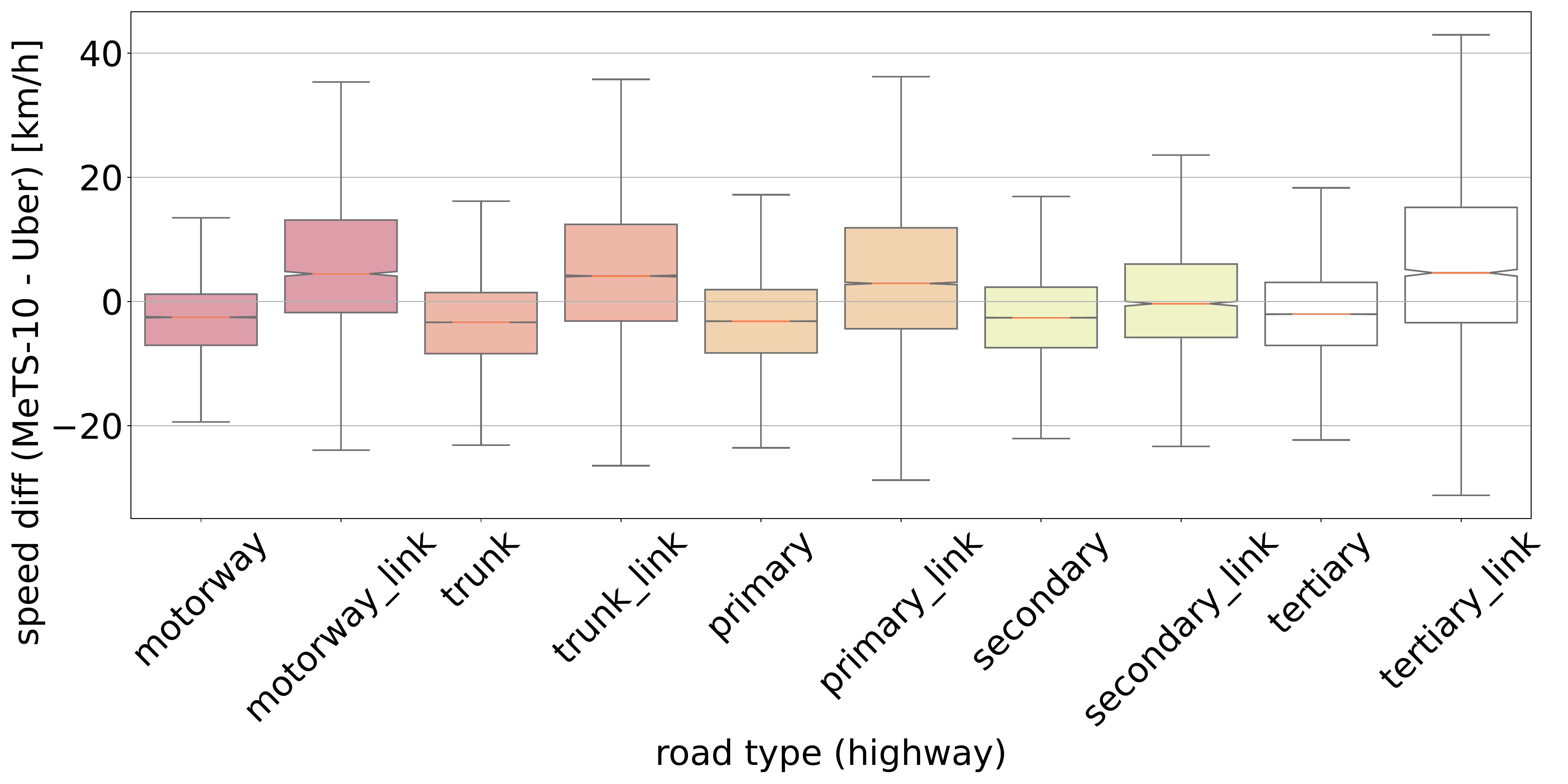}
  \caption{Speed differences Uber and \mcswts{} for London daytime (8am--6pm, \t4c bounding box only) by road type. 
  Color scheme: OSM Carto.}
  \label{fig:London_Uber_speed_diff}
  \vspace{-2mm}
\end{figure}

Figure~\ref{fig:London_Uber_speed_diff} shows the mean speed differences of the matching data, \ie{} within \mcswts{} bounding box only and where data is available at the same time and segment. Positive speed difference means higher values in \mcswts.
We see that our speeds are slightly lower, except for the link road classes, where our speeds tend to be much higher.
We suspect this is due to the aggregation strategy: \cite{uber-movement-speeds-calculation-methodology} seems to take the mean of the trajectory speeds, resulting in a time-mean speed (arithmetic mean of vehicle speeds), whereas our spot binning aggregation results in an approximation of space-mean speed (harmonic mean of vehicle speeds) as discussed above in Section~\ref{sec:methods_and_data:traffic_map_movies} and Supplement~B~\cite{neun2023metropolitansupplementary}.
We hypothesize that the higher speeds on link roads are due to links running in parallel to the corresponding higher-class roads they link to; as the Uber map-matching follows the vehicle trajectory, Uber can make a distinction here whether the vehicle is on the link or not.
These findings are confirmed in a quantitative analysis.
We use absolute percentage error as used in the traffic simulation calibration literature \cite{fhwa}. We show differences in APE between complex and non-complex road segments, differentiated by road type.

For additional material and the analysis of Barcelona and Berlin, we refer to the Supplementary Material~\cite{neun2023metropolitansupplementary} and our code repository.

\subsection{Comparison with Stationary Vehicle Detector Data}\label{sec:validation_counters}

Stationary vehicle detectors such as inductive loop detectors or road cameras are commonly used to monitor traffic. Earlier work for Berlin \cite{wagner_observations_2021} has already shown that there is a good global alignment between the \t4c data used in \mcswts{} and the traffic volumes in vehicle detector data. The good overlap between floating car data and stationary vehicle detector ground truth has also been shown in the Netherlands \cite{haak_fcd_validations_2016}. In this section, we compare vehicle detector data speed readings with the \mcswts{} values in three cities (Berlin, London, and Madrid) to highlight the differences between the two datasets and their suitability in different applications.

Table \ref{tab:counter-details} gives an overview of openly available vehicle sensor data we use in our comparisons. We focus on detectors that directly measure the speed of passing vehicles in addition to the traffic flow.
In Berlin all sensors provide speed measurements at a temporal resolution of 60 minutes while in London and Madrid fewer counters do have speed measurements but at a temporal resolution of 15 minutes.

\begin{table}[!t]
\caption{Overview of stationary vehicle detector sources.\label{tab:counter-details}}
\centering
\begin{tabular}{p{2.2cm} V{2} p{1.5cm}|p{1.5cm}|p{1.5cm}}
\toprule
 & Berlin & London & Madrid\\
\midrule
provider & VIZ Berlin \cite{berlin-counters} & TfL \cite{tfl-tims} / Highways England \cite{highways-england-webtris} & Ayuntamiento de Madrid \cite{madrid-counters}\\ 
temporal resolution & 60 min & 15 min & 15 min\\
\# all sensors & 547 & 3819 & 4413\\
\# speed sensors & 547 & 236 & 404\\
speed sensor coverage & entire city & motorways in outskirts & inner-city motorways\\
\bottomrule
\end{tabular}
\vspace{-3mm}
\end{table}

\begin{figure}[th!]
  \centering
  \includegraphics[width=0.45\textwidth]{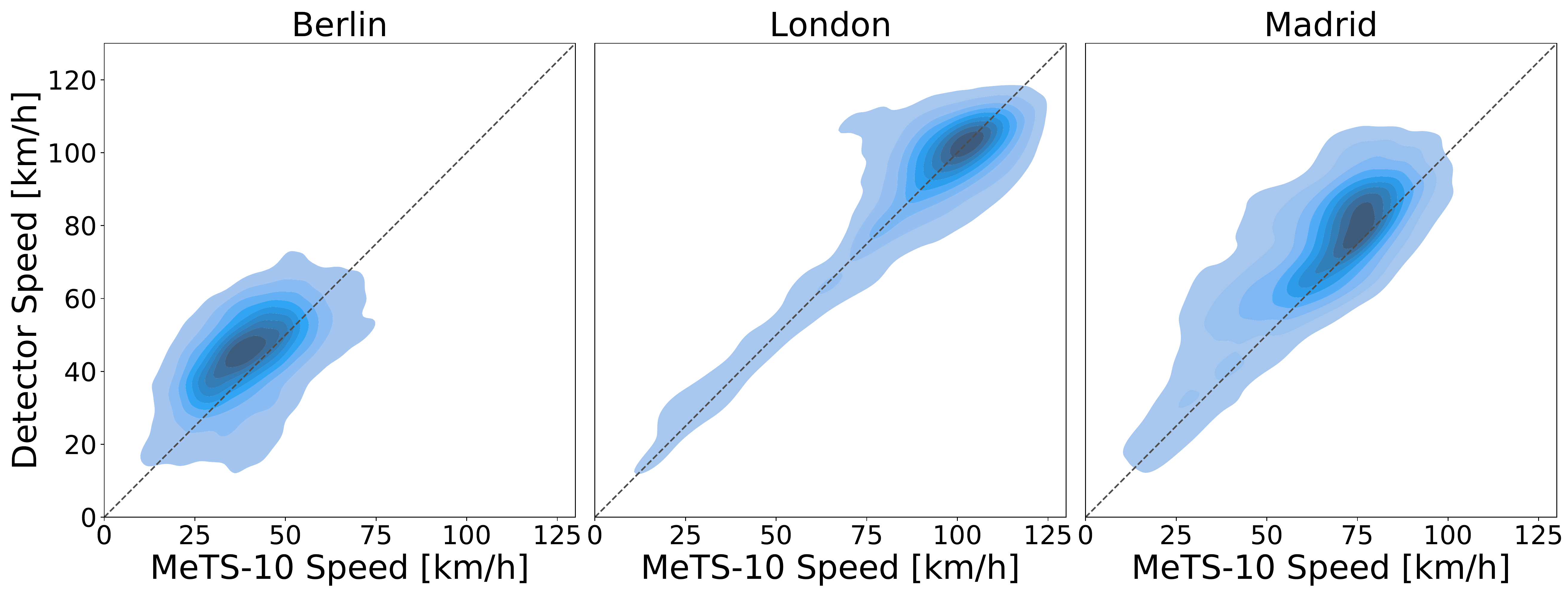}
  \caption{Binned Kernel Distribution Estimation (KDE) Plots of speeds of \mcswts{} and sensors. 
  }
  \label{fig:05val02-counter-speed-kde}
  \vspace{-3mm}
\end{figure}

Figure \ref{fig:05val02-counter-speed-kde} shows the distribution of the measured speed values of an entire day in the three cities. \mcswts{} speeds are shown on the x-axis and corresponding sensor speeds on the y-axis using 10 bins for clustering. Speed readings are from a full day between 6am and 11pm in 15-minute (resp. 60 minutes for Berlin) intervals.

On the y-axis, the average sensor speeds per time interval are used and on the x-axis the \mcswts{} speed readings on the corresponding segment. Hence, the best correspondence of the speeds is found on the diagonal line. All three cities show a good alignment of the majority of the points along the diagonals. The shapes of the KDE plots reflect the differences in the city sensor placement (\cf{} \autoref{tab:counter-details} and Figure \ref{fig:05val02-counter-diff-boxplot}, and placements in Supplement~C~\cite{neun2023metropolitansupplementary}).
In Berlin, most sensors are placed on major streets (primary/secondary) with speed limits between 50 and 60 km/h. In contrast, speed sensors in London are all along motorways
yielding a high speed density at around 100 km/h in \autoref{fig:05val02-counter-speed-kde}. In Madrid, most sensors are also along motorways, but all are within the city on the main ring road. Hence, here lower speed limits are common and there is more variance of speeds caused by the general traffic situation (traffic lights or alike).

For London, Madrid, and non-motorway/trunk highways in Berlin, the sensor speeds usually are a bit higher than the \mcswts{} speeds (see Figure \ref{fig:05val02-counter-diff-boxplot}). This is in line with the differences observed in the comparison with Uber speeds (see \autoref{sec:validation_uber}). In particular, speed sensors measure every vehicle only once, while the space mean speed of GPS data weighs slow vehicles more often than fast vehicles along the same segment (spot binning resulting in harmonic mean under idealized conditions, see Supplement~B~\cite{neun2023metropolitansupplementary}). On motorway or trunk roads in Berlin, sensors show a 10 to 15 km/h lower speed level. However, this only affects 10 sensors (2\%, see Supplement~C.B~\cite{neun2023metropolitansupplementary}) and was also observed in the comparison between Uber and sensor data (see \autoref{sec:validation_uber_counters}).

\begin{figure}[t!]
  \centering
  \includegraphics[width=0.45\textwidth]{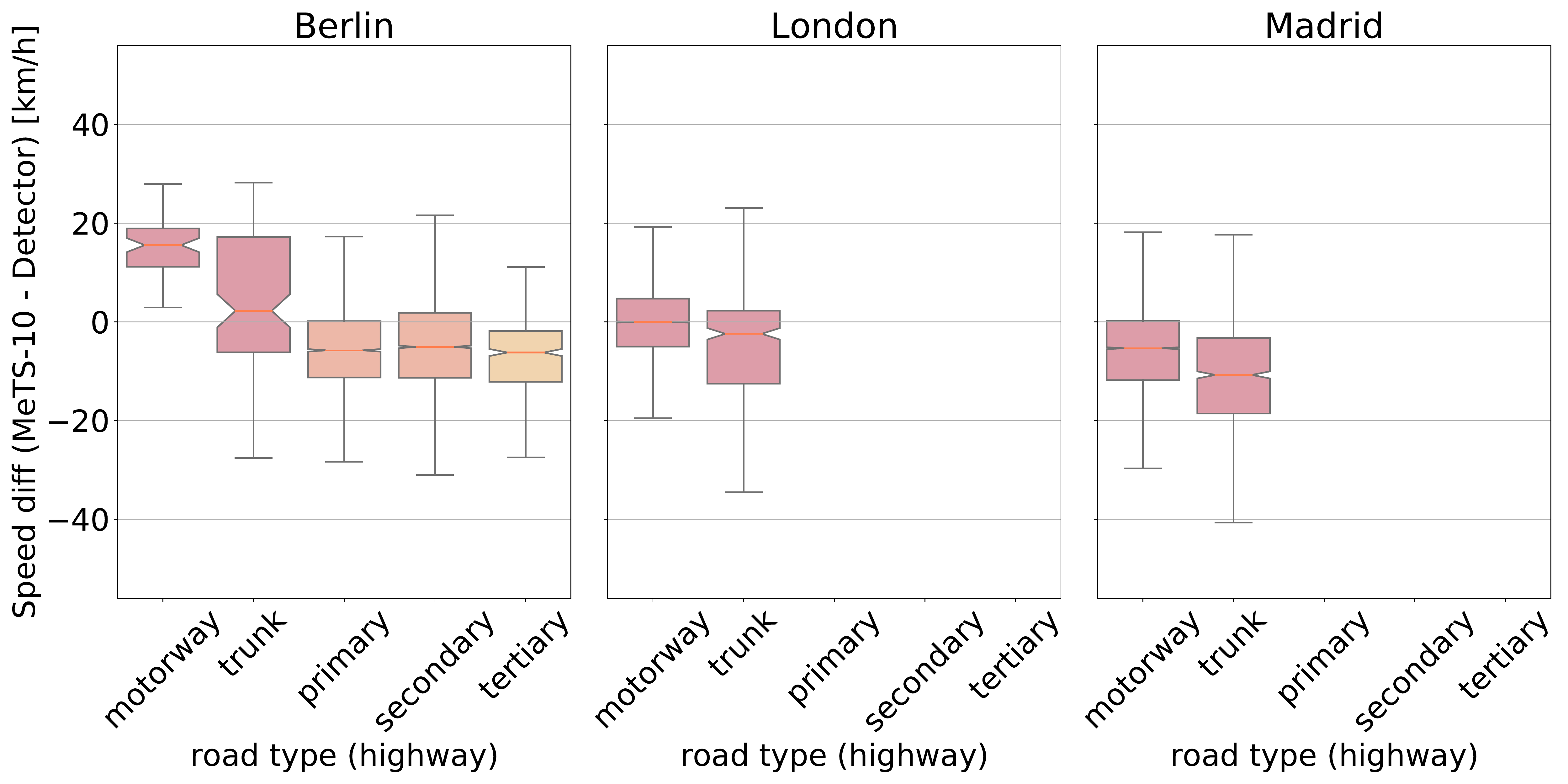}
  \caption{Box plots of speed differences between \mcswts{} 
 and vehicle detectors by highway type.
 Color scheme: OSM Carto.}
  \label{fig:05val02-counter-diff-boxplot}
  \vspace{-6mm}
\end{figure}

\begin{figure}[t!]
    \centering
    \vspace{-3mm}
    \begin{subfigure}[b]{0.44\textwidth}
    \centering
    \caption{Sensor in an undisturbed situation.}
    \includegraphics[clip, trim=0.0cm 0.3cm 0.0cm 0.6cm, width=\textwidth]{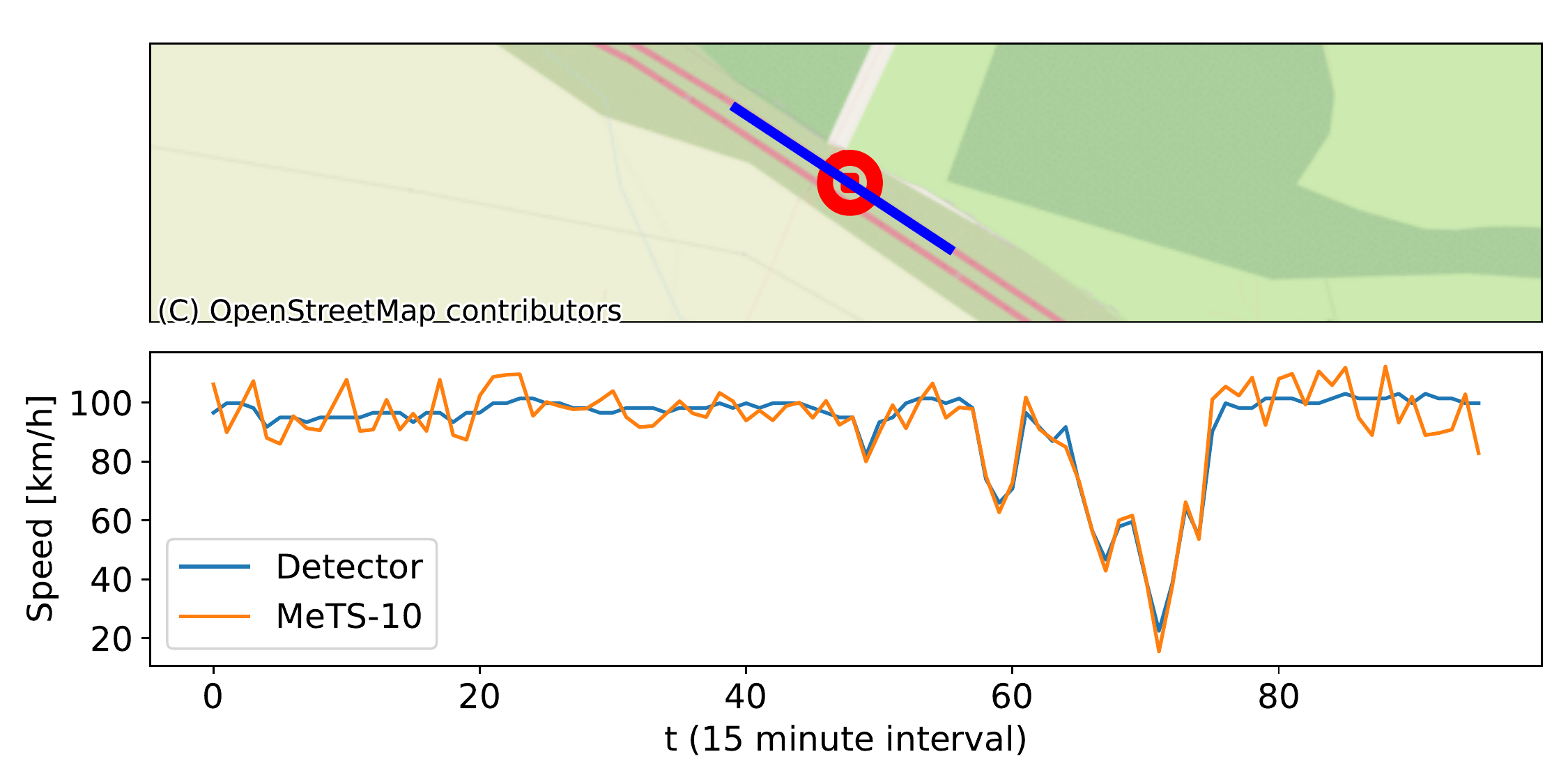}
    \label{fig:05val02-counter-situations1}
    \vspace{-5mm}
    \end{subfigure}
    \vspace{-4mm}
    \begin{subfigure}[b]{0.44\textwidth}
    \centering
    \caption{Sensor near motorway exit.}
    \includegraphics[clip, trim=0.0cm 0.3cm 0.0cm 0.6cm, width=\textwidth]{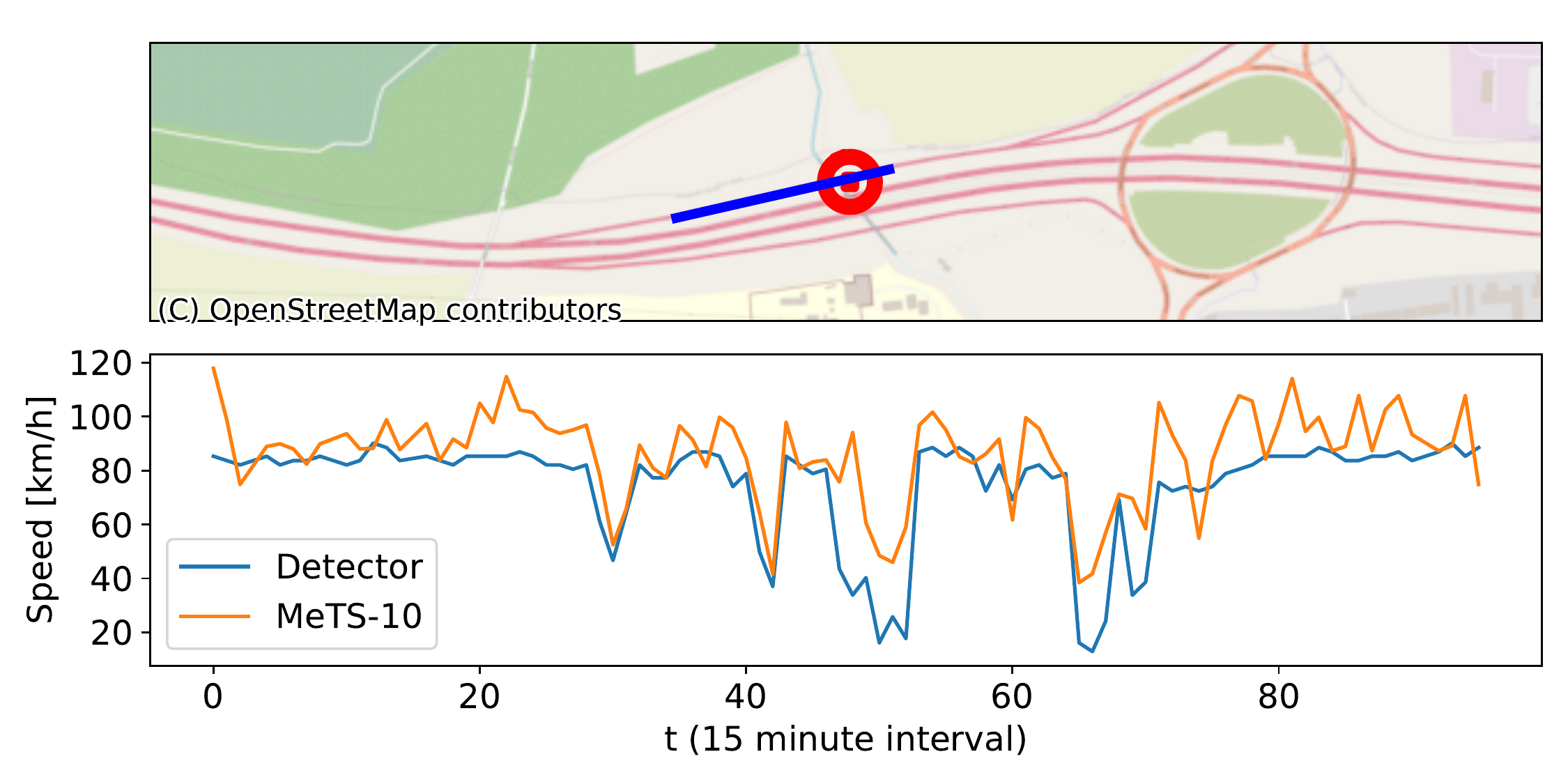}
    \label{fig:05val02-counter-situations2}
    \end{subfigure}
  \caption{Sensor placement examples along motorways in Greater London with corresponding speeds 
  during one day.
  }
  \label{fig:05val02-counter-situations}
  \vspace{-5mm}
\end{figure}

The alignment of speed readings during an entire day can be seen in Figure \ref{fig:05val02-counter-situations} with examples on motorways in London. The general alignment between the \mcswts{} speed (yellow) and the sensor speed (blue) is good but \mcswts{} shows a higher variance and responsiveness to changes. For example, during evening and night times in the undisturbed situation with no intersecting roads or lane changes (Figure \ref{fig:05val02-counter-situations1}) the sensor speed is significantly smoother. A possible reason is the fact that the sensor speed is the mean over 15 minutes in exactly the same location while the \mcswts{} speed contains a median over multiple values along the segment.

In addition, in situations where the sensor is near a motorway exit (Figure \ref{fig:05val02-counter-situations2}), the general \mcswts{} speed level might be increased and smoothed by signals from the nearby motorway segments which results in less accentuated slowdowns during rushhour ($t\approx50$ and $t\approx70)$ with the lowest \mcswts{} speed approx 20km/h higher than the sensor speed.

Overall the comparisons and examples in the three cities show that \mcswts{} speeds show a good general correspondence with the sensor speeds. The differences seen can be explained sufficiently through the differences in data collection and aggregation methods.

\subsection{Baseline comparison of Uber Movement Speeds with Stationary Vehicle Detector Data}\label{sec:validation_uber_counters}

As a baseline and additional sanity check of the used ground-truth-like data, we compared the Uber Movement Speeds data with the stationary vehicle detector speeds in Berlin and London. The matching approach was the same as for the comparison with \mcswts{} in Section \ref{sec:validation_counters}.
The detailed plots of the analysis in Supplement~C.C~\cite{neun2023metropolitansupplementary} show similar effects and differences as discussed with regards to Figures \ref{fig:05val02-counter-speed-kde} and \ref{fig:05val02-counter-diff-boxplot}. For example, the speed differences for motorways in Berlin are also present
with average speed differences 12.7$\pm 7.2$ km/h for \mcswts{} and 13.5$\pm$7.8 km/h for Uber.
Over all road types, we have -2.9$\pm$13.4 km/h for \mcswts{} and -2.6$\pm$14.2 km/h for Uber. 

\section{Discussion}\label{sec:discussion}
In this section, we discuss the opportunities and limitations of our design choices of our method and validations.

\subsection{Discussion of Method Design Choices}

\paragraph{
Traffic Map Movies} 
The GPS probe data 
originates from connected vehicles from different probe data providers \cite{here-launches-advanced-real-time-traffic-service,here-real-time-traffic}. HERE Technologies leverage many different probe data providers in collaboration with leading automotive companies, logistics providers,  city governments and transportation agencies \cite{here-launches-advanced-real-time-traffic-service}. Information on the exact set of probe data providers and provider changes in each city is not publicly available. The HERE Traffic Map Movies are not restricted to a single GPS probe provider, e.g., taxis or buses, as other floating car data. 
%
%
Hence, as for other floating vehicle datasets \cite{bruwer_fcd_bias_2022}, the dataset only reflects a (biased) subset of road traffic in the cities. 
Clearly, volumes do not reflect total traffic and we assume they are also biased towards the vehicle fleet of connected cars providing GPS probes; Uber, for instance, does not publish probe volume for this reason \cite{uber-movement-faqs}.
On the other hand, speeds can be expected to be accurate and representative apart from notorious situations like lower speed limitations for transport vehicles on motorways and vehicles stopping for delivery of goods or people boarding or alighting.


From a more theoretical point of view, it would be interesting to explore the characteristics of spot binning under different assumptions (spatial and temporal resolution, GPS emittance rate, sampling bias and flow conditions, road network complexity), \eg{} using traffic simulations.


Whereas the Traffic Map Movies dataset does not cover long enough periods for long-term trend analysis, it could still be useful for daily and/or seasonal traffic analysis.


\paragraph{
OpenStreetMap Road Graph
(dp03)} 
We chose to use OpenStreetMap data due to its availability for research. The method and the source data can be applied to any directed road graph. For instance, we use our pipeline to also map the Traffic Map Movies to the non-simplified historic OSM road graph used by Uber.
The flexibility of using any road graph helps to prevent practical usage limitations when the road graph evolves or a different type of road graph is preferred.


\paragraph{
Spatial Intersection 
of Road Graph and City Cells 
(dp04)}

In order to investigate whether our method suffers from having only 4 headings in the Traffic Map Movies, we compare the coverage on segments mainly along diagonals and along the horizontal/vertical main axes.
We see only a slightly higher coverage on the segments along the main axes.
For more details, we refer to the Supplementary Material~\cite{neun2023metropolitansupplementary}.

\paragraph{
Temporal Aggregation of 
Traffic Map Movies (dp01)} 
We chose to default to 15-minute aggregation of Traffic Map Movies as it is the least common multiple of the temporal resolution of many public stationary vehicle detector datasets. 
Depending on the use-case, the raw 5-minute of HERE Traffic Map Movies can also be used.


\paragraph{
Spatial Join (dp06a)}
We chose the median of all intersecting cells for its robustness against outlier data, \eg{} data from a higher-speed road distorting the value of a lower-speed road nearby. 
The described temporal-spatial join gives equal weight to all cells that intersect a road segment -- we chose it 
for its simplicity. Our method could be refined by giving different weights to these intersecting cells based on volume or geometric properties, \eg{} intersection length or  alignment with spot binning heading. 


\paragraph{
Confidence Filtering of Segment Speeds (dp06b)}

The confidence-based filtering approach of \mcswts{} is a pragmatic implementation for the 
matching of the aggregated speed values. 
The congestion factor thresholds (as coming from \cite{here_mapping_traffic_congestion}) and derived free flow values do not need to be perfectly calibrated as they are only used during filtering but not directly for the computation of the median segment speeds.
We used sample segments to validate the chosen setting, aiming at a balance between reducing noise through filtering and ensuring high data quality.

The filtering is not overly aggressive.
For instance, in Madrid (moderate data availability), around 20\% of input speeds are discarded.
The chosen thresholds do work well for our use-case of a general-purpose segment speed dataset as well as in the comparisons with Uber and the stationary vehicle counters. 



From a methodological point of view, the effect of this filtering would merit additional investigations and refinements, \eg{} for data smoothing and data imputation. In particular, we never filter speeds much higher than signalized speeds, which could be done during aggregation.

\subsection{Discussion of Validations}

There is no existing baseline method for mapping spot-binned traffic data such as the HERE Traffic Map Movies to a road graph. 
Also, there is no per-se ground-truth data: First, there is only partial spatio-temporal overlap with the Uber dataset in addition to the collection method being different. Second, stationary vehicle detectors are spatially sparse and may not capture the full road segment (relevant at traffic lights) and have no fleet-bias as GPS probe based datasets. Third, the vehicle trajectories used by HERE to create the Traffic Map Movies are not openly available for research. Hence, we resort to comparing to two other available datasets with partial temporal-spatial overlap.

The differences between \mcswts{} from loop counters are not higher than between Uber and loop counters. This shows that the spot binning method of Traffic Map Movies gives similar results as trajectory-based methods used by Uber, at lower computational costs.

While the comparison to Uber and vehicle detector data demonstrated good representativeness and coverage, \mcswts{} is of course not free of biases. 
The speed channel can be biased in complex road network situations 
(where probes from parallel roads closeby may interfere) 
and under low-volume conditions.
(where a single stopped car emits many signals). 
The effects are stable and can be mitigated using \eg{} the standard deviation, which is provided in addition to the median speeds, for filtering. 
The volumes provided are biased by the provider's vehicle fleet composition; however, they can be a valuable additional signal for sanity checks if interpreted carefully. 
Similarly, our free flow speeds are data-driven and could be compared with other data sources, and the divergences between them and signalized speeds could be used for map correction or roadworks detection. 

Furthermore, the daily temporal coverage is highly dependent on the road class and is usually very good for highway types such as motorways. There is also a bias depending on the mix of the city's providing vehicle fleet and on the location within the city.

In order to further validate our speed dataset, corridors could be sampled and ETAs derived from our speeds could be compared with ETAs from routing engines or real ETAs from taxi fleets for typical daytime situations.

\section{Conclusion}\label{sec:conclusion}

In this paper, we have presented the new \mcswtslong{} (\mcswts{}). The source data covers 10 large metropolitan areas with 108 to 361 days of sampled data per city and with a temporal resolution of 5 minutes. In contrast to other datasets, \mcswts{} neither suffers from low spatial coverage as common vehicle detector datasets, nor is it 
restricted to a single GPS probe provider, e.g., taxis or buses, as other floating car data. 
Specifically, our validation experiments showed that \mcswts{} has better spatial and temporal coverage than the Uber speeds within the main city area. Therefore, depending on the use-case, there can be an additional gain from combining the two data sources.
Floating car data availability is also often limited due to privacy constraints. The Traffic Map Movie format of the source data circumvents such restrictions through its spatio-temporal aggregation. With the proposed pipeline, we are offering an efficient and standardized way of disaggregating and matching such aggregated floating car data to a road graph.

The comparisons with the Uber dataset on the one hand as well as the stationary vehicle detector datasets on the other hand show the good representativeness and coverage of the \mcswts{} dataset. The limitations \eg{} for complex network situations, low-volume situations, and location and road class dependent effects of vehicle fleet composition are understood and discussed. 
Spot binning, the aggregation method of Traffic Map Movies, is shown to correspond to space-mean speeds (harmonic mean of vehicle speeds) under idealized conditions, resulting in potentially lower speeds than time-mean speed (arithmetic mean of vehicle speeds).
The effects are stable and researchers can use the additional attributes 
at their location of interest for filtering, \eg{} using the standard deviation and volume channel. 

The large spatial and temporal coverage offers the opportunity to also complement other datasets and extend on the disaggregation, matching, or fusing techniques for other data-driven use-cases.
The sources and the complete code for creating the dataset are publicly available. This open approach allows the extension and modification of the matching algorithm as well the use of other or customized road graphs, preventing many practical limitations, for example when the road graph evolves or with unavailable historic road graph data. It can also be a suitable approach for other GPS data sources and potentially even for coarser data such as mobile phone location information.
\mcswts{} is also a candidate for the creation or extension of benchmark datasets for common domain specific tasks such as short-term traffic prediction and also for more general machine learning tasks \eg{} in the area of graph neural networks. This can help to compare the performances of different approaches, 
as well as to identify potential biases and limitations in models and datasets.

\section*{Data and Code Availability}\label{sec:data_and_code_availability}

The \t4c movies input data is available from \cite{here:sample-data}.
The code to create the dataset and to reproduce the analysis and figures in this paper is available in our GitHub repository\footnote{\url{https://github.com/iarai/MeTS-10}}. The Uber data is available from \cite{uber-movement}, the OSM data for the historic road graphs is available from Geofabrik\footnote{\url{https://download.geofabrik.de/}}, and the vehicle detector data from VIZ Berlin \cite{berlin-counters}, TfL \cite{tfl-tims},  Highways England \cite{highways-england-webtris} and  Ayuntamiento de Madrid \cite{madrid-counters}.

\section*{Acknowledgment}


\addcontentsline{toc}{section}{Acknowledgment}
The authors would like to thank HERE Technologies for making the \t4c competition data available.

\section*{Author Contributions Statement}

\addcontentsline{toc}{section}{Author Contributions Statement}
Following CRediT (Contributor Roles Taxonomy\footnote{\url{https://credit.niso.org/}}), the authors have contributed as follows.

{\it Writing -- original draft (integral), Software, Data Curation}: M.N. and Ch.E.
{\it Validation, Visualization, Writing -- original draft (Validation Section)}: Y.X., Ch.F., M.N., and Ch.E.
{\it Writing -- original draft (Introduction and Related Work)}: N.W. and H.M.
{\it Conceptualization, Writing -- review \& editing}: all.

\bibliographystyle{IEEEtran}
\bibliography{references}


\begin{IEEEbiographynophoto}{Moritz Neun}
Researcher and director of research engineering at IARAI. His background is geo-spatial big data processing and analytics, geo-spatial search and transport optimization.
\end{IEEEbiographynophoto}
\vspace{-10mm}
%
%
\begin{IEEEbiographynophoto}{Christian Eichenberger}
Researcher and engineer at IARAI. His background is theoretical computer science (algebraic information theory) and transport engineering.
\end{IEEEbiographynophoto}
\vspace{-10mm}

\begin{IEEEbiographynophoto}{Yanan Xin}
Lead at the Mobility Information Engineering(MIE) Lab as a postdoctoral researcher at the Chair of Geoinformation Engineering, ETH Zurich. Her background is in Geographic Information Science.
\end{IEEEbiographynophoto}
\vspace{-10mm}

\begin{IEEEbiographynophoto}{Cheng Fu}
Group Leader of Urban Geoinformatics and Lecturer in the Department of Geography, University of Zurich (UZH), Switzerland. His background is in Geography, Remote Sensing and GIS.
\end{IEEEbiographynophoto}
\vspace{-10mm}

\begin{IEEEbiographynophoto}{Nina Wiedemann}
PhD student at the Mobility Information Engineering (MIE) lab at ETH Zurich. Her Background is in Cognitive Science and Data Science with a focus on machine learning and optimisation theory.
\end{IEEEbiographynophoto}
\vspace{-10mm}

\begin{IEEEbiographynophoto}{Henry Martin}
PhD student at the Mobility Information Engineering (MIE) lab at ETH Zurich and at IARAI. He has a Master's in Electrical Engineering and Information Technology from the Technical University of Munich.
\end{IEEEbiographynophoto}
\vspace{-10mm}

\begin{IEEEbiographynophoto}{Martin Tomko}
Associate Professor in Spatial Information Science at the University of Melbourne. His research covers spatial data science, trajectory analytics and spatial data management.
\end{IEEEbiographynophoto}
\vspace{-10mm}

\begin{IEEEbiographynophoto}{Lukas Ambühl}
Post-Doctoral Researcher at Research Group of Traffic Engineering, Institute for Transport Planning and Systems, ETHZ. His background is civil engineering and management, technology, and economics.
\end{IEEEbiographynophoto}
\vspace{-10mm}

\begin{IEEEbiographynophoto}{Luca Hermes}
PhD Student at Machine Learning Group, Bielefeld University. His background is in machine learning with a focus on graph representation learning.
\end{IEEEbiographynophoto}
\vspace{-10mm}

\begin{IEEEbiographynophoto}{Michael Kopp}
Founding director of IARAI.
His background is in pure mathematics, mathematical modelling in finance and machine learning.
\end{IEEEbiographynophoto}

\clearpage
\pagebreak
\newpage
\newpage
\newpage
%

\setcounter{page}{1}
{\huge\centering \it Supplementary Material: \\\mcswtslong{}}

\renewcommand\appendixname{Supplement}
\appendices

\section{Complements on Input Data and Data Pipeline}\label{appendix:methods_and_data}

\subsection{Input Data}
Here, we describe the two input data sources in terms of their data model, collection, provisioning, and license.

\subsubsection{Traffic Map Movies}\label{app:methods_and_data:traffic_map_movies}
\paragraph{Dataset presentation}

Raw GPS data face privacy issues if individual users' behavior can be deduced from the data \cite{montjoye_privacy_2013,krause_privacy_2008, krumm_privacy_2007}. There are methods to preserve privacy by transforming the data, such as obfuscation, aggregation, privacy thresholds or snipping \cite{rass_privacy_2008,ardagna_privacy_2011,bo_privacy_2018}. HERE Traffic Map Movies use spatio-temporal aggregation and privacy thresholding -- we use the term \emph{spot binning} for  this aggregation method. Another motivation for framing a geo-spatial time series forecasting as a video prediction task \cite{pmlr-v123-kreil20a} is to leverage state-of-the-art deep-learning methodologies in image and video processing \cite{jiang_geospatial_2019}. The data was made available for the \t4c competition series \cite{pmlr-v123-kreil20a,pmlr-v133-kopp21a,pmlr-v176-eichenberger22a} at NeurIPS, an annual machine learning conference.
A third motivation for bringing up GPS data in a map-free form is that spatio-temporal binning of GPS probes is computationally cheap in contrast to map-matching, and that it is also applicable in situations where no appropriate maps are available or not accurate enough \cite{saki_mapmatching_2022}.

Traffic Map Movies Data are provisioned in HDF5 (.h5), a format for typed multidimensional arrays \cite{hdfgroup:solutions:hdf5,wikipedia:hierarchical-data-format}.
It can be downloaded for free from the HERE sample data website \cite{here:sample-data}.
The data must be used solely for academic and non-commercial purposes and under standard HERE terms \& conditions \cite{here:terms-and-conditions}, which in particular explicitly forbid re-distribution of HERE materials or derivatives thereof in combination with any open source or open data licenses.

\paragraph{Dataset generation}

GPS probes are binned spatially for each heading direction quadrant of North--East (heading 0\textdegree--90\textdegree), North--West (heading 270\textdegree--0\textdegree), South--East (heading 90\textdegree--180\textdegree), and South--West (heading 180\textdegree--270\textdegree)  into an $8$-channel encoding 
 (see Figure~\ref{fig:intersecting_cells_overview}), where two features are calculated :
\begin{itemize}
  \item Volume: The number of probe points recorded from the collection of HERE sources capped both above and below and normalized and discretized to an integer number between $0$ and $255$ ($\ZZZ_{256}$). 
  \item Mean speed: The average speed from the collected probe points. The values are capped at a maximum level and then discretized to $\{1,2,\ldots, 255\}$, by linearly scaling the capping speed to $255$ and rounding the resulting values to the nearest integer. If no probes were collected (\ie{} the volume is $0$), the speed value is $0$. This has to be taken into account when averaging speeds.
\end{itemize}
More formally, a GPS probe $(t,x,y,\alpha,v) \in \RRR \times \RRR \times [0,360] \times \RRR^+$ consists of a timestamp $t$, a position $(x,y)$, an angle $\alpha$ and a speed $v$.
A spatio-temporal binning is a projection  $\pi$ to bins $(day,t,row,col,heading)\in \NNN \times  \ZZZ_{288} \times \ZZZ_{495} \times \ZZZ_{436} \times \{\textrm{NE},\textrm{NW},\textrm{SE},\textrm{SW}\} $.
The aggregation of a set $P$ of GPS probes produces 
\begin{eqnarray}
vol_{d,t,r,c,h} = {\ulcorner\left((\left \vert \{\substack{(tt,x,y,\alpha,v) \in P: \\\pi(tt,x,y,\alpha,v)=(d,t,r,c)}\} \right \vert - \theta)_{\bot 0}^{\top \kappa} \frac{255}{co} \right)\lrcorner}
\end{eqnarray}
and
\begin{eqnarray}
speed_{d,t,r,c,h} = 
\begin{cases} 
    {\ulcorner\left(\overline{\{v_{\bot 0}^{\top 120} : \substack{(tt,x,y,\alpha,v) \in P, \\\pi(tt,x,y,\alpha,v)=(d,t,r,c)}\}}\frac{255}{120}\right)\lrcorner} \\ 
    \qquad\textrm{     if } vol_{d,t,r,c,h} >0\\
    0 \textrm{     if } vol_{d,t,r,c,h} =0
\end{cases}
\end{eqnarray}
where clipping below is denoted by $x_{\bot b} = \max(x,b)$, clipping above by $x^{\top a} = \min(x,a)$, integer rounding as $\ulcorner \cdot \lrcorner$, and where $\theta$ is a privacy volume threshold and $\kappa$ a volume cutoff.
This means volume is set to $0$ if the probe volume does not reach the privacy threshold.

\IEEEpubidadjcol

Intuitively, this \emph{spot binning} favors lower speeds as slower cars stay longer in the same spatio-temporal bin and are counted multiple times.
Under idealized conditions (see Supplement~B), 
the spatio-temporally aggregated speed $speed_B$ represents the total distance divided by the total travel time (see Eq.~(\ref{eq:totaldistancetotaltraveltime_main_text})), which is the harmonic sum of the speeds of Eq.~(\ref{eq:harmonic_main_text}), \ie{}
\begin{eqnarray}
speed_B &=& \frac{\sum_k 1}{\sum_k v_k^{-1}}\label{eq:harmonic_main_text}\\
&\stackrel{}{=}& \frac{s \cdot \sum_k 1}{\sum_k t_k}\label{eq:totaldistancetotaltraveltime_main_text}
\end{eqnarray}
for a bin $B$ covering a road segment of length $s$ and for virtual vehicles indexed by $k$ at speeds $v_k$ taking time $t_k$. 
In particular, this requires controlling the probe rate of vehicles and depends on traffic volume and homogeneity.
Future work could establish bounds to control these factors, \eg{} through simulations.

\subsubsection{OpenStreetMap}

OpenStreetMap (OSM) is a database of GIS data built by an open community of contributors \cite{OSM:about}.
Its data model \cite{OSM:elements} has three main elements: \emph{nodes} represent a specific point on the earth's surface (id number and a pair of coordinates),
\emph{ways} define polylines (ordered list of nodes), and \emph{relations} 
between two or more data elements (nodes, ways, and/or other relations), optionally with different \emph{roles}.
Hence, the OSM data model does not directly describe a road graph, a traversable graph needs to be derived from the OSM elements. 
OSM data comes under the Open Database License (ODbL) \cite{OSM:Licensce}, requiring attribution of public use, share-alike (under the same license) and open redistribution. 
We use the Overpass API \cite{OSM:OverpassAPI} to download OSM data.

\subsection{Data Pipeline}



\begin{figure}[!t]
\centering
\includegraphics[width=3.5in]{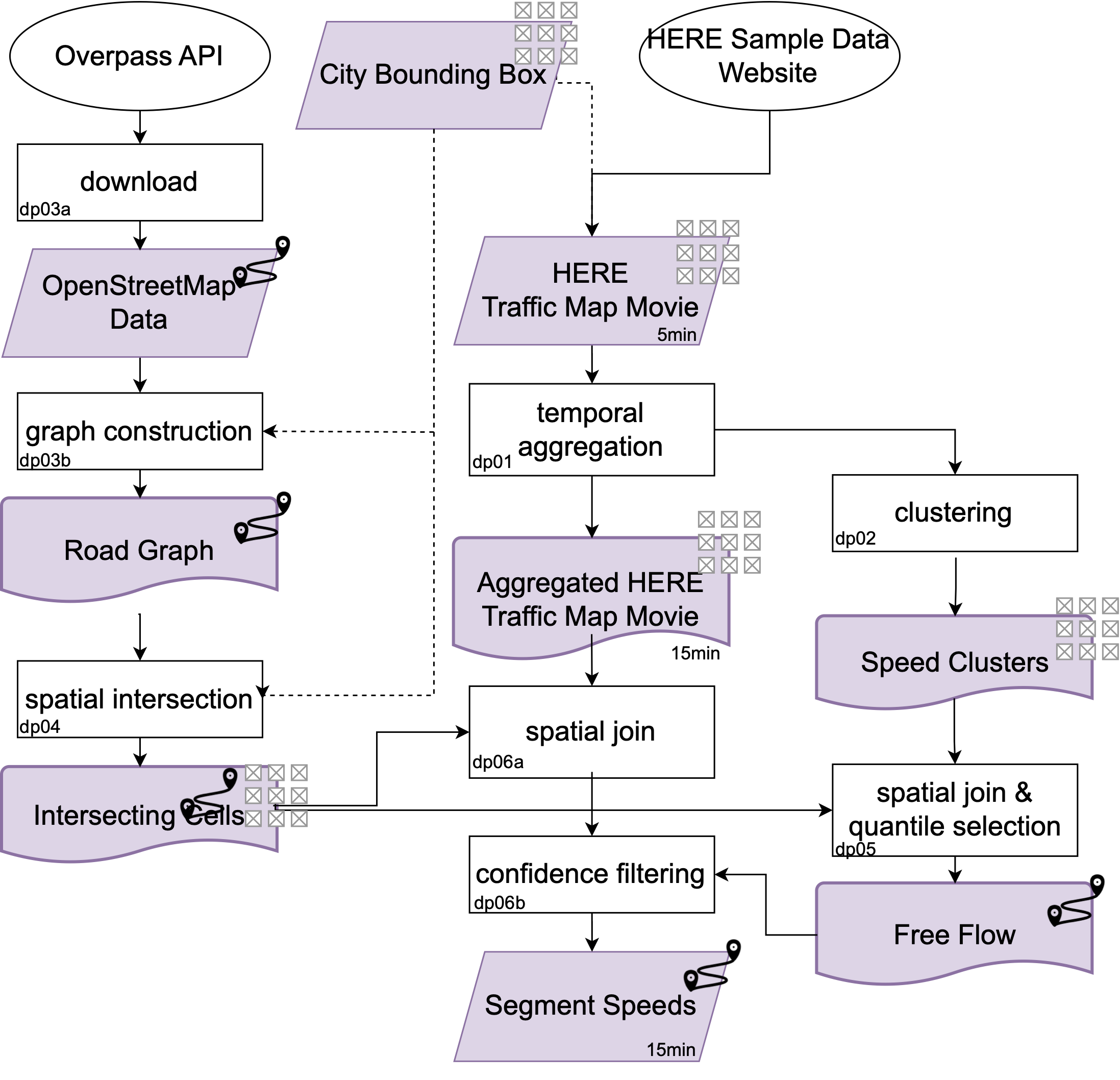}
\caption{Methods and data flowchart. Rectangles represent processing steps; ovals represent data repositories; rhomboids represent input and output; 
rectangles with wavy base represent data artifacts. There are two reference schemes (top-right of rectangles): spatial grid and road graph; both are geo-referenced to allow for spatial joining. The temporal resolution is marked bottom-right on rectangles. Arrows represent data flow. Processing steps are labeled by \texttt{dp<N>} bottom left referring to the prefix of the corresponding Python script of the data pipeline (some processing steps are implemented in the same script, indicated by suffixes \texttt{a}/\texttt{b}).
}
\label{fig:04_methods_and_data/pipeline_overview}
\end{figure}

Referring to \autoref{fig:04_methods_and_data/pipeline_overview}, we now describe the steps of our data pipeline in more detail. Each step is prefixed by \texttt{dp<N>} and corresponds to a standalone Python script in our public GitHub repo (see \textit{Data and Code Availability} below). Our code is released under Apache License 2.0 \cite{apache:license20}. This allows our method and code to be used and improved permissively in research and even in commercial use cases. The GitHub repo also contains a more technical data specification of all the (intermediate) data formats.

\subsubsection{OpenStreetMap Data Download and Road Graph Construction (dp03)} 

The resulting graph is a primal (road junctions are vertices and road links are edges), non-planar, weighted multidigraph with self-loops and preserves one-way directionality \cite{BOEING2017126,Marshall2018-ob}.
Formally, an edge is uniquely identified by a triplet $(u,v,g)$ where $u,v$ are node ids and where $g$ is a road geometry; we use OSM node IDs for $u,v$ and an integer hash of the road geometry for $g$.

In addition, we add the legal speed limit as an edge attribute. The source is the OSM \verb|maxspeed| tag.
Due to graph simplification, multiple such \verb|maxspeed| values can be present per edge or can be missing. 
By default, we use the \texttt{OSMnx} implementation which assigns the mean of multiple values.
In some cases (\eg{} in Madrid) the \texttt{OSMnx} implementation sees parsing errors. Therefore, we also provide an alternative implementation that takes the max if multiple values are present.
In the presence of missing values, the \texttt{OSMnx} implementation imputes the mean of the corresponding OSM highway type in the data, whereas our implementation uses hard-coded defaults for different OSM highway types.

Note that any other road graph present in this form could be used instead of a road graph downloaded from OSM.

\subsubsection{Spatial Intersection of Road Graph and City Cells (dp04)}

\begin{figure}[t!]
  \centering
  \includegraphics[width=0.45\textwidth]{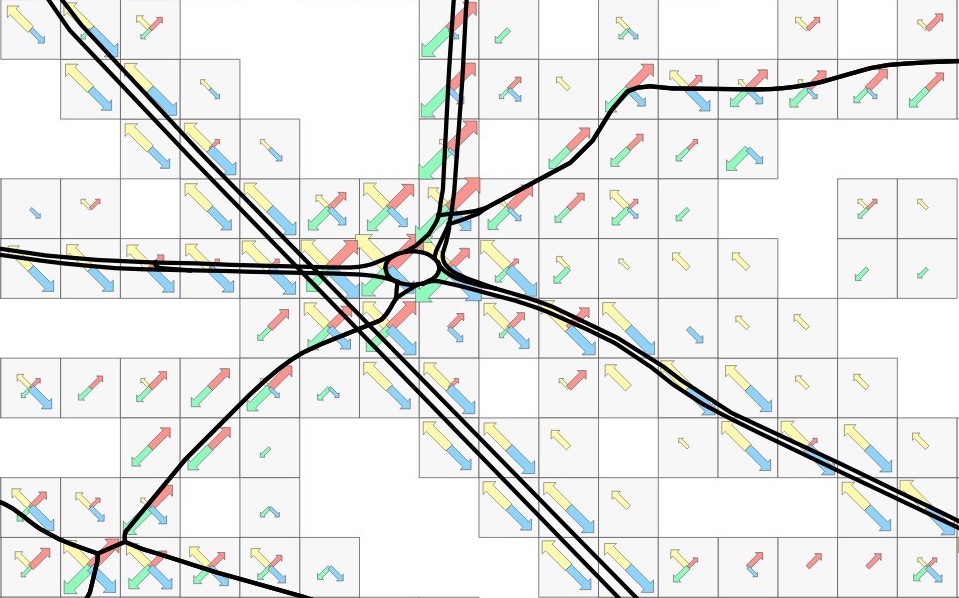}
  \caption{Probe data and road network example from London. Black lines are the road centerline segments, grey boxes show grid cells with probe data, the sizes of the arrows correspond to the summed volumes of a sample day along the 4 headings, each heading with a different color. Only the data aligned with the road are mapped to the edge. There are some cells with data but without a road graph intersecting.}
  \label{fig:intersecting_cells_overview}
\end{figure}

\begin{figure}[t!]
    \centering
    \begin{subfigure}[b]{0.45\textwidth}
    \centering
    \includegraphics[width=\textwidth]{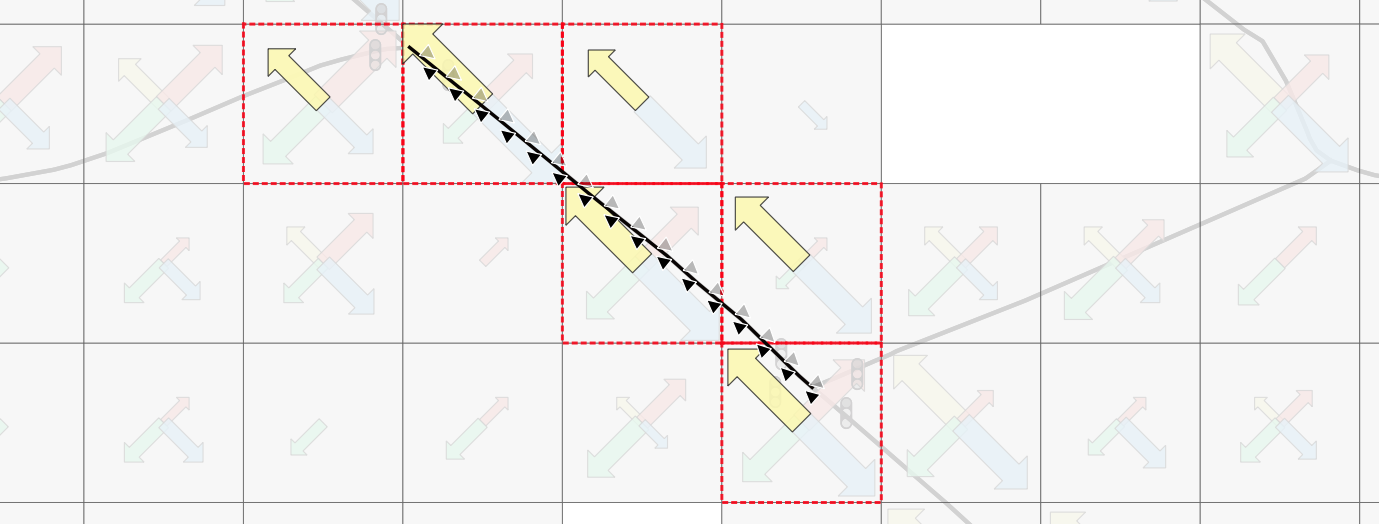}
    \caption{Intersecting cells for one road segment in only one direction (bottom right to top left as indicated by the black arrow heads). The red boxes and yellow arrows in the foreground show the intersecting cells of this road segment. The data from non-intersecting cells of this road segment are greyed out -- only the data along the NW heading are mapped on this road segment as the road segment is well aligned with the NW heading.}
    \label{fig:intersecting_cells_detailed1}
    \end{subfigure}
    \\
    \begin{subfigure}[b]{0.45\textwidth}
    \centering
    \includegraphics[width=\textwidth]{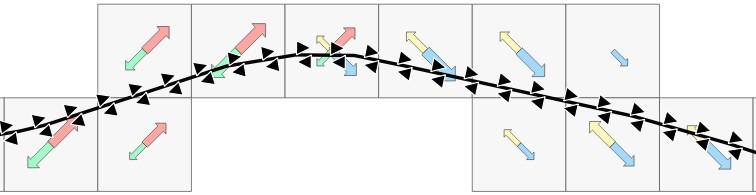}
    \caption{Intersecting cells of both the two road segments going from left to right and from right to left (as indicated by the arrow heads in both directions). When the road is close to horizontal, data from NE and SE are used for the direction left to right while both NW and SW are used for right to left.
    When the road is close to the diagonal, there is only data from the corresponding headings. On the upper right, there is also one cell not intersecting with the road segment but within the 5m margin, so its data is mapped onto the corresponding direction of the edge as well. In this example, all data shown are mapped to either one of the two road segments.}
    \label{fig:intersecting_cells_detailed2}
    \end{subfigure}
  \caption{Intersecting cells example from London illustrating the mapping for individual roads.}
  \label{fig:intersecting_cells_detailed}
\end{figure}

This step generates lists of intersecting cells for each road segment in the corresponding Traffic Movie Grid.
By interpolation over the edge geometries with a constant step size, we get a list of  $(row, column, heading, fraction)$ where $(row,column,heading)$ denotes a directed cell and $fraction$ is the percentage of the length of the segment overlapping with this cell. The fraction can be zero as we add data from neighboring cells (up to a margin of $0.0005^\circ\sim 5\,\textrm{m}$ by default), and the sum of fractions in the intersecting cells can be larger than one as we add data from neighboring headings (margin of $10^\circ$ by default); for edges going over the city boundary, the sum of fractions can also be smaller than one, obviously. See illustration in Figures~\ref{fig:intersecting_cells_overview}--\ref{fig:intersecting_cells_detailed}. These fractions are currently not used in our pipeline, but they could used for a weighting the contributions of different intersecting cells.

\subsubsection{Temporal Aggregation of HERE Traffic Map Movies (dp01)} 

By default, we aggregate 3 consecutive 5-minute movie time bins into 15-minute bins by summing volumes and taking the mean of speeds after invalidating speeds with zero volume. 


\subsubsection{Confidence Filtering of Segment Speeds (dp06b) 
}



The confidence-based filtering $\mathit{conff}$ is given by
\begin{equation}
\mathit{conff}(\mathit{ssp}, \mathit{vol}, \mathit{ff}) = 
\begin{cases}
\texttt{nan} & \text{ if } \mathit{vol} < 5 \textrm{ or } \mathit{cf} < 0.4 \\
\texttt{nan} & \text{ elif } \mathit{vol} < 3 \textrm{ or } \mathit{cf} < 0.8 \\
\texttt{nan} & \text{ elif } \mathit{vol} < 1 \\
ssp & \text{else}
\end{cases}
\end{equation}
for median segment speed $\mathit{ssp}$, probe volume $\mathit{vol}$, free flow $\mathit{ff}$ (opt. normalized) and congestion factor 
\begin{equation}
\mathit{cf} = \frac{\mathit{ssp}}{\mathit{ff}}.
\end{equation}
In order not to need to keep the intermediate results, confidence filtering is implemented in the same script dp06 as the derivation of median speeds.

\subsubsection{Speed Clustering (dp02) and Free Flow speeds from Spatial Join and Quantile Selection (dp05)}

Here, we describe the derivation of free flow speeds from the data. This step derives free-flow speeds for each road segment using the speeds clusters in the Traffic4cast Movie Grid data.

We compute the 5 most dominant speeds clusters for every cell and heading from the aggregated 15 minute traffic
map movies. By default, all speed values from 20 days of data are clustered using the K-means clustering
algorithm for every cell/heading combination (495x436x4$\approx$860K). Hence, this process is computationally expensive
and can easily take 2--3 hours per city on a standard consumer laptop.  We then take the cluster median as the center representation of the speed clusters.
More formally,  for each directed cell, the output is a list of 5 pairs consisting of $(center,size)$, \ie{} a $(495,436,4,5,2)$ tensor per city.


For each road segment, we start by merging the speed clusters from the intersecting cells by collecting all speed cluster medians from all the intersecting cells and sorting them. 
We then take as free flow the 80\% percentile of these cluster medians based on the corresponding volumes; we default to 20 if there is no data; and we clip above to the signalized speed limit from OSM.
More formally,  for an edge with intersecting cells $intersecting\_cells$, speed limit $sl$ and for $cl(r,ch)$ denoting the 5 clusters (structures with attributes center and size) computed above, we derive its free flow speed as
\begin{equation}
center\left(q_{0.8,20}(\bigcup_{ \substack{(r,c,h) \in \\intersecting\_cells}} cl(r,c,h), size) \right)^{\top sl}
\end{equation}
where $q_{v,d}(X,a)$ is the $v$ quantile of the set $X$ based on attribute $a$ defaulting to $d$ if $X=\emptyset$ and $(X)^{\top b} = \{\min(x,b): x \in X\}$.

Optionally, we normalize free flow speed before computing the congestion factor. In normalization, we first use the signalized speed limit to make sure the free flow speed $\mathit{ff}$ is not below the signalized limit $\mathit{sl}$ from OSM and not above 60\% of that speed limit: 
\begin{equation}
\begin{cases}
\mathrm{max}(\mathrm{clip}(\mathit{ff})_{20}^{sl}),{0.6 \cdot \mathit{sl}}) &\text{if } \mathit{sl} \geq 5\\
\mathrm{max}(\mathrm{clip}(\mathit{ff})_{20}),{0.6 \cdot \mathit{sl}}) &\text{else}
\end{cases}
\end{equation}
where $\mathrm{clip}(x)^u_l = \min(\max(x,l),u)$.

\section{Analysis of Spot Binning under Idealized Conditions}\label{appendix:analysis_of_spot_binning}

Here, we give details on Spot Binning as introduced in Section~IV-A1 of the main text. 
Consider the following idealized conditions. For vehicles $i$ in a bin $B$ and $r_i$ the number of their readings within $B$,
\begin{description}
    \item[$(i)$] all vehicles travel the same distance $s=s_i$
    \item[$(ii)$] constant speed $v_{ij} = v_i$ for all readings $ij$ of vehicle $i$
    \item[$(iii)$] vehicles not fully covering the distance negligible, \ie{} for every vehicle not traveling the full distance $s_i$ within the temporal extension of $B$, there is another vehicle compensating with the same speed (not having left is compensated by another vehicle leaving). Technically,
    we have classes of vehicles $[i]$ merged together, so we can re-index vehicles $i$ (possibly partially covering the full distance within the temporal extension of $B$) to virtual vehicles $k$ going the full distance during the temporal extension of $B$,  $v_k=v_{[i]}=v_i$) and     $r_k = \sum_{ij: [i]=k} 1$.
    \item[$(iv)$] same probe rate $R$, \ie{} a vehicle with speed $v$ will have $r = c \cdot v^{-1}$ readings
\end{description}
Then,
\begin{eqnarray}
speed_B &=& \frac{\sum_{ij}v_{ij} }{\sum_{ij}1} \\
&\stackrel{(ii)}{=}& \frac{\sum_i r_i v_i}{\sum_i r_i} \\
&\stackrel{(iii)}{\approx}& \frac{\sum_k r_k v_k}{\sum_k r_k} \\
&\stackrel{(iv)}{=}& \frac{\sum_k \frac{c}{v_k} v_k}{\sum_k\frac{c}{v_k}}\\
&=& \frac{\sum_k 1}{\sum_k v_k^{-1}}\label{eq:harmonic}\\
&\stackrel{(i)}{=}& \frac{\sum_k 1}{\sum_k \frac{t_k}{s}}\\
&\stackrel{}{=}& \frac{s \cdot \sum_k 1}{\sum_k t_k}\label{eq:totaldistancetotaltraveltime}
\end{eqnarray}
Hence, under these assumptions, the spatio-temporally aggregated speed $speed_B$ represents the total distance divided by the total travel time as per Eq.~(\ref{eq:totaldistancetotaltraveltime}), which is the harmonic sum of the speeds of Eq.~(\ref{eq:harmonic}).
Assumption $(iii)$ depends on the temporal and spatial extension of bins and on the probe rate.
Assumption $(i)$ of the same distance is satisfied if the binning comes from the same underlying road segment.
The assumption $(ii)$ of constant speed can be relaxed as long as the probe rate represents the speed changes well and as long as the number of vehicles not fully counted in a bin is negligible (implying a trivial variant of $(iii)$).
Condition $(iv)$ is hard to control in many settings; if vehicles do not have the same probe rate $(iv)$ or even a correlation between speed and probe rate, then this can have a substantial impact in low-flow situations.
This analysis shows the importance to control homogeneity of probe rates, the probe frequency and the bin extensions in time and space, especially under low-flow conditions.


\section{Complements on Validation}\label{appendix:validation}

\subsection{Comparison with Uber Movement Speeds}\label{appendix:validation_uber}

\subsubsection{Historic Road Graph}
Uber matches their data to the OSM data of the time of collection \cite{uber-movement-speeds-calculation-methodology}, only storing the OSM node and way IDs of the segments without feature attributes like position. Therefore, we download the historic OSM data closest to the collection period and take OSM start node ID, end node ID, and OSM way ID from the Uber data and match it with the OSM data to construct the same structure as issuing from dp03 above. We then run our pipeline on this road graph. Notice that this road graph is potentially incomplete due to divergences between the OSM IDs in Uber and the OSM data snapshot; in addition, as this approach does not use the \texttt{OSMnx} road graph simplification, its segments are shorter in general.
We refer to the Supplementary Material \cite{neun2023metropolitansupplementary} for the key figures of the road graph properties and the \mcswts{} speeds.

\subsubsection{Spatial and Temporal Coverage}

\begin{figure}[tbp]
  \centering
  \includegraphics[width=0.48\textwidth]{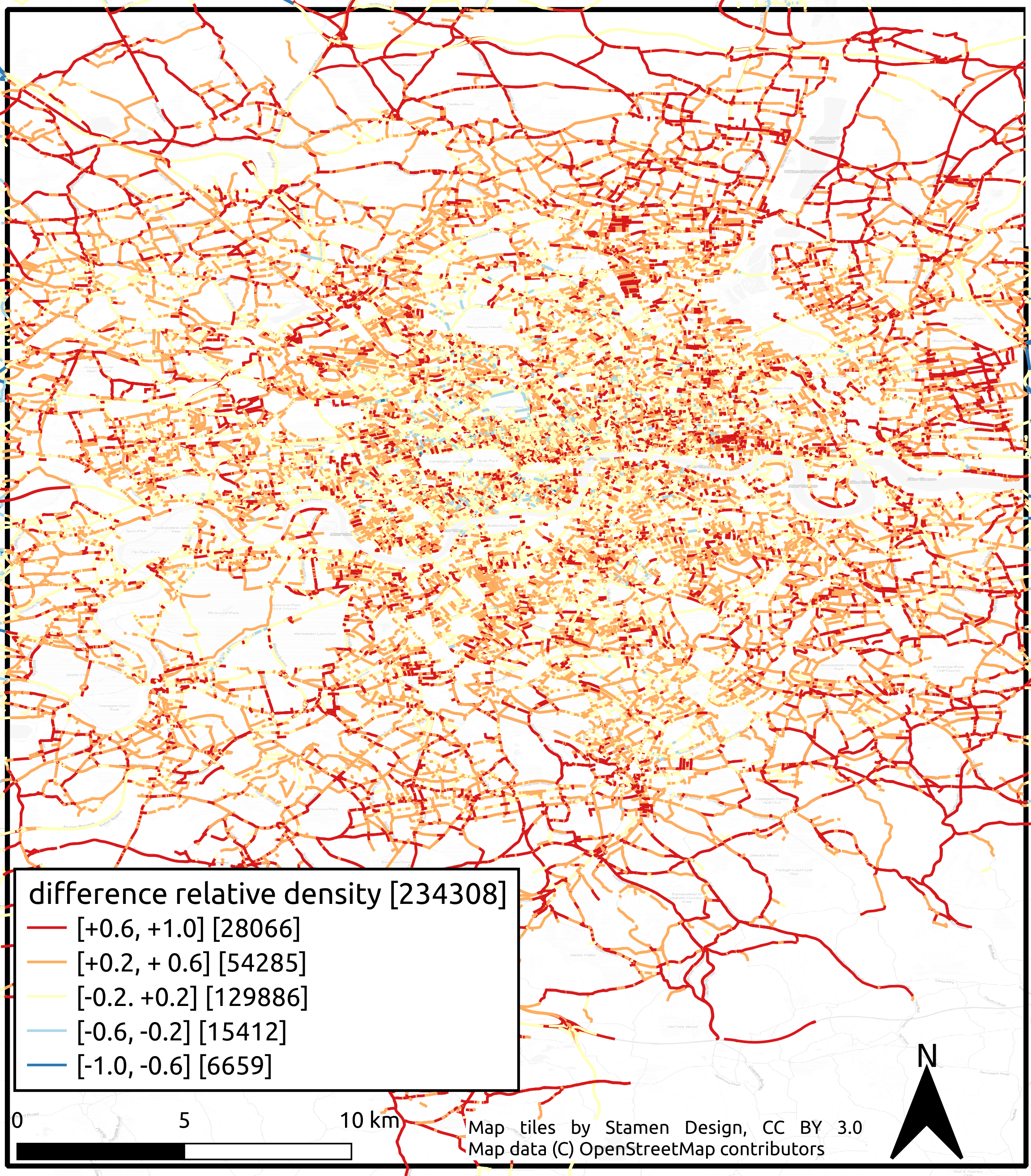}
  \caption{Segment density differences of Uber and \mcswts{} on the historic road graph London (8am--6pm). The color encoding shows the edge density difference, higher values mean higher temporal coverage of \mcswts{}. In particular, the negative values outside of the bounding box are due only Uber having data there. The numbers of edges are shown in brackets.}
  \label{fig:London_Uber_density_diff}
\end{figure}

Figure~\ref{fig:London_Uber_density_diff} shows the difference in spatial temporal coverage between the Uber and \mcswts{} datasets for London (color coded). Spatially, we see that Uber has data outside of the \mcswts{} bounding box, mainly including roads that lead to the airports outside of the city. Roughly speaking, we see similar temporal coverage in the city center (light yellow and light green colors), higher coverage of \mcswts{} in the outer city areas, and negative values outside of the \t4c bounding box.

\begin{figure}[thb!]
  \centering
  \includegraphics[width=0.45\textwidth]{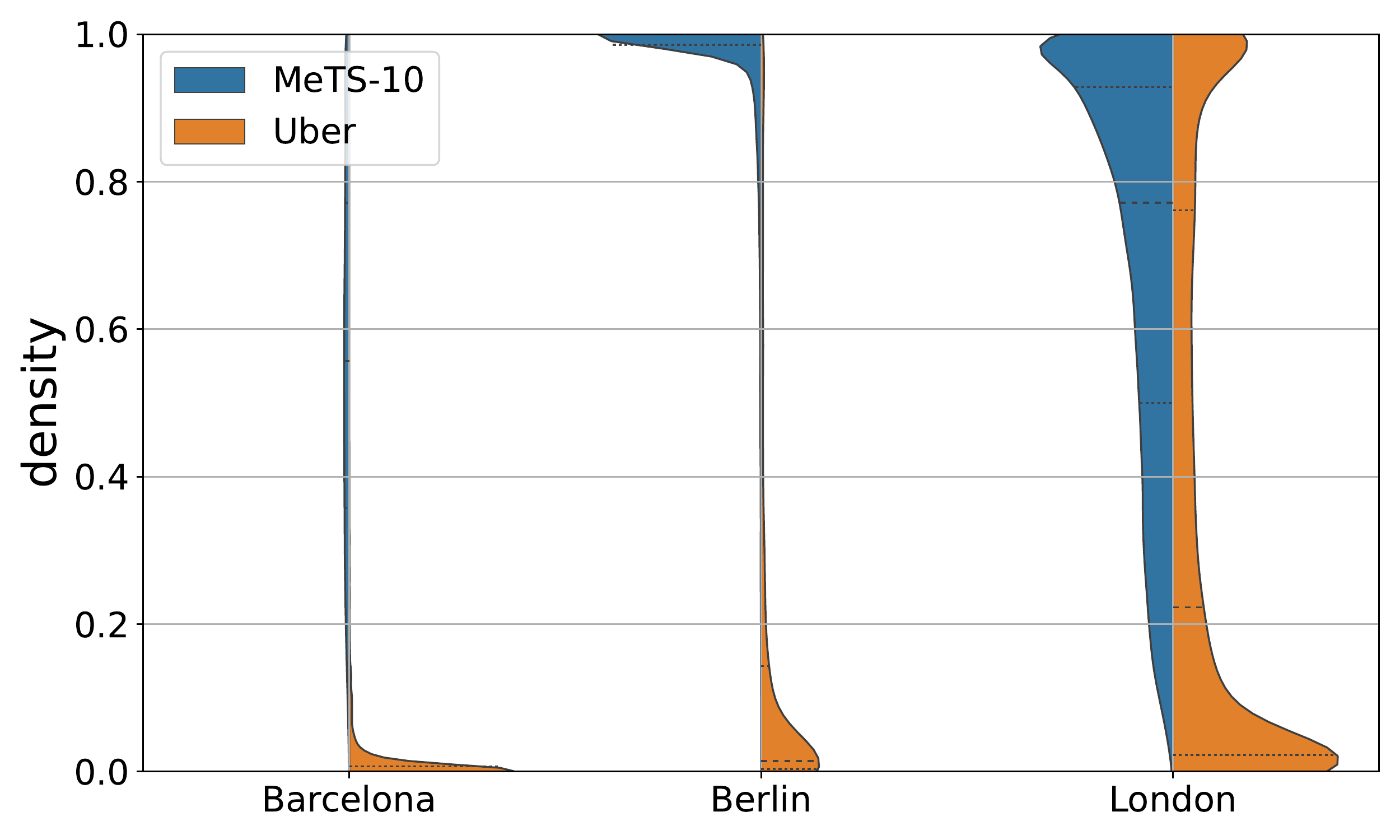}

  \caption{Distribution of segment-wise temporal coverage of Uber and \mcswts{} data
  for the 3 cities Barcelona, Berlin and London, bounding box only. 
  }
  \label{fig:uber03_spatial_coverage_city_comparison_bb_relative}
\end{figure}

Figure~\ref{fig:uber03_spatial_coverage_city_comparison_bb_relative} shows a comparison of the temporal coverage of all three cities. It shows that temporal coverage in Barcelona and Berlin is in general higher in \mcswts{} than in Uber. \mcswts{} has high coverage for many segments in Berlin and low coverage for Uber. For London, many segments have similar temporal coverage, but also a considerable amount of segments with differences -- slightly more segments with better coverage in \mcswts{}. We refer to the Supplementary Material \cite{neun2023metropolitansupplementary} for additional figures.

\subsubsection{Speed Differences}

\begin{figure}[ht!]
  \centering
  \includegraphics[width=0.48\textwidth]{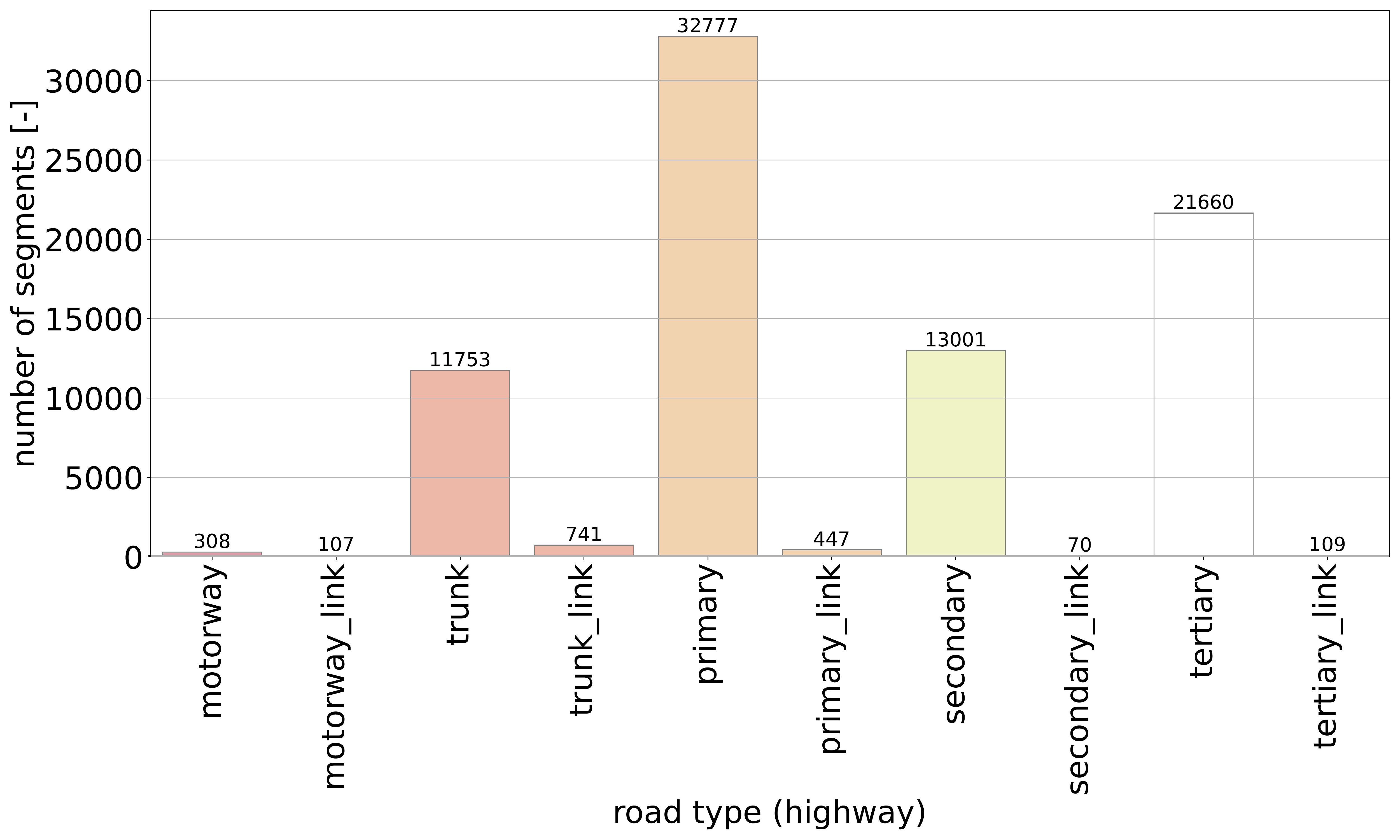}
  \caption{Segment counts \mcswts{} -- Uber matched data.}
  \label{fig:London_Uber_segment_counts}
\end{figure}

We match the 1h Uber and \mcswts{} segment speeds for one week of data in London, achieving a high number of samples that overlap spatio-temporally, namely 3.1 M data points. This corresponds to 77\% of the 4.67\,M data points for Uber within the bounding box, and to 47\% of 6.63\,M data points for \mcswts; the higher absolute number of data points and the lower matching rate for \mcswts{} is plausible in light of the previous paragraph. See \autoref{fig:London_Uber_segment_counts}.

\begin{figure}[ht!]
  \centering
  \includegraphics[width=0.48\textwidth]{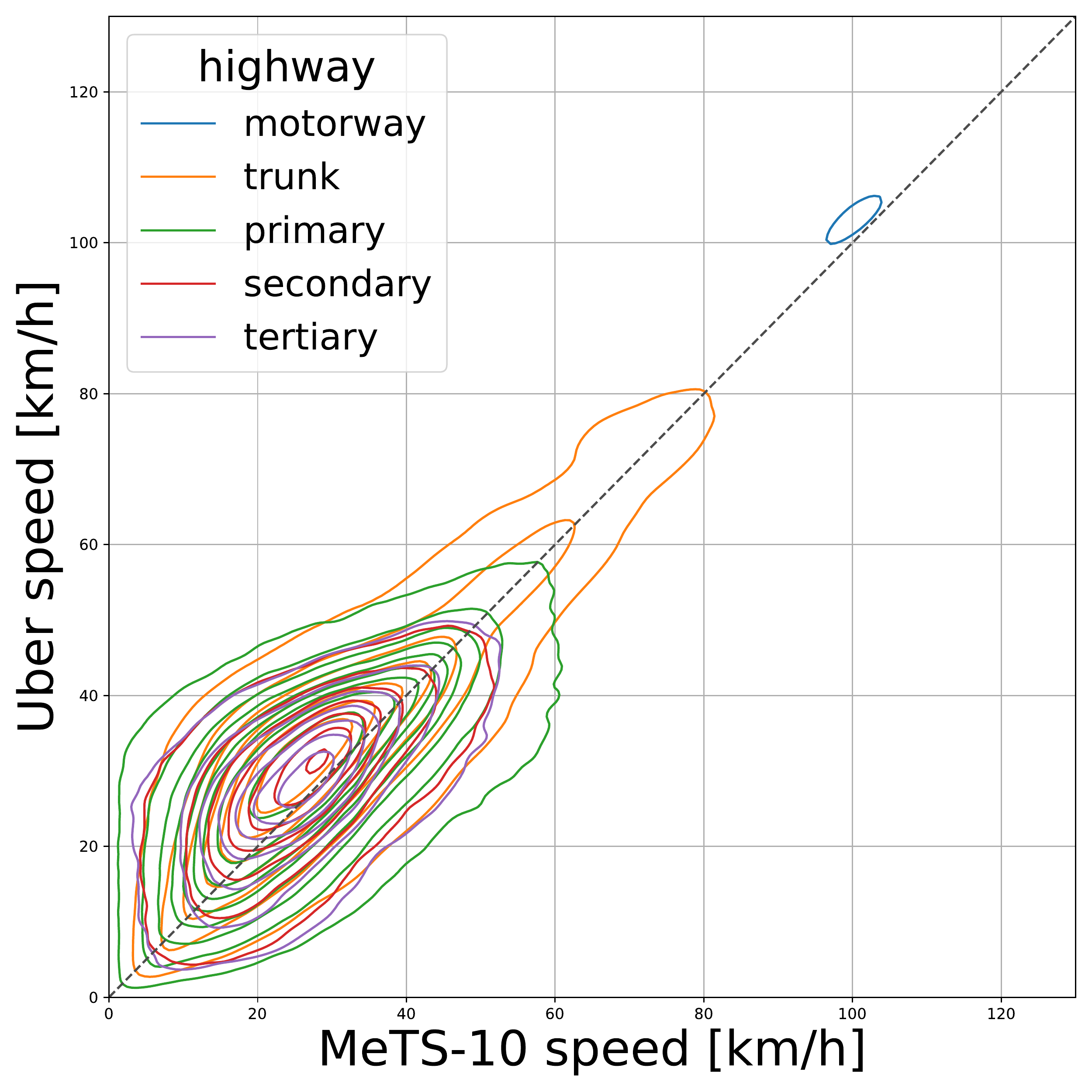}
  \caption{Kernel density estimation of speeds of \mcswts{} and Uber on the historic road graph London daytime (8am--6pm) on the matching data, \ie{} within \mcswts{} bounding box only and where data is available at the same time and segment, for the most important road types.}
  \label{fig:London_Uber_kde_highway}
\end{figure}

\paragraph{Speed Differences: Qualitative Analysis}
Figure~\ref{fig:London_Uber_kde_highway} shows the concentration of  Uber and \mcswts{} speeds along the diagonal. There are few cases with much higher \mcswts{} speeds (bumps lower right) -- inspection of examples shows that these tend to be due to high-speed GPS outliers above the signalized speed in low-flow situations which are not smoothed out by our median approach. On the other hand, there are many cases with \mcswts{} close to zero and higher Uber speeds (bumps on the left) -- here, an inspection of examples shows these tend to happen in junction situations, where spot binning of GPS probes from standing vehicles leads to much lower aggregated speed values compared to the per-vehicle mean speeds on the segment.

\begin{figure}[ht!]
  \centering
  \includegraphics[width=0.48\textwidth]{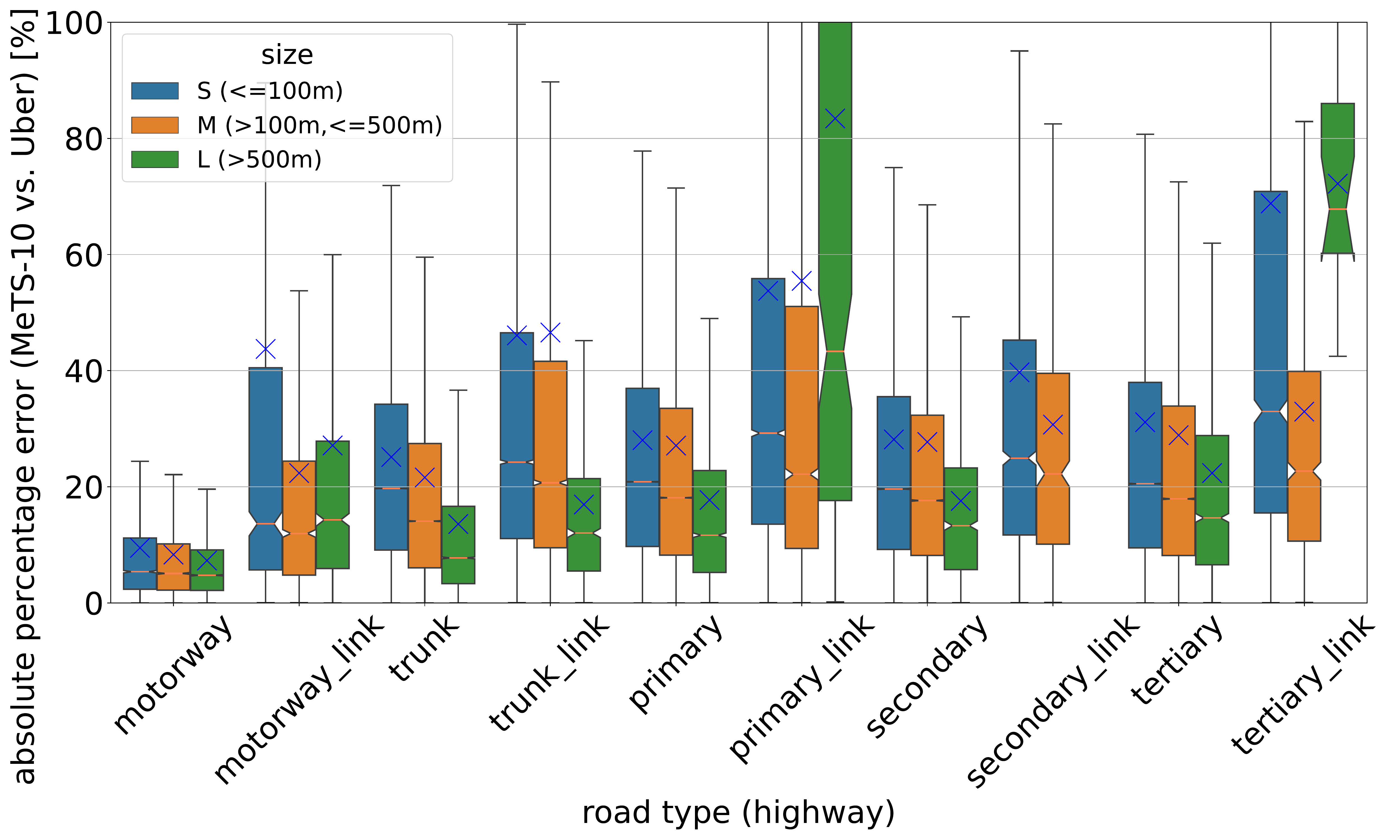}
  \caption{London absolute percentage error \mcswts{} vs. Uber by road type and segment length. Blue crosses indicate the mean per road type.}
  \label{fig:London_Uber_ape_size}
\end{figure}

\paragraph{Speed Differences: Quantitative Analysis}
\begin{figure}[ht]
  \centering
  \includegraphics[width=0.48\textwidth]{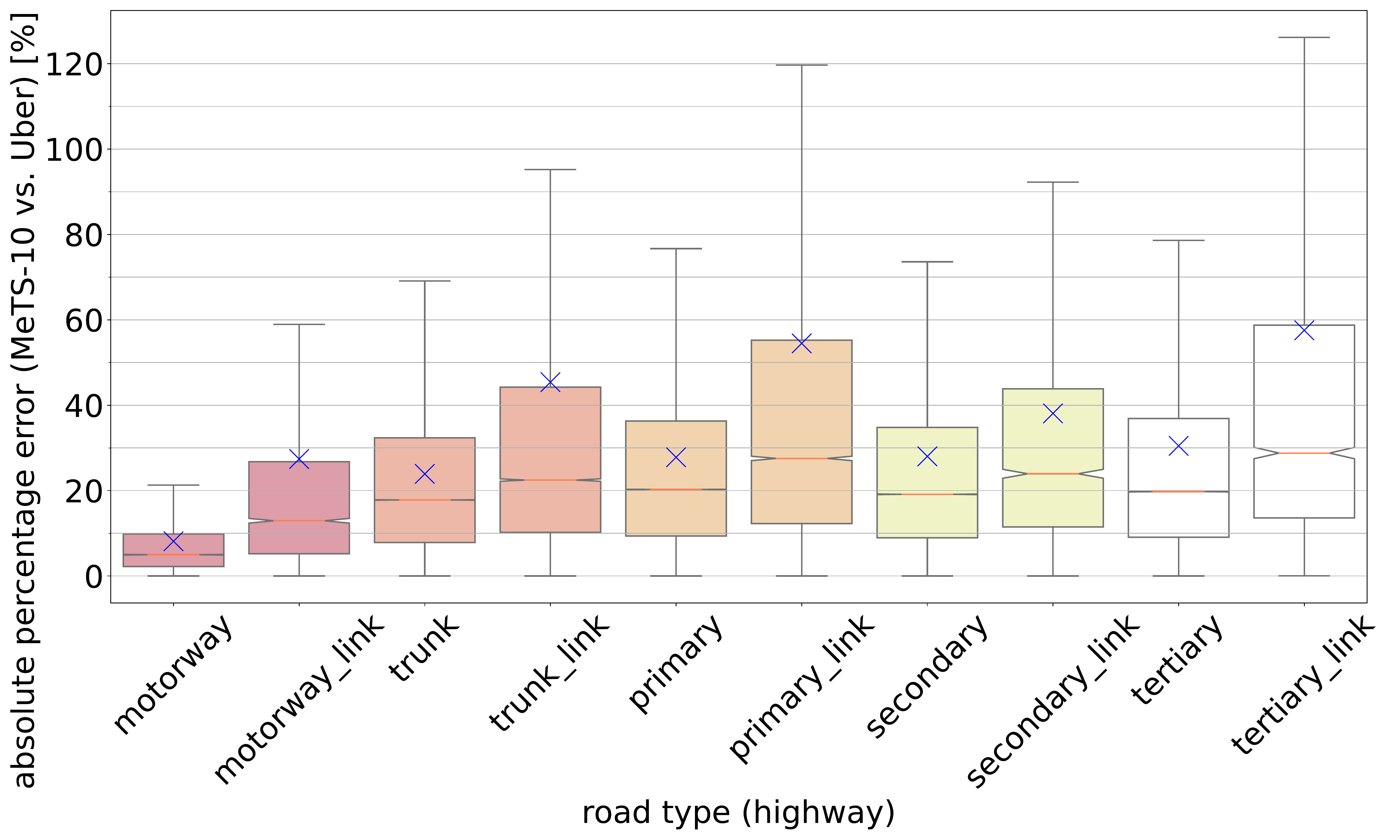}
  \caption{London absolute percentage error \mcswts{} vs. Uber by road type. Blue crosses indicate the mean per road type.}
  \label{fig:London_Uber_ape}
\end{figure}
The qualitative interpretation is confirmed by the quantitative analysis of \autoref{fig:London_Uber_ape} and \autoref{fig:London_Uber_ape_size}: longer and non-link segments tend to have lower absolute percentage error comparing \mcswts{} to Uber.
We use absolute percentage error as used in the traffic simulation calibration literature \cite{fhwa}:
in this context, a rule of thumbs asks for 85\% of the counts measured in the simulation should deviate less than 15\%.

\paragraph{Speed Differences: Effect of Segment Length}

In order to also quantitatively assert the effect of segment length on APE, we further distinguish three ad-hoc sizes in \autoref{fig:London_Uber_ape_size}: S for segments up to 100~m, M for segments between 100~m and 500~m, and L for segments longer than 500~m. Apart from the link road types (with higher variance as discussed in Section~V and low number of segments as per \autoref{fig:London_Uber_segment_counts}), we see that longer segments tend to be better aligned with Uber than shorter ones.

\paragraph{Speed Differences: Effect of Road Segment Complexity}

\begin{figure}[ht!]
  \centering
  \includegraphics[width=0.48\textwidth]{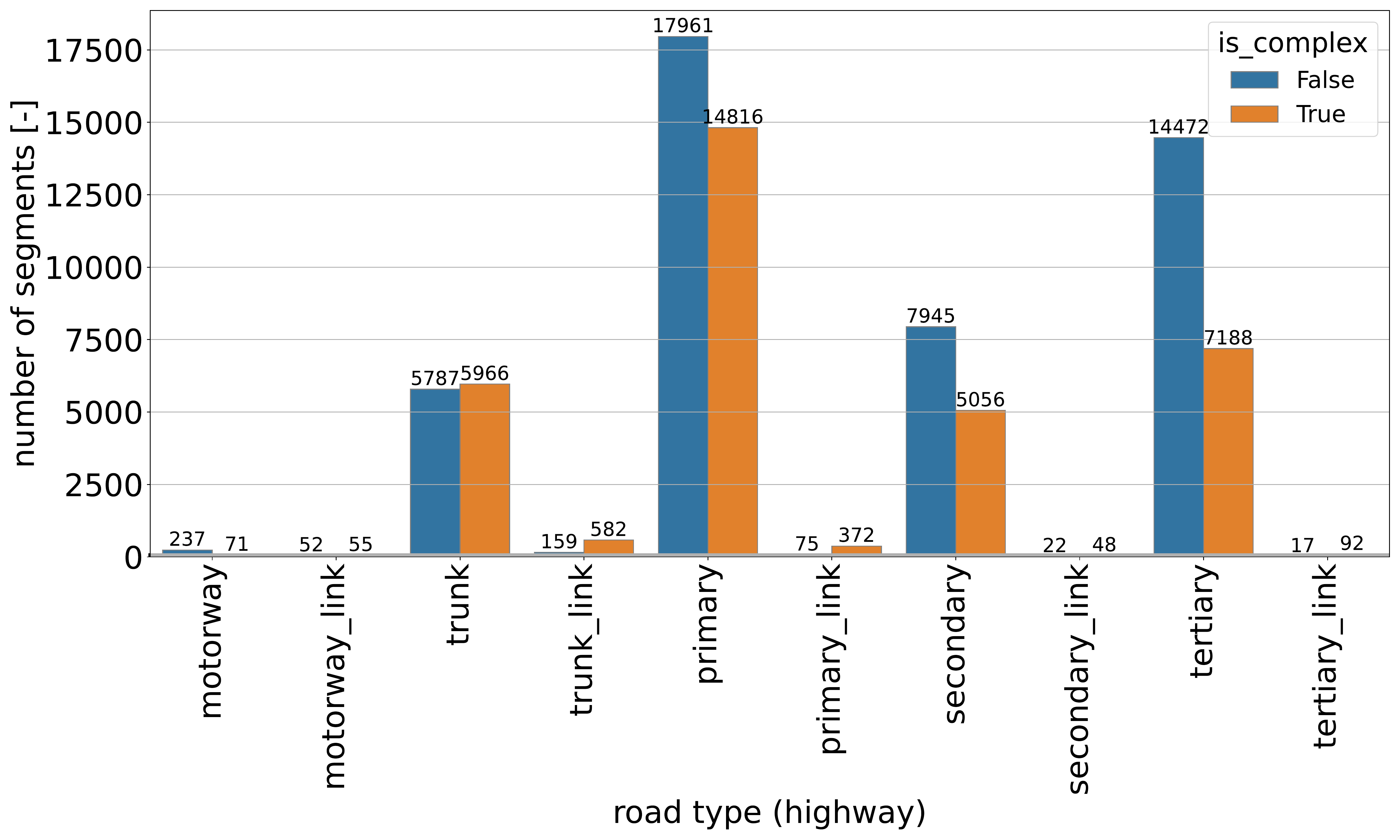}
  \caption{Segment counts \mcswts{} -- Uber matched data.}
  \label{fig:London_Uber_segment_counts_complex_non_complex}
\end{figure}

\begin{figure}[ht!]
  \centering
  \begin{subfigure}[b]{0.45\textwidth}
  \includegraphics[width=1.0\textwidth]{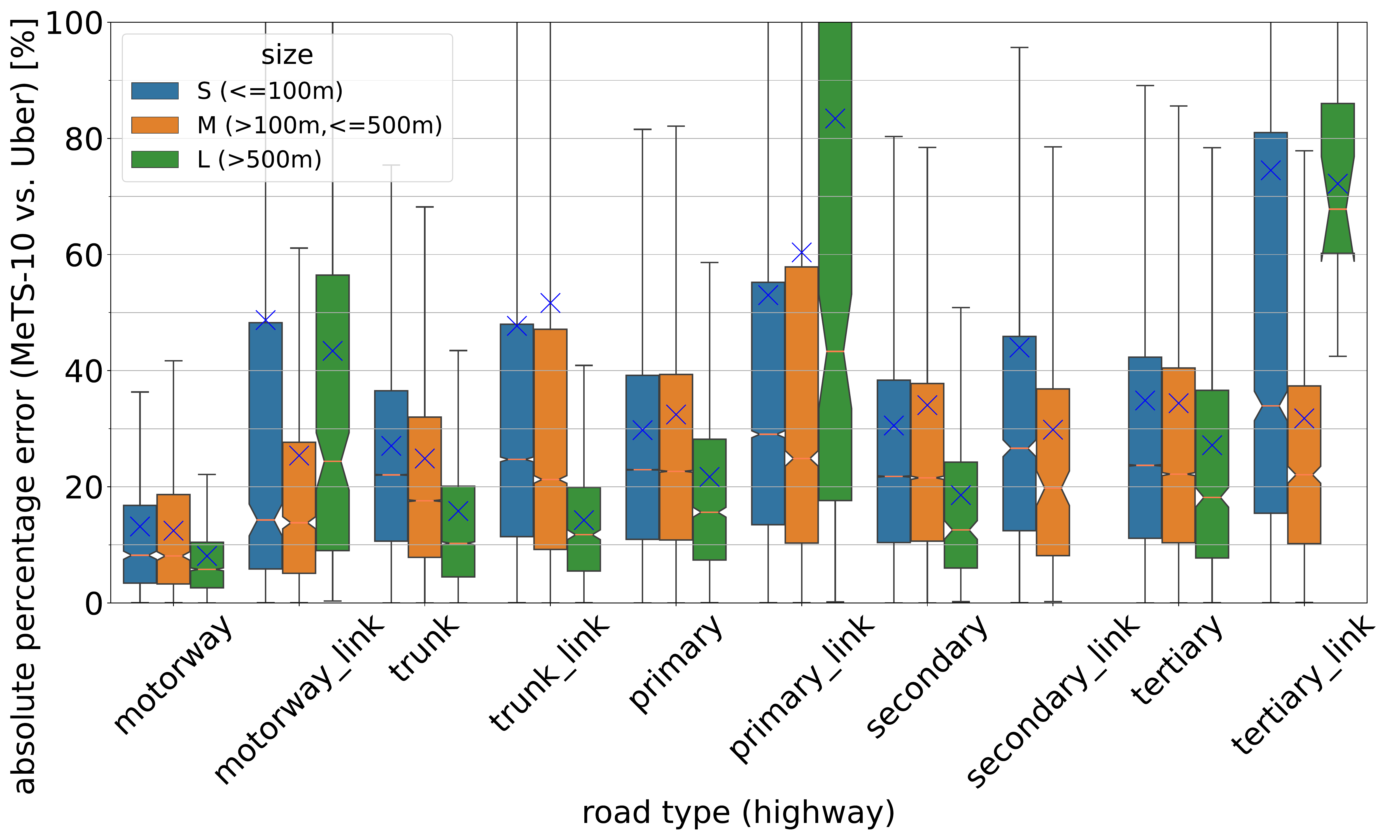}
  \caption{complex road segments}
  \label{fig:London_Uber_ape_complex}
  \end{subfigure}
  \begin{subfigure}[b]{0.45\textwidth}
  \includegraphics[width=1.0\textwidth]{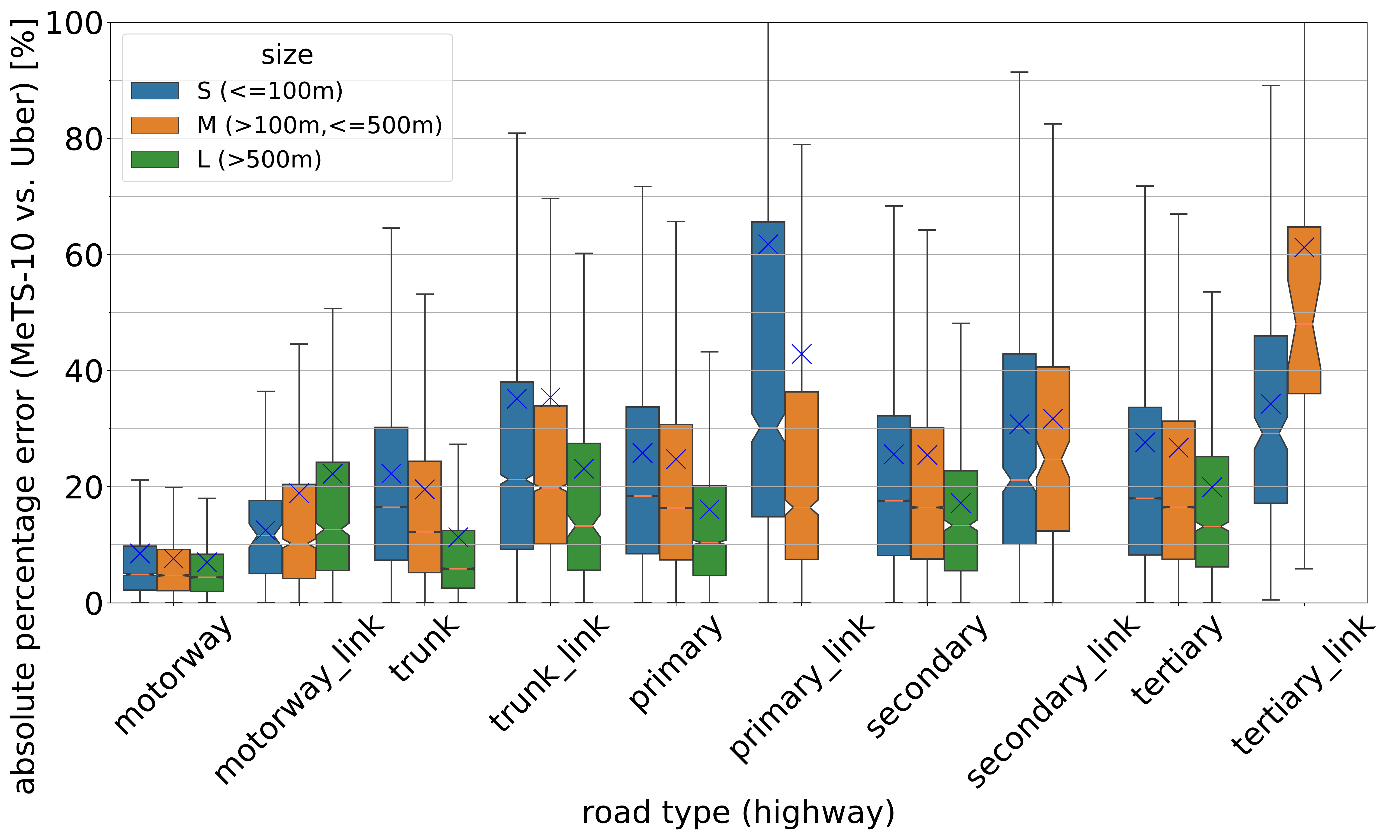}
  \label{fig:London_Uber_ape_non_complex}
  \caption{non-complex road segments}
  \end{subfigure}
  \caption{London absolute percentage error \mcswts{} vs. Uber by road type and segment length. Blue crosses indicate the mean per road type.}
  \label{fig:London_Uber_ape_complex_non_complex}
\end{figure}

In order to quantiatively assert the sensitivity to complex road situations,
we differentiate between complex and non-complex road segments: we consider those road segments as complex which have at least one intersecting cell shared by at least 4 further road segments. The case of sharing with 1 further segment is trivial as this happens for every node not exactly at a cell border. 
See \autoref{fig:London_Uber_segment_counts_complex_non_complex} and \autoref{fig:London_Uber_ape_complex_non_complex}.

\subsection{Comparison with Stationary Vehicle Detector Data}\label{appendix:validation_counters}

The sensor locations were matched to the \mcswts{} road graph in order to find the corresponding road segment. The matching logic uses the distance as well as, if provided, the heading angle and/or the name of the segment for determining the best match.

\begin{table}[!t]
\caption{Percentage and absolute count of matched stationary vehicle detector with \mcswts{} speed measurements by OpenStreetMap highway type \cite{osm-highway}). \label{tab:city-sensors}}
\centering
\begin{tabular}{
p{1.4cm}|>
{\raggedleft\arraybackslash}p{0.7cm}|>
{\raggedleft\arraybackslash}p{0.7cm}|>{\raggedleft\arraybackslash}p{0.7cm}|>
{\raggedleft\arraybackslash}p{0.7cm}|>
{\raggedleft\arraybackslash}p{0.7cm}|>{\raggedleft\arraybackslash}p{0.7cm}}
\toprule
 & \multicolumn{2}{c}{Berlin} & \multicolumn{2}{|c|}{London} & \multicolumn{2}{c}{Madrid}\\
 \midrule
motorway & 1 \% & 4 & 95 \% & 237 & 85 \% & 345 \\
trunk & 1 \% & 6 & 5 \% & 12 & 5 \% & 21\\
primary & 41 \% & 218 & & & 4 \% & 16\\
secondary & 52 \% & 280 & & & &\\
tertiary & 4 \% & 22 & & & 1 \% & 4\\
other & 1 \% & 8 & & & 4 \% & 18\\
\bottomrule
\end{tabular}
\end{table}

\begin{figure}[th!]
  \centering
  \includegraphics[clip, trim=0.8cm 1.9cm 0.1cm 0.6cm, width=0.48\textwidth]{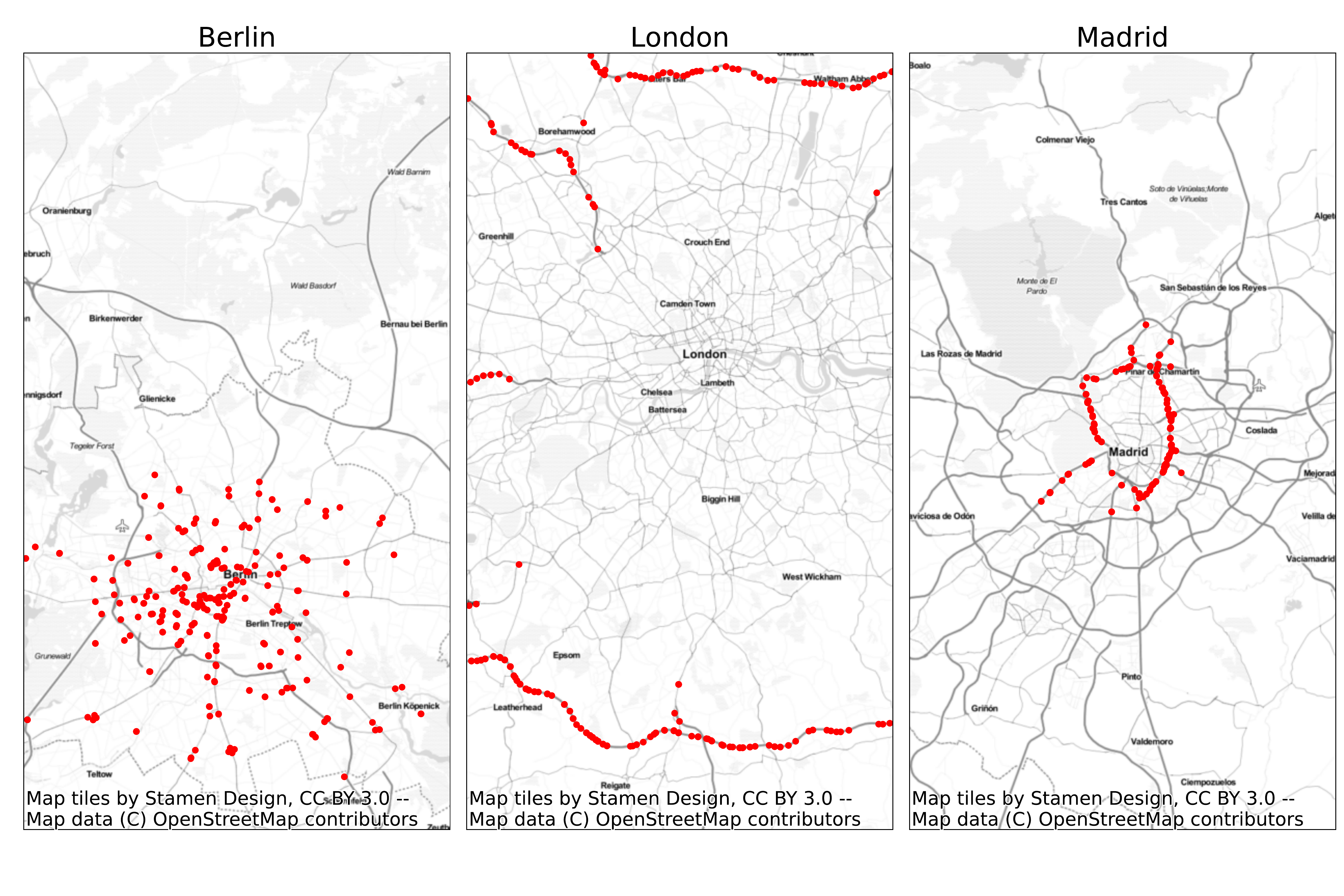}
  \caption{Distribution of Vehicle Detectors with speed measurement in the corresponding \mcswts{} bounding boxes.}
  \label{fig:05val02-counter-distribution}
\end{figure}

The spatial distribution of the speed sensors can be seen in Figure \ref{fig:05val02-counter-distribution}. These different distributions are also clearly visible in the differences in covered highway types in Table \ref{tab:city-sensors}. In Berlin, sensors cover the whole city with the majority of them located on primary and secondary roads. In contrast, in London and Madrid, speed sensors are mostly located on motorways and trunks. In London, these motorways are only located on the outskirts of the city where roads have fewer tunnels and overall higher speeds. In Madrid, the available speed counter data is mostly for motorways of the inner-city ring road, where we see more significant speed limits and other effects of traffic control. We consider sensors on motorways that are not disturbed by traffic lights or other effects to be close to real ground truth values. This ignores the quality of the sensor, malfunctioning sensors, and potential temporal aggregation effects.

Erroneous sensor readings were filtered before comparing the speed measurements. There were two main sources of erroneous readings: a) values at night due to maintenance or impaired averaging when no speed was detected in the measurement interval. Hence, we only looked at readings between 6 am and 11 pm. b) readings in situations where GPS signals are disturbed, such as tunnels. Hence, sensors in tunnels were filtered out as well. For more details on the pre-processing, see our code base referenced in \textit{Data and Code Availability} below.

\subsection{Baseline comparison of Uber Movement Speeds with Stationary Vehicle Detector Data}\label{appendix:validation_counters_Uber}

As a baseline of the used validation datasets we compare here the Uber Movement Speeds with the speed readings from the Stationary Vehicle Detectors.
This was done for Berlin and London were we had overlapping time ranges for the comparison.

The same matching procedure was used as for the comparison of the \mcswts{} speeds with the stationary vehicle detectors.
Table \ref{tab:city-sensors-uber} shows the statistics of the used vehicle detectors with matching Uber speeds.
Uber coverage is lower, hence compared to \mcswts{} (see also Table \ref{tab:city-sensors}) less counters did have a corresponding Uber speed on the associated street segment.

\begin{table}[!t]
\caption{Percentage and absolute count (matched/total) of stationary vehicle detector with Uber Movement Speeds by OpenStreetMap highway type \cite{osm-highway}).
\label{tab:city-sensors-uber}}
\centering
\begin{tabular}{
p{1.4cm}|>
{\raggedleft\arraybackslash}p{1.0cm}|>
{\raggedleft\arraybackslash}p{1.0cm}|>{\raggedleft\arraybackslash}p{1.0cm}|>
{\raggedleft\arraybackslash}p{1.0cm}}
\toprule
 & \multicolumn{2}{c}{Berlin} & \multicolumn{2}{|c}{London}\\
 \midrule
motorway & 1 \% & 4 & 95 \% & 210/237\\
trunk & & 0/6 & 5 \% & 11/12\\
primary & 41 \% & 192/218 & &\\
secondary & 53 \% & 249/280 & &\\
tertiary & 4 \% & 19/22 & &\\
other & 1 \% & 3 & &\\
\bottomrule
\end{tabular}
\end{table}

\begin{figure}[th!]
  \centering
  \includegraphics[clip, trim=0.4cm 0.4cm 0.1cm 0.4cm, width=0.42\textwidth]{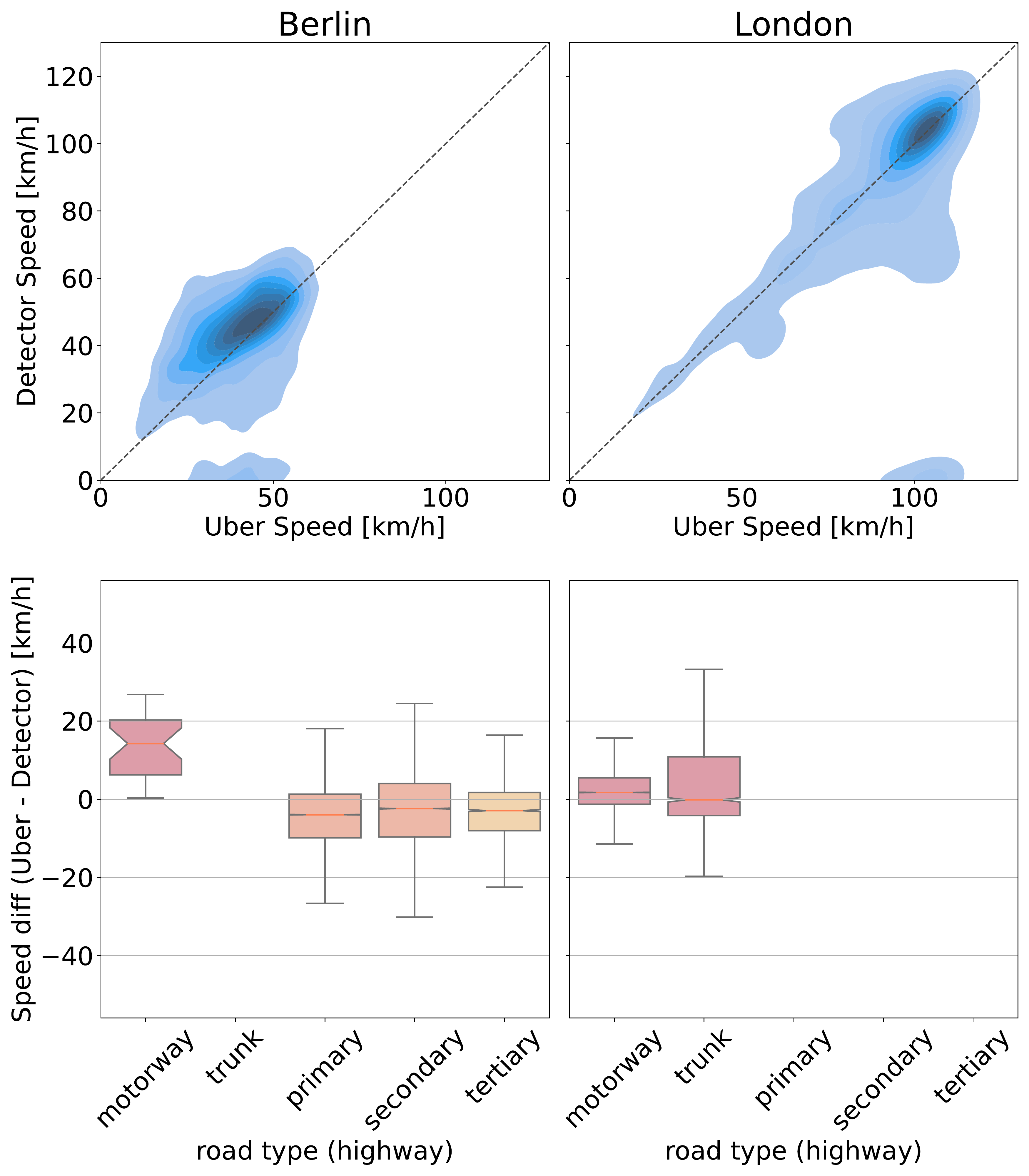}
  \caption{Comparisons of Uber Movement Speeds and vehicle detector speeds (30 days of data matched to the Uber OSM road graph). Top: Binned Kernel Distribution Estimation Plots of speeds; Bottom: Box plots of speed differences by highway type (OSM carto color scheme)}
  \label{fig:05val02-counter_uber_diff}
\end{figure}

Figure \ref{fig:05val02-counter_uber_diff} shows the Binned Kernel Distribution Estimation (KDE) Plots of Uber and the detectors as well as box plots of the speed differences between Uber Movement Speeds and vehicle detector speeds.

\begin{figure}[th!]
  \centering
  \begin{subfigure}[b]{0.48\textwidth}
  \includegraphics[clip, width=0.98\textwidth]{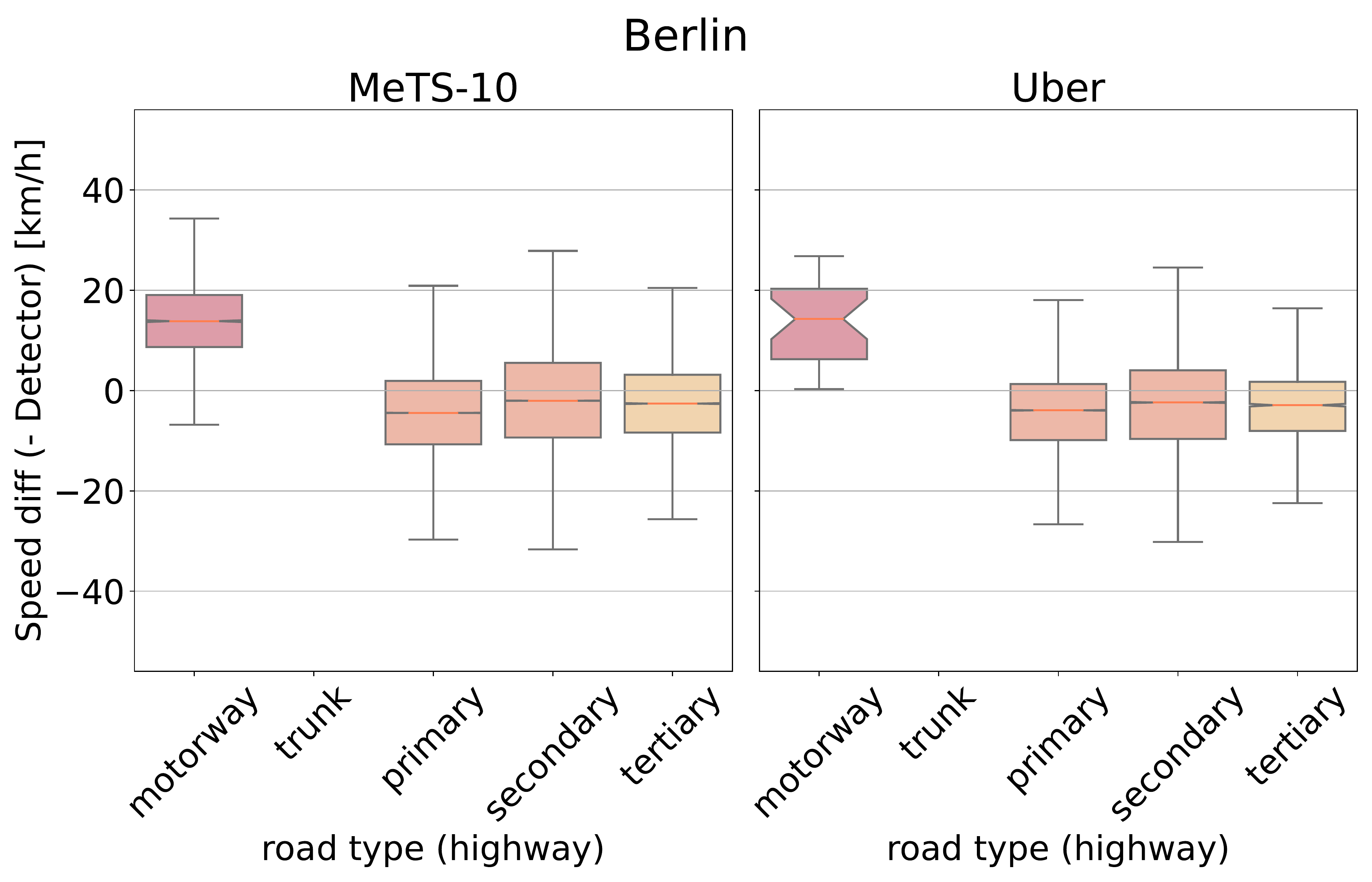
}
  \end{subfigure}
  \\
  \begin{subfigure}[b]{0.48\textwidth}
  \includegraphics[clip, width=0.98\textwidth]{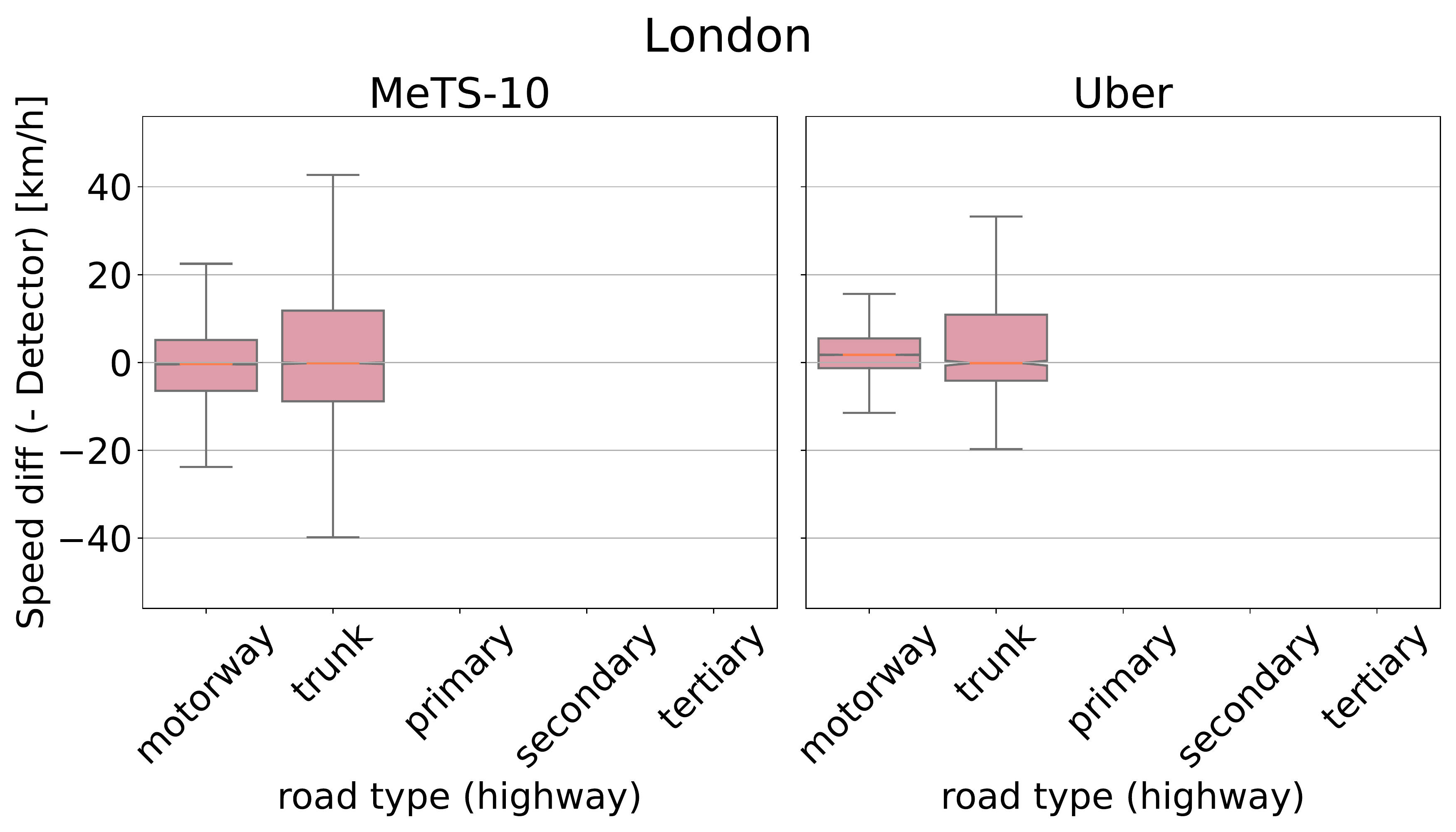
}
  \end{subfigure}
  \caption{SxS of \mcswts{}--Detector and Uber--Detector Differences by highway type (OSM carto color scheme).}
  \label{fig:05val02-counter_diffs_sxs_onemonth}
\end{figure}

In both cities we see a good alignment with the majority of the points along the diagonals in the KDE plots. The shapes of the plots are very similar to the KDE plots in the comparison between \mcswts{} and the detectors. Hence, also here we see the differences in sensor placement on predominantly inner city roads (Berlin) vs mostly motorways (London) as shown in Table \ref{tab:city-sensors-uber}.
In the box plots by highway type (Figure \ref{fig:05val02-counter_uber_diff} bottom; Figure \ref{fig:05val02-counter_diffs_sxs_onemonth}) we can see a very good alignment for this higher class highway types in London as well as for the lower class highway types in Berlin.

In the side-by-side (SxS) comparisons in \autoref{fig:05val02-counter_diffs_sxs_onemonth}, both \mcswts{} and Uber show the same speed level difference for motorways in Berlin. This only affects 4 motorway counters and hence seems to be a systematic difference for the type of counter or location. Note, the plots here only show counters on segments where both Uber and \mcswts{} had data points. Therefore, for example the trunk type segments are not shown as they only had data in the \mcswts{} dataset.

\begin{figure}[th!]
  \centering
  \begin{subfigure}[b]{0.48\textwidth}
  \includegraphics[clip, trim=0.1cm 0.0cm 0.1cm 0.2cm, width=0.98\textwidth]{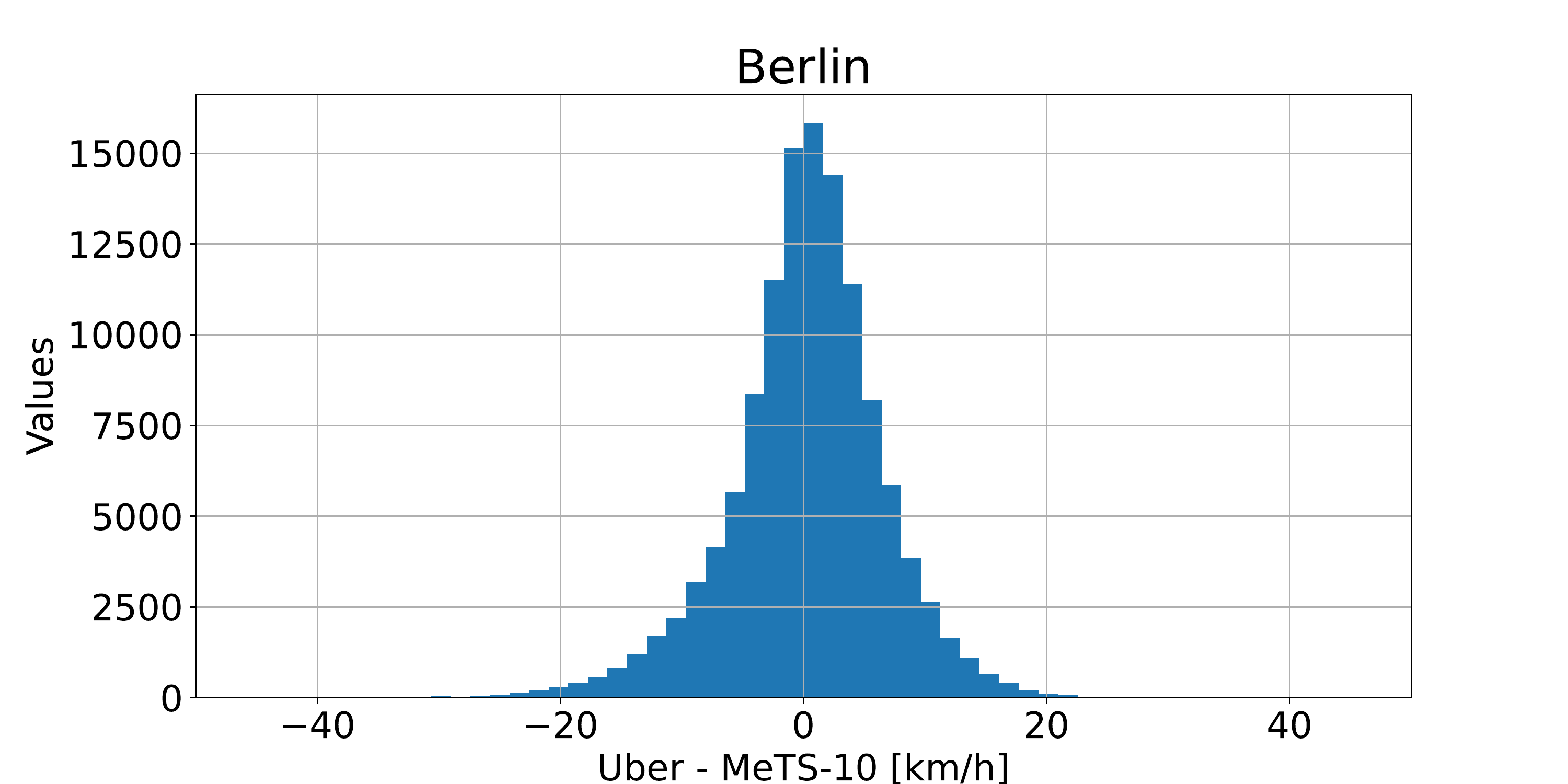
}
  \end{subfigure}
  \\
  \begin{subfigure}[b]{0.48\textwidth}
  \includegraphics[clip, trim=0.1cm 0.0cm 0.1cm 0.2cm, width=0.98\textwidth]{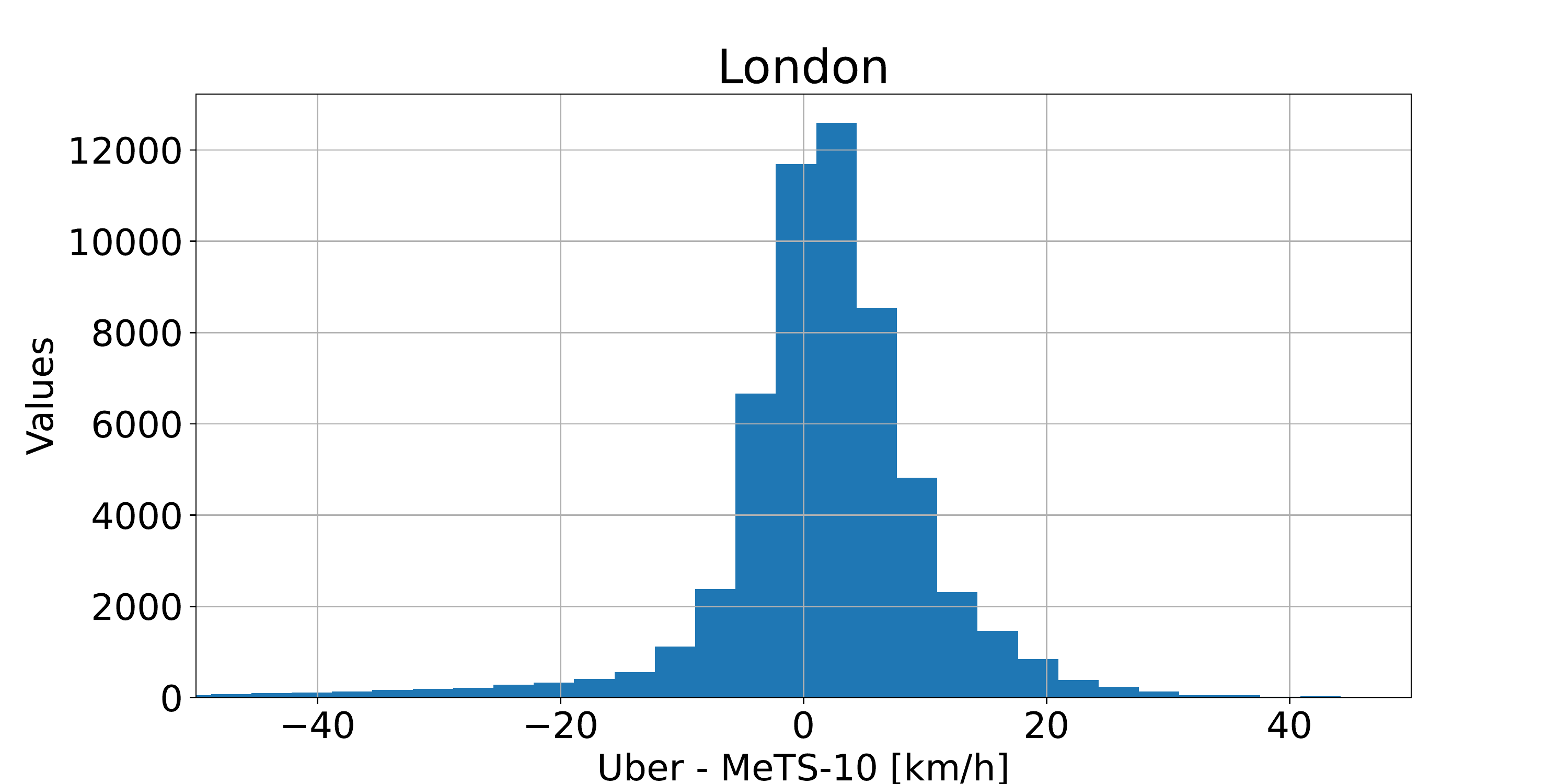
}
  \end{subfigure}
  \caption{Histogram of Uber/\mcswts{} speed differences at stationary vehicle detector locations (Uber - \mcswts{})}
  \label{fig:05val02-counter_diff_stats_onemonth}
\end{figure}

The good general visual alignment in the box plots is also confirmed by the quantitative comparisons in \autoref{tab:counterdiffdiff_berlin} and \ref{tab:counterdiffdiff_london} as well as in the histograms in \autoref{fig:05val02-counter_diff_stats_onemonth}.
These comparisons show the speed differences between Uber, \mcswts{} and detector readings for all detectors with both Uber and \mcswts{} data available during 30 consecutive days.

In Berlin (see \autoref{tab:counterdiffdiff_berlin}), the majority of the detectors is on roads of type primary and secondary. On these two road types, we see also a very good alignment of the Uber and \mcswts{} speeds with a speed diff of 0.5$\pm$6.4~km/h and -0.1$\pm$6.2~km/h (average$\pm$standard deviation), respectively.

In London, almost all counters are along motorways. Here, we see a small but stable difference of 3.1$\pm$6.9~km/h.

\begin{table*}[!t]
\caption{Berlin  differences of Uber/\mcswts{} speeds  at stationary vehicle detector locations \label{tab:counterdiffdiff_berlin}}
\centering
\begin{tabular}{p{1.7cm}rrrrrrrrrr}
\toprule
    & \multicolumn{3}{c}{Uber -- Detector} & \multicolumn{3}{c}{\mcswts{} -- Detector} & \multicolumn{3}{c}{Uber -- \mcswts{}} & \\
    highway & diff mean & diff std & diff median & diff mean & diff std & diff median & diff mean & diff std &  diff median & \# values \\
\midrule
   motorway &               13.483067 &               7.795273 &                   14.2730 &               12.686667 &               7.246002 &                 12.182353 &             0.796400 &            5.414309 &              -0.223706 &           30 \\
    primary &               -3.084538 &              13.618173 &                   -3.9420 &               -3.610773 &              12.903032 &                 -5.332353 &             0.526235 &            6.378054 &               0.787451 &        69214 \\
residential &               -4.955351 &               7.833748 &                   -2.3000 &               -7.877973 &               7.149825 &                 -5.766667 &             2.922622 &            4.547528 &               2.402353 &           57 \\
  secondary &               -2.004949 &              15.230711 &                   -2.3785 &               -1.881549 &              14.190849 &                 -2.976471 &            -0.123400 &            6.207306 &               0.194863 &        49526 \\
   tertiary &               -2.247903 &              10.411396 &                   -2.9145 &               -2.196423 &              10.097512 &                 -2.988235 &            -0.051480 &            5.857588 &              -0.034078 &         3466 \\
      TOTAL &               -2.620424 &              14.225456 &                   -3.3330 &               -2.868381 &              13.397577 &                 -4.319608 &             0.247957 &            6.302485 &               0.503647 &       122293 \\
\bottomrule
\end{tabular}
\end{table*}


\begin{table*}[!t]
\caption{London differences of Uber/\mcswts{} speeds at stationary vehicle detector locations \label{tab:counterdiffdiff_london}}
\centering
\begin{tabular}{p{1.7cm}rrrrrrrrrr}
\toprule
    & \multicolumn{3}{c}{Uber -- Detector} & \multicolumn{3}{c}{\mcswts{} -- Detector} & \multicolumn{3}{c}{Uber -- \mcswts{}} & \\
    highway & diff mean & diff std & diff median & diff mean & diff std & diff median & diff mean & diff std &  diff median & \# values \\
\midrule
     motorway &                5.253279 &              17.986759 &                  1.739697 &                2.196420 &              18.520102 &                 -0.463783 &             3.056858 &            6.926557 &               2.273585 &        50777 \\
motorway\_link &              -10.143697 &              10.655478 &                 -9.506371 &                8.156388 &              13.872980 &                  9.457262 &           -18.300085 &           15.978163 &             -16.496598 &         3275 \\
        trunk &                8.395831 &              22.285910 &                 -0.150473 &                4.489668 &              24.107821 &                 -0.287161 &             3.906163 &           11.074098 &               2.583295 &         1834 \\
   trunk\_link &              -12.372454 &              20.987359 &                -14.653041 &               15.267473 &              22.729794 &                 15.082768 &           -27.639927 &           12.654695 &             -27.497865 &          339 \\
        TOTAL &                4.352669 &              18.243336 &                  1.384032 &                2.697190 &              18.602136 &                 -0.276012 &             1.655479 &            9.667750 &               1.873293 &        56225 \\
\bottomrule
\end{tabular}
\end{table*}


\section{Complements on Discussion}\label{appendix:discussion}
\subsection{Complements Spatial Intersection (dp04)}

In \autoref{tab:headings}, we contrast edges whose geometry is along the N/E/S/W axis (horizontal/diagonal $\pm 10^{\circ}$) or along the NE/SE/SW/NE (diagonal $\pm 10^{\circ}$).
We sample 10 days of speed data. 
We see that the coverage along the diagonal is slightly lower. This is plausible in light of the horizontal/diagonal edges getting values from two headings.
\begin{table*}[!t]
\caption{Coverage for segments along the diagonals vs. segments along the horizontal/vertical axes. \label{tab:headings}}
\centering
\begin{tabular}{lrrrrrrr}
\toprule
 city (year) & \#edges & \#N/E/S/W & \#NE/SE/SW/NW & coverage & coverage N/E/S/W & coverage NE/SE/SW/NW & difference \\
\midrule
Antwerp (2021) & 81667 & 17349 (21.2\%) & 21443 (26.3\%) & 0.13 & 0.16 & 0.11 & +0.03\\
Bangkok (2021) & 694818 & 201316 (29.0\%) & 91609 (13.2\%) & 0.03 & 0.04 & 0.03 & +0.00\\
Barcelona (2021) & 118813 & 22003 (18.5\%) & 34724 (29.2\%) & 0.06 & 0.07 & 0.06 & +0.01\\
Berlin (2021) & 88882 & 20875 (23.5\%) & 19940 (22.4\%) & 0.29 & 0.37 & 0.29 & +0.08\\
Chicago (2021) & 187570 & 118970 (63.4\%) & 9399 (5.0\%) & 0.07 & 0.07 & 0.10 & -0.00\\
Istanbul (2021) & 270109 & 61255 (22.7\%) & 61564 (22.8\%) & 0.52 & 0.61 & 0.48 & +0.09\\
Melbourne (2021) & 230654 & 103833 (45.0\%) & 31915 (13.8\%) & 0.05 & 0.05 & 0.04 & +0.00\\
Moscow (2021) & 47877 & 10177 (21.3\%) & 13259 (27.7\%) & 0.69 & 0.71 & 0.67 & +0.03\\
London (2022) & 271075 & 59654 (22.0\%) & 65380 (24.1\%) & 0.15 & 0.19 & 0.15 & +0.04\\
Madrid (2022) & 143402 & 32018 (22.3\%) & 35551 (24.8\%) & 0.33 & 0.39 & 0.32 & +0.06\\
Melbourne (2022) & 230654 & 103833 (45.0\%) & 31915 (13.8\%) & 0.05 & 0.06 & 0.04 & +0.01\\
\bottomrule
\end{tabular}
\end{table*}

\section{Extended Dataset Overview}\label{appendix:extended_dataset_overview}
Referring to \autoref{tab:dataset_extended}, the dataset comprises 10 cities with data from 108 up to 361 days publicly available:
        \textbf{t4c year} is the competition year which the was published for by HERE (relevant for download);
        \textbf{days} is the number of full days of data (288 5 minute bins of full bounding box each);
        \textbf{date ranges} are the corresponding date ranges -- in the 2022 competition, every second week in the date range was held back for tests in the competition, so there is not a consecutive range of dates available;
        \t4c bounding box is given by \textbf{lat\_min},\textbf{lat\_max},\textbf{lon\_min},\textbf{lon\_max};
        number of \textbf{nodes} and \textbf{edges} in the road graph;
        \textbf{total segment length} is the sum of all lengths of all directed edges;
        \textbf{ratio covered edges} is the ratio of edges in the road graph with at least one speed value (from 20 sampled days);
        \textbf{mapped ratio} is the ratio of GPS probes mapped to the road graph;
        \textbf{daily fcd volume} is the daily GPS probe volume sum (from 20 sampled days); 
        \textbf{8--18 coverage} is the ratio of edges with speed values between 8am and 6pm (from 20 sampled days);
        \textbf{mean segment volume} is mean volume sum of all intersecting cells for segments with speed value (from 20 sample days).
\nopagebreak
\begin{table*}[ht]
    \caption{Extended dataset overview.
    }
    \label{tab:dataset_extended}
    \centering
    \resizebox{\textwidth}{!}{
        \begin{tabular}{llp{2.2cm}L{2.5cm}rrR{1.5cm}R{1.5cm}R{1cm}R{1cm}R{1cm}R{1cm}R{1cm}}
        \toprule
        city (t4c year) & days & date ranges & lat\_min, lat\_max, lon\_min, lon\_max & nodes & edges & total segment length [m] & mean segment length [m] &  ratio covered edges & mapped ratio  & daily fcd data & 8--18 coverage & mean segment volume  \\
        \midrule
        Antwerp (2021) & 361  & 2019-01--2019-06, 2020-01--2020-06  & (51.001, 51.437, 4.153, 4.648) & 34722 & 81667 & 13833313.7 & 169.4 &  0.99 & 0.89 & 3.247e+06& 0.17 & 7.10 \\
        Bangkok (2021) & 361  & 2019-01--2019-06, 2020-01--2020-06  & (13.554, 14.049, 100.308, 100.744) & 317797 & 694818 & 58792833.8 & 84.6 &  0.76 & 0.91 & 4.576e+06& 0.05 & 7.50 \\
        Barcelona (2021) & 361  & 2019-01--2019-06, 2020-01--2020-06  & (41.253, 41.748, 1.925, 2.361) & 58106 & 118813 & 14081313.2 & 118.5 &  0.97 & 0.81 & 1.811e+06& 0.09 & 6.24 \\
        Berlin (2021) & 180  & 2019-01--2019-06  & (52.359, 52.854, 13.189, 13.625) & 34308 & 88882 & 14045121.9 & 158.0 &  1.00 & 0.94 & 6.270e+07& 0.36 & 57.31 \\
        Chicago (2021) & 180  & 2019-01--2019-06  & (41.601, 42.096, -87.945, -87.509) & 68430 & 187570 & 27117716.0 & 144.6 &  0.98 & 0.93 & 4.684e+06& 0.11 & 9.82 \\
        Istanbul (2021) & 180  & 2019-01--2019-06  & (40.81, 41.305, 28.794, 29.23) & 102754 & 270109 & 22126616.3 & 81.9 &  1.00 & 0.96 & 7.065e+07& 0.63 & 24.53 \\
        Melbourne (2021) & 180  & 2019-01--2019-06  & (-38.106, -37.611, 144.757, 145.193) & 103062 & 230654 & 24277388.0 & 105.3 &  0.95 & 0.93 & 2.519e+06& 0.08 & 5.56 \\
        Moscow (2021) & 361  & 2019-01--2019-06, 2020-01--2020-06  & (55.506, 55.942, 37.358, 37.853) & 22627 & 47877 & 10906823.2 & 227.8 &  1.00 & 0.81 & 7.292e+07& 0.71 & 46.88 \\
        London (2022) & 110  & 2019-07--2019-12, 2020-01  & (51.205, 51.7, -0.369, 0.067) & 116304 & 271075 & 26738400.4 & 98.6 &  0.99 & 0.95 & 1.119e+07& 0.24 & 8.15 \\
        Madrid (2022) & 109  & 2021-06--2021-12  & (40.177, 40.672, -3.927, -3.491) & 71757 & 143402 & 15799502.4 & 110.2 &  0.99 & 0.93 & 2.977e+07& 0.36 & 22.05 \\
        Melbourne (2022) & 108  & 2020-06--2020-12  & (-38.106, -37.611, 144.757, 145.193) & 103062 & 230654 & 24277388.0 & 105.3 &  0.95 & 0.90 & 2.086e+06& 0.08 & 4.52 \\
        \bottomrule
        \end{tabular}
    }
\end{table*}

\onecolumn
\section{Key Figures}\label{appendix:key_figures}

When a city has appeared in multiple competition years, we add the competition year for data download. The code to reproduce the figures, as well as additional figures can be found in our code repository.
\clearpage

\subsection{Key Figures Antwerp (2021)}
\subsubsection{Road graph map Antwerp (2021)}
\mbox{}
\nopagebreak{}
\begin{figure}[H]
\centering
\includegraphics[width=0.85\textwidth]{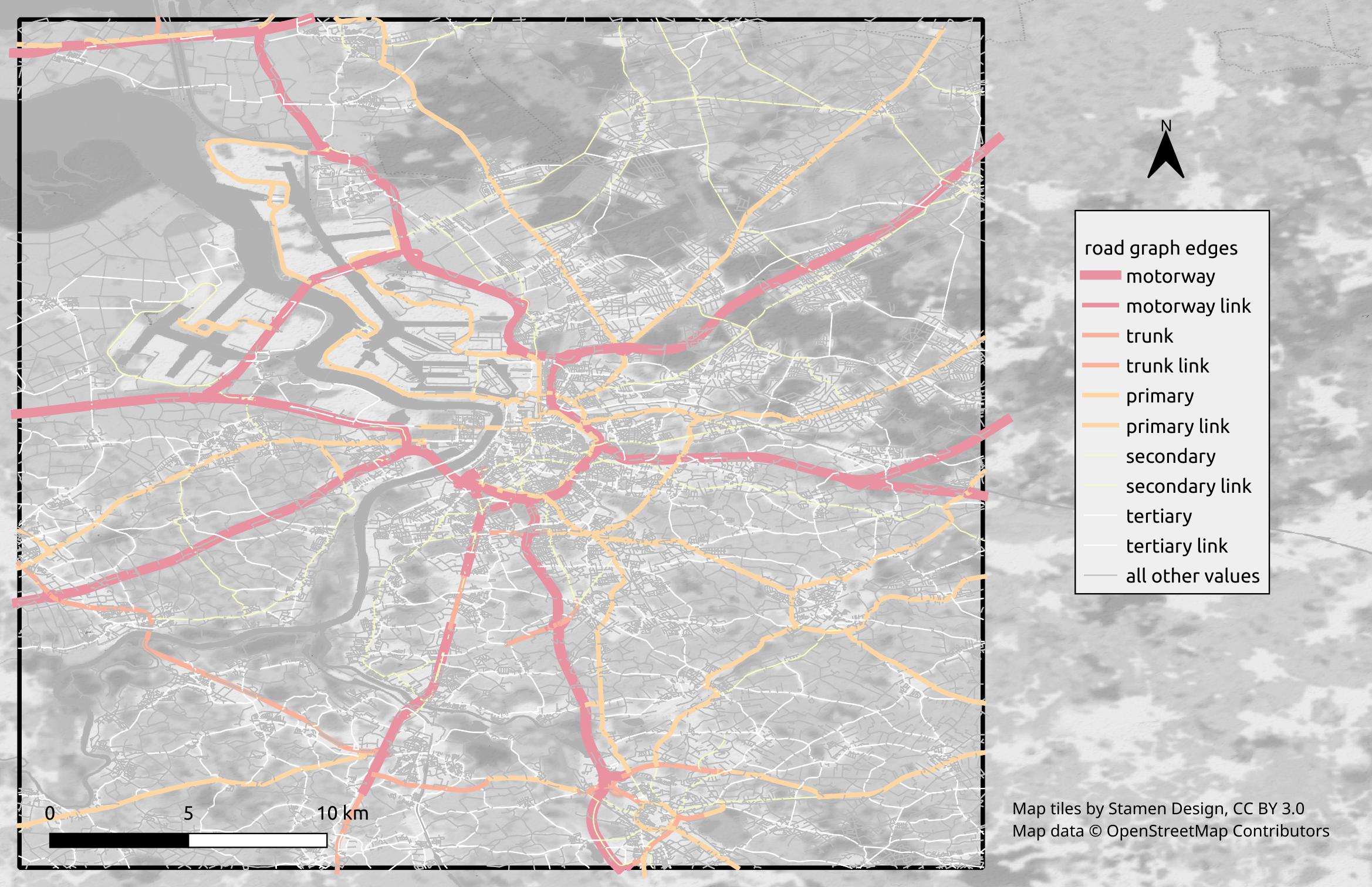}
\caption[Road graph Antwerp]{Road graph Antwerp, OSM color scheme (2021).}
\label{figures/speed_stats/road_graph_antwerp_2021.jpg}
\end{figure}
\subsubsection{Static data  Antwerp  (2021) }
\mbox{}\nopagebreak
\begin{small}
\begin{longtable}{p{4cm}rrrrrrrr}
\toprule
Attribute      & {mean} &{std} & {median}  & {q01} & {q99} & {data points} & {sum}  \\
\midrule
 bounding box                &  &  &  &  &  &  4.153--4.648 / 51.001--51.437 &                                                \\
 num\_edges                &  &  &  &  &  &  81'667 &                                                \\
 \hspace{10pt}  motorway               &  &  &  &  &  &  232 &                                                \\
 \hspace{10pt}  motorway\_link               &  &  &  &  &  &  483 &                                                \\
 \hspace{10pt}  trunk               &  &  &  &  &  &  274 &                                                \\
 \hspace{10pt}  trunk\_link               &  &  &  &  &  &  67 &                                                \\
 \hspace{10pt}  primary               &  &  &  &  &  &  2798 &                                                \\
 \hspace{10pt}  primary\_link               &  &  &  &  &  &  152 &                                                \\
 \hspace{10pt}  secondary               &  &  &  &  &  &  3348 &                                                \\
 \hspace{10pt}  secondary\_link               &  &  &  &  &  &  69 &                                                \\
 \hspace{10pt}  tertiary               &  &  &  &  &  &  10929 &                                                \\
 \hspace{10pt}  tertiary\_link               &  &  &  &  &  &  42 &                                                \\
 \hspace{10pt}  unclassified               &  &  &  &  &  &  11700 &                                                \\
 \hspace{10pt}  residential               &  &  &  &  &  &  51573 &                                                \\
 num\_nodes                &  &  &  &  &  &  34722 &                                                 \\
 num\_edges\_per\_cell                & 1.0 & 0.2 & 1.0 & 1.0 & 2.0 &  385'170 &                                                 \\
 num\_intersecting\_cells                & 4.9 & 5.1 & 4.0 & 1.0 & 25.0 &  81'667 &                                                 \\
 node\_degree                & 2.7 & 0.9 & 3.0 & 1.0 & 4.0 &  34'722 &                                                 \\
 length\_meters                & 169.4 & 240.4 & 100.8 & 6.6 & 1'113.6 &  81'667 & 1.4e+07                                                \\
 \hspace{10pt}  motorway               & 1'496.8 & 1'426.3 & 954.9 & 93.1 & 6'717.8 &  232  & 3.5e+05                                                \\
 \hspace{10pt}  motorway\_link               & 313.5 & 330.0 & 218.0 & 11.0 & 1'684.1 &  483  & 1.5e+05                                                \\
 \hspace{10pt}  trunk               & 370.6 & 468.1 & 184.0 & 7.6 & 2'112.0 &  274  & 1.0e+05                                                \\
 \hspace{10pt}  trunk\_link               & 162.4 & 156.1 & 121.7 & 15.4 & 646.1 &  67  & 1.1e+04                                                \\
 \hspace{10pt}  primary               & 201.6 & 273.3 & 106.4 & 5.4 & 1'393.4 &  2'798  & 5.6e+05                                                \\
 \hspace{10pt}  primary\_link               & 49.6 & 66.0 & 24.4 & 4.1 & 288.3 &  152  & 7.5e+03                                                \\
 \hspace{10pt}  secondary               & 171.1 & 246.4 & 95.5 & 5.5 & 1'142.0 &  3'348  & 5.7e+05                                                \\
 \hspace{10pt}  secondary\_link               & 63.1 & 69.7 & 42.6 & 6.7 & 300.2 &  69  & 4.4e+03                                                \\
 \hspace{10pt}  tertiary               & 176.7 & 237.1 & 103.4 & 5.4 & 1'116.8 &  10'929  & 1.9e+06                                                \\
 \hspace{10pt}  tertiary\_link               & 52.2 & 42.7 & 39.7 & 15.8 & 207.5 &  42  & 2.2e+03                                                \\
 \hspace{10pt}  unclassified               & 299.4 & 348.2 & 184.3 & 7.2 & 1'541.1 &  11'700  & 3.5e+06                                                \\
 \hspace{10pt}  residential               & 128.7 & 137.1 & 92.4 & 6.9 & 650.5 &  51'573  & 6.6e+06                                                \\
 speed\_kph                & 42.0 & 12.0 & 36.3 & 30.0 & 80.0 &  81'667 &                                                 \\
 \hspace{10pt}  motorway               & 109.2 & 15.9 & 120.0 & 50.0 & 120.0 &  232  &                                                 \\
 \hspace{10pt}  motorway\_link               & 82.6 & 16.6 & 81.4 & 50.0 & 120.0 &  483  &                                                 \\
 \hspace{10pt}  trunk               & 77.9 & 14.4 & 70.0 & 50.0 & 120.0 &  274  &                                                 \\
 \hspace{10pt}  trunk\_link               & 61.5 & 9.0 & 61.2 & 50.0 & 90.0 &  67  &                                                 \\
 \hspace{10pt}  primary               & 61.7 & 11.0 & 66.5 & 30.0 & 80.0 &  2'798  &                                                 \\
 \hspace{10pt}  primary\_link               & 55.5 & 5.7 & 55.5 & 50.0 & 70.0 &  152  &                                                 \\
 \hspace{10pt}  secondary               & 56.0 & 9.4 & 50.0 & 30.0 & 70.0 &  3'348  &                                                 \\
 \hspace{10pt}  secondary\_link               & 58.4 & 7.9 & 59.2 & 43.6 & 76.4 &  69  &                                                 \\
 \hspace{10pt}  tertiary               & 49.6 & 9.7 & 50.0 & 30.0 & 70.0 &  10'929  &                                                 \\
 \hspace{10pt}  tertiary\_link               & 49.7 & 7.0 & 48.9 & 30.0 & 70.0 &  42  &                                                 \\
 \hspace{10pt}  unclassified               & 48.1 & 6.6 & 48.2 & 30.0 & 70.0 &  11'700  &                                                 \\
 \hspace{10pt}  residential               & 36.1 & 6.8 & 36.3 & 30.0 & 50.0 &  51'573  &                                                 \\
 free\_flow\_kph                & 38.9 & 18.0 & 35.8 & 5.6 & 116.7 &  77'109 &                                                 \\
 \hspace{10pt}  motorway               & 109.6 & 12.9 & 117.1 & 71.5 & 120.0 &  232  &                                                 \\
 \hspace{10pt}  motorway\_link               & 87.4 & 28.5 & 93.9 & 26.3 & 120.0 &  483  &                                                 \\
 \hspace{10pt}  trunk               & 66.5 & 22.9 & 65.2 & 23.3 & 120.0 &  272  &                                                 \\
 \hspace{10pt}  trunk\_link               & 74.1 & 25.1 & 74.6 & 27.1 & 120.0 &  67  &                                                 \\
 \hspace{10pt}  primary               & 52.2 & 16.3 & 50.5 & 24.0 & 95.8 &  2'798  &                                                 \\
 \hspace{10pt}  primary\_link               & 41.4 & 23.1 & 36.8 & 7.4 & 92.0 &  148  &                                                 \\
 \hspace{10pt}  secondary               & 49.4 & 16.8 & 46.6 & 20.7 & 108.8 &  3'344  &                                                 \\
 \hspace{10pt}  secondary\_link               & 53.8 & 31.4 & 41.8 & 13.6 & 118.4 &  68  &                                                 \\
 \hspace{10pt}  tertiary               & 46.0 & 15.5 & 43.8 & 19.8 & 104.0 &  10'915  &                                                 \\
 \hspace{10pt}  tertiary\_link               & 43.9 & 18.6 & 41.2 & 21.1 & 95.0 &  42  &                                                 \\
 \hspace{10pt}  unclassified               & 44.1 & 21.7 & 41.9 & 3.5 & 119.5 &  10'392  &                                                 \\
 \hspace{10pt}  residential               & 33.6 & 13.9 & 32.0 & 3.3 & 80.0 &  48'348  &                                                 \\
 free\_flow\_kph-speed\_kph                & -3.3 & 15.8 & -3.9 & -36.6 & 53.6 &  77'109 &                                                 \\
 \hspace{10pt}  motorway               & 0.4 & 11.9 & -1.4 & -25.2 & 42.8 &  232  &                                                 \\
 \hspace{10pt}  motorway\_link               & 4.9 & 28.8 & 6.8 & -55.6 & 62.1 &  483  &                                                 \\
 \hspace{10pt}  trunk               & -11.5 & 19.8 & -7.2 & -57.4 & 30.5 &  272  &                                                 \\
 \hspace{10pt}  trunk\_link               & 12.6 & 24.8 & 10.8 & -34.1 & 70.0 &  67  &                                                 \\
 \hspace{10pt}  primary               & -9.5 & 14.5 & -7.9 & -44.3 & 23.9 &  2'798  &                                                 \\
 \hspace{10pt}  primary\_link               & -14.1 & 23.7 & -15.5 & -55.7 & 38.9 &  148  &                                                 \\
 \hspace{10pt}  secondary               & -6.7 & 16.1 & -7.2 & -39.3 & 48.9 &  3'344  &                                                 \\
 \hspace{10pt}  secondary\_link               & -4.6 & 30.2 & -17.6 & -45.6 & 58.4 &  68  &                                                 \\
 \hspace{10pt}  tertiary               & -3.6 & 14.2 & -3.9 & -32.4 & 48.6 &  10'915  &                                                 \\
 \hspace{10pt}  tertiary\_link               & -5.9 & 20.2 & -7.7 & -37.1 & 46.1 &  42  &                                                 \\
 \hspace{10pt}  unclassified               & -4.0 & 21.6 & -6.1 & -45.4 & 71.1 &  10'392  &                                                 \\
 \hspace{10pt}  residential               & -2.5 & 14.1 & -3.2 & -33.1 & 42.3 &  48'348  &                                                 \\
\bottomrule

        \caption[Key figures Antwerp ]{Key figures Antwerp for the generated data from 20 randomly sampled days.
        \textbf{num\_edges} number of edges in the street network graph;
        \textbf{num\_nodes} number of nodes in the street network graph;
        \textbf{num\_edges\_per\_cell} number of edges a cell (row,col,heading) has in its intersecting cells;
        \textbf{num\_intersecting\_cells} number of cells (row,col,heading) in an edge's intersecting cells;
        \textbf{node\_degree} number of (unique) neighbor nodes per node;
        \textbf{length\_meters} free flow speed derived from data;
        \textbf{speed\_kph} signalled speed;
        \textbf{free\_flow\_kph} free flow speed derived from data;
        \textbf{free\_flow\_kph-speed\_kph} difference
        }
    \label{tab:key_figures:/iarai/public/t4c/data_pipeline/release20221026_residential_unclassified/2021:Antwerp:}
    \end{longtable}
    \end{small}
    
\subsubsection{Segment density map  Antwerp (2021)}
\mbox{}
\nopagebreak{}
\begin{figure}[H]
\centering
\includegraphics[width=0.85\textwidth]{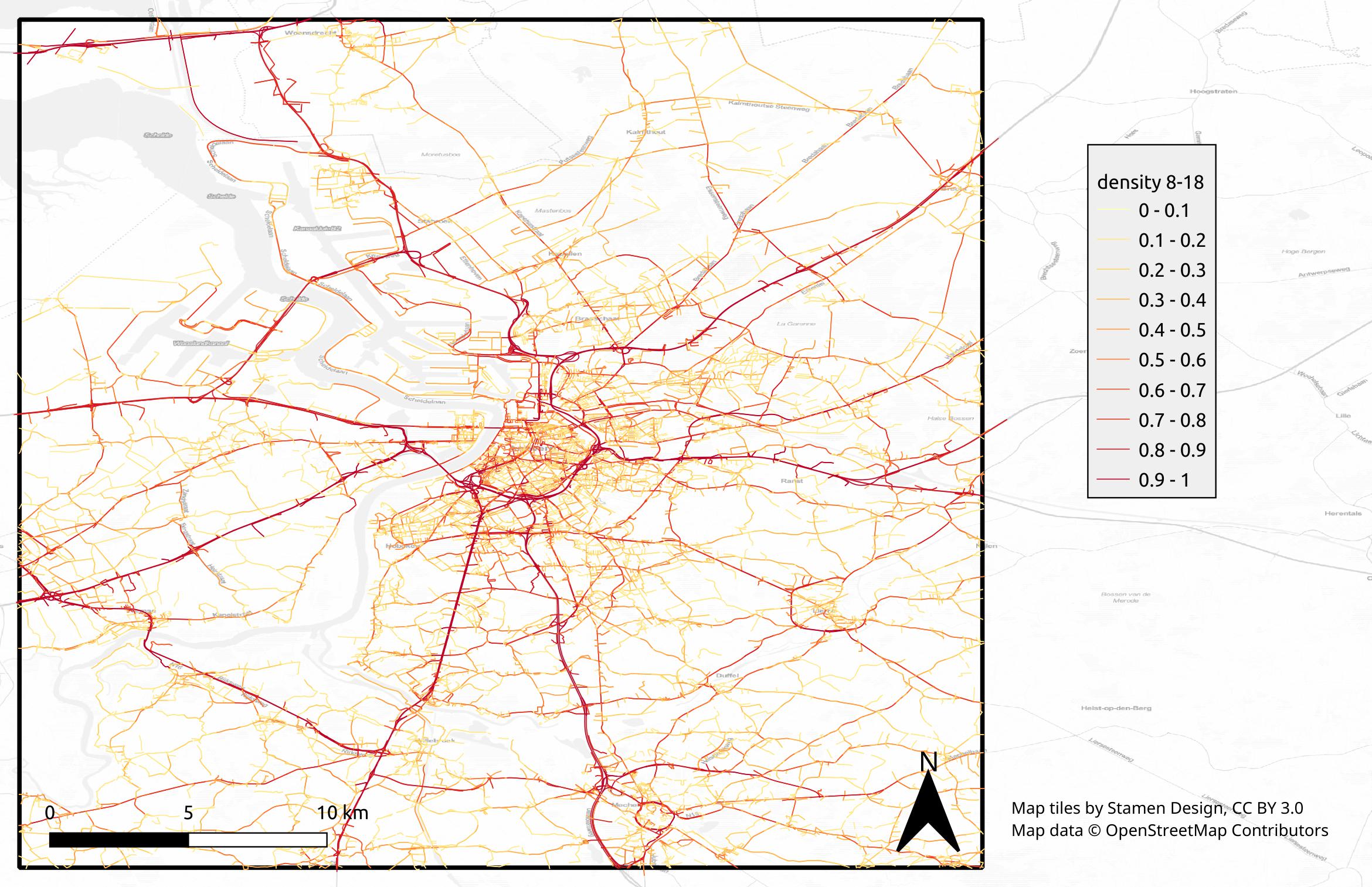}
\caption[Segment-wise density 8am--6pm Antwerp]{Segment-wise density 8am--6pm Antwerp from 20 randomly sampled days.}
\label{figures/speed_stats/density_8_18_antwerp_2021.jpg}
\end{figure}
\clearpage
\subsubsection{Daily density profile  Antwerp  (2021) } 
\mbox{}
\nopagebreak{}
\begin{figure}[H]
\centering
\includegraphics[width=0.85\textwidth]{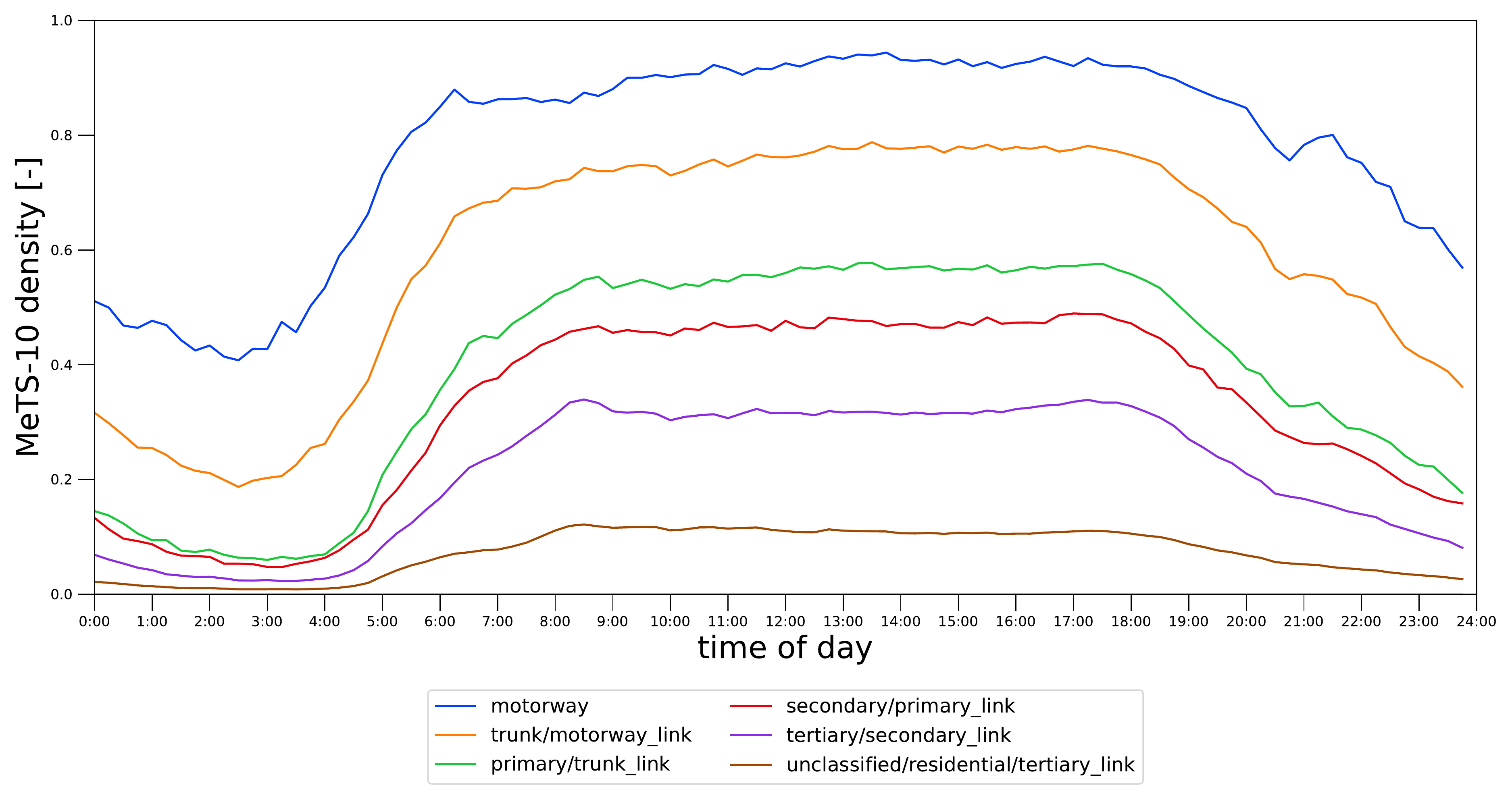}
\caption[Daily density profile Antwerp]{Daily density profile for different road types for Antwerp . Data from 20 randomly sampled days.}
\label{figures/speed_stats/speed_stats_coverage_antwerp_2021_by_highway.pdf}
\end{figure}
\subsubsection{Daily speed profile  Antwerp  (2021) }
\mbox{}
\nopagebreak{}
\begin{figure}[H]
\centering
\includegraphics[width=0.85\textwidth]{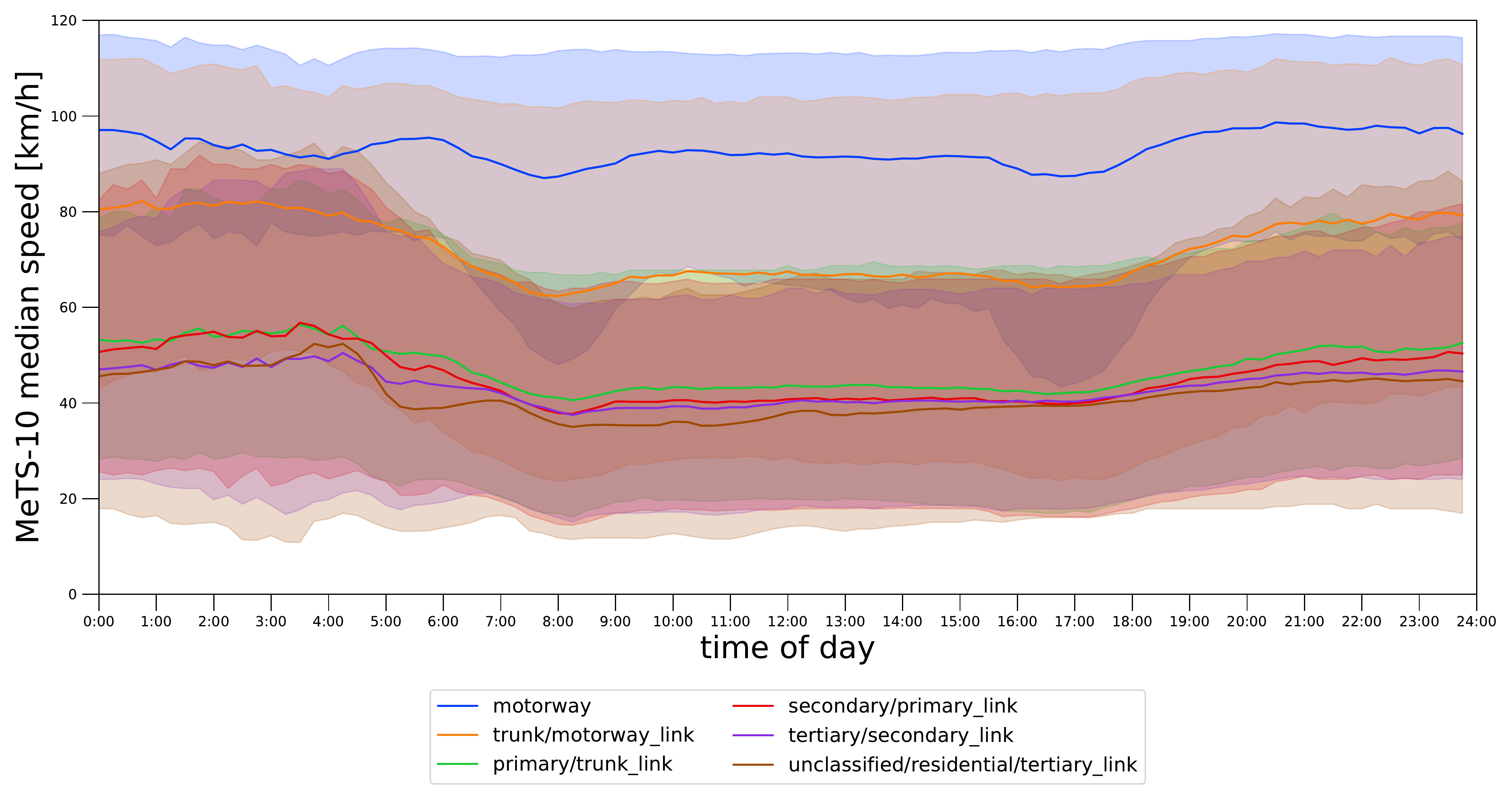}
\caption[Daily median 15 min speeds of all intersecting cells profile Antwerp]{Daily median 15 min speeds of all intersecting cells profile for different road types for Antwerp . The error hull is the 80\% data interval [10.0--90.0 percentiles] of daily means from 20 randomly sampled days.}
\label{figures/speed_stats/speed_stats_median_speed_kph_antwerp_2021_by_highway.pdf}
\end{figure}
\clearpage

\subsection{Key Figures Bangkok (2021)}
\subsubsection{Road graph map Bangkok (2021)}
\mbox{}
\nopagebreak{}
\begin{figure}[H]
\centering
\includegraphics[width=0.85\textwidth]{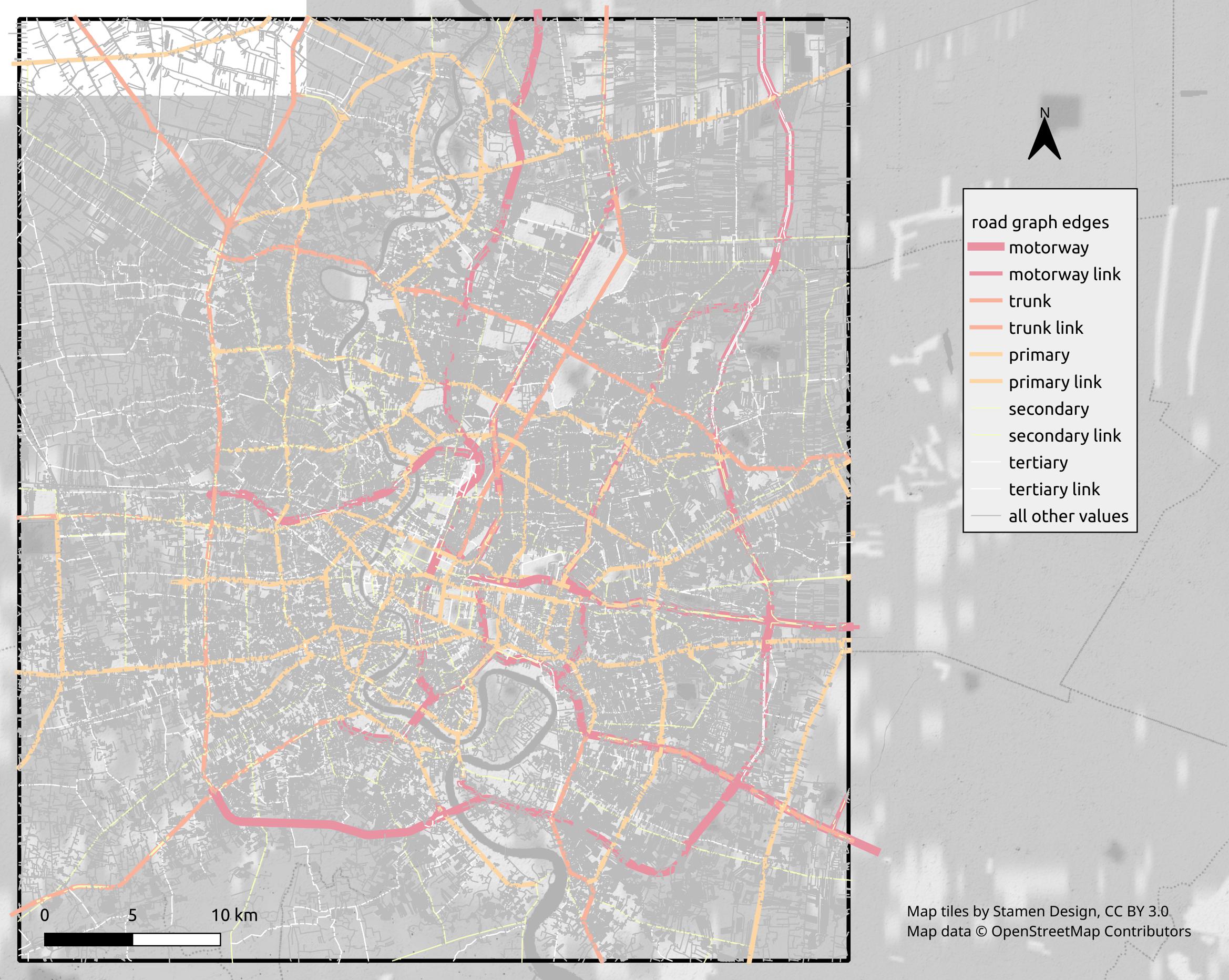}
\caption[Road graph Bangkok]{Road graph Bangkok, OSM color scheme (2021).}
\label{figures/speed_stats/road_graph_bangkok_2021.jpg}
\end{figure}
\subsubsection{Static data  Bangkok  (2021) }
\mbox{}\nopagebreak
\begin{small}
\begin{longtable}{p{4cm}rrrrrrrr}
\toprule
Attribute      & {mean} &{std} & {median}  & {q01} & {q99} & {data points} & {sum}  \\
\midrule
 bounding box                &  &  &  &  &  &  100.308--100.744 / 13.554--14.049 &                                                \\
 num\_edges                &  &  &  &  &  &  694'818 &                                                \\
 \hspace{10pt}  motorway               &  &  &  &  &  &  378 &                                                \\
 \hspace{10pt}  motorway\_link               &  &  &  &  &  &  878 &                                                \\
 \hspace{10pt}  trunk               &  &  &  &  &  &  1579 &                                                \\
 \hspace{10pt}  trunk\_link               &  &  &  &  &  &  682 &                                                \\
 \hspace{10pt}  primary               &  &  &  &  &  &  7160 &                                                \\
 \hspace{10pt}  primary\_link               &  &  &  &  &  &  1974 &                                                \\
 \hspace{10pt}  secondary               &  &  &  &  &  &  15319 &                                                \\
 \hspace{10pt}  secondary\_link               &  &  &  &  &  &  1510 &                                                \\
 \hspace{10pt}  tertiary               &  &  &  &  &  &  25096 &                                                \\
 \hspace{10pt}  tertiary\_link               &  &  &  &  &  &  342 &                                                \\
 \hspace{10pt}  unclassified               &  &  &  &  &  &  16290 &                                                \\
 \hspace{10pt}  residential               &  &  &  &  &  &  623610 &                                                \\
 num\_nodes                &  &  &  &  &  &  317797 &                                                 \\
 num\_edges\_per\_cell                & 1.2 & 0.7 & 1.0 & 1.0 & 5.0 &  1'827'614 &                                                 \\
 num\_intersecting\_cells                & 3.1 & 2.5 & 2.0 & 1.0 & 12.0 &  694'818 &                                                 \\
 node\_degree                & 2.3 & 1.1 & 3.0 & 1.0 & 4.0 &  317'797 &                                                 \\
 length\_meters                & 84.6 & 118.3 & 49.6 & 5.2 & 538.4 &  694'818 & 5.9e+07                                                \\
 \hspace{10pt}  motorway               & 1'443.2 & 1'511.9 & 961.7 & 36.7 & 7'006.2 &  378  & 5.5e+05                                                \\
 \hspace{10pt}  motorway\_link               & 366.4 & 357.8 & 297.5 & 9.2 & 1'518.2 &  878  & 3.2e+05                                                \\
 \hspace{10pt}  trunk               & 299.6 & 439.6 & 115.9 & 6.4 & 2'025.4 &  1'579  & 4.7e+05                                                \\
 \hspace{10pt}  trunk\_link               & 138.9 & 189.5 & 76.1 & 11.1 & 914.0 &  682  & 9.5e+04                                                \\
 \hspace{10pt}  primary               & 159.8 & 263.3 & 78.3 & 5.1 & 1'190.7 &  7'160  & 1.1e+06                                                \\
 \hspace{10pt}  primary\_link               & 112.9 & 202.2 & 52.6 & 6.6 & 864.2 &  1'974  & 2.2e+05                                                \\
 \hspace{10pt}  secondary               & 107.7 & 148.3 & 59.3 & 4.0 & 773.0 &  15'319  & 1.6e+06                                                \\
 \hspace{10pt}  secondary\_link               & 111.4 & 178.4 & 45.2 & 6.5 & 837.9 &  1'510  & 1.7e+05                                                \\
 \hspace{10pt}  tertiary               & 87.9 & 125.2 & 49.5 & 3.5 & 635.1 &  25'096  & 2.2e+06                                                \\
 \hspace{10pt}  tertiary\_link               & 57.2 & 121.8 & 14.8 & 4.9 & 752.7 &  342  & 2.0e+04                                                \\
 \hspace{10pt}  unclassified               & 124.7 & 175.4 & 63.6 & 3.7 & 877.5 &  16'290  & 2.0e+06                                                \\
 \hspace{10pt}  residential               & 80.0 & 95.0 & 49.0 & 5.5 & 463.3 &  623'610  & 5.0e+07                                                \\
 speed\_kph                & 31.8 & 7.5 & 29.7 & 29.7 & 67.0 &  694'818 &                                                 \\
 \hspace{10pt}  motorway               & 108.3 & 10.6 & 109.0 & 40.6 & 120.0 &  378  &                                                 \\
 \hspace{10pt}  motorway\_link               & 40.6 & 0.0 & 40.6 & 40.6 & 40.6 &  878  &                                                 \\
 \hspace{10pt}  trunk               & 54.9 & 1.4 & 54.9 & 54.9 & 54.9 &  1'579  &                                                 \\
 \hspace{10pt}  trunk\_link               & 54.9 & 0.0 & 54.9 & 54.9 & 54.9 &  682  &                                                 \\
 \hspace{10pt}  primary               & 62.8 & 2.3 & 62.9 & 50.0 & 62.9 &  7'160  &                                                 \\
 \hspace{10pt}  primary\_link               & 43.8 & 1.4 & 43.8 & 43.8 & 43.8 &  1'974  &                                                 \\
 \hspace{10pt}  secondary               & 67.0 & 2.1 & 67.0 & 67.0 & 67.0 &  15'319  &                                                 \\
 \hspace{10pt}  secondary\_link               & 54.9 & 0.8 & 54.9 & 54.9 & 54.9 &  1'510  &                                                 \\
 \hspace{10pt}  tertiary               & 36.4 & 1.9 & 36.4 & 30.0 & 36.4 &  25'096  &                                                 \\
 \hspace{10pt}  tertiary\_link               & 54.9 & 0.0 & 54.9 & 54.9 & 54.9 &  342  &                                                 \\
 \hspace{10pt}  unclassified               & 49.7 & 1.6 & 49.7 & 49.7 & 49.7 &  16'290  &                                                 \\
 \hspace{10pt}  residential               & 29.7 & 0.6 & 29.7 & 29.7 & 29.7 &  623'610  &                                                 \\
 free\_flow\_kph                & 34.9 & 24.3 & 29.6 & 0.0 & 120.0 &  350'957 &                                                 \\
 \hspace{10pt}  motorway               & 83.3 & 17.3 & 81.5 & 37.8 & 120.0 &  378  &                                                 \\
 \hspace{10pt}  motorway\_link               & 74.4 & 21.3 & 77.4 & 16.9 & 120.0 &  877  &                                                 \\
 \hspace{10pt}  trunk               & 57.1 & 16.0 & 53.6 & 28.5 & 91.3 &  1'579  &                                                 \\
 \hspace{10pt}  trunk\_link               & 63.6 & 19.9 & 68.7 & 9.3 & 93.6 &  682  &                                                 \\
 \hspace{10pt}  primary               & 54.5 & 17.6 & 51.8 & 19.8 & 89.9 &  7'156  &                                                 \\
 \hspace{10pt}  primary\_link               & 58.5 & 22.3 & 60.2 & 8.5 & 93.3 &  1'965  &                                                 \\
 \hspace{10pt}  secondary               & 54.1 & 20.1 & 50.8 & 18.8 & 100.6 &  15'314  &                                                 \\
 \hspace{10pt}  secondary\_link               & 59.0 & 22.4 & 60.7 & 8.5 & 96.9 &  1'507  &                                                 \\
 \hspace{10pt}  tertiary               & 44.3 & 19.9 & 40.0 & 15.6 & 120.0 &  24'594  &                                                 \\
 \hspace{10pt}  tertiary\_link               & 50.9 & 21.8 & 49.2 & 10.7 & 96.9 &  327  &                                                 \\
 \hspace{10pt}  unclassified               & 41.8 & 20.9 & 37.6 & 5.6 & 120.0 &  13'506  &                                                 \\
 \hspace{10pt}  residential               & 31.5 & 23.7 & 26.4 & 0.0 & 119.1 &  283'072  &                                                 \\
 free\_flow\_kph-speed\_kph                & 1.2 & 23.5 & -3.3 & -34.1 & 83.6 &  350'957 &                                                 \\
 \hspace{10pt}  motorway               & -25.0 & 20.7 & -28.1 & -71.2 & 37.4 &  378  &                                                 \\
 \hspace{10pt}  motorway\_link               & 33.8 & 21.3 & 36.8 & -23.7 & 79.4 &  877  &                                                 \\
 \hspace{10pt}  trunk               & 2.2 & 16.1 & -1.3 & -26.5 & 36.4 &  1'579  &                                                 \\
 \hspace{10pt}  trunk\_link               & 8.7 & 19.9 & 13.8 & -45.6 & 38.7 &  682  &                                                 \\
 \hspace{10pt}  primary               & -8.3 & 17.4 & -11.1 & -43.1 & 27.0 &  7'156  &                                                 \\
 \hspace{10pt}  primary\_link               & 14.7 & 22.3 & 16.4 & -35.3 & 49.5 &  1'965  &                                                 \\
 \hspace{10pt}  secondary               & -12.9 & 20.2 & -16.4 & -48.2 & 40.0 &  15'314  &                                                 \\
 \hspace{10pt}  secondary\_link               & 4.0 & 22.4 & 5.8 & -46.4 & 42.0 &  1'507  &                                                 \\
 \hspace{10pt}  tertiary               & 7.9 & 19.8 & 3.6 & -20.9 & 83.6 &  24'594  &                                                 \\
 \hspace{10pt}  tertiary\_link               & -4.0 & 21.8 & -5.7 & -44.2 & 42.0 &  327  &                                                 \\
 \hspace{10pt}  unclassified               & -7.9 & 20.9 & -12.1 & -44.1 & 70.3 &  13'506  &                                                 \\
 \hspace{10pt}  residential               & 1.8 & 23.7 & -3.3 & -29.7 & 89.4 &  283'072  &                                                 \\
\bottomrule

        \caption[Key figures Bangkok ]{Key figures Bangkok for the generated data from 20 randomly sampled days.
        \textbf{num\_edges} number of edges in the street network graph;
        \textbf{num\_nodes} number of nodes in the street network graph;
        \textbf{num\_edges\_per\_cell} number of edges a cell (row,col,heading) has in its intersecting cells;
        \textbf{num\_intersecting\_cells} number of cells (row,col,heading) in an edge's intersecting cells;
        \textbf{node\_degree} number of (unique) neighbor nodes per node;
        \textbf{length\_meters} free flow speed derived from data;
        \textbf{speed\_kph} signalled speed;
        \textbf{free\_flow\_kph} free flow speed derived from data;
        \textbf{free\_flow\_kph-speed\_kph} difference
        }
    \label{tab:key_figures:/iarai/public/t4c/data_pipeline/release20221026_residential_unclassified/2021:Bangkok:}
    \end{longtable}
    \end{small}
    
\subsubsection{Segment density map  Bangkok (2021)}
\mbox{}
\nopagebreak{}
\begin{figure}[H]
\centering
\includegraphics[width=0.85\textwidth]{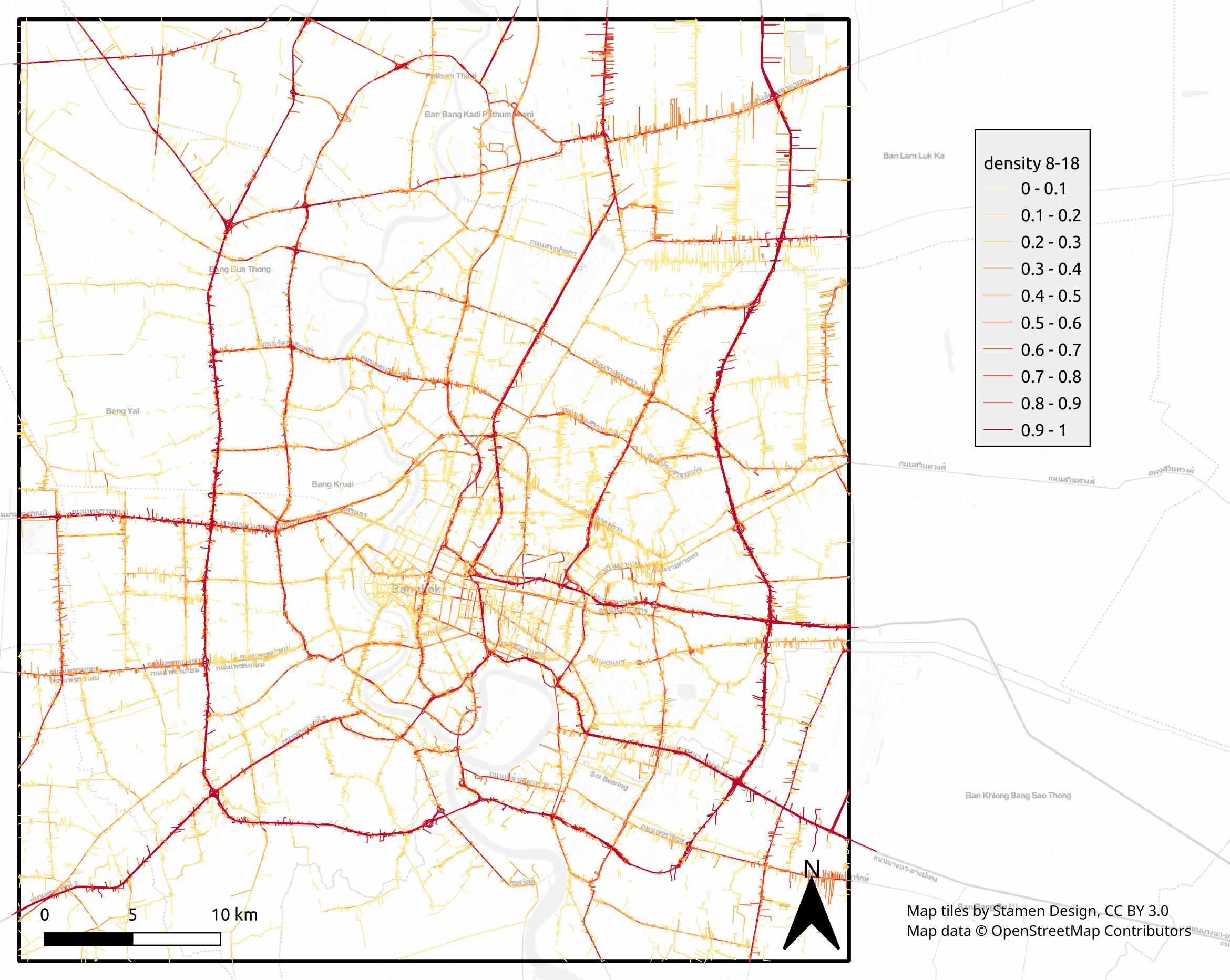}
\caption[Segment-wise density 8am--6pm Bangkok]{Segment-wise density 8am--6pm Bangkok from 20 randomly sampled days.}
\label{figures/speed_stats/density_8_18_bangkok_2021.jpg}
\end{figure}
\clearpage
\subsubsection{Daily density profile  Bangkok  (2021) } 
\mbox{}
\nopagebreak{}
\begin{figure}[H]
\centering
\includegraphics[width=0.85\textwidth]{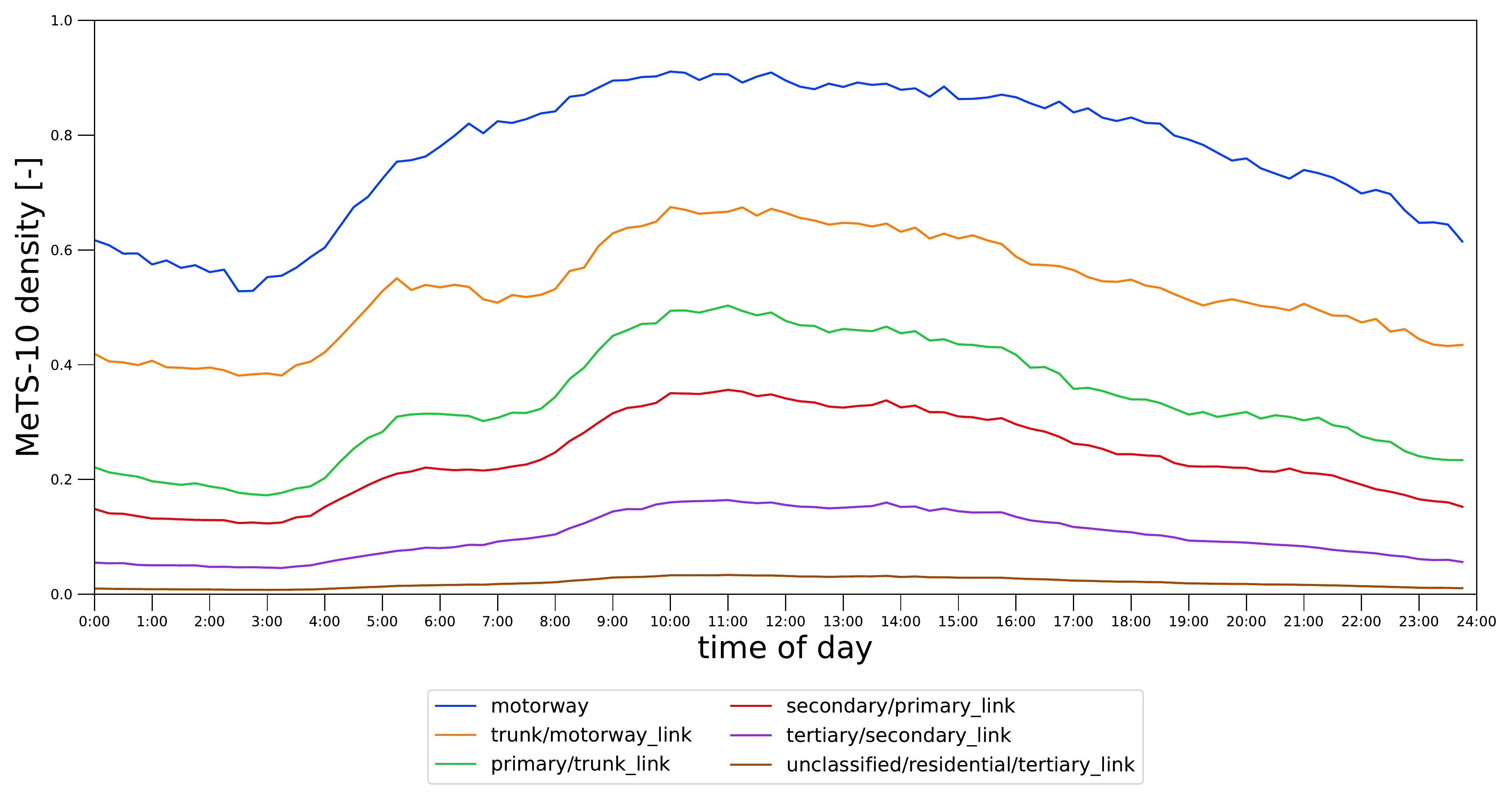}
\caption[Daily density profile Bangkok]{Daily density profile for different road types for Bangkok . Data from 20 randomly sampled days.}
\label{figures/speed_stats/speed_stats_coverage_bangkok_2021_by_highway.pdf}
\end{figure}
\subsubsection{Daily speed profile  Bangkok  (2021) }
\mbox{}
\nopagebreak{}
\begin{figure}[H]
\centering
\includegraphics[width=0.85\textwidth]{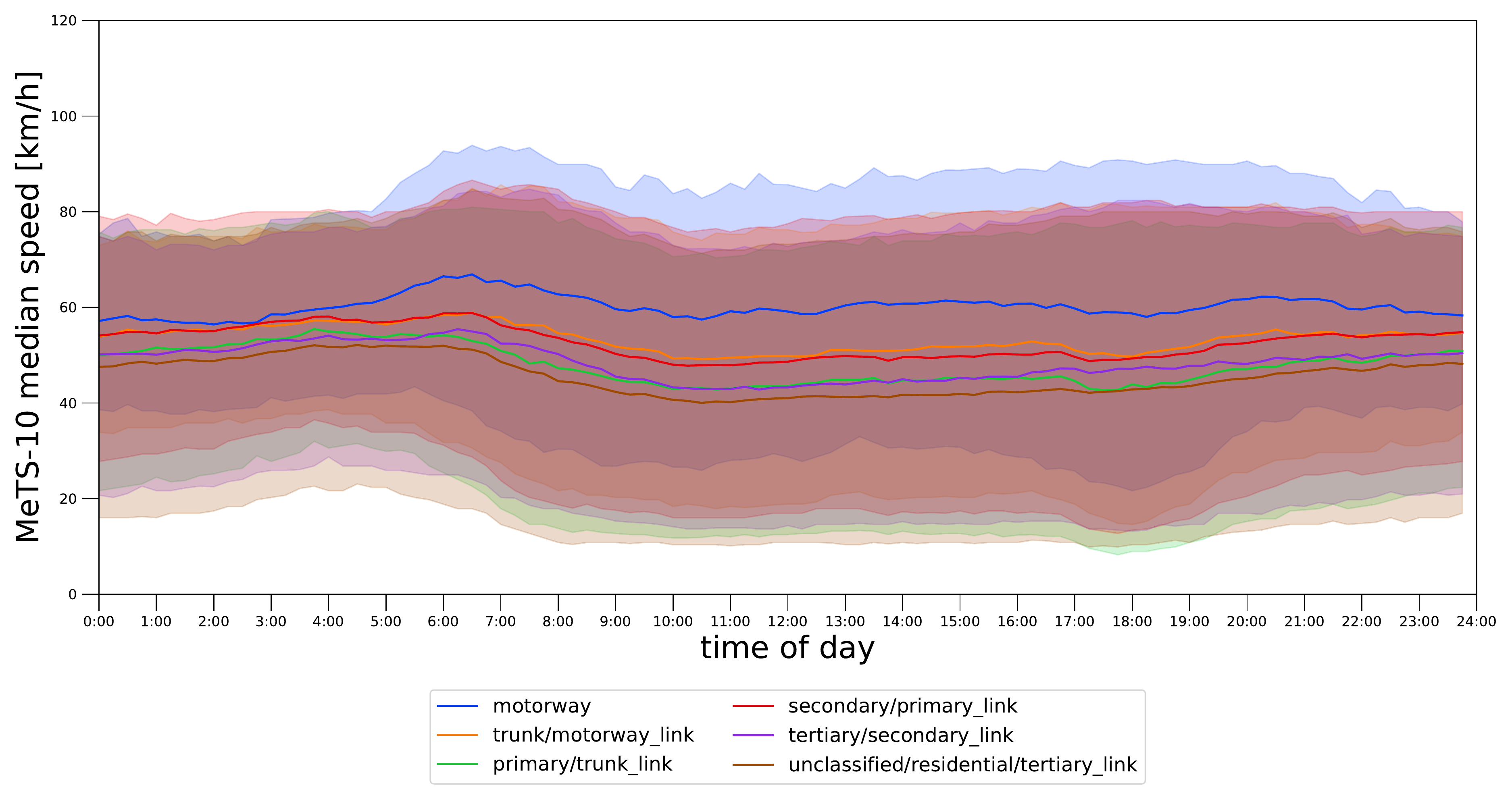}
\caption[Daily median 15 min speeds of all intersecting cells profile Bangkok]{Daily median 15 min speeds of all intersecting cells profile for different road types for Bangkok . The error hull is the 80\% data interval [10.0--90.0 percentiles] of daily means from 20 randomly sampled days.}
\label{figures/speed_stats/speed_stats_median_speed_kph_bangkok_2021_by_highway.pdf}
\end{figure}
\clearpage

\subsection{Key Figures Barcelona (2021)}
\subsubsection{Road graph map Barcelona (2021)}
\mbox{}
\nopagebreak{}
\begin{figure}[H]
\centering
\includegraphics[width=0.85\textwidth]{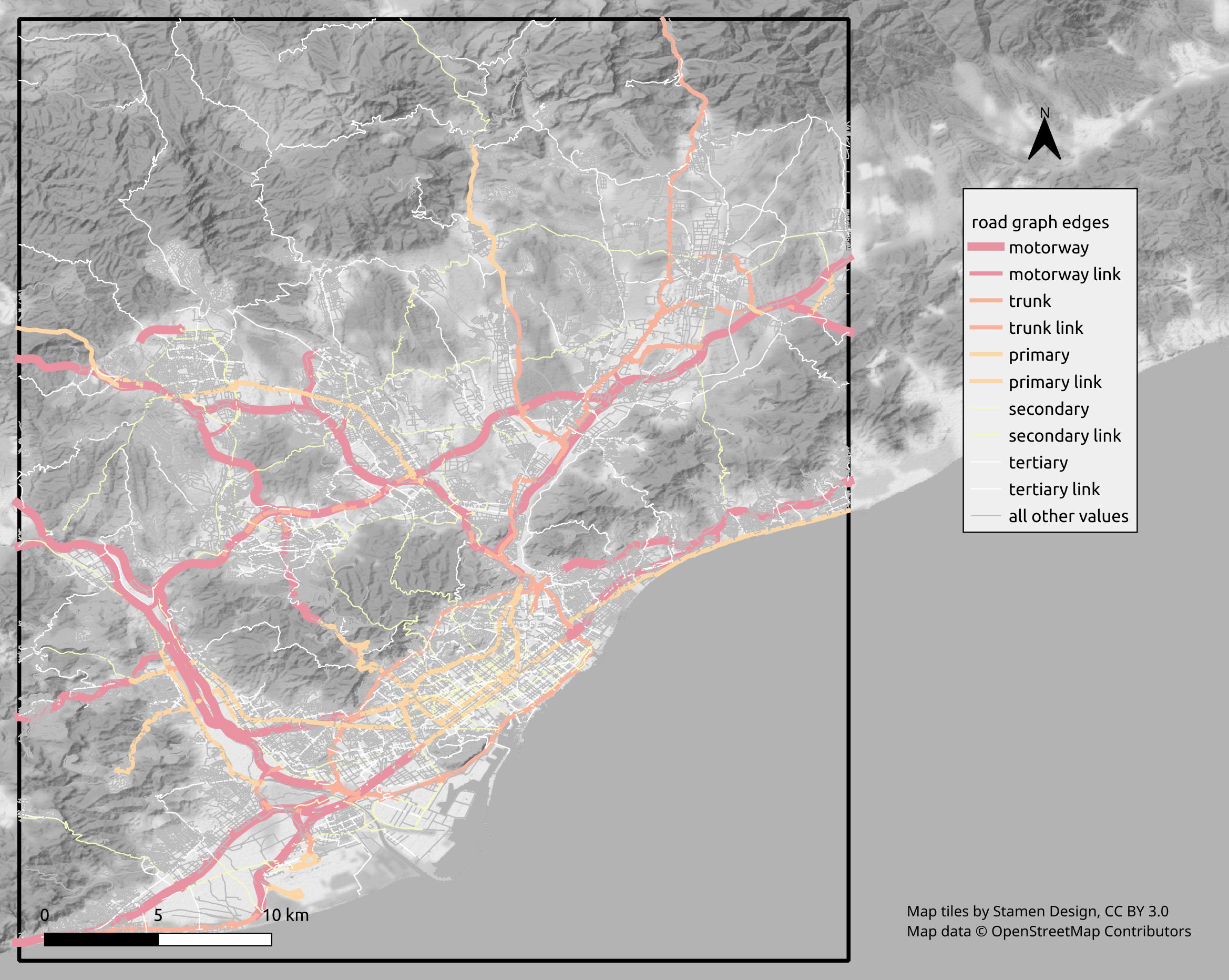}
\caption[Road graph Barcelona]{Road graph Barcelona, OSM color scheme (2021).}
\label{figures/speed_stats/road_graph_barcelona_2021.jpg}
\end{figure}
\subsubsection{Static data  Barcelona  (2021) }
\mbox{}\nopagebreak
\begin{small}
\begin{longtable}{p{4cm}rrrrrrrr}
\toprule
Attribute      & {mean} &{std} & {median}  & {q01} & {q99} & {data points} & {sum}  \\
\midrule
 bounding box                &  &  &  &  &  &  1.925--2.361 / 41.253--41.748 &                                                \\
 num\_edges                &  &  &  &  &  &  118'813 &                                                \\
 \hspace{10pt}  motorway               &  &  &  &  &  &  442 &                                                \\
 \hspace{10pt}  motorway\_link               &  &  &  &  &  &  700 &                                                \\
 \hspace{10pt}  trunk               &  &  &  &  &  &  613 &                                                \\
 \hspace{10pt}  trunk\_link               &  &  &  &  &  &  860 &                                                \\
 \hspace{10pt}  primary               &  &  &  &  &  &  2126 &                                                \\
 \hspace{10pt}  primary\_link               &  &  &  &  &  &  690 &                                                \\
 \hspace{10pt}  secondary               &  &  &  &  &  &  5525 &                                                \\
 \hspace{10pt}  secondary\_link               &  &  &  &  &  &  879 &                                                \\
 \hspace{10pt}  tertiary               &  &  &  &  &  &  14875 &                                                \\
 \hspace{10pt}  tertiary\_link               &  &  &  &  &  &  1250 &                                                \\
 \hspace{10pt}  unclassified               &  &  &  &  &  &  3840 &                                                \\
 \hspace{10pt}  residential               &  &  &  &  &  &  87013 &                                                \\
 num\_nodes                &  &  &  &  &  &  58106 &                                                 \\
 num\_edges\_per\_cell                & 1.1 & 0.4 & 1.0 & 1.0 & 3.0 &  410'314 &                                                 \\
 num\_intersecting\_cells                & 3.7 & 4.6 & 3.0 & 1.0 & 18.0 &  118'813 &                                                 \\
 node\_degree                & 3.0 & 0.8 & 3.0 & 1.0 & 4.0 &  58'106 &                                                 \\
 length\_meters                & 118.5 & 215.0 & 74.0 & 4.5 & 808.4 &  118'813 & 1.4e+07                                                \\
 \hspace{10pt}  motorway               & 987.6 & 954.8 & 708.9 & 38.7 & 5'246.0 &  442  & 4.4e+05                                                \\
 \hspace{10pt}  motorway\_link               & 315.6 & 295.8 & 262.0 & 12.6 & 1'313.0 &  700  & 2.2e+05                                                \\
 \hspace{10pt}  trunk               & 456.4 & 428.8 & 355.4 & 13.5 & 2'071.3 &  613  & 2.8e+05                                                \\
 \hspace{10pt}  trunk\_link               & 199.6 & 163.0 & 165.8 & 9.5 & 734.7 &  860  & 1.7e+05                                                \\
 \hspace{10pt}  primary               & 122.2 & 227.6 & 63.1 & 3.4 & 1'054.5 &  2'126  & 2.6e+05                                                \\
 \hspace{10pt}  primary\_link               & 81.5 & 106.4 & 45.4 & 4.4 & 450.3 &  690  & 5.6e+04                                                \\
 \hspace{10pt}  secondary               & 122.6 & 268.5 & 52.9 & 3.7 & 1'083.5 &  5'525  & 6.8e+05                                                \\
 \hspace{10pt}  secondary\_link               & 54.9 & 80.4 & 29.9 & 3.1 & 365.3 &  879  & 4.8e+04                                                \\
 \hspace{10pt}  tertiary               & 122.3 & 361.2 & 55.2 & 3.1 & 1'192.1 &  14'875  & 1.8e+06                                                \\
 \hspace{10pt}  tertiary\_link               & 46.2 & 60.3 & 27.8 & 3.4 & 301.6 &  1'250  & 5.8e+04                                                \\
 \hspace{10pt}  unclassified               & 238.8 & 458.7 & 101.7 & 5.1 & 2'086.6 &  3'840  & 9.2e+05                                                \\
 \hspace{10pt}  residential               & 105.0 & 112.4 & 76.6 & 5.2 & 550.9 &  87'013  & 9.1e+06                                                \\
 speed\_kph                & 37.2 & 9.4 & 33.9 & 30.0 & 80.0 &  118'813 &                                                 \\
 \hspace{10pt}  motorway               & 100.0 & 19.9 & 100.0 & 40.0 & 120.0 &  442  &                                                 \\
 \hspace{10pt}  motorway\_link               & 63.7 & 13.6 & 63.3 & 40.0 & 120.0 &  700  &                                                 \\
 \hspace{10pt}  trunk               & 81.5 & 15.5 & 80.0 & 40.0 & 100.0 &  613  &                                                 \\
 \hspace{10pt}  trunk\_link               & 58.5 & 10.9 & 58.6 & 30.0 & 100.0 &  860  &                                                 \\
 \hspace{10pt}  primary               & 48.0 & 9.6 & 50.0 & 20.0 & 80.0 &  2'126  &                                                 \\
 \hspace{10pt}  primary\_link               & 46.8 & 7.0 & 50.0 & 20.0 & 60.0 &  690  &                                                 \\
 \hspace{10pt}  secondary               & 49.5 & 10.7 & 50.0 & 30.0 & 90.0 &  5'525  &                                                 \\
 \hspace{10pt}  secondary\_link               & 46.2 & 6.5 & 46.7 & 30.0 & 60.0 &  879  &                                                 \\
 \hspace{10pt}  tertiary               & 43.3 & 8.1 & 44.2 & 30.0 & 60.0 &  14'875  &                                                 \\
 \hspace{10pt}  tertiary\_link               & 38.7 & 6.9 & 38.5 & 30.0 & 50.0 &  1'250  &                                                 \\
 \hspace{10pt}  unclassified               & 40.9 & 5.1 & 41.0 & 30.0 & 50.0 &  3'840  &                                                 \\
 \hspace{10pt}  residential               & 33.7 & 3.7 & 33.9 & 30.0 & 50.0 &  87'013  &                                                 \\
 free\_flow\_kph                & 34.5 & 21.6 & 29.6 & 0.0 & 117.4 &  103'177 &                                                 \\
 \hspace{10pt}  motorway               & 105.3 & 12.3 & 104.0 & 71.4 & 120.0 &  442  &                                                 \\
 \hspace{10pt}  motorway\_link               & 94.3 & 22.7 & 99.8 & 35.3 & 120.0 &  699  &                                                 \\
 \hspace{10pt}  trunk               & 87.4 & 16.0 & 89.9 & 39.9 & 118.6 &  613  &                                                 \\
 \hspace{10pt}  trunk\_link               & 81.6 & 20.2 & 86.6 & 24.1 & 119.3 &  858  &                                                 \\
 \hspace{10pt}  primary               & 47.1 & 18.6 & 42.8 & 18.5 & 100.7 &  2'126  &                                                 \\
 \hspace{10pt}  primary\_link               & 52.9 & 21.9 & 49.9 & 14.5 & 109.1 &  686  &                                                 \\
 \hspace{10pt}  secondary               & 43.6 & 17.3 & 39.5 & 17.9 & 102.3 &  5'519  &                                                 \\
 \hspace{10pt}  secondary\_link               & 44.4 & 19.4 & 40.9 & 14.0 & 102.8 &  865  &                                                 \\
 \hspace{10pt}  tertiary               & 39.0 & 18.4 & 34.8 & 12.7 & 104.9 &  14'706  &                                                 \\
 \hspace{10pt}  tertiary\_link               & 42.9 & 22.7 & 36.7 & 10.9 & 120.0 &  1'215  &                                                 \\
 \hspace{10pt}  unclassified               & 42.1 & 26.0 & 35.3 & 0.0 & 120.0 &  3'228  &                                                 \\
 \hspace{10pt}  residential               & 29.7 & 18.6 & 26.4 & 0.0 & 104.9 &  72'220  &                                                 \\
 free\_flow\_kph-speed\_kph                & -3.1 & 19.4 & -6.6 & -33.9 & 70.0 &  103'177 &                                                 \\
 \hspace{10pt}  motorway               & 5.3 & 15.7 & 0.0 & -24.0 & 40.0 &  442  &                                                 \\
 \hspace{10pt}  motorway\_link               & 30.6 & 22.6 & 35.1 & -27.5 & 70.0 &  699  &                                                 \\
 \hspace{10pt}  trunk               & 5.9 & 13.1 & 4.0 & -25.0 & 43.2 &  613  &                                                 \\
 \hspace{10pt}  trunk\_link               & 23.1 & 21.1 & 28.0 & -34.5 & 61.4 &  858  &                                                 \\
 \hspace{10pt}  primary               & -0.9 & 17.5 & -3.9 & -30.7 & 53.6 &  2'126  &                                                 \\
 \hspace{10pt}  primary\_link               & 6.0 & 22.3 & 0.9 & -33.1 & 58.8 &  686  &                                                 \\
 \hspace{10pt}  secondary               & -5.9 & 17.3 & -8.4 & -42.2 & 53.4 &  5'519  &                                                 \\
 \hspace{10pt}  secondary\_link               & -1.7 & 19.1 & -4.4 & -35.9 & 59.2 &  865  &                                                 \\
 \hspace{10pt}  tertiary               & -4.3 & 18.8 & -7.8 & -34.0 & 64.9 &  14'706  &                                                 \\
 \hspace{10pt}  tertiary\_link               & 4.1 & 23.0 & -1.8 & -30.7 & 80.0 &  1'215  &                                                 \\
 \hspace{10pt}  unclassified               & 1.0 & 26.5 & -5.7 & -41.0 & 79.0 &  3'228  &                                                 \\
 \hspace{10pt}  residential               & -3.9 & 18.7 & -7.1 & -33.9 & 71.0 &  72'220  &                                                 \\
\bottomrule

        \caption[Key figures Barcelona ]{Key figures Barcelona for the generated data from 20 randomly sampled days.
        \textbf{num\_edges} number of edges in the street network graph;
        \textbf{num\_nodes} number of nodes in the street network graph;
        \textbf{num\_edges\_per\_cell} number of edges a cell (row,col,heading) has in its intersecting cells;
        \textbf{num\_intersecting\_cells} number of cells (row,col,heading) in an edge's intersecting cells;
        \textbf{node\_degree} number of (unique) neighbor nodes per node;
        \textbf{length\_meters} free flow speed derived from data;
        \textbf{speed\_kph} signalled speed;
        \textbf{free\_flow\_kph} free flow speed derived from data;
        \textbf{free\_flow\_kph-speed\_kph} difference
        }
    \label{tab:key_figures:/iarai/public/t4c/data_pipeline/release20221026_residential_unclassified/2021:Barcelona:}
    \end{longtable}
    \end{small}
    
\subsubsection{Segment density map  Barcelona (2021)}
\mbox{}
\nopagebreak{}
\begin{figure}[H]
\centering
\includegraphics[width=0.85\textwidth]{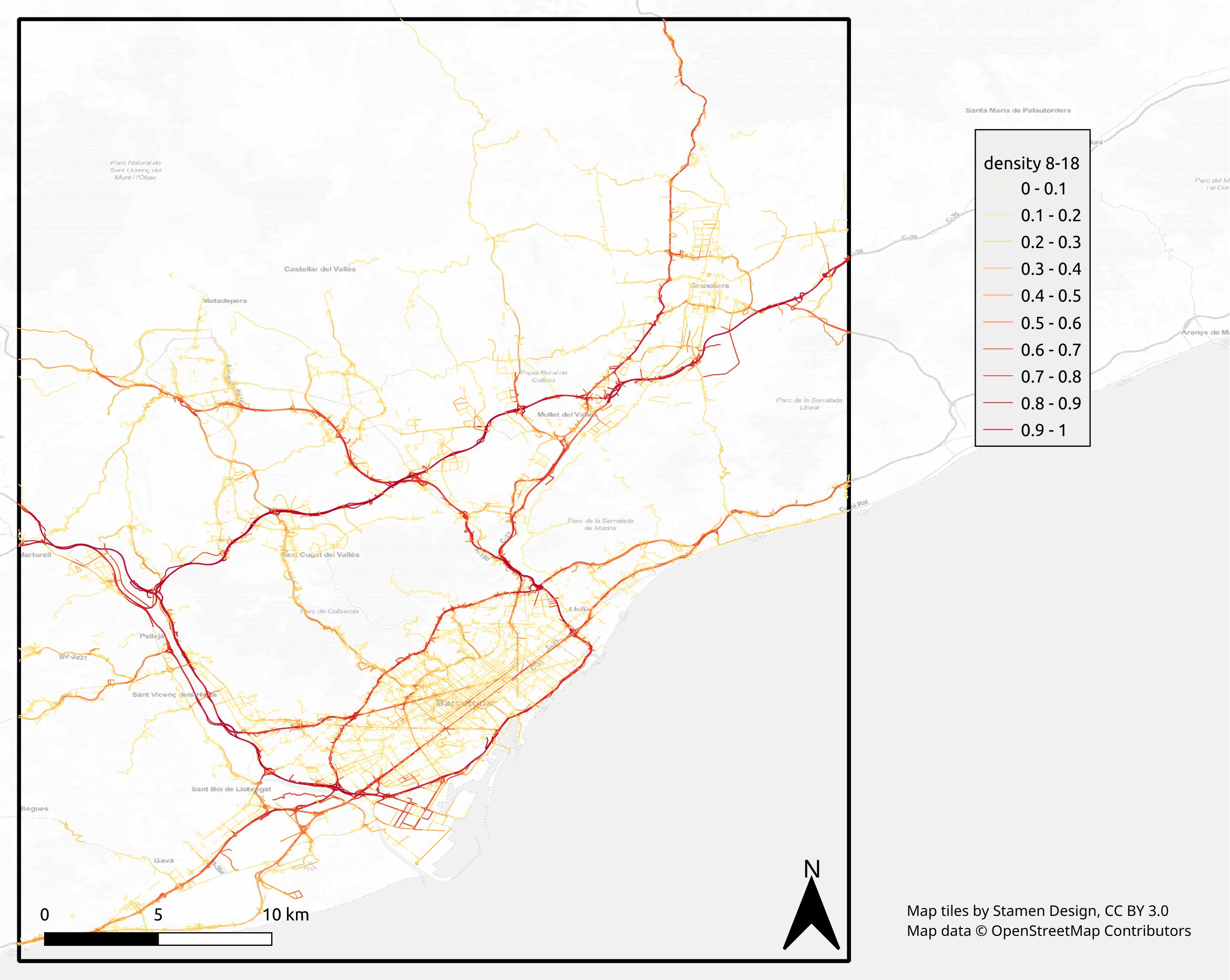}
\caption[Segment-wise density 8am--6pm Barcelona]{Segment-wise density 8am--6pm Barcelona from 20 randomly sampled days.}
\label{figures/speed_stats/density_8_18_barcelona_2021.jpg}
\end{figure}
\clearpage
\subsubsection{Daily density profile  Barcelona  (2021) } 
\mbox{}
\nopagebreak{}
\begin{figure}[H]
\centering
\includegraphics[width=0.85\textwidth]{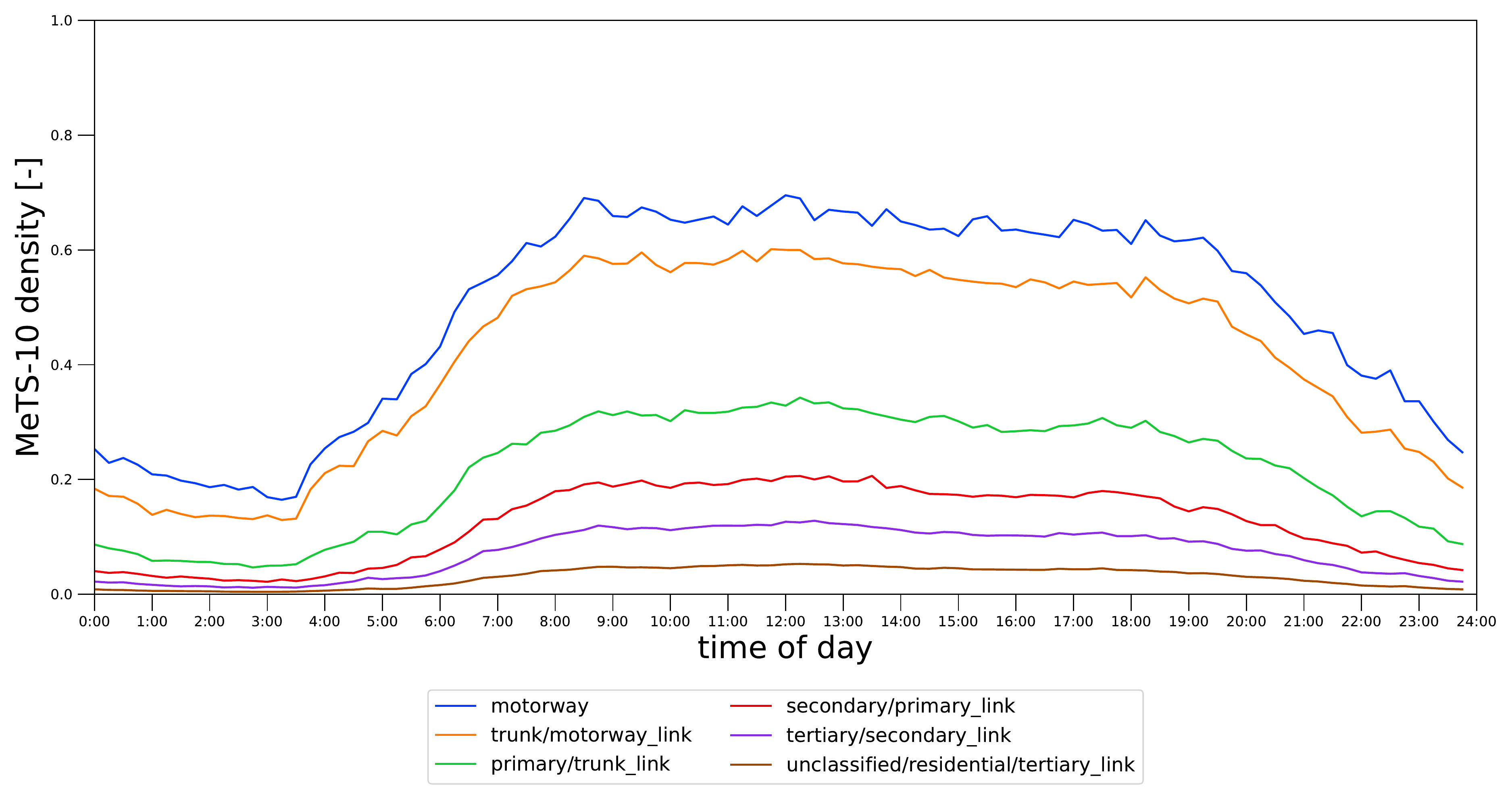}
\caption[Daily density profile Barcelona]{Daily density profile for different road types for Barcelona . Data from 20 randomly sampled days.}
\label{figures/speed_stats/speed_stats_coverage_barcelona_2021_by_highway.pdf}
\end{figure}
\subsubsection{Daily speed profile  Barcelona  (2021) }
\mbox{}
\nopagebreak{}
\begin{figure}[H]
\centering
\includegraphics[width=0.85\textwidth]{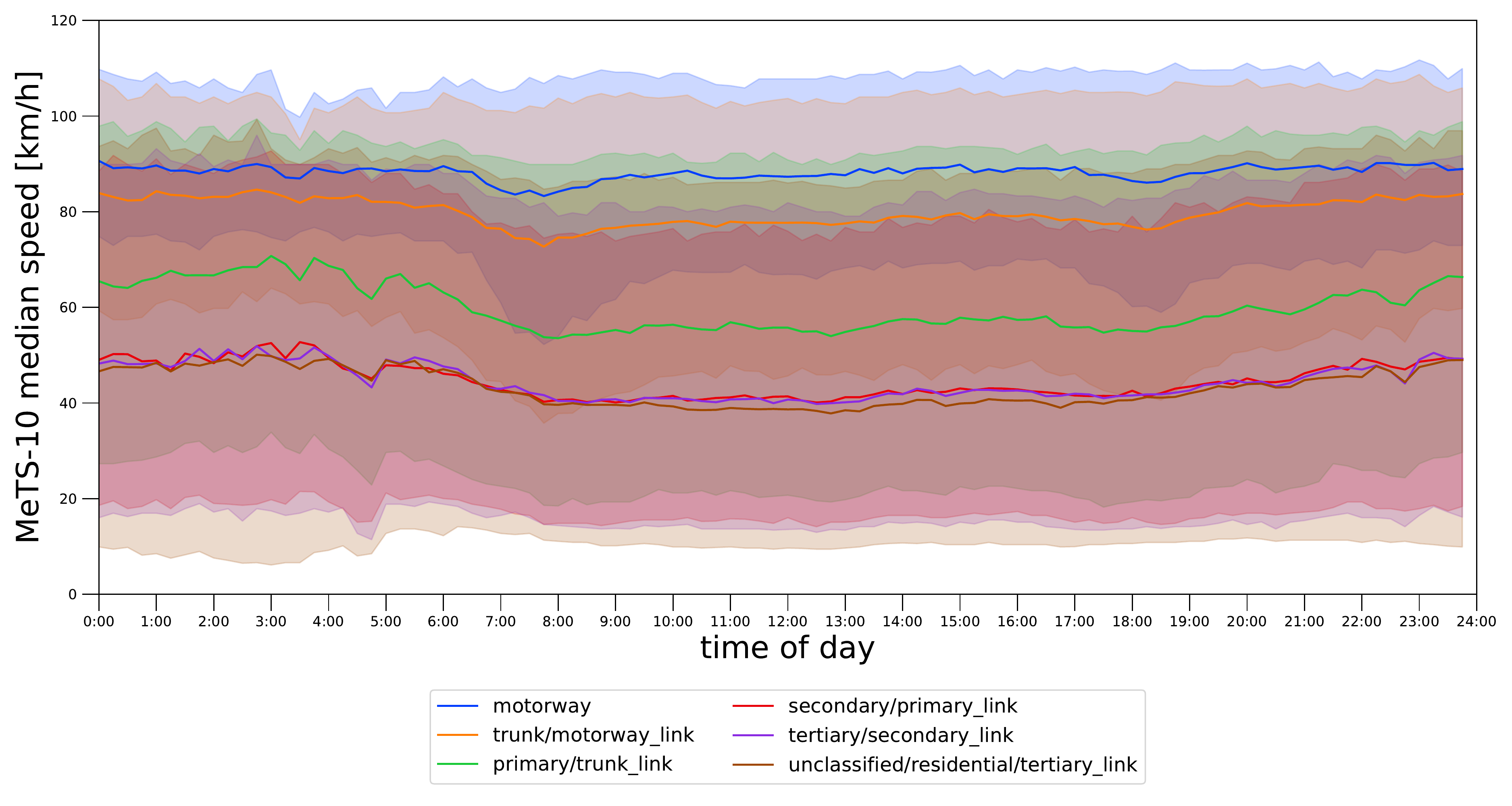}
\caption[Daily median 15 min speeds of all intersecting cells profile Barcelona]{Daily median 15 min speeds of all intersecting cells profile for different road types for Barcelona . The error hull is the 80\% data interval [10.0--90.0 percentiles] of daily means from 20 randomly sampled days.}
\label{figures/speed_stats/speed_stats_median_speed_kph_barcelona_2021_by_highway.pdf}
\end{figure}
\clearpage

\subsection{Key Figures Berlin (2021)}
\subsubsection{Road graph map Berlin (2021)}
\mbox{}
\nopagebreak{}
\begin{figure}[H]
\centering
\includegraphics[width=0.85\textwidth]{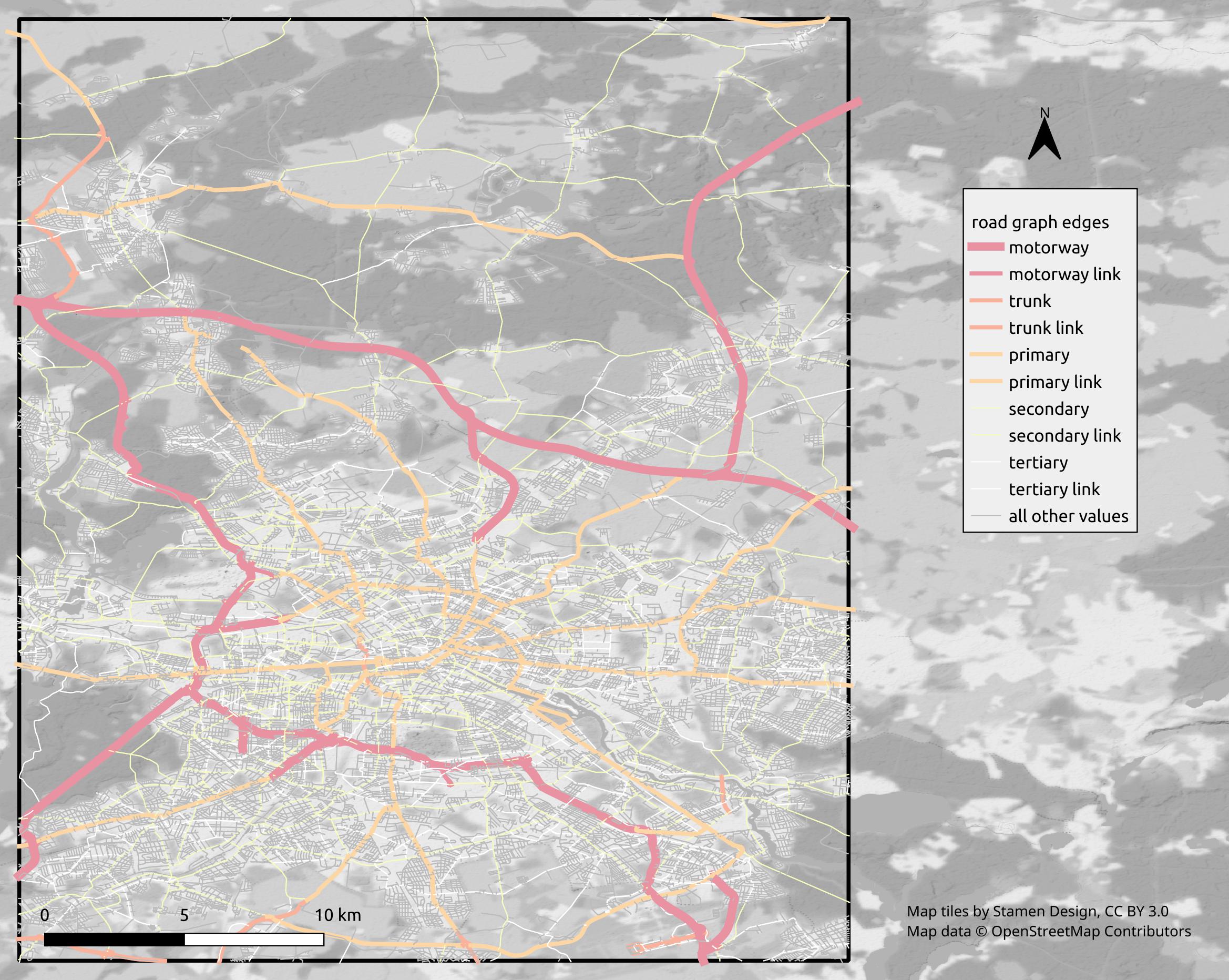}
\caption[Road graph Berlin]{Road graph Berlin, OSM color scheme (2021).}
\label{figures/speed_stats/road_graph_berlin_2021.jpg}
\end{figure}
\subsubsection{Static data  Berlin  (2021) }
\mbox{}\nopagebreak
\begin{small}
\begin{longtable}{p{4cm}rrrrrrrr}
\toprule
Attribute      & {mean} &{std} & {median}  & {q01} & {q99} & {data points} & {sum}  \\
\midrule
 bounding box                &  &  &  &  &  &  13.189--13.625 / 52.359--52.854 &                                                \\
 num\_edges                &  &  &  &  &  &  88'882 &                                                \\
 \hspace{10pt}  motorway               &  &  &  &  &  &  279 &                                                \\
 \hspace{10pt}  motorway\_link               &  &  &  &  &  &  469 &                                                \\
 \hspace{10pt}  trunk               &  &  &  &  &  &  75 &                                                \\
 \hspace{10pt}  trunk\_link               &  &  &  &  &  &  77 &                                                \\
 \hspace{10pt}  primary               &  &  &  &  &  &  2754 &                                                \\
 \hspace{10pt}  primary\_link               &  &  &  &  &  &  150 &                                                \\
 \hspace{10pt}  secondary               &  &  &  &  &  &  10292 &                                                \\
 \hspace{10pt}  secondary\_link               &  &  &  &  &  &  280 &                                                \\
 \hspace{10pt}  tertiary               &  &  &  &  &  &  8486 &                                                \\
 \hspace{10pt}  tertiary\_link               &  &  &  &  &  &  61 &                                                \\
 \hspace{10pt}  unclassified               &  &  &  &  &  &  1766 &                                                \\
 \hspace{10pt}  residential               &  &  &  &  &  &  64193 &                                                \\
 num\_nodes                &  &  &  &  &  &  34308 &                                                 \\
 num\_edges\_per\_cell                & 1.0 & 0.2 & 1.0 & 1.0 & 2.0 &  405'500 &                                                 \\
 num\_intersecting\_cells                & 4.7 & 4.3 & 4.0 & 1.0 & 20.0 &  88'882 &                                                 \\
 node\_degree                & 2.9 & 0.9 & 3.0 & 1.0 & 4.0 &  34'308 &                                                 \\
 length\_meters                & 158.0 & 202.0 & 119.3 & 8.5 & 784.2 &  88'882 & 1.4e+07                                                \\
 \hspace{10pt}  motorway               & 1'058.1 & 1'232.4 & 585.3 & 88.1 & 5'875.6 &  279  & 3.0e+05                                                \\
 \hspace{10pt}  motorway\_link               & 256.9 & 257.1 & 216.8 & 14.3 & 1'456.1 &  469  & 1.2e+05                                                \\
 \hspace{10pt}  trunk               & 846.4 & 959.8 & 516.8 & 44.1 & 5'038.1 &  75  & 6.3e+04                                                \\
 \hspace{10pt}  trunk\_link               & 214.3 & 117.7 & 220.9 & 13.5 & 532.3 &  77  & 1.6e+04                                                \\
 \hspace{10pt}  primary               & 191.6 & 292.9 & 119.1 & 7.7 & 1'583.0 &  2'754  & 5.3e+05                                                \\
 \hspace{10pt}  primary\_link               & 52.6 & 73.4 & 18.0 & 7.2 & 302.2 &  150  & 7.9e+03                                                \\
 \hspace{10pt}  secondary               & 174.3 & 300.9 & 111.5 & 7.6 & 1'554.5 &  10'292  & 1.8e+06                                                \\
 \hspace{10pt}  secondary\_link               & 25.5 & 31.5 & 14.6 & 7.0 & 149.6 &  280  & 7.1e+03                                                \\
 \hspace{10pt}  tertiary               & 155.1 & 216.5 & 113.0 & 7.3 & 951.6 &  8'486  & 1.3e+06                                                \\
 \hspace{10pt}  tertiary\_link               & 23.3 & 34.8 & 12.9 & 7.2 & 160.2 &  61  & 1.4e+03                                                \\
 \hspace{10pt}  unclassified               & 310.2 & 439.8 & 171.5 & 9.5 & 2'386.5 &  1'766  & 5.5e+05                                                \\
 \hspace{10pt}  residential               & 145.6 & 116.1 & 120.1 & 9.0 & 547.1 &  64'193  & 9.3e+06                                                \\
 speed\_kph                & 35.7 & 10.1 & 30.0 & 30.0 & 60.0 &  88'882 &                                                 \\
 \hspace{10pt}  motorway               & 85.4 & 17.1 & 80.0 & 60.0 & 120.0 &  279  &                                                 \\
 \hspace{10pt}  motorway\_link               & 61.7 & 14.7 & 60.0 & 40.0 & 120.0 &  469  &                                                 \\
 \hspace{10pt}  trunk               & 87.5 & 32.5 & 100.0 & 30.0 & 120.0 &  75  &                                                 \\
 \hspace{10pt}  trunk\_link               & 52.6 & 8.5 & 52.5 & 30.0 & 72.4 &  77  &                                                 \\
 \hspace{10pt}  primary               & 49.5 & 8.3 & 50.0 & 30.0 & 80.0 &  2'754  &                                                 \\
 \hspace{10pt}  primary\_link               & 48.9 & 7.8 & 50.0 & 30.0 & 70.0 &  150  &                                                 \\
 \hspace{10pt}  secondary               & 48.7 & 6.5 & 50.0 & 30.0 & 70.0 &  10'292  &                                                 \\
 \hspace{10pt}  secondary\_link               & 48.2 & 5.2 & 50.0 & 30.0 & 60.0 &  280  &                                                 \\
 \hspace{10pt}  tertiary               & 46.4 & 8.5 & 50.0 & 30.0 & 60.0 &  8'486  &                                                 \\
 \hspace{10pt}  tertiary\_link               & 46.4 & 8.3 & 50.0 & 22.0 & 54.0 &  61  &                                                 \\
 \hspace{10pt}  unclassified               & 42.6 & 10.0 & 43.1 & 10.0 & 70.0 &  1'766  &                                                 \\
 \hspace{10pt}  residential               & 30.8 & 4.4 & 30.0 & 20.0 & 50.0 &  64'193  &                                                 \\
 free\_flow\_kph                & 37.4 & 14.1 & 35.3 & 12.0 & 87.5 &  85'074 &                                                 \\
 \hspace{10pt}  motorway               & 92.2 & 15.7 & 87.8 & 68.4 & 119.5 &  279  &                                                 \\
 \hspace{10pt}  motorway\_link               & 78.4 & 20.0 & 80.6 & 31.4 & 119.0 &  469  &                                                 \\
 \hspace{10pt}  trunk               & 90.5 & 25.2 & 99.1 & 47.4 & 118.6 &  75  &                                                 \\
 \hspace{10pt}  trunk\_link               & 83.7 & 26.8 & 82.4 & 18.6 & 118.3 &  77  &                                                 \\
 \hspace{10pt}  primary               & 48.8 & 12.1 & 48.5 & 24.0 & 89.8 &  2'754  &                                                 \\
 \hspace{10pt}  primary\_link               & 44.8 & 18.5 & 41.9 & 13.2 & 96.4 &  150  &                                                 \\
 \hspace{10pt}  secondary               & 46.9 & 11.3 & 47.5 & 21.5 & 84.7 &  10'283  &                                                 \\
 \hspace{10pt}  secondary\_link               & 34.8 & 16.7 & 33.6 & 11.8 & 80.9 &  280  &                                                 \\
 \hspace{10pt}  tertiary               & 44.2 & 10.7 & 43.8 & 21.6 & 83.0 &  8'484  &                                                 \\
 \hspace{10pt}  tertiary\_link               & 32.8 & 15.1 & 33.4 & 11.3 & 61.8 &  59  &                                                 \\
 \hspace{10pt}  unclassified               & 43.9 & 17.0 & 42.8 & 16.0 & 105.4 &  1'689  &                                                 \\
 \hspace{10pt}  residential               & 33.4 & 11.8 & 32.0 & 10.4 & 73.5 &  60'475  &                                                 \\
 free\_flow\_kph-speed\_kph                & 1.5 & 12.6 & 0.6 & -28.4 & 42.2 &  85'074 &                                                 \\
 \hspace{10pt}  motorway               & 6.9 & 16.6 & 6.6 & -48.4 & 37.0 &  279  &                                                 \\
 \hspace{10pt}  motorway\_link               & 16.7 & 20.9 & 19.2 & -29.0 & 59.1 &  469  &                                                 \\
 \hspace{10pt}  trunk               & 3.0 & 13.4 & -0.1 & -35.2 & 34.3 &  75  &                                                 \\
 \hspace{10pt}  trunk\_link               & 31.1 & 26.1 & 29.8 & -33.9 & 65.8 &  77  &                                                 \\
 \hspace{10pt}  primary               & -0.7 & 10.8 & -0.5 & -26.2 & 31.8 &  2'754  &                                                 \\
 \hspace{10pt}  primary\_link               & -4.1 & 19.6 & -7.2 & -52.1 & 39.7 &  150  &                                                 \\
 \hspace{10pt}  secondary               & -1.8 & 11.5 & -1.1 & -29.3 & 33.2 &  10'283  &                                                 \\
 \hspace{10pt}  secondary\_link               & -13.4 & 17.9 & -15.3 & -42.1 & 30.9 &  280  &                                                 \\
 \hspace{10pt}  tertiary               & -2.2 & 11.4 & -2.2 & -28.4 & 31.9 &  8'484  &                                                 \\
 \hspace{10pt}  tertiary\_link               & -13.5 & 19.6 & -14.6 & -37.8 & 40.4 &  59  &                                                 \\
 \hspace{10pt}  unclassified               & 1.3 & 18.1 & -0.3 & -32.1 & 59.7 &  1'689  &                                                 \\
 \hspace{10pt}  residential               & 2.5 & 12.3 & 1.3 & -27.2 & 42.9 &  60'475  &                                                 \\
\bottomrule

        \caption[Key figures Berlin ]{Key figures Berlin for the generated data from 20 randomly sampled days.
        \textbf{num\_edges} number of edges in the street network graph;
        \textbf{num\_nodes} number of nodes in the street network graph;
        \textbf{num\_edges\_per\_cell} number of edges a cell (row,col,heading) has in its intersecting cells;
        \textbf{num\_intersecting\_cells} number of cells (row,col,heading) in an edge's intersecting cells;
        \textbf{node\_degree} number of (unique) neighbor nodes per node;
        \textbf{length\_meters} free flow speed derived from data;
        \textbf{speed\_kph} signalled speed;
        \textbf{free\_flow\_kph} free flow speed derived from data;
        \textbf{free\_flow\_kph-speed\_kph} difference
        }
    \label{tab:key_figures:/iarai/public/t4c/data_pipeline/release20221026_residential_unclassified/2021:Berlin:}
    \end{longtable}
    \end{small}
    
\subsubsection{Segment density map  Berlin (2021)}
\mbox{}
\nopagebreak{}
\begin{figure}[H]
\centering
\includegraphics[width=0.85\textwidth]{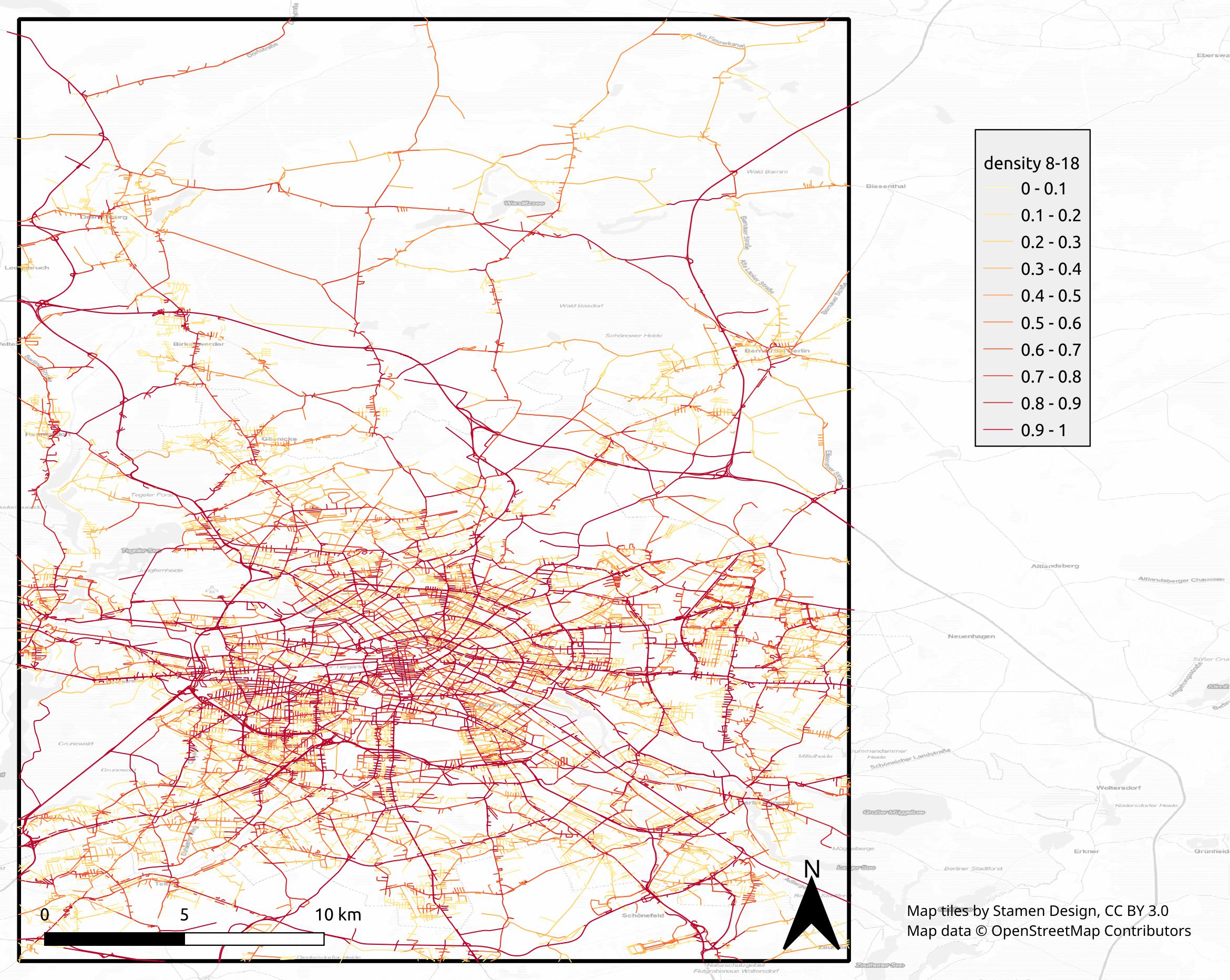}
\caption[Segment-wise density 8am--6pm Berlin]{Segment-wise density 8am--6pm Berlin from 20 randomly sampled days.}
\label{figures/speed_stats/density_8_18_berlin_2021.jpg}
\end{figure}
\clearpage
\subsubsection{Daily density profile  Berlin  (2021) } 
\mbox{}
\nopagebreak{}
\begin{figure}[H]
\centering
\includegraphics[width=0.85\textwidth]{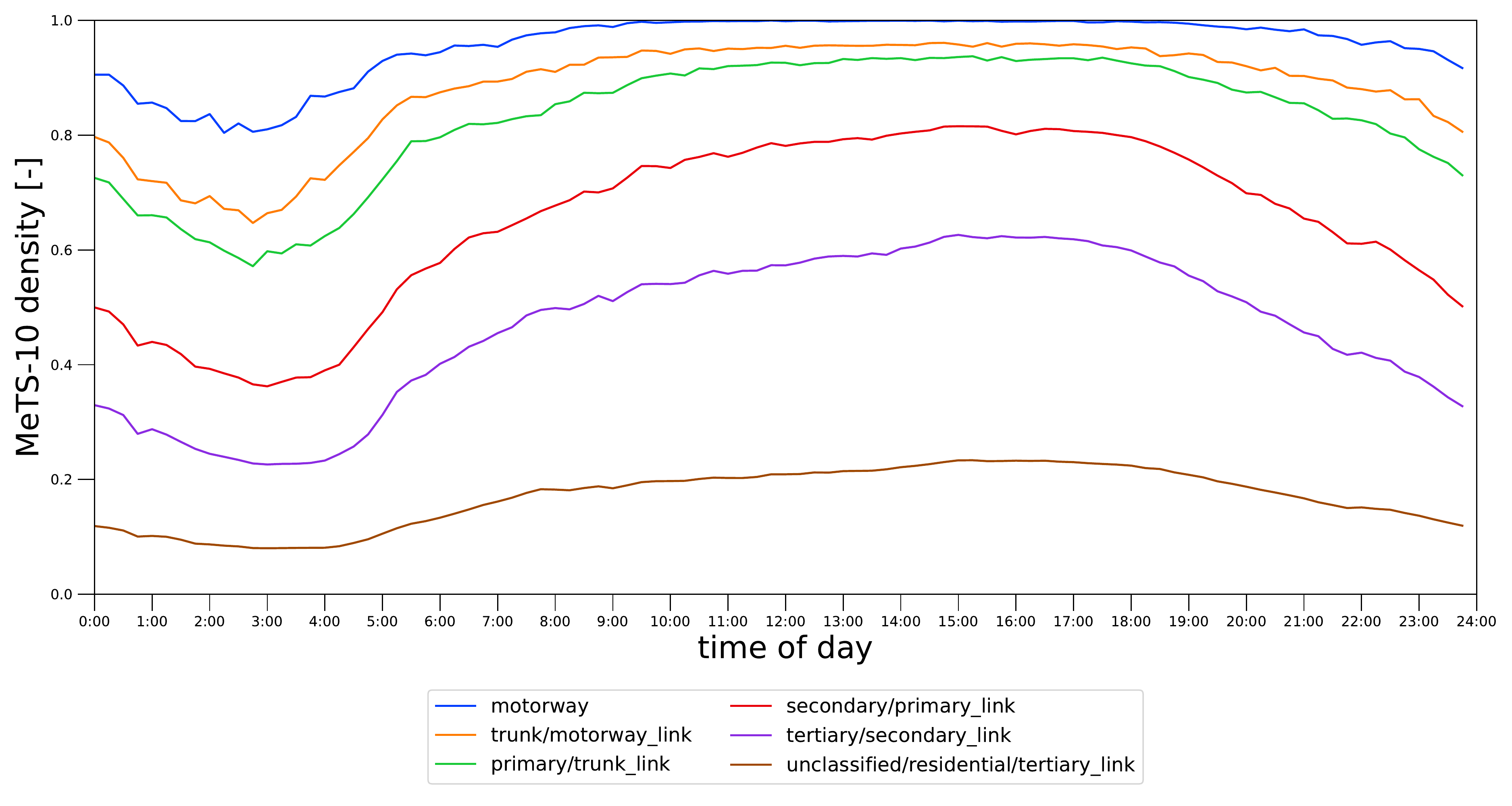}
\caption[Daily density profile Berlin]{Daily density profile for different road types for Berlin . Data from 20 randomly sampled days.}
\label{figures/speed_stats/speed_stats_coverage_berlin_2021_by_highway.pdf}
\end{figure}
\subsubsection{Daily speed profile  Berlin  (2021) }
\mbox{}
\nopagebreak{}
\begin{figure}[H]
\centering
\includegraphics[width=0.85\textwidth]{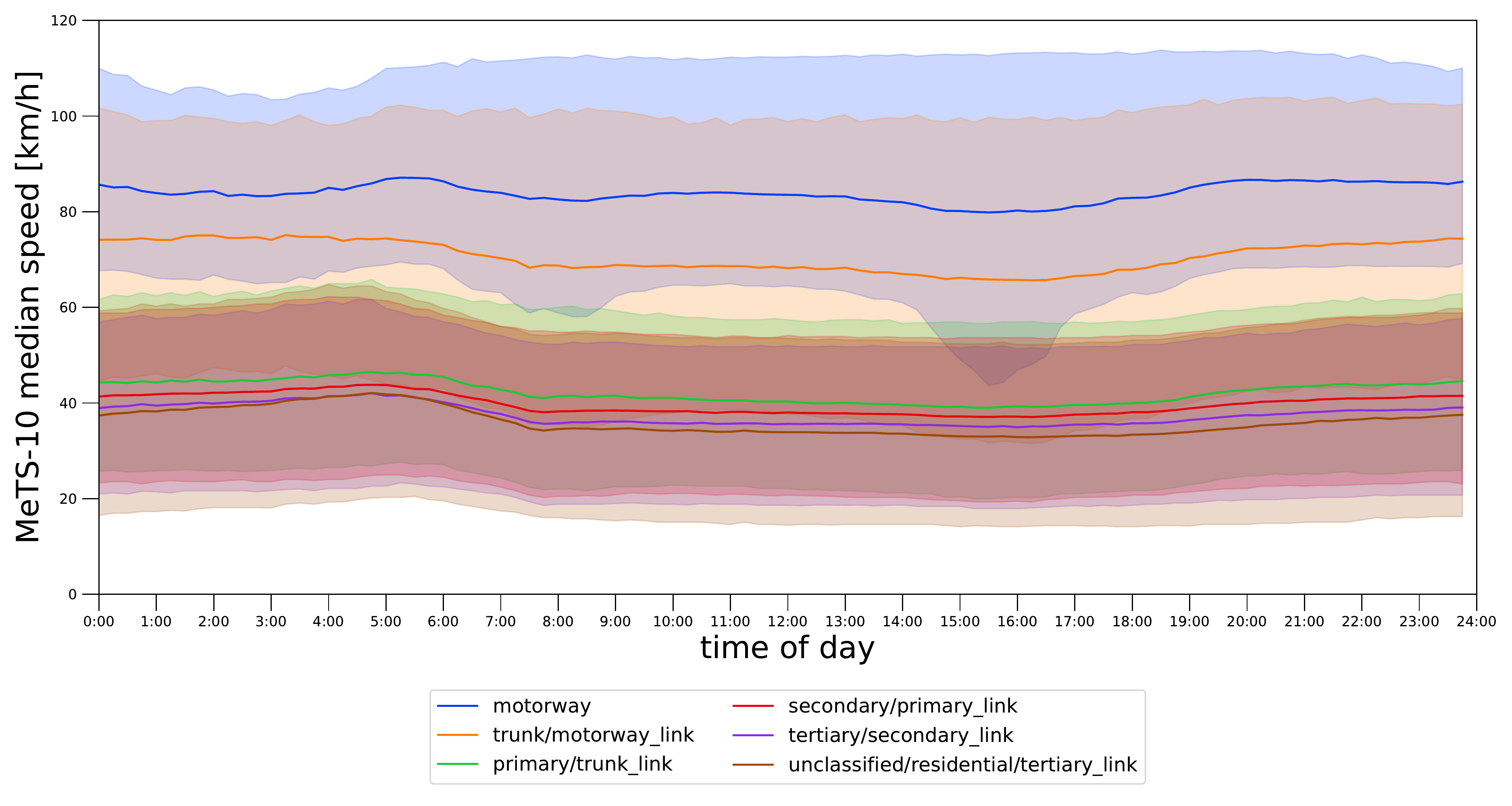}
\caption[Daily median 15 min speeds of all intersecting cells profile Berlin]{Daily median 15 min speeds of all intersecting cells profile for different road types for Berlin . The error hull is the 80\% data interval [10.0--90.0 percentiles] of daily means from 20 randomly sampled days.}
\label{figures/speed_stats/speed_stats_median_speed_kph_berlin_2021_by_highway.pdf}
\end{figure}
\clearpage

\subsection{Key Figures Chicago (2021)}
\subsubsection{Road graph map Chicago (2021)}
\mbox{}
\nopagebreak{}
\begin{figure}[H]
\centering
\includegraphics[width=0.85\textwidth]{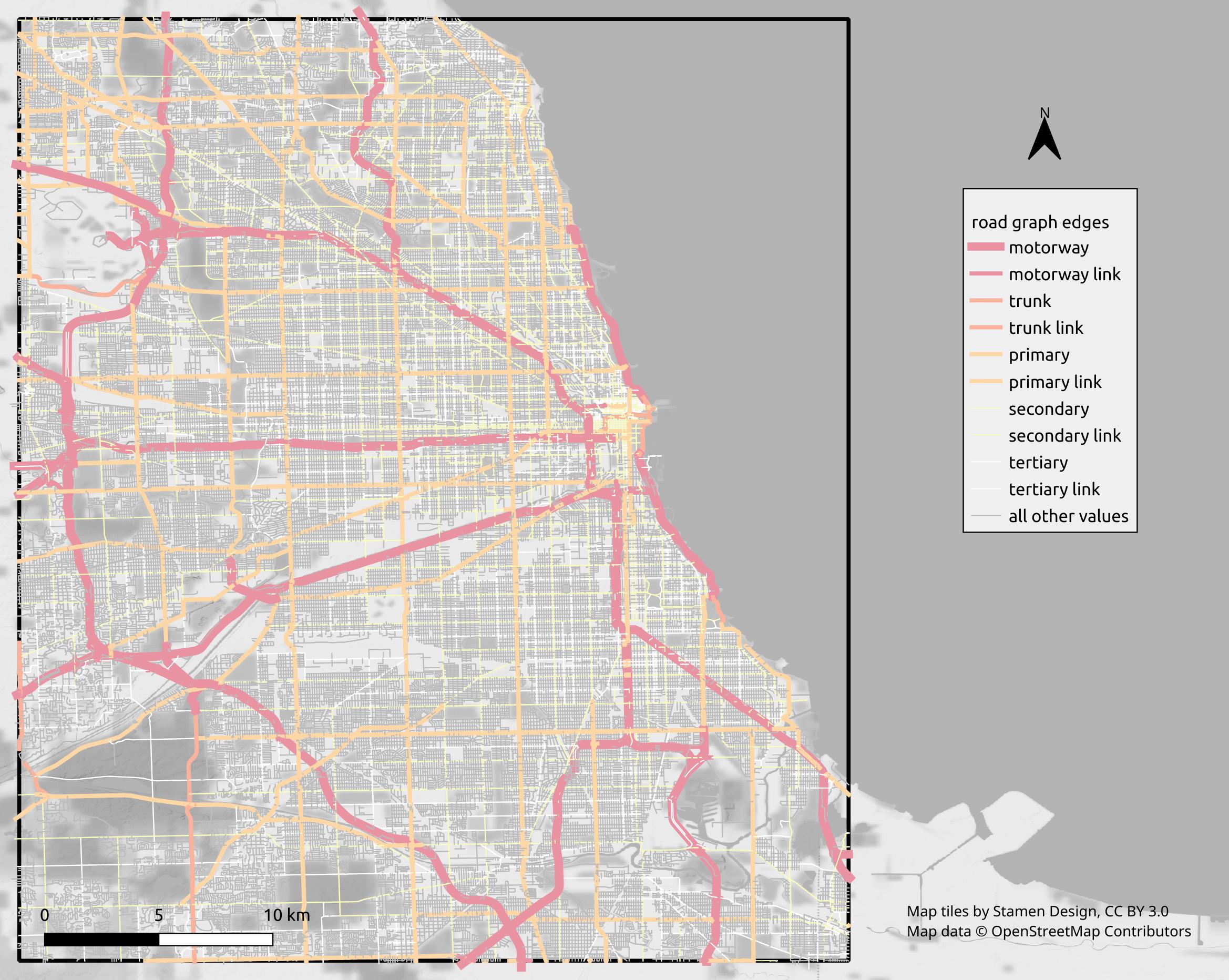}
\caption[Road graph Chicago]{Road graph Chicago, OSM color scheme (2021).}
\label{figures/speed_stats/road_graph_chicago_2021.jpg}
\end{figure}
\subsubsection{Static data  Chicago  (2021) }
\mbox{}\nopagebreak
\begin{small}
\begin{longtable}{p{4cm}rrrrrrrr}
\toprule
Attribute      & {mean} &{std} & {median}  & {q01} & {q99} & {data points} & {sum}  \\
\midrule
 bounding box                &  &  &  &  &  &  -87.945---87.509 / 41.601--42.096 &                                                \\
 num\_edges                &  &  &  &  &  &  187'570 &                                                \\
 \hspace{10pt}  motorway               &  &  &  &  &  &  822 &                                                \\
 \hspace{10pt}  motorway\_link               &  &  &  &  &  &  1178 &                                                \\
 \hspace{10pt}  trunk               &  &  &  &  &  &  207 &                                                \\
 \hspace{10pt}  trunk\_link               &  &  &  &  &  &  73 &                                                \\
 \hspace{10pt}  primary               &  &  &  &  &  &  10439 &                                                \\
 \hspace{10pt}  primary\_link               &  &  &  &  &  &  387 &                                                \\
 \hspace{10pt}  secondary               &  &  &  &  &  &  22812 &                                                \\
 \hspace{10pt}  secondary\_link               &  &  &  &  &  &  208 &                                                \\
 \hspace{10pt}  tertiary               &  &  &  &  &  &  16603 &                                                \\
 \hspace{10pt}  tertiary\_link               &  &  &  &  &  &  71 &                                                \\
 \hspace{10pt}  unclassified               &  &  &  &  &  &  1069 &                                                \\
 \hspace{10pt}  residential               &  &  &  &  &  &  133701 &                                                \\
 num\_nodes                &  &  &  &  &  &  68430 &                                                 \\
 num\_edges\_per\_cell                & 1.1 & 0.3 & 1.0 & 1.0 & 3.0 &  966'289 &                                                 \\
 num\_intersecting\_cells                & 5.5 & 3.3 & 4.0 & 1.0 & 16.0 &  187'570 &                                                 \\
 node\_degree                & 3.2 & 0.9 & 3.0 & 1.0 & 4.0 &  68'430 &                                                 \\
 length\_meters                & 144.6 & 134.2 & 103.9 & 9.7 & 583.7 &  187'570 & 2.7e+07                                                \\
 \hspace{10pt}  motorway               & 778.1 & 872.2 & 559.2 & 22.6 & 4'295.7 &  822  & 6.4e+05                                                \\
 \hspace{10pt}  motorway\_link               & 329.9 & 265.3 & 301.3 & 10.2 & 1'229.2 &  1'178  & 3.9e+05                                                \\
 \hspace{10pt}  trunk               & 370.4 & 431.5 & 262.0 & 7.6 & 1'872.6 &  207  & 7.7e+04                                                \\
 \hspace{10pt}  trunk\_link               & 106.8 & 102.1 & 59.3 & 4.6 & 371.3 &  73  & 7.8e+03                                                \\
 \hspace{10pt}  primary               & 141.9 & 158.8 & 102.1 & 6.9 & 797.0 &  10'439  & 1.5e+06                                                \\
 \hspace{10pt}  primary\_link               & 70.4 & 72.8 & 52.5 & 10.2 & 404.8 &  387  & 2.7e+04                                                \\
 \hspace{10pt}  secondary               & 128.0 & 111.6 & 101.5 & 7.2 & 588.6 &  22'812  & 2.9e+06                                                \\
 \hspace{10pt}  secondary\_link               & 61.2 & 66.2 & 43.8 & 8.7 & 346.1 &  208  & 1.3e+04                                                \\
 \hspace{10pt}  tertiary               & 140.2 & 148.4 & 102.3 & 8.2 & 674.1 &  16'603  & 2.3e+06                                                \\
 \hspace{10pt}  tertiary\_link               & 40.3 & 31.5 & 30.6 & 6.4 & 131.5 &  71  & 2.9e+03                                                \\
 \hspace{10pt}  unclassified               & 229.4 & 297.9 & 126.5 & 4.8 & 1'519.1 &  1'069  & 2.5e+05                                                \\
 \hspace{10pt}  residential               & 142.0 & 95.5 & 106.8 & 10.3 & 452.4 &  133'701  & 1.9e+07                                                \\
 speed\_kph                & 41.0 & 8.0 & 36.8 & 36.8 & 83.8 &  187'570 &                                                 \\
 \hspace{10pt}  motorway               & 87.7 & 6.9 & 88.5 & 72.4 & 112.7 &  822  &                                                 \\
 \hspace{10pt}  motorway\_link               & 83.7 & 3.0 & 83.8 & 83.8 & 83.8 &  1'178  &                                                 \\
 \hspace{10pt}  trunk               & 75.1 & 8.8 & 74.4 & 48.3 & 88.5 &  207  &                                                 \\
 \hspace{10pt}  trunk\_link               & 57.1 & 0.0 & 57.1 & 57.1 & 57.1 &  73  &                                                 \\
 \hspace{10pt}  primary               & 55.6 & 3.4 & 55.8 & 48.3 & 64.4 &  10'439  &                                                 \\
 \hspace{10pt}  primary\_link               & 57.1 & 0.0 & 57.1 & 57.1 & 57.1 &  387  &                                                 \\
 \hspace{10pt}  secondary               & 50.9 & 1.9 & 50.9 & 48.3 & 56.3 &  22'812  &                                                 \\
 \hspace{10pt}  secondary\_link               & 32.2 & 0.0 & 32.2 & 32.2 & 32.2 &  208  &                                                 \\
 \hspace{10pt}  tertiary               & 44.7 & 2.5 & 44.7 & 32.2 & 56.3 &  16'603  &                                                 \\
 \hspace{10pt}  tertiary\_link               & 48.3 & 0.0 & 48.3 & 48.3 & 48.3 &  71  &                                                 \\
 \hspace{10pt}  unclassified               & 56.3 & 2.7 & 56.3 & 56.3 & 56.3 &  1'069  &                                                 \\
 \hspace{10pt}  residential               & 36.8 & 0.6 & 36.8 & 36.8 & 36.8 &  133'701  &                                                 \\
 free\_flow\_kph                & 45.1 & 19.0 & 42.8 & 4.2 & 111.2 &  162'191 &                                                 \\
 \hspace{10pt}  motorway               & 99.4 & 12.9 & 98.1 & 52.3 & 118.6 &  822  &                                                 \\
 \hspace{10pt}  motorway\_link               & 91.4 & 20.9 & 95.5 & 29.6 & 118.9 &  1'178  &                                                 \\
 \hspace{10pt}  trunk               & 68.8 & 17.9 & 70.6 & 32.1 & 98.8 &  207  &                                                 \\
 \hspace{10pt}  trunk\_link               & 67.4 & 17.8 & 76.7 & 26.4 & 91.4 &  73  &                                                 \\
 \hspace{10pt}  primary               & 55.2 & 13.6 & 54.1 & 30.1 & 98.4 &  10'427  &                                                 \\
 \hspace{10pt}  primary\_link               & 52.4 & 17.2 & 51.8 & 12.7 & 96.4 &  384  &                                                 \\
 \hspace{10pt}  secondary               & 49.6 & 14.6 & 46.6 & 25.9 & 99.8 &  22'787  &                                                 \\
 \hspace{10pt}  secondary\_link               & 51.7 & 21.0 & 48.5 & 13.7 & 110.0 &  208  &                                                 \\
 \hspace{10pt}  tertiary               & 48.9 & 16.7 & 45.6 & 21.2 & 111.1 &  16'538  &                                                 \\
 \hspace{10pt}  tertiary\_link               & 48.4 & 20.2 & 46.6 & 19.4 & 110.9 &  69  &                                                 \\
 \hspace{10pt}  unclassified               & 47.6 & 24.9 & 46.5 & 3.3 & 118.0 &  988  &                                                 \\
 \hspace{10pt}  residential               & 41.5 & 18.5 & 39.2 & 2.7 & 107.3 &  108'510  &                                                 \\
 free\_flow\_kph-speed\_kph                & 3.4 & 17.7 & 0.9 & -33.5 & 64.5 &  162'191 &                                                 \\
 \hspace{10pt}  motorway               & 11.8 & 12.8 & 10.6 & -35.5 & 40.4 &  822  &                                                 \\
 \hspace{10pt}  motorway\_link               & 7.7 & 21.0 & 11.7 & -54.2 & 36.2 &  1'178  &                                                 \\
 \hspace{10pt}  trunk               & -6.3 & 14.9 & -3.7 & -44.6 & 18.3 &  207  &                                                 \\
 \hspace{10pt}  trunk\_link               & 10.3 & 17.8 & 19.6 & -30.7 & 34.3 &  73  &                                                 \\
 \hspace{10pt}  primary               & -0.4 & 13.3 & -1.2 & -26.2 & 42.1 &  10'427  &                                                 \\
 \hspace{10pt}  primary\_link               & -4.7 & 17.2 & -5.3 & -44.4 & 39.3 &  384  &                                                 \\
 \hspace{10pt}  secondary               & -1.2 & 14.6 & -4.1 & -25.0 & 48.9 &  22'787  &                                                 \\
 \hspace{10pt}  secondary\_link               & 19.5 & 21.0 & 16.3 & -18.5 & 77.8 &  208  &                                                 \\
 \hspace{10pt}  tertiary               & 4.2 & 16.7 & 1.2 & -23.5 & 66.4 &  16'538  &                                                 \\
 \hspace{10pt}  tertiary\_link               & 0.1 & 20.2 & -1.7 & -28.9 & 62.6 &  69  &                                                 \\
 \hspace{10pt}  unclassified               & -8.7 & 24.8 & -9.7 & -53.0 & 61.7 &  988  &                                                 \\
 \hspace{10pt}  residential               & 4.7 & 18.5 & 2.3 & -34.0 & 70.6 &  108'510  &                                                 \\
\bottomrule

        \caption[Key figures Chicago ]{Key figures Chicago for the generated data from 20 randomly sampled days.
        \textbf{num\_edges} number of edges in the street network graph;
        \textbf{num\_nodes} number of nodes in the street network graph;
        \textbf{num\_edges\_per\_cell} number of edges a cell (row,col,heading) has in its intersecting cells;
        \textbf{num\_intersecting\_cells} number of cells (row,col,heading) in an edge's intersecting cells;
        \textbf{node\_degree} number of (unique) neighbor nodes per node;
        \textbf{length\_meters} free flow speed derived from data;
        \textbf{speed\_kph} signalled speed;
        \textbf{free\_flow\_kph} free flow speed derived from data;
        \textbf{free\_flow\_kph-speed\_kph} difference
        }
    \label{tab:key_figures:/iarai/public/t4c/data_pipeline/release20221026_residential_unclassified/2021:Chicago:}
    \end{longtable}
    \end{small}
    
\subsubsection{Segment density map  Chicago (2021)}
\mbox{}
\nopagebreak{}
\begin{figure}[H]
\centering
\includegraphics[width=0.85\textwidth]{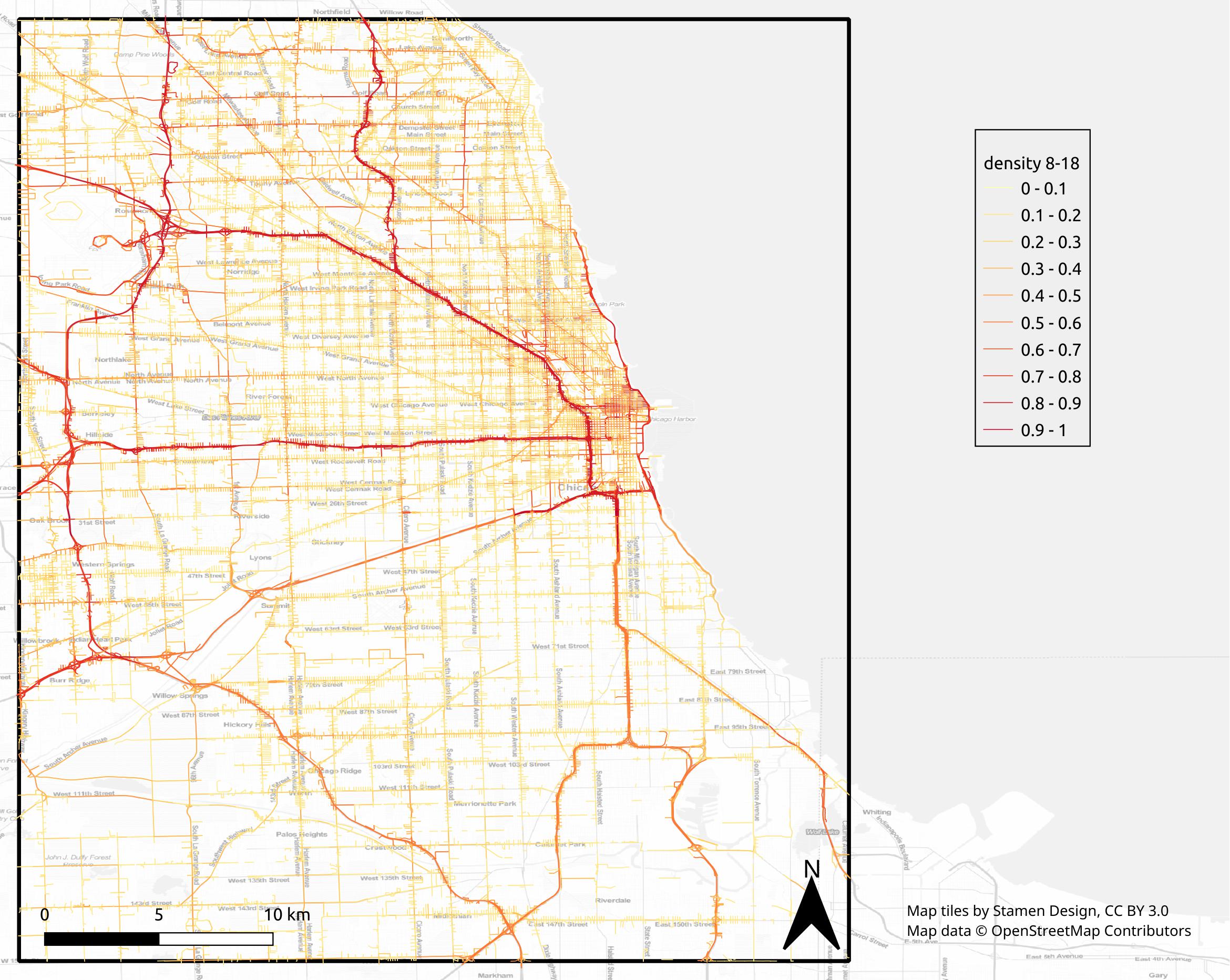}
\caption[Segment-wise density 8am--6pm Chicago]{Segment-wise density 8am--6pm Chicago from 20 randomly sampled days.}
\label{figures/speed_stats/density_8_18_chicago_2021.jpg}
\end{figure}
\clearpage
\subsubsection{Daily density profile  Chicago  (2021) } 
\mbox{}
\nopagebreak{}
\begin{figure}[H]
\centering
\includegraphics[width=0.85\textwidth]{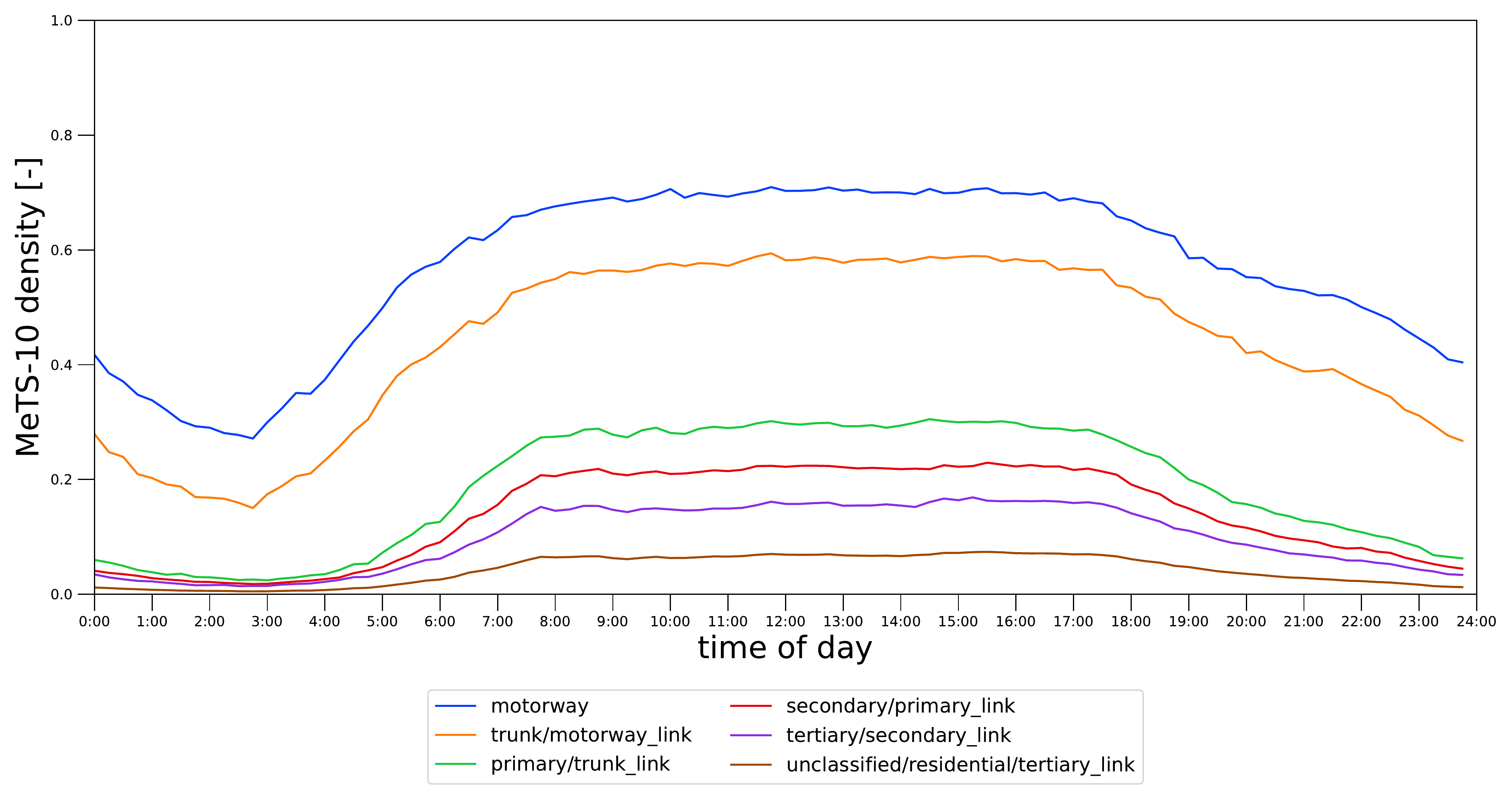}
\caption[Daily density profile Chicago]{Daily density profile for different road types for Chicago . Data from 20 randomly sampled days.}
\label{figures/speed_stats/speed_stats_coverage_chicago_2021_by_highway.pdf}
\end{figure}
\subsubsection{Daily speed profile  Chicago  (2021) }
\mbox{}
\nopagebreak{}
\begin{figure}[H]
\centering
\includegraphics[width=0.85\textwidth]{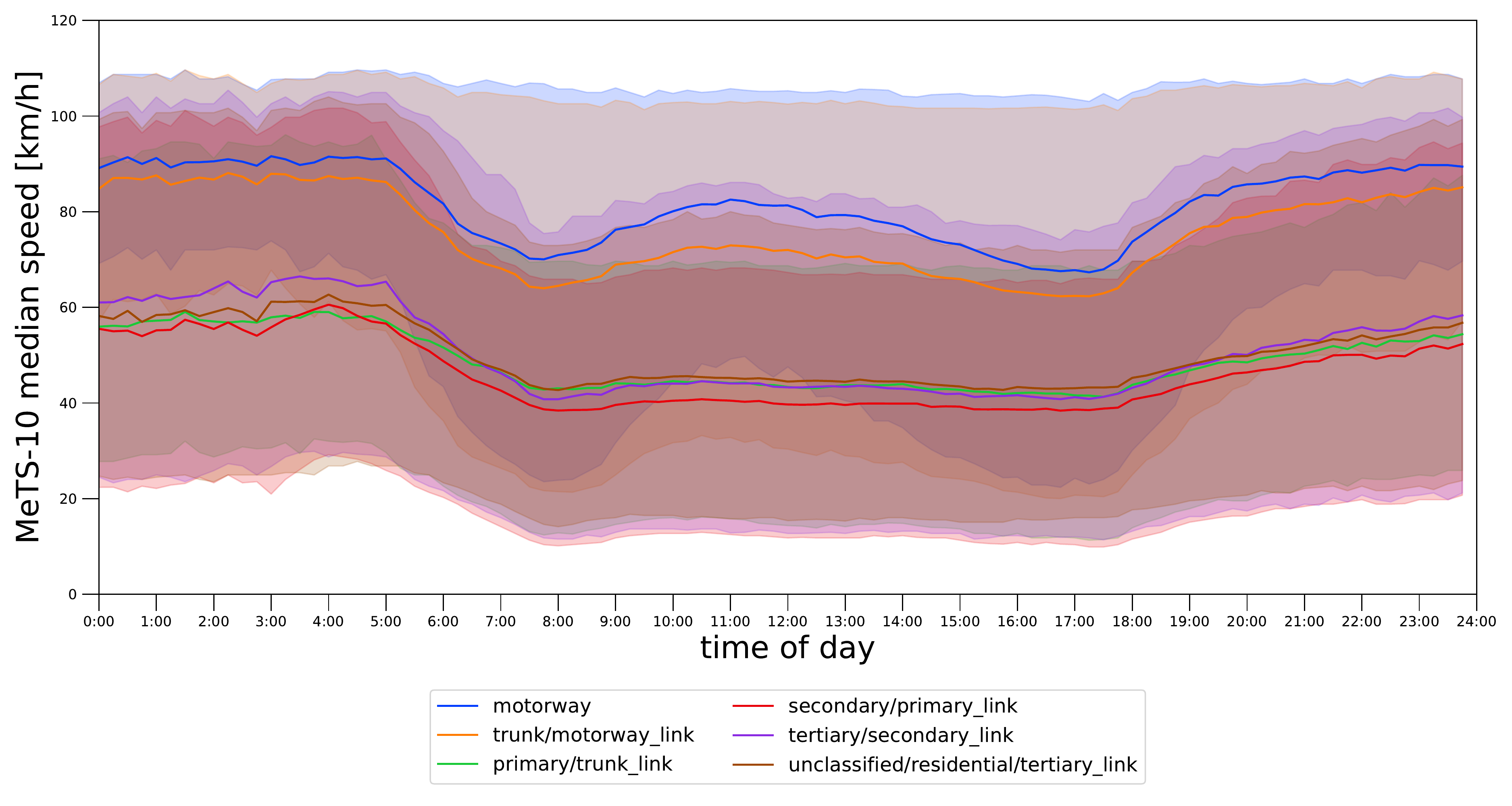}
\caption[Daily median 15 min speeds of all intersecting cells profile Chicago]{Daily median 15 min speeds of all intersecting cells profile for different road types for Chicago . The error hull is the 80\% data interval [10.0--90.0 percentiles] of daily means from 20 randomly sampled days.}
\label{figures/speed_stats/speed_stats_median_speed_kph_chicago_2021_by_highway.pdf}
\end{figure}
\clearpage

\subsection{Key Figures Istanbul (2021)}
\subsubsection{Road graph map Istanbul (2021)}
\mbox{}
\nopagebreak{}
\begin{figure}[H]
\centering
\includegraphics[width=0.85\textwidth]{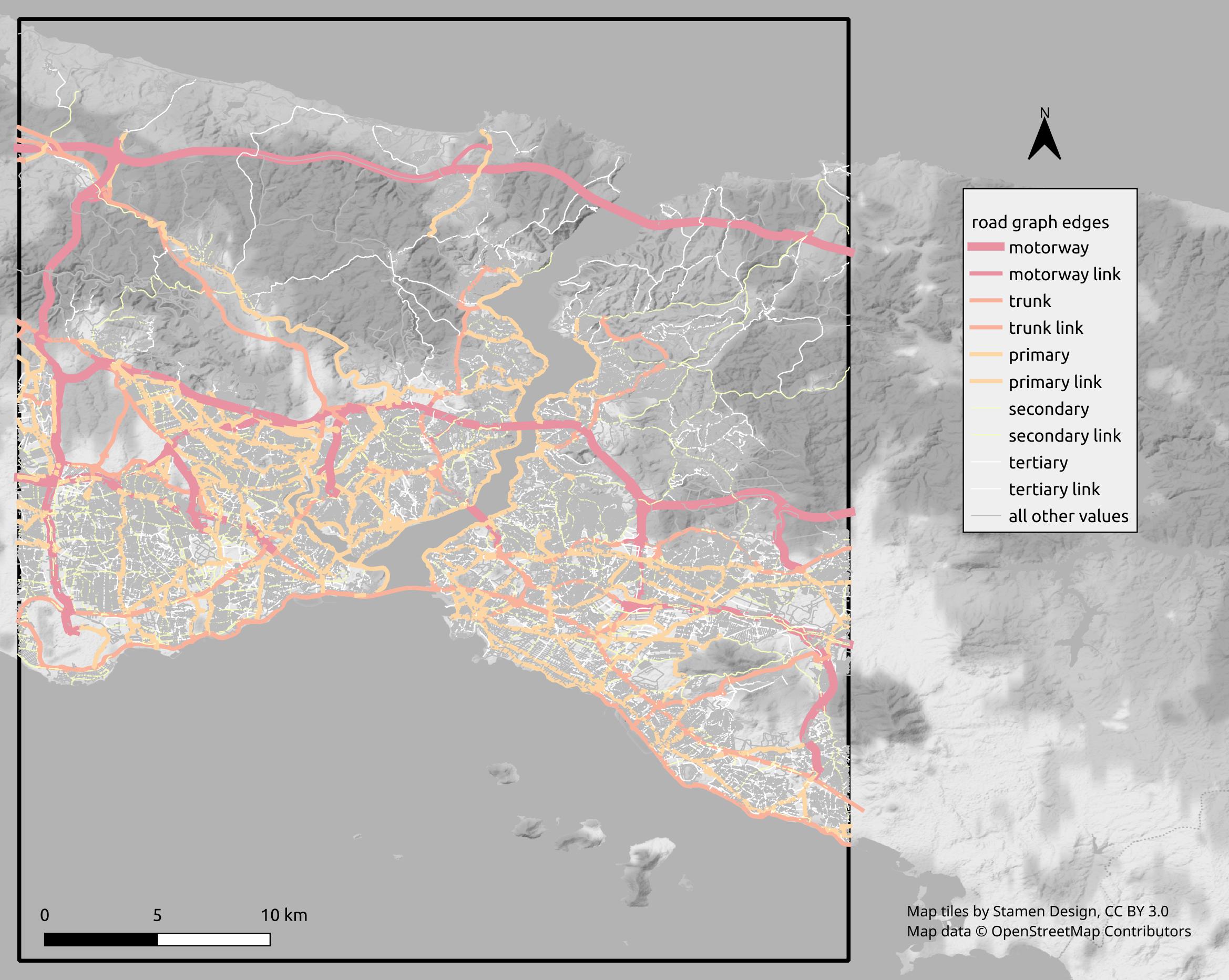}
\caption[Road graph Istanbul]{Road graph Istanbul, OSM color scheme (2021).}
\label{figures/speed_stats/road_graph_istanbul_2021.jpg}
\end{figure}
\subsubsection{Static data  Istanbul  (2021) }
\mbox{}\nopagebreak
\begin{small}
\begin{longtable}{p{4cm}rrrrrrrr}
\toprule
Attribute      & {mean} &{std} & {median}  & {q01} & {q99} & {data points} & {sum}  \\
\midrule
 bounding box                &  &  &  &  &  &  28.794--29.23 / 40.81--41.305 &                                                \\
 num\_edges                &  &  &  &  &  &  270'109 &                                                \\
 \hspace{10pt}  motorway               &  &  &  &  &  &  338 &                                                \\
 \hspace{10pt}  motorway\_link               &  &  &  &  &  &  699 &                                                \\
 \hspace{10pt}  trunk               &  &  &  &  &  &  1156 &                                                \\
 \hspace{10pt}  trunk\_link               &  &  &  &  &  &  1529 &                                                \\
 \hspace{10pt}  primary               &  &  &  &  &  &  9242 &                                                \\
 \hspace{10pt}  primary\_link               &  &  &  &  &  &  2430 &                                                \\
 \hspace{10pt}  secondary               &  &  &  &  &  &  17014 &                                                \\
 \hspace{10pt}  secondary\_link               &  &  &  &  &  &  1403 &                                                \\
 \hspace{10pt}  tertiary               &  &  &  &  &  &  30221 &                                                \\
 \hspace{10pt}  tertiary\_link               &  &  &  &  &  &  625 &                                                \\
 \hspace{10pt}  unclassified               &  &  &  &  &  &  5422 &                                                \\
 \hspace{10pt}  residential               &  &  &  &  &  &  200030 &                                                \\
 num\_nodes                &  &  &  &  &  &  102754 &                                                 \\
 num\_edges\_per\_cell                & 1.1 & 0.6 & 1.0 & 1.0 & 4.0 &  734'722 &                                                 \\
 num\_intersecting\_cells                & 3.1 & 2.9 & 2.0 & 1.0 & 12.0 &  270'109 &                                                 \\
 node\_degree                & 3.0 & 0.8 & 3.0 & 1.0 & 4.0 &  102'754 &                                                 \\
 length\_meters                & 81.9 & 135.0 & 53.3 & 5.9 & 479.4 &  270'109 & 2.2e+07                                                \\
 \hspace{10pt}  motorway               & 1'066.1 & 1'693.6 & 647.9 & 42.4 & 8'557.5 &  338  & 3.6e+05                                                \\
 \hspace{10pt}  motorway\_link               & 361.1 & 312.1 & 277.1 & 11.8 & 1'505.5 &  699  & 2.5e+05                                                \\
 \hspace{10pt}  trunk               & 394.2 & 522.6 & 216.3 & 9.5 & 2'387.7 &  1'156  & 4.6e+05                                                \\
 \hspace{10pt}  trunk\_link               & 162.4 & 148.9 & 126.8 & 7.7 & 677.8 &  1'529  & 2.5e+05                                                \\
 \hspace{10pt}  primary               & 104.8 & 160.3 & 57.2 & 5.6 & 723.8 &  9'242  & 9.7e+05                                                \\
 \hspace{10pt}  primary\_link               & 65.0 & 78.0 & 37.4 & 5.7 & 378.0 &  2'430  & 1.6e+05                                                \\
 \hspace{10pt}  secondary               & 74.6 & 128.1 & 42.2 & 5.0 & 507.6 &  17'014  & 1.3e+06                                                \\
 \hspace{10pt}  secondary\_link               & 48.6 & 75.7 & 28.1 & 4.6 & 308.9 &  1'403  & 6.8e+04                                                \\
 \hspace{10pt}  tertiary               & 73.5 & 153.4 & 43.8 & 4.8 & 515.8 &  30'221  & 2.2e+06                                                \\
 \hspace{10pt}  tertiary\_link               & 30.4 & 32.4 & 20.1 & 3.9 & 170.2 &  625  & 1.9e+04                                                \\
 \hspace{10pt}  unclassified               & 217.9 & 370.2 & 104.7 & 7.1 & 1'861.4 &  5'422  & 1.2e+06                                                \\
 \hspace{10pt}  residential               & 74.6 & 67.7 & 55.0 & 6.5 & 317.6 &  200'030  & 1.5e+07                                                \\
 speed\_kph                & 35.4 & 6.2 & 33.2 & 22.3 & 50.0 &  270'109 &                                                 \\
 \hspace{10pt}  motorway               & 107.1 & 9.9 & 107.3 & 80.0 & 120.0 &  338  &                                                 \\
 \hspace{10pt}  motorway\_link               & 47.5 & 8.8 & 46.6 & 30.0 & 90.0 &  699  &                                                 \\
 \hspace{10pt}  trunk               & 75.3 & 10.8 & 76.4 & 30.0 & 100.0 &  1'156  &                                                 \\
 \hspace{10pt}  trunk\_link               & 42.3 & 4.8 & 42.4 & 30.0 & 50.0 &  1'529  &                                                 \\
 \hspace{10pt}  primary               & 44.9 & 4.5 & 44.9 & 30.0 & 50.0 &  9'242  &                                                 \\
 \hspace{10pt}  primary\_link               & 32.4 & 2.8 & 32.4 & 30.0 & 32.4 &  2'430  &                                                 \\
 \hspace{10pt}  secondary               & 38.1 & 2.9 & 38.1 & 30.0 & 50.0 &  17'014  &                                                 \\
 \hspace{10pt}  secondary\_link               & 30.5 & 1.8 & 30.6 & 20.0 & 30.6 &  1'403  &                                                 \\
 \hspace{10pt}  tertiary               & 45.1 & 2.2 & 45.2 & 33.2 & 50.0 &  30'221  &                                                 \\
 \hspace{10pt}  tertiary\_link               & 31.6 & 1.4 & 31.6 & 30.0 & 31.6 &  625  &                                                 \\
 \hspace{10pt}  unclassified               & 22.4 & 1.2 & 22.3 & 22.3 & 33.2 &  5'422  &                                                 \\
 \hspace{10pt}  residential               & 33.2 & 1.1 & 33.2 & 33.2 & 33.2 &  200'030  &                                                 \\
 free\_flow\_kph                & 24.7 & 12.4 & 20.9 & 10.4 & 75.8 &  268'548 &                                                 \\
 \hspace{10pt}  motorway               & 81.4 & 14.2 & 83.9 & 18.9 & 106.3 &  338  &                                                 \\
 \hspace{10pt}  motorway\_link               & 69.4 & 16.7 & 71.7 & 24.4 & 98.1 &  698  &                                                 \\
 \hspace{10pt}  trunk               & 61.7 & 14.7 & 64.5 & 19.6 & 91.1 &  1'156  &                                                 \\
 \hspace{10pt}  trunk\_link               & 53.9 & 18.2 & 56.5 & 10.8 & 87.4 &  1'529  &                                                 \\
 \hspace{10pt}  primary               & 36.4 & 12.4 & 34.4 & 16.5 & 75.9 &  9'242  &                                                 \\
 \hspace{10pt}  primary\_link               & 33.6 & 13.9 & 31.5 & 11.3 & 77.1 &  2'430  &                                                 \\
 \hspace{10pt}  secondary               & 31.2 & 14.4 & 26.8 & 14.3 & 82.3 &  17'014  &                                                 \\
 \hspace{10pt}  secondary\_link               & 32.0 & 16.9 & 26.8 & 10.7 & 82.5 &  1'403  &                                                 \\
 \hspace{10pt}  tertiary               & 26.1 & 10.2 & 24.0 & 12.7 & 67.8 &  30'195  &                                                 \\
 \hspace{10pt}  tertiary\_link               & 24.7 & 10.7 & 22.0 & 10.1 & 76.5 &  625  &                                                 \\
 \hspace{10pt}  unclassified               & 29.7 & 16.4 & 25.6 & 8.0 & 89.1 &  4'735  &                                                 \\
 \hspace{10pt}  residential               & 22.5 & 10.4 & 19.8 & 9.9 & 68.2 &  199'183  &                                                 \\
 free\_flow\_kph-speed\_kph                & -10.7 & 12.1 & -13.7 & -29.5 & 37.7 &  268'548 &                                                 \\
 \hspace{10pt}  motorway               & -25.8 & 15.6 & -23.3 & -88.4 & 2.1 &  338  &                                                 \\
 \hspace{10pt}  motorway\_link               & 21.9 & 17.6 & 24.5 & -26.7 & 53.6 &  698  &                                                 \\
 \hspace{10pt}  trunk               & -13.6 & 15.8 & -11.4 & -57.6 & 25.1 &  1'156  &                                                 \\
 \hspace{10pt}  trunk\_link               & 11.6 & 18.0 & 14.5 & -31.6 & 44.6 &  1'529  &                                                 \\
 \hspace{10pt}  primary               & -8.5 & 12.6 & -10.3 & -30.3 & 29.7 &  9'242  &                                                 \\
 \hspace{10pt}  primary\_link               & 1.2 & 14.1 & -0.6 & -22.2 & 43.6 &  2'430  &                                                 \\
 \hspace{10pt}  secondary               & -6.9 & 14.6 & -11.3 & -25.2 & 44.4 &  17'014  &                                                 \\
 \hspace{10pt}  secondary\_link               & 1.5 & 16.9 & -3.5 & -19.9 & 51.9 &  1'403  &                                                 \\
 \hspace{10pt}  tertiary               & -18.9 & 10.4 & -21.0 & -33.0 & 22.1 &  30'195  &                                                 \\
 \hspace{10pt}  tertiary\_link               & -6.9 & 10.8 & -9.6 & -22.5 & 44.9 &  625  &                                                 \\
 \hspace{10pt}  unclassified               & 7.3 & 16.5 & 3.3 & -15.9 & 66.8 &  4'735  &                                                 \\
 \hspace{10pt}  residential               & -10.8 & 10.4 & -13.4 & -23.3 & 35.0 &  199'183  &                                                 \\
\bottomrule

        \caption[Key figures Istanbul ]{Key figures Istanbul for the generated data from 20 randomly sampled days.
        \textbf{num\_edges} number of edges in the street network graph;
        \textbf{num\_nodes} number of nodes in the street network graph;
        \textbf{num\_edges\_per\_cell} number of edges a cell (row,col,heading) has in its intersecting cells;
        \textbf{num\_intersecting\_cells} number of cells (row,col,heading) in an edge's intersecting cells;
        \textbf{node\_degree} number of (unique) neighbor nodes per node;
        \textbf{length\_meters} free flow speed derived from data;
        \textbf{speed\_kph} signalled speed;
        \textbf{free\_flow\_kph} free flow speed derived from data;
        \textbf{free\_flow\_kph-speed\_kph} difference
        }
    \label{tab:key_figures:/iarai/public/t4c/data_pipeline/release20221026_residential_unclassified/2021:Istanbul:}
    \end{longtable}
    \end{small}
    
\subsubsection{Segment density map  Istanbul (2021)}
\mbox{}
\nopagebreak{}
\begin{figure}[H]
\centering
\includegraphics[width=0.85\textwidth]{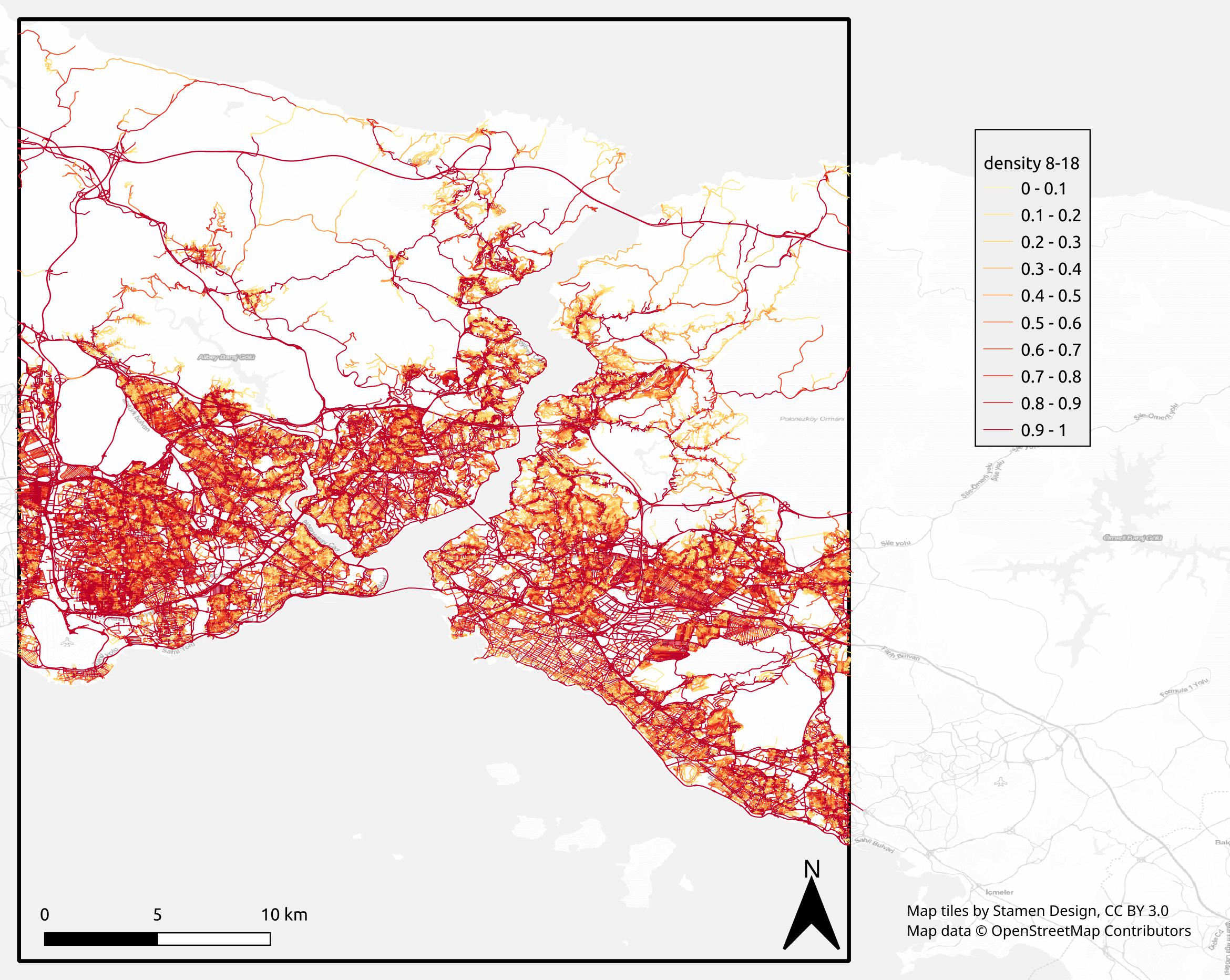}
\caption[Segment-wise density 8am--6pm Istanbul]{Segment-wise density 8am--6pm Istanbul from 20 randomly sampled days.}
\label{figures/speed_stats/density_8_18_istanbul_2021.jpg}
\end{figure}
\clearpage
\subsubsection{Daily density profile  Istanbul  (2021) } 
\mbox{}
\nopagebreak{}
\begin{figure}[H]
\centering
\includegraphics[width=0.85\textwidth]{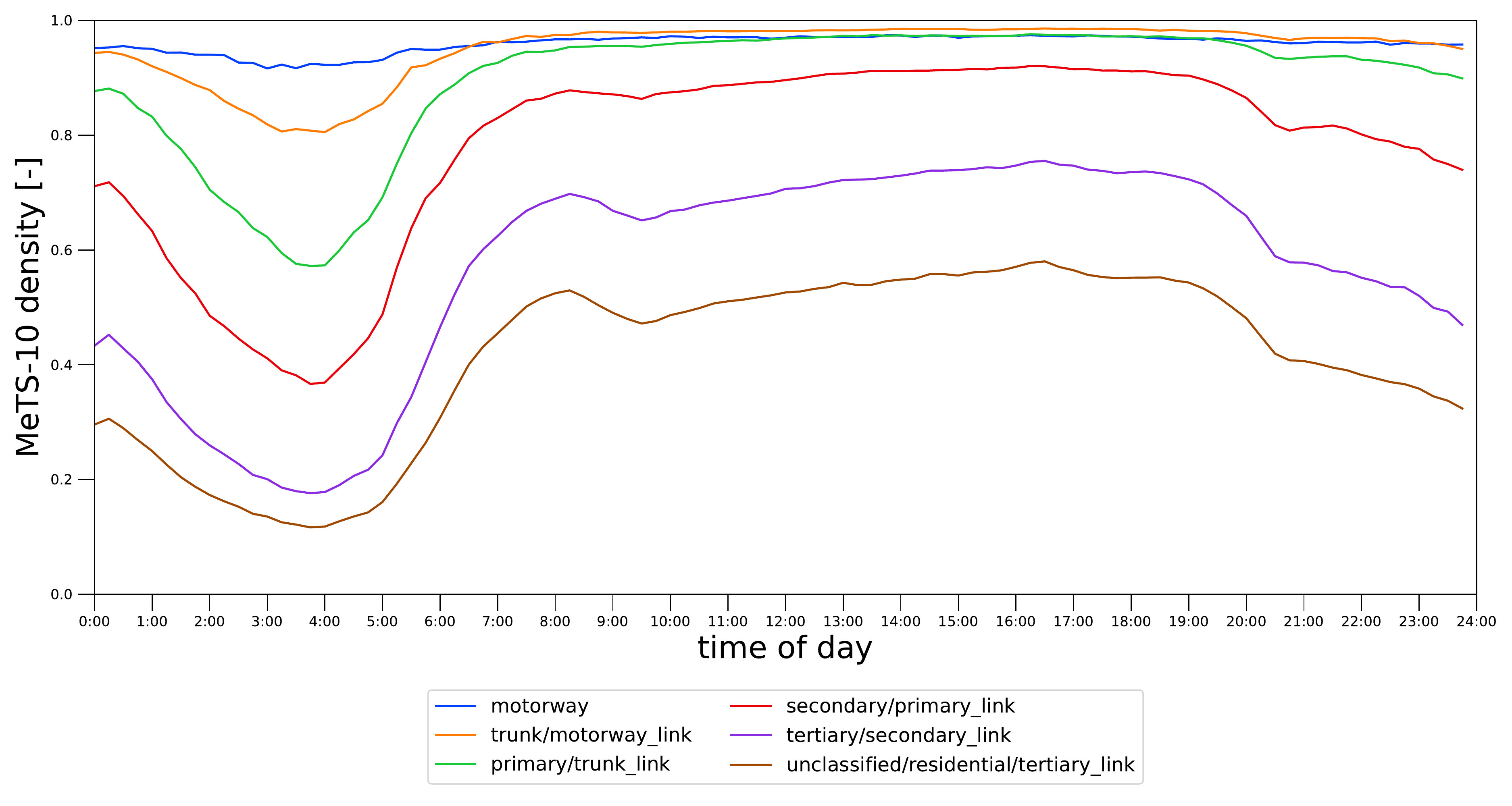}
\caption[Daily density profile Istanbul]{Daily density profile for different road types for Istanbul . Data from 20 randomly sampled days.}
\label{figures/speed_stats/speed_stats_coverage_istanbul_2021_by_highway.pdf}
\end{figure}
\subsubsection{Daily speed profile  Istanbul  (2021) }
\mbox{}
\nopagebreak{}
\begin{figure}[H]
\centering
\includegraphics[width=0.85\textwidth]{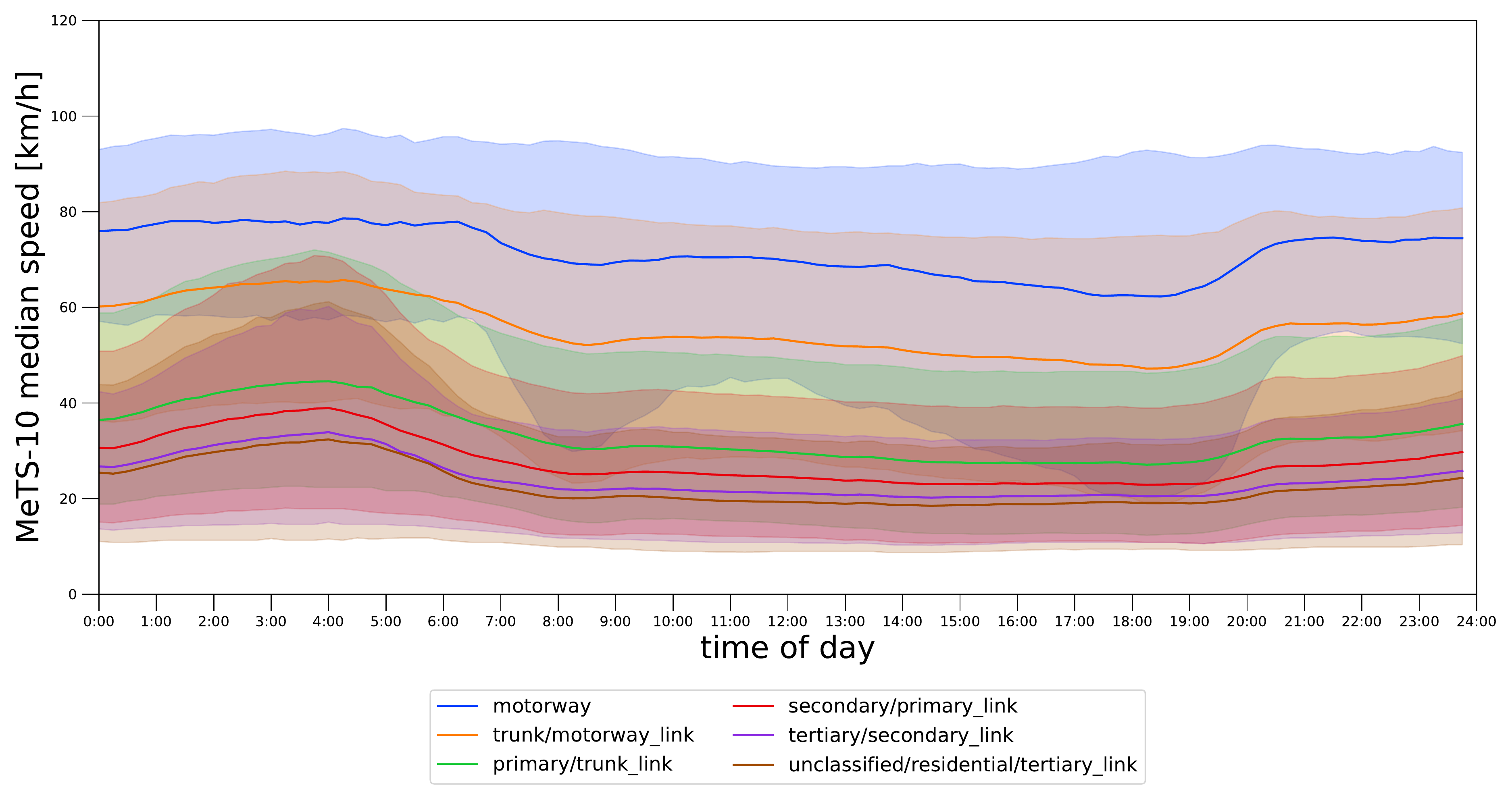}
\caption[Daily median 15 min speeds of all intersecting cells profile Istanbul]{Daily median 15 min speeds of all intersecting cells profile for different road types for Istanbul . The error hull is the 80\% data interval [10.0--90.0 percentiles] of daily means from 20 randomly sampled days.}
\label{figures/speed_stats/speed_stats_median_speed_kph_istanbul_2021_by_highway.pdf}
\end{figure}
\clearpage

\subsection{Key Figures Melbourne (2021)}
\subsubsection{Road graph map Melbourne (2021)}
\mbox{}
\nopagebreak{}
\begin{figure}[H]
\centering
\includegraphics[width=0.85\textwidth]{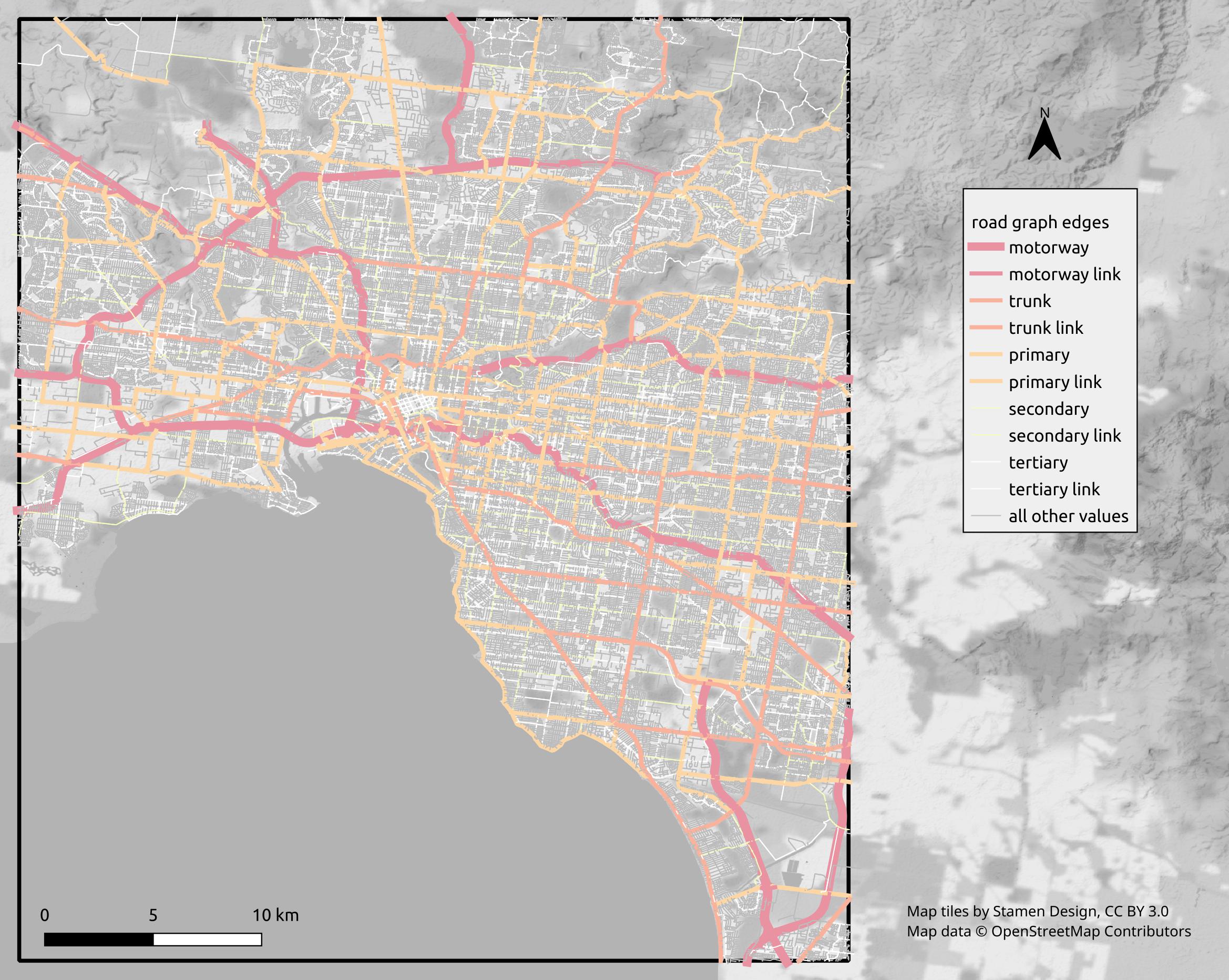}
\caption[Road graph Melbourne]{Road graph Melbourne, OSM color scheme (2021).}
\label{figures/speed_stats/road_graph_melbourne_2021.jpg}
\end{figure}
\subsubsection{Static data  Melbourne  (2021) }
\mbox{}\nopagebreak
\begin{small}
\begin{longtable}{p{4cm}rrrrrrrr}
\toprule
Attribute      & {mean} &{std} & {median}  & {q01} & {q99} & {data points} & {sum}  \\
\midrule
 bounding box                &  &  &  &  &  &  144.757--145.193 / -38.106---37.611 &                                                \\
 num\_edges                &  &  &  &  &  &  230'654 &                                                \\
 \hspace{10pt}  motorway               &  &  &  &  &  &  354 &                                                \\
 \hspace{10pt}  motorway\_link               &  &  &  &  &  &  891 &                                                \\
 \hspace{10pt}  trunk               &  &  &  &  &  &  5383 &                                                \\
 \hspace{10pt}  trunk\_link               &  &  &  &  &  &  766 &                                                \\
 \hspace{10pt}  primary               &  &  &  &  &  &  13917 &                                                \\
 \hspace{10pt}  primary\_link               &  &  &  &  &  &  1574 &                                                \\
 \hspace{10pt}  secondary               &  &  &  &  &  &  10342 &                                                \\
 \hspace{10pt}  secondary\_link               &  &  &  &  &  &  394 &                                                \\
 \hspace{10pt}  tertiary               &  &  &  &  &  &  30552 &                                                \\
 \hspace{10pt}  tertiary\_link               &  &  &  &  &  &  1007 &                                                \\
 \hspace{10pt}  unclassified               &  &  &  &  &  &  7301 &                                                \\
 \hspace{10pt}  residential               &  &  &  &  &  &  158173 &                                                \\
 num\_nodes                &  &  &  &  &  &  103062 &                                                 \\
 num\_edges\_per\_cell                & 1.1 & 0.4 & 1.0 & 1.0 & 3.0 &  891'475 &                                                 \\
 num\_intersecting\_cells                & 4.2 & 3.1 & 4.0 & 1.0 & 14.0 &  230'654 &                                                 \\
 node\_degree                & 2.8 & 0.8 & 3.0 & 1.0 & 4.0 &  103'062 &                                                 \\
 length\_meters                & 105.3 & 123.1 & 81.5 & 2.7 & 497.8 &  230'654 & 2.4e+07                                                \\
 \hspace{10pt}  motorway               & 1'102.6 & 774.5 & 976.5 & 26.1 & 4'090.9 &  354  & 3.9e+05                                                \\
 \hspace{10pt}  motorway\_link               & 261.5 & 319.2 & 102.2 & 10.1 & 1'525.0 &  891  & 2.3e+05                                                \\
 \hspace{10pt}  trunk               & 105.4 & 139.8 & 68.1 & 3.9 & 657.1 &  5'383  & 5.7e+05                                                \\
 \hspace{10pt}  trunk\_link               & 36.2 & 39.8 & 22.9 & 7.8 & 182.2 &  766  & 2.8e+04                                                \\
 \hspace{10pt}  primary               & 99.4 & 115.9 & 69.4 & 3.0 & 565.5 &  13'917  & 1.4e+06                                                \\
 \hspace{10pt}  primary\_link               & 30.5 & 34.3 & 17.0 & 6.2 & 146.5 &  1'574  & 4.8e+04                                                \\
 \hspace{10pt}  secondary               & 86.5 & 104.6 & 64.2 & 2.3 & 457.6 &  10'342  & 9.0e+05                                                \\
 \hspace{10pt}  secondary\_link               & 40.3 & 28.6 & 37.1 & 5.1 & 124.3 &  394  & 1.6e+04                                                \\
 \hspace{10pt}  tertiary               & 78.5 & 111.2 & 54.4 & 1.6 & 421.8 &  30'552  & 2.4e+06                                                \\
 \hspace{10pt}  tertiary\_link               & 23.2 & 25.0 & 12.7 & 2.0 & 117.5 &  1'007  & 2.3e+04                                                \\
 \hspace{10pt}  unclassified               & 153.4 & 235.0 & 89.3 & 3.7 & 1'066.3 &  7'301  & 1.1e+06                                                \\
 \hspace{10pt}  residential               & 108.6 & 99.5 & 87.7 & 3.1 & 444.9 &  158'173  & 1.7e+07                                                \\
 speed\_kph                & 51.1 & 6.8 & 48.7 & 40.0 & 80.0 &  230'654 &                                                 \\
 \hspace{10pt}  motorway               & 93.6 & 9.5 & 100.0 & 80.0 & 100.0 &  354  &                                                 \\
 \hspace{10pt}  motorway\_link               & 77.2 & 11.0 & 78.9 & 50.0 & 100.0 &  891  &                                                 \\
 \hspace{10pt}  trunk               & 68.6 & 9.5 & 70.0 & 40.0 & 80.0 &  5'383  &                                                 \\
 \hspace{10pt}  trunk\_link               & 65.2 & 5.0 & 65.4 & 50.0 & 80.0 &  766  &                                                 \\
 \hspace{10pt}  primary               & 62.5 & 8.3 & 60.0 & 40.0 & 80.0 &  13'917  &                                                 \\
 \hspace{10pt}  primary\_link               & 57.1 & 3.3 & 57.0 & 40.0 & 70.0 &  1'574  &                                                 \\
 \hspace{10pt}  secondary               & 58.8 & 5.9 & 60.0 & 40.0 & 80.0 &  10'342  &                                                 \\
 \hspace{10pt}  secondary\_link               & 59.4 & 2.5 & 59.6 & 50.0 & 70.0 &  394  &                                                 \\
 \hspace{10pt}  tertiary               & 51.4 & 5.8 & 51.4 & 40.0 & 70.0 &  30'552  &                                                 \\
 \hspace{10pt}  tertiary\_link               & 54.4 & 3.0 & 54.6 & 40.0 & 60.0 &  1'007  &                                                 \\
 \hspace{10pt}  unclassified               & 49.1 & 4.8 & 49.0 & 20.0 & 60.0 &  7'301  &                                                 \\
 \hspace{10pt}  residential               & 48.7 & 2.4 & 48.7 & 40.0 & 50.0 &  158'173  &                                                 \\
 free\_flow\_kph                & 40.6 & 18.7 & 39.5 & 0.0 & 96.7 &  187'487 &                                                 \\
 \hspace{10pt}  motorway               & 91.4 & 10.0 & 96.0 & 65.0 & 100.7 &  348  &                                                 \\
 \hspace{10pt}  motorway\_link               & 76.9 & 22.7 & 82.6 & 24.2 & 99.8 &  879  &                                                 \\
 \hspace{10pt}  trunk               & 56.4 & 13.4 & 56.9 & 26.8 & 80.8 &  5'380  &                                                 \\
 \hspace{10pt}  trunk\_link               & 47.2 & 19.8 & 49.9 & 4.3 & 85.6 &  747  &                                                 \\
 \hspace{10pt}  primary               & 51.0 & 13.1 & 52.7 & 23.1 & 82.8 &  13'913  &                                                 \\
 \hspace{10pt}  primary\_link               & 46.4 & 19.6 & 47.4 & 3.8 & 94.4 &  1'516  &                                                 \\
 \hspace{10pt}  secondary               & 49.5 & 13.3 & 51.1 & 20.8 & 94.6 &  10'301  &                                                 \\
 \hspace{10pt}  secondary\_link               & 43.1 & 15.7 & 41.2 & 7.7 & 93.9 &  387  &                                                 \\
 \hspace{10pt}  tertiary               & 42.1 & 13.0 & 41.6 & 16.9 & 84.7 &  29'619  &                                                 \\
 \hspace{10pt}  tertiary\_link               & 39.8 & 16.4 & 38.6 & 6.6 & 76.2 &  941  &                                                 \\
 \hspace{10pt}  unclassified               & 37.9 & 18.5 & 33.9 & 3.3 & 94.6 &  6'868  &                                                 \\
 \hspace{10pt}  residential               & 37.1 & 19.4 & 35.8 & 0.0 & 96.9 &  116'588  &                                                 \\
 free\_flow\_kph-speed\_kph                & -11.0 & 17.6 & -11.2 & -48.7 & 44.5 &  187'487 &                                                 \\
 \hspace{10pt}  motorway               & -2.1 & 9.0 & -2.1 & -30.0 & 17.2 &  348  &                                                 \\
 \hspace{10pt}  motorway\_link               & -0.3 & 22.8 & 0.4 & -54.7 & 38.8 &  879  &                                                 \\
 \hspace{10pt}  trunk               & -12.3 & 12.2 & -9.2 & -45.6 & 12.7 &  5'380  &                                                 \\
 \hspace{10pt}  trunk\_link               & -18.0 & 20.0 & -15.4 & -61.1 & 19.7 &  747  &                                                 \\
 \hspace{10pt}  primary               & -11.5 & 12.5 & -8.5 & -44.0 & 19.5 &  13'913  &                                                 \\
 \hspace{10pt}  primary\_link               & -10.7 & 19.8 & -9.1 & -53.2 & 36.5 &  1'516  &                                                 \\
 \hspace{10pt}  secondary               & -9.3 & 13.1 & -7.8 & -38.6 & 36.9 &  10'301  &                                                 \\
 \hspace{10pt}  secondary\_link               & -16.4 & 16.0 & -18.7 & -51.9 & 34.3 &  387  &                                                 \\
 \hspace{10pt}  tertiary               & -9.3 & 12.7 & -9.5 & -35.4 & 32.8 &  29'619  &                                                 \\
 \hspace{10pt}  tertiary\_link               & -14.6 & 16.4 & -15.8 & -50.0 & 21.6 &  941  &                                                 \\
 \hspace{10pt}  unclassified               & -11.2 & 18.8 & -15.1 & -47.1 & 44.6 &  6'868  &                                                 \\
 \hspace{10pt}  residential               & -11.5 & 19.5 & -12.8 & -48.7 & 47.8 &  116'588  &                                                 \\
\bottomrule

        \caption[Key figures Melbourne ]{Key figures Melbourne for the generated data from 20 randomly sampled days.
        \textbf{num\_edges} number of edges in the street network graph;
        \textbf{num\_nodes} number of nodes in the street network graph;
        \textbf{num\_edges\_per\_cell} number of edges a cell (row,col,heading) has in its intersecting cells;
        \textbf{num\_intersecting\_cells} number of cells (row,col,heading) in an edge's intersecting cells;
        \textbf{node\_degree} number of (unique) neighbor nodes per node;
        \textbf{length\_meters} free flow speed derived from data;
        \textbf{speed\_kph} signalled speed;
        \textbf{free\_flow\_kph} free flow speed derived from data;
        \textbf{free\_flow\_kph-speed\_kph} difference
        }
    \label{tab:key_figures:/iarai/public/t4c/data_pipeline/release20221026_residential_unclassified/2021:Melbourne:}
    \end{longtable}
    \end{small}
    
\subsubsection{Segment density map  Melbourne (2021)}
\mbox{}
\nopagebreak{}
\begin{figure}[H]
\centering
\includegraphics[width=0.85\textwidth]{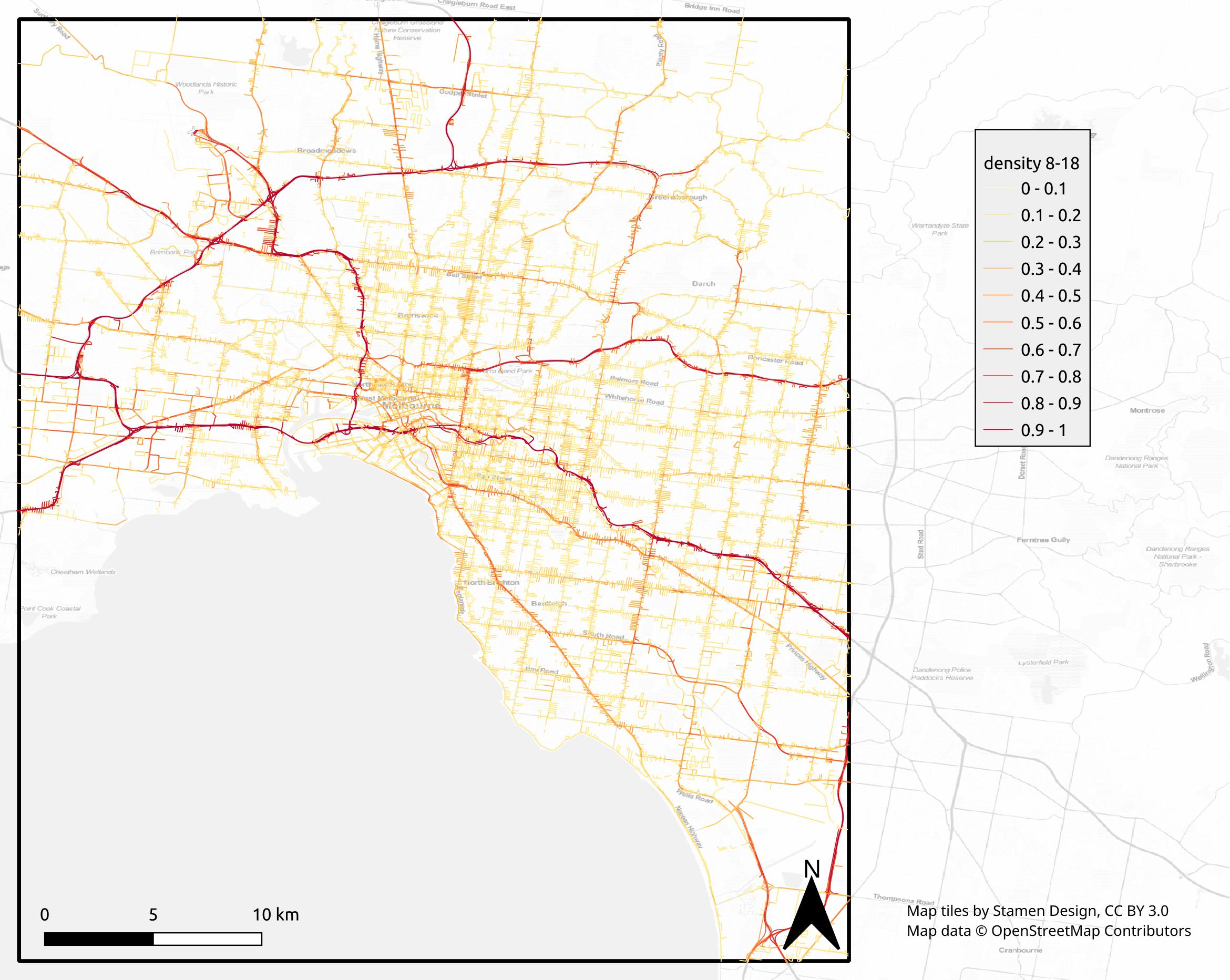}
\caption[Segment-wise density 8am--6pm Melbourne]{Segment-wise density 8am--6pm Melbourne from 20 randomly sampled days.}
\label{figures/speed_stats/density_8_18_melbourne_2021.jpg}
\end{figure}
\clearpage
\subsubsection{Daily density profile  Melbourne  (2021) } 
\mbox{}
\nopagebreak{}
\begin{figure}[H]
\centering
\includegraphics[width=0.85\textwidth]{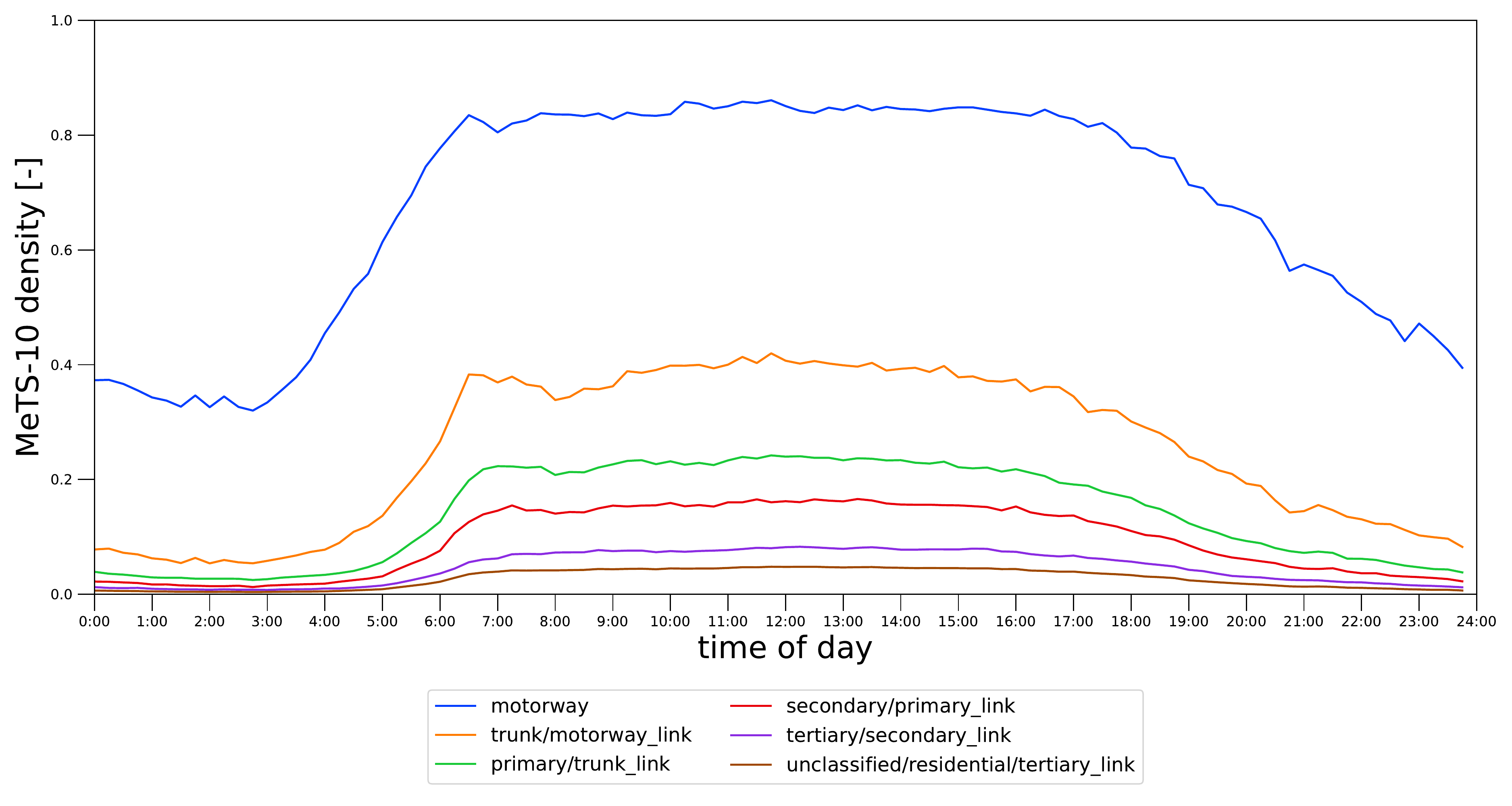}
\caption[Daily density profile Melbourne]{Daily density profile for different road types for Melbourne . Data from 20 randomly sampled days.}
\label{figures/speed_stats/speed_stats_coverage_melbourne_2021_by_highway.pdf}
\end{figure}
\subsubsection{Daily speed profile  Melbourne  (2021) }
\mbox{}
\nopagebreak{}
\begin{figure}[H]
\centering
\includegraphics[width=0.85\textwidth]{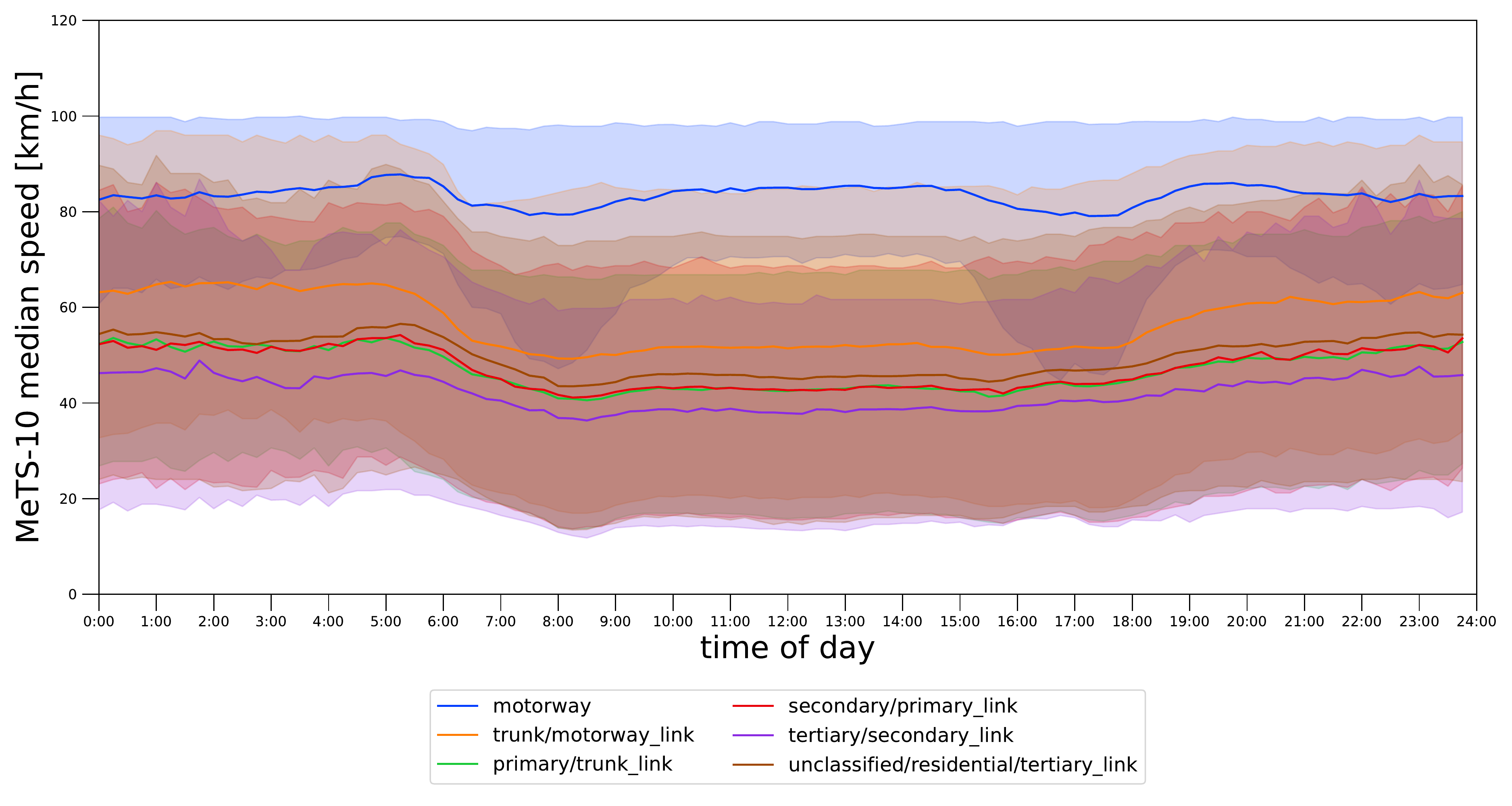}
\caption[Daily median 15 min speeds of all intersecting cells profile Melbourne]{Daily median 15 min speeds of all intersecting cells profile for different road types for Melbourne . The error hull is the 80\% data interval [10.0--90.0 percentiles] of daily means from 20 randomly sampled days.}
\label{figures/speed_stats/speed_stats_median_speed_kph_melbourne_2021_by_highway.pdf}
\end{figure}
\clearpage

\subsection{Key Figures Moscow (2021)}
\subsubsection{Road graph map Moscow (2021)}
\mbox{}
\nopagebreak{}
\begin{figure}[H]
\centering
\includegraphics[width=0.85\textwidth]{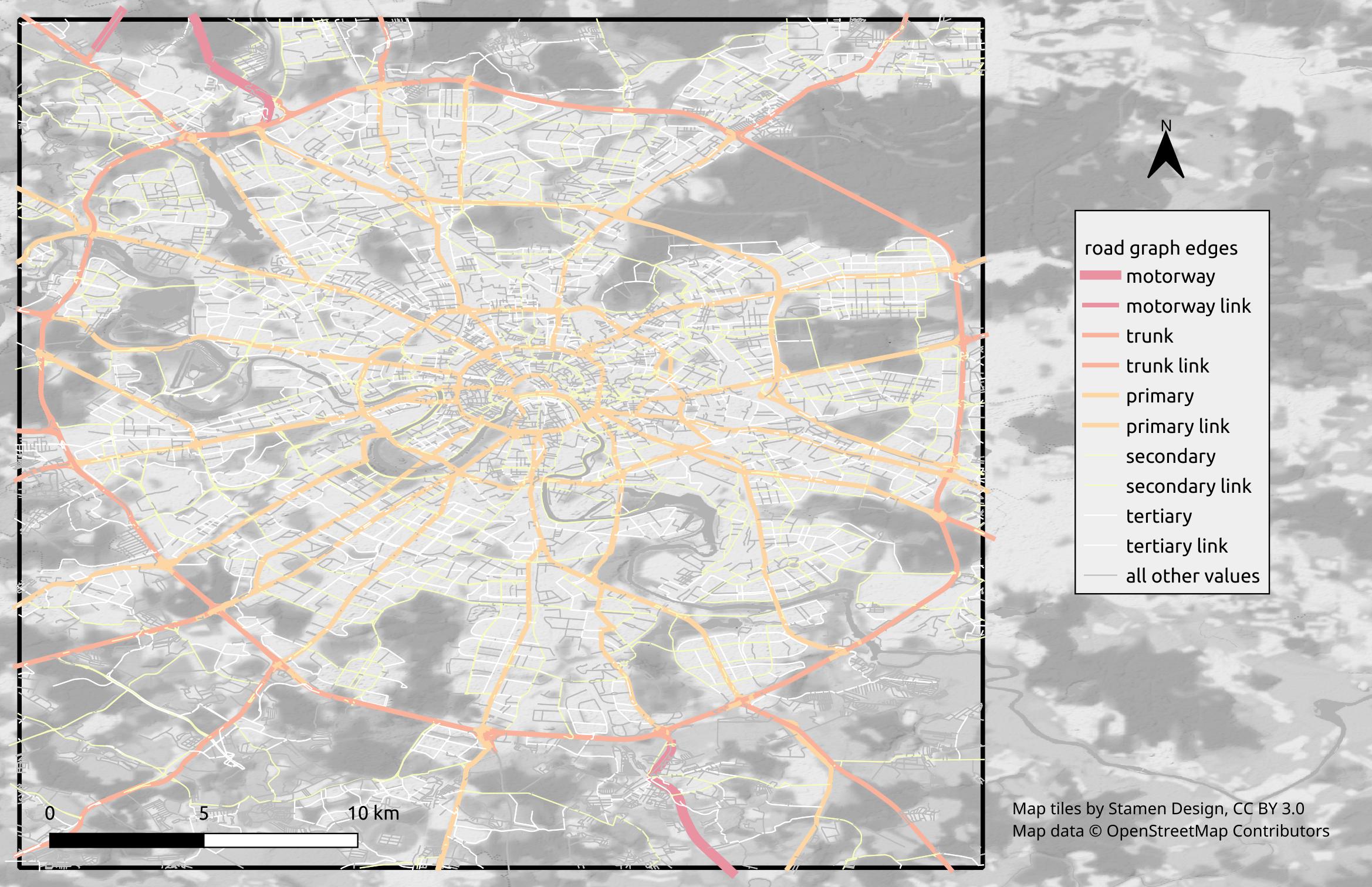}
\caption[Road graph Moscow]{Road graph Moscow, OSM color scheme (2021).}
\label{figures/speed_stats/road_graph_moscow_2021.jpg}
\end{figure}
\subsubsection{Static data  Moscow  (2021) }
\mbox{}\nopagebreak
\begin{small}
\begin{longtable}{p{4cm}rrrrrrrr}
\toprule
Attribute      & {mean} &{std} & {median}  & {q01} & {q99} & {data points} & {sum}  \\
\midrule
 bounding box                &  &  &  &  &  &  37.358--37.853 / 55.506--55.942 &                                                \\
 num\_edges                &  &  &  &  &  &  47'877 &                                                \\
 \hspace{10pt}  motorway               &  &  &  &  &  &  14 &                                                \\
 \hspace{10pt}  trunk               &  &  &  &  &  &  653 &                                                \\
 \hspace{10pt}  trunk\_link               &  &  &  &  &  &  150 &                                                \\
 \hspace{10pt}  primary               &  &  &  &  &  &  2766 &                                                \\
 \hspace{10pt}  primary\_link               &  &  &  &  &  &  575 &                                                \\
 \hspace{10pt}  secondary               &  &  &  &  &  &  9406 &                                                \\
 \hspace{10pt}  secondary\_link               &  &  &  &  &  &  1461 &                                                \\
 \hspace{10pt}  tertiary               &  &  &  &  &  &  11198 &                                                \\
 \hspace{10pt}  tertiary\_link               &  &  &  &  &  &  806 &                                                \\
 \hspace{10pt}  unclassified               &  &  &  &  &  &  6407 &                                                \\
 \hspace{10pt}  residential               &  &  &  &  &  &  14441 &                                                \\
 num\_nodes                &  &  &  &  &  &  22627 &                                                 \\
 num\_edges\_per\_cell                & 1.0 & 0.2 & 1.0 & 1.0 & 2.0 &  293'771 &                                                 \\
 num\_intersecting\_cells                & 6.3 & 6.2 & 4.0 & 1.0 & 30.0 &  47'877 &                                                 \\
 node\_degree                & 2.9 & 0.8 & 3.0 & 1.0 & 4.0 &  22'627 &                                                 \\
 length\_meters                & 227.8 & 280.2 & 140.5 & 7.6 & 1'272.5 &  47'877 & 1.1e+07                                                \\
 \hspace{10pt}  motorway               & 1'970.2 & 1'889.5 & 869.5 & 175.3 & 4'800.2 &  14  & 2.8e+04                                                \\
 \hspace{10pt}  trunk               & 561.1 & 840.2 & 293.3 & 18.5 & 3'519.5 &  653  & 3.7e+05                                                \\
 \hspace{10pt}  trunk\_link               & 363.9 & 332.4 & 268.5 & 15.6 & 1'560.7 &  150  & 5.5e+04                                                \\
 \hspace{10pt}  primary               & 294.2 & 371.6 & 174.6 & 8.3 & 1'723.0 &  2'766  & 8.1e+05                                                \\
 \hspace{10pt}  primary\_link               & 238.1 & 242.6 & 164.0 & 9.4 & 1'165.3 &  575  & 1.4e+05                                                \\
 \hspace{10pt}  secondary               & 209.8 & 255.6 & 115.8 & 6.2 & 1'204.2 &  9'406  & 2.0e+06                                                \\
 \hspace{10pt}  secondary\_link               & 138.4 & 165.0 & 75.8 & 9.2 & 810.4 &  1'461  & 2.0e+05                                                \\
 \hspace{10pt}  tertiary               & 231.5 & 270.7 & 134.7 & 7.8 & 1'276.1 &  11'198  & 2.6e+06                                                \\
 \hspace{10pt}  tertiary\_link               & 97.5 & 123.9 & 54.8 & 8.5 & 619.4 &  806  & 7.9e+04                                                \\
 \hspace{10pt}  unclassified               & 220.6 & 277.7 & 131.1 & 6.5 & 1'348.2 &  6'407  & 1.4e+06                                                \\
 \hspace{10pt}  residential               & 224.9 & 217.1 & 161.8 & 8.8 & 1'070.7 &  14'441  & 3.2e+06                                                \\
 speed\_kph                & 48.6 & 11.3 & 49.0 & 20.0 & 94.7 &  47'877 &                                                 \\
 \hspace{10pt}  motorway               & 71.1 & 16.4 & 73.2 & 40.0 & 105.2 &  14  &                                                 \\
 \hspace{10pt}  trunk               & 94.1 & 12.0 & 100.0 & 50.0 & 110.0 &  653  &                                                 \\
 \hspace{10pt}  trunk\_link               & 54.1 & 5.6 & 54.5 & 30.0 & 60.0 &  150  &                                                 \\
 \hspace{10pt}  primary               & 73.8 & 6.5 & 74.1 & 50.0 & 90.0 &  2'766  &                                                 \\
 \hspace{10pt}  primary\_link               & 45.2 & 4.7 & 45.1 & 27.4 & 60.0 &  575  &                                                 \\
 \hspace{10pt}  secondary               & 53.5 & 5.5 & 53.5 & 30.0 & 60.0 &  9'406  &                                                 \\
 \hspace{10pt}  secondary\_link               & 50.6 & 3.8 & 50.4 & 40.0 & 60.0 &  1'461  &                                                 \\
 \hspace{10pt}  tertiary               & 49.6 & 5.3 & 49.0 & 30.0 & 60.0 &  11'198  &                                                 \\
 \hspace{10pt}  tertiary\_link               & 52.6 & 3.2 & 52.4 & 40.0 & 60.0 &  806  &                                                 \\
 \hspace{10pt}  unclassified               & 36.9 & 5.3 & 36.6 & 20.0 & 60.0 &  6'407  &                                                 \\
 \hspace{10pt}  residential               & 42.5 & 4.5 & 42.6 & 20.0 & 60.0 &  14'441  &                                                 \\
 free\_flow\_kph                & 35.4 & 15.8 & 32.5 & 8.9 & 79.8 &  47'501 &                                                 \\
 \hspace{10pt}  motorway               & 93.0 & 9.7 & 92.0 & 76.9 & 105.9 &  14  &                                                 \\
 \hspace{10pt}  trunk               & 73.8 & 10.7 & 76.5 & 43.5 & 92.4 &  653  &                                                 \\
 \hspace{10pt}  trunk\_link               & 63.0 & 11.7 & 62.4 & 37.2 & 85.0 &  150  &                                                 \\
 \hspace{10pt}  primary               & 51.6 & 12.5 & 51.8 & 23.4 & 77.2 &  2'766  &                                                 \\
 \hspace{10pt}  primary\_link               & 53.4 & 13.3 & 53.9 & 22.1 & 82.1 &  575  &                                                 \\
 \hspace{10pt}  secondary               & 41.5 & 12.5 & 40.5 & 19.6 & 72.9 &  9'406  &                                                 \\
 \hspace{10pt}  secondary\_link               & 47.1 & 15.3 & 47.4 & 14.6 & 81.8 &  1'461  &                                                 \\
 \hspace{10pt}  tertiary               & 35.9 & 11.9 & 34.0 & 16.5 & 73.2 &  11'198  &                                                 \\
 \hspace{10pt}  tertiary\_link               & 44.3 & 18.0 & 42.8 & 11.8 & 83.3 &  806  &                                                 \\
 \hspace{10pt}  unclassified               & 29.4 & 14.1 & 26.4 & 7.1 & 73.8 &  6'363  &                                                 \\
 \hspace{10pt}  residential               & 25.7 & 12.3 & 23.3 & 6.6 & 68.2 &  14'109  &                                                 \\
 free\_flow\_kph-speed\_kph                & -13.3 & 14.2 & -15.6 & -40.4 & 28.8 &  47'501 &                                                 \\
 \hspace{10pt}  motorway               & 21.9 & 22.0 & 18.6 & -12.9 & 65.9 &  14  &                                                 \\
 \hspace{10pt}  trunk               & -20.3 & 12.1 & -20.9 & -47.6 & 11.5 &  653  &                                                 \\
 \hspace{10pt}  trunk\_link               & 8.9 & 12.9 & 7.7 & -19.7 & 37.0 &  150  &                                                 \\
 \hspace{10pt}  primary               & -22.2 & 12.1 & -21.0 & -50.8 & 1.8 &  2'766  &                                                 \\
 \hspace{10pt}  primary\_link               & 8.2 & 14.0 & 8.7 & -25.1 & 37.2 &  575  &                                                 \\
 \hspace{10pt}  secondary               & -12.0 & 13.2 & -13.0 & -36.3 & 20.9 &  9'406  &                                                 \\
 \hspace{10pt}  secondary\_link               & -3.5 & 15.5 & -3.3 & -35.8 & 33.8 &  1'461  &                                                 \\
 \hspace{10pt}  tertiary               & -13.7 & 12.6 & -15.4 & -36.7 & 24.4 &  11'198  &                                                 \\
 \hspace{10pt}  tertiary\_link               & -8.4 & 18.0 & -9.6 & -40.6 & 31.0 &  806  &                                                 \\
 \hspace{10pt}  unclassified               & -7.5 & 14.5 & -10.2 & -35.1 & 36.8 &  6'363  &                                                 \\
 \hspace{10pt}  residential               & -16.8 & 13.0 & -19.5 & -38.3 & 26.3 &  14'109  &                                                 \\
\bottomrule

        \caption[Key figures Moscow ]{Key figures Moscow for the generated data from 20 randomly sampled days.
        \textbf{num\_edges} number of edges in the street network graph;
        \textbf{num\_nodes} number of nodes in the street network graph;
        \textbf{num\_edges\_per\_cell} number of edges a cell (row,col,heading) has in its intersecting cells;
        \textbf{num\_intersecting\_cells} number of cells (row,col,heading) in an edge's intersecting cells;
        \textbf{node\_degree} number of (unique) neighbor nodes per node;
        \textbf{length\_meters} free flow speed derived from data;
        \textbf{speed\_kph} signalled speed;
        \textbf{free\_flow\_kph} free flow speed derived from data;
        \textbf{free\_flow\_kph-speed\_kph} difference
        }
    \label{tab:key_figures:/iarai/public/t4c/data_pipeline/release20221026_residential_unclassified/2021:Moscow:}
    \end{longtable}
    \end{small}
    
\subsubsection{Segment density map  Moscow (2021)}
\mbox{}
\nopagebreak{}
\begin{figure}[H]
\centering
\includegraphics[width=0.85\textwidth]{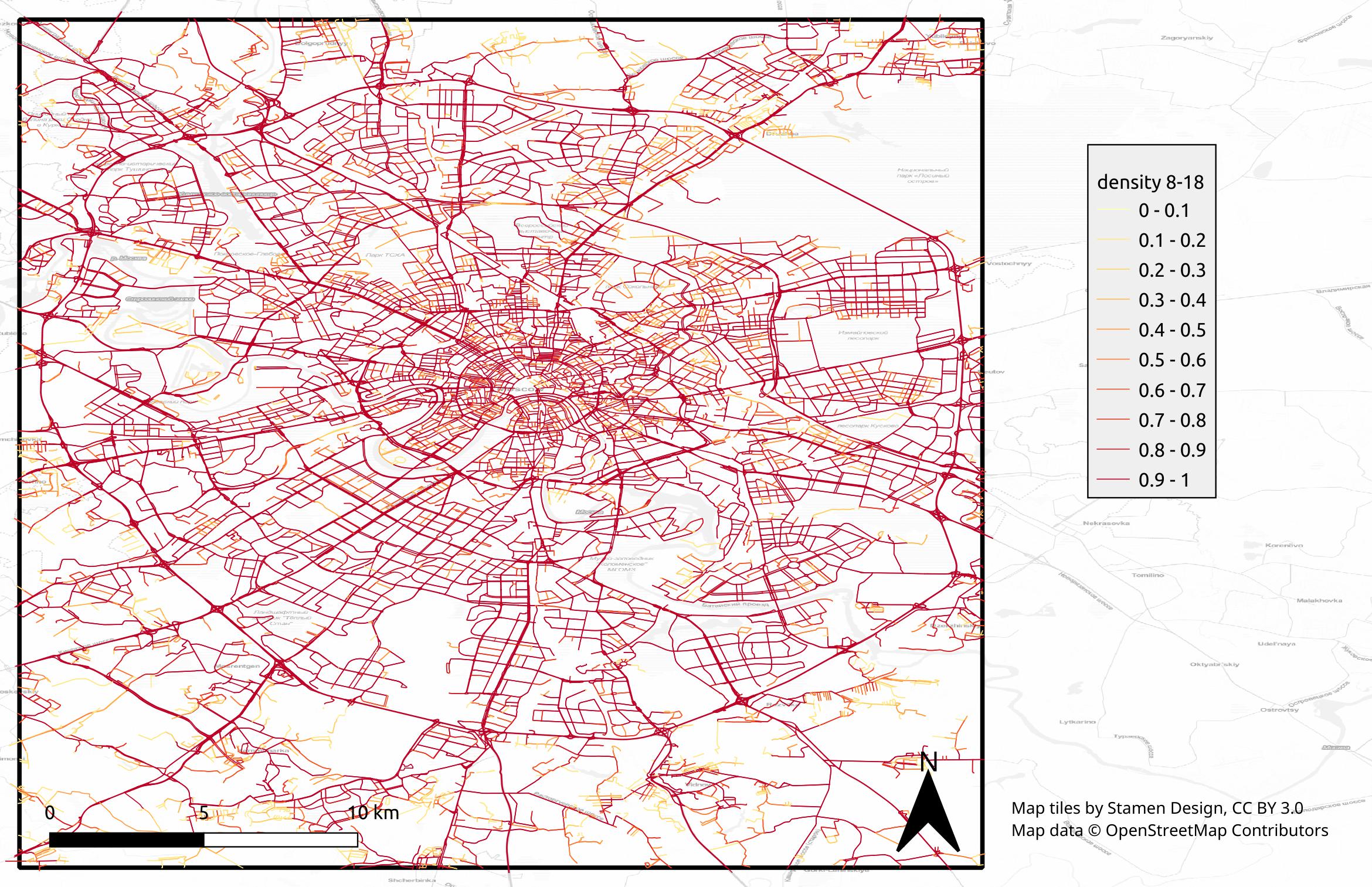}
\caption[Segment-wise density 8am--6pm Moscow]{Segment-wise density 8am--6pm Moscow from 20 randomly sampled days.}
\label{figures/speed_stats/density_8_18_moscow_2021.jpg}
\end{figure}
\clearpage
\subsubsection{Daily density profile  Moscow  (2021) } 
\mbox{}
\nopagebreak{}
\begin{figure}[H]
\centering
\includegraphics[width=0.85\textwidth]{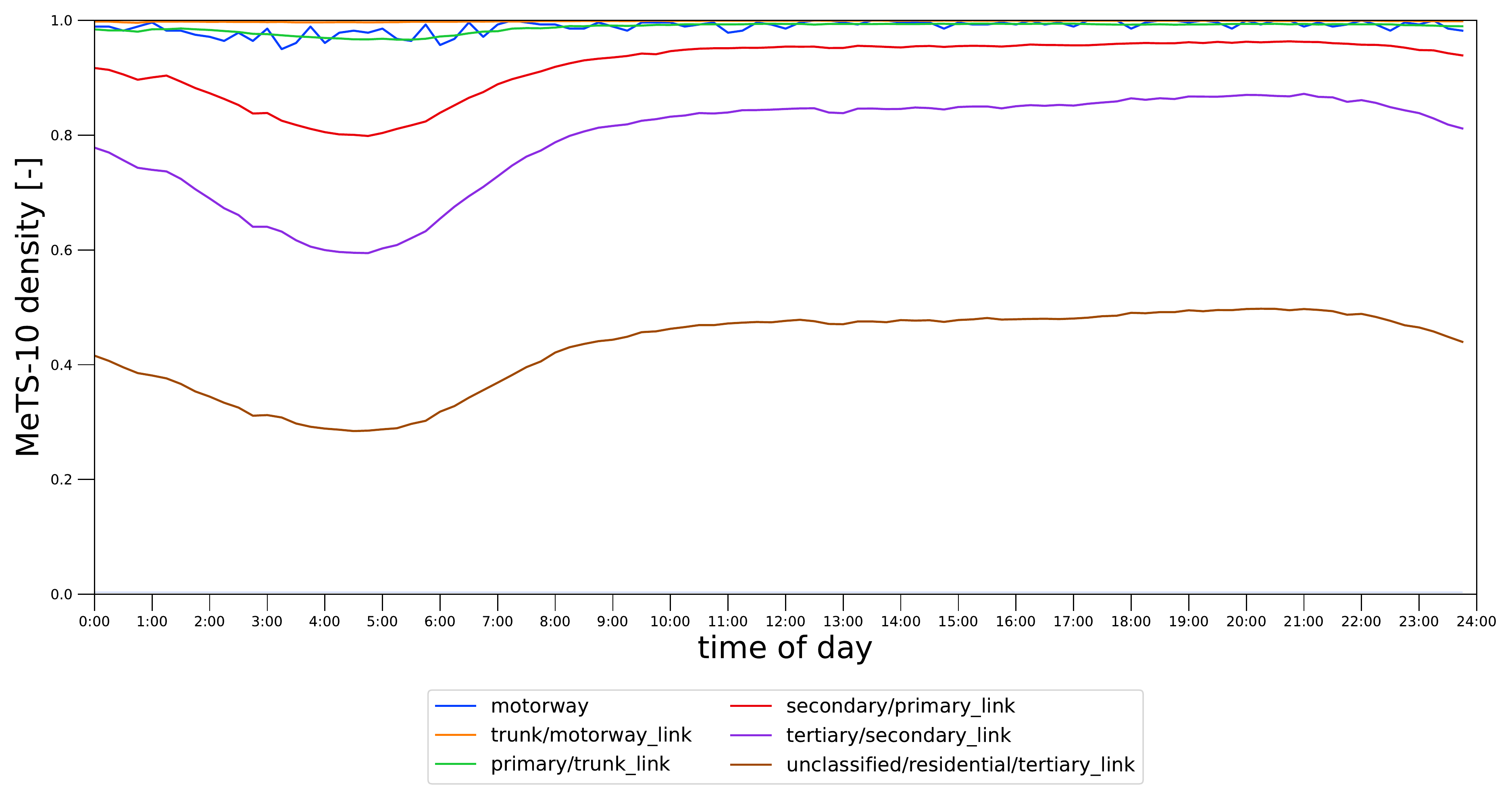}
\caption[Daily density profile Moscow]{Daily density profile for different road types for Moscow . Data from 20 randomly sampled days.}
\label{figures/speed_stats/speed_stats_coverage_moscow_2021_by_highway.pdf}
\end{figure}
\subsubsection{Daily speed profile  Moscow  (2021) }
\mbox{}
\nopagebreak{}
\begin{figure}[H]
\centering
\includegraphics[width=0.85\textwidth]{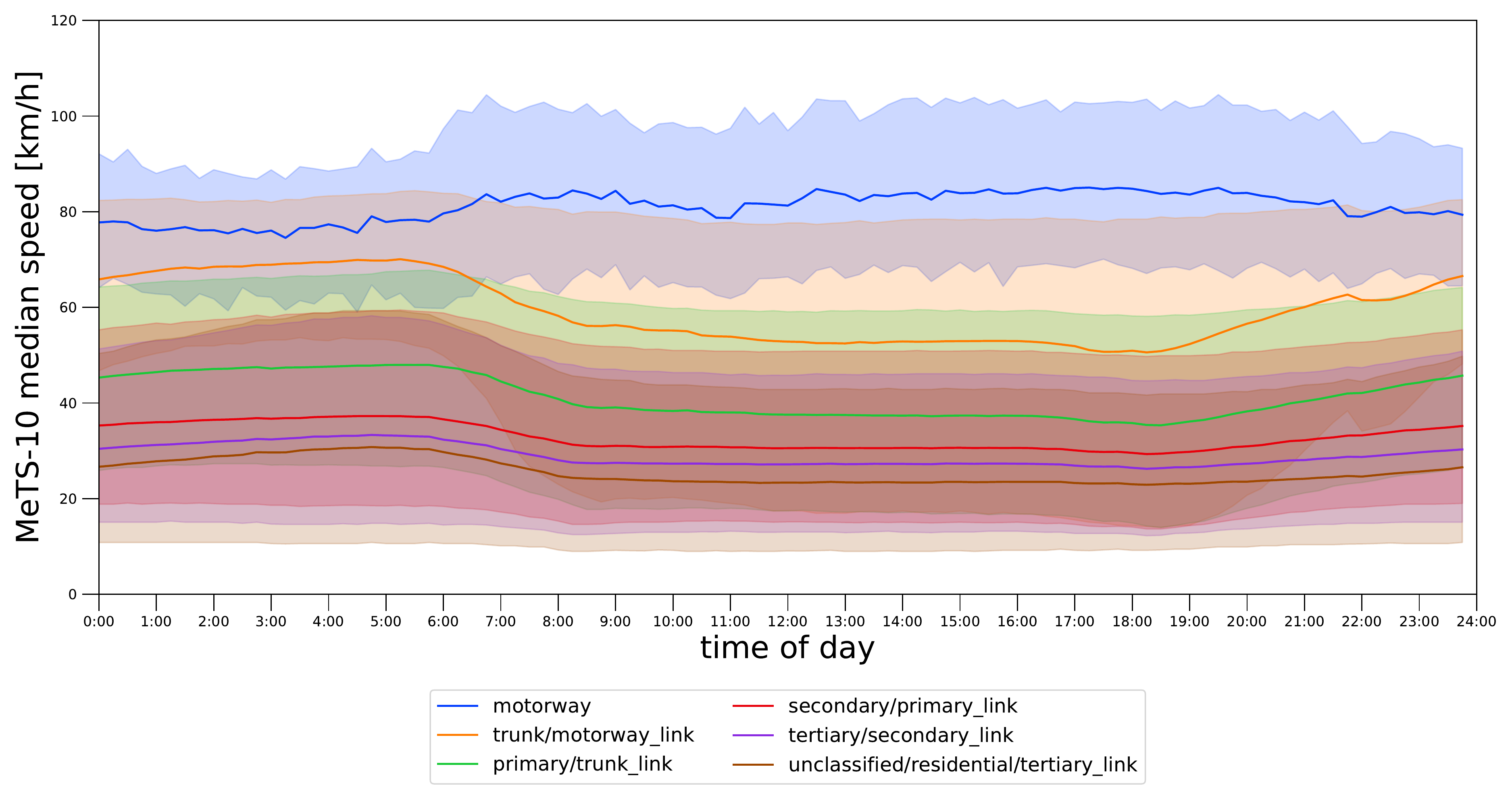}
\caption[Daily median 15 min speeds of all intersecting cells profile Moscow]{Daily median 15 min speeds of all intersecting cells profile for different road types for Moscow . The error hull is the 80\% data interval [10.0--90.0 percentiles] of daily means from 20 randomly sampled days.}
\label{figures/speed_stats/speed_stats_median_speed_kph_moscow_2021_by_highway.pdf}
\end{figure}
\clearpage

\subsection{Key Figures London (2022)}
\subsubsection{Road graph map London (2022)}
\mbox{}
\nopagebreak{}
\begin{figure}[H]
\centering
\includegraphics[width=0.85\textwidth]{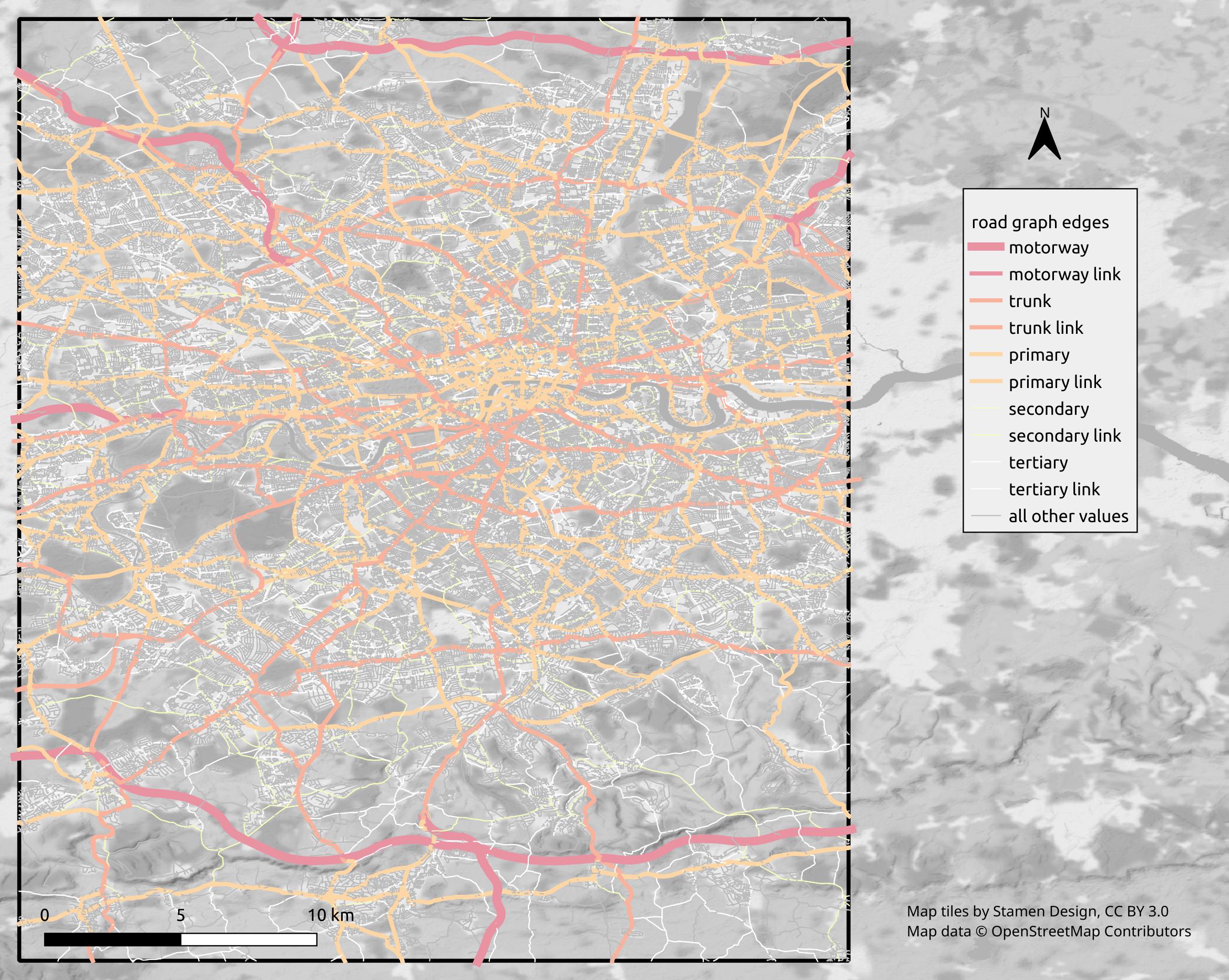}
\caption[Road graph London]{Road graph London, OSM color scheme (2022).}
\label{figures/speed_stats/road_graph_london_2022.jpg}
\end{figure}
\subsubsection{Static data  London  (2022) }
\mbox{}\nopagebreak
\begin{small}
\begin{longtable}{p{4cm}rrrrrrrr}
\toprule
Attribute      & {mean} &{std} & {median}  & {q01} & {q99} & {data points} & {sum}  \\
\midrule
 bounding box                &  &  &  &  &  &  -0.369--0.067 / 51.205--51.7 &                                                \\
 num\_edges                &  &  &  &  &  &  271'075 &                                                \\
 \hspace{10pt}  motorway               &  &  &  &  &  &  79 &                                                \\
 \hspace{10pt}  motorway\_link               &  &  &  &  &  &  82 &                                                \\
 \hspace{10pt}  trunk               &  &  &  &  &  &  8629 &                                                \\
 \hspace{10pt}  trunk\_link               &  &  &  &  &  &  686 &                                                \\
 \hspace{10pt}  primary               &  &  &  &  &  &  25189 &                                                \\
 \hspace{10pt}  primary\_link               &  &  &  &  &  &  411 &                                                \\
 \hspace{10pt}  secondary               &  &  &  &  &  &  11275 &                                                \\
 \hspace{10pt}  secondary\_link               &  &  &  &  &  &  110 &                                                \\
 \hspace{10pt}  tertiary               &  &  &  &  &  &  23878 &                                                \\
 \hspace{10pt}  tertiary\_link               &  &  &  &  &  &  124 &                                                \\
 \hspace{10pt}  unclassified               &  &  &  &  &  &  16513 &                                                \\
 \hspace{10pt}  residential               &  &  &  &  &  &  184099 &                                                \\
 num\_nodes                &  &  &  &  &  &  116304 &                                                 \\
 num\_edges\_per\_cell                & 1.1 & 0.3 & 1.0 & 1.0 & 3.0 &  912'914 &                                                 \\
 num\_intersecting\_cells                & 3.6 & 3.0 & 3.0 & 1.0 & 13.0 &  271'075 &                                                 \\
 node\_degree                & 2.5 & 0.9 & 3.0 & 1.0 & 4.0 &  116'304 &                                                 \\
 length\_meters                & 98.6 & 127.4 & 69.4 & 5.2 & 496.3 &  271'075 & 2.7e+07                                                \\
 \hspace{10pt}  motorway               & 2'516.4 & 2'181.2 & 1'699.3 & 332.5 & 10'264.8 &  79  & 2.0e+05                                                \\
 \hspace{10pt}  motorway\_link               & 520.9 & 373.5 & 463.0 & 28.6 & 1'615.7 &  82  & 4.3e+04                                                \\
 \hspace{10pt}  trunk               & 114.3 & 205.3 & 57.4 & 3.7 & 1'001.6 &  8'629  & 9.9e+05                                                \\
 \hspace{10pt}  trunk\_link               & 124.8 & 135.0 & 56.8 & 6.6 & 521.9 &  686  & 8.6e+04                                                \\
 \hspace{10pt}  primary               & 83.2 & 116.4 & 56.4 & 3.6 & 505.1 &  25'189  & 2.1e+06                                                \\
 \hspace{10pt}  primary\_link               & 40.8 & 48.2 & 28.1 & 6.2 & 242.2 &  411  & 1.7e+04                                                \\
 \hspace{10pt}  secondary               & 96.8 & 131.4 & 66.5 & 3.9 & 597.7 &  11'275  & 1.1e+06                                                \\
 \hspace{10pt}  secondary\_link               & 28.3 & 21.2 & 24.9 & 7.2 & 119.8 &  110  & 3.1e+03                                                \\
 \hspace{10pt}  tertiary               & 112.9 & 166.6 & 73.2 & 5.0 & 835.6 &  23'878  & 2.7e+06                                                \\
 \hspace{10pt}  tertiary\_link               & 40.4 & 34.5 & 29.9 & 8.1 & 189.0 &  124  & 5.0e+03                                                \\
 \hspace{10pt}  unclassified               & 117.2 & 188.0 & 68.2 & 5.1 & 943.9 &  16'513  & 1.9e+06                                                \\
 \hspace{10pt}  residential               & 95.5 & 85.8 & 71.6 & 6.1 & 402.9 &  184'099  & 1.8e+07                                                \\
 speed\_kph                & 36.8 & 7.9 & 32.2 & 32.2 & 64.4 &  271'075 &                                                 \\
 \hspace{10pt}  motorway               & 106.2 & 15.6 & 112.7 & 60.9 & 112.7 &  79  &                                                 \\
 \hspace{10pt}  motorway\_link               & 104.4 & 16.6 & 112.7 & 48.3 & 112.7 &  82  &                                                 \\
 \hspace{10pt}  trunk               & 51.1 & 12.4 & 48.3 & 32.2 & 96.6 &  8'629  &                                                 \\
 \hspace{10pt}  trunk\_link               & 57.0 & 14.3 & 48.3 & 32.2 & 112.7 &  686  &                                                 \\
 \hspace{10pt}  primary               & 42.0 & 9.6 & 48.3 & 32.2 & 64.4 &  25'189  &                                                 \\
 \hspace{10pt}  primary\_link               & 44.2 & 8.1 & 48.3 & 32.2 & 64.4 &  411  &                                                 \\
 \hspace{10pt}  secondary               & 38.9 & 8.5 & 32.2 & 32.2 & 64.4 &  11'275  &                                                 \\
 \hspace{10pt}  secondary\_link               & 39.4 & 8.5 & 40.0 & 32.2 & 64.4 &  110  &                                                 \\
 \hspace{10pt}  tertiary               & 38.5 & 8.4 & 32.2 & 32.2 & 64.4 &  23'878  &                                                 \\
 \hspace{10pt}  tertiary\_link               & 46.1 & 11.9 & 47.0 & 32.2 & 96.6 &  124  &                                                 \\
 \hspace{10pt}  unclassified               & 35.4 & 6.9 & 32.2 & 24.1 & 64.4 &  16'513  &                                                 \\
 \hspace{10pt}  residential               & 35.1 & 5.5 & 32.2 & 32.2 & 48.3 &  184'099  &                                                 \\
 free\_flow\_kph                & 30.0 & 11.8 & 28.7 & 6.8 & 68.7 &  263'309 &                                                 \\
 \hspace{10pt}  motorway               & 104.3 & 14.4 & 110.4 & 60.6 & 118.3 &  79  &                                                 \\
 \hspace{10pt}  motorway\_link               & 92.3 & 20.2 & 99.0 & 36.6 & 117.6 &  82  &                                                 \\
 \hspace{10pt}  trunk               & 39.5 & 12.9 & 36.2 & 19.2 & 81.8 &  8'629  &                                                 \\
 \hspace{10pt}  trunk\_link               & 53.5 & 20.7 & 57.9 & 13.6 & 85.9 &  682  &                                                 \\
 \hspace{10pt}  primary               & 35.3 & 9.6 & 32.9 & 17.9 & 64.9 &  25'189  &                                                 \\
 \hspace{10pt}  primary\_link               & 31.6 & 14.3 & 28.7 & 8.9 & 76.6 &  411  &                                                 \\
 \hspace{10pt}  secondary               & 34.9 & 9.7 & 32.9 & 17.9 & 65.9 &  11'275  &                                                 \\
 \hspace{10pt}  secondary\_link               & 27.4 & 9.4 & 26.8 & 4.7 & 50.3 &  110  &                                                 \\
 \hspace{10pt}  tertiary               & 35.0 & 10.2 & 32.9 & 17.9 & 67.3 &  23'872  &                                                 \\
 \hspace{10pt}  tertiary\_link               & 34.7 & 14.0 & 30.2 & 18.0 & 76.7 &  124  &                                                 \\
 \hspace{10pt}  unclassified               & 28.0 & 13.0 & 26.4 & 6.1 & 72.0 &  16'213  &                                                 \\
 \hspace{10pt}  residential               & 27.8 & 11.1 & 26.8 & 5.6 & 63.1 &  176'643  &                                                 \\
 free\_flow\_kph-speed\_kph                & -6.9 & 11.5 & -6.4 & -33.7 & 27.5 &  263'309 &                                                 \\
 \hspace{10pt}  motorway               & -1.9 & 12.5 & -1.6 & -33.7 & 35.9 &  79  &                                                 \\
 \hspace{10pt}  motorway\_link               & -12.1 & 17.4 & -11.1 & -68.8 & 13.2 &  82  &                                                 \\
 \hspace{10pt}  trunk               & -11.6 & 10.6 & -11.4 & -45.7 & 10.9 &  8'629  &                                                 \\
 \hspace{10pt}  trunk\_link               & -3.5 & 18.8 & -3.6 & -45.6 & 32.6 &  682  &                                                 \\
 \hspace{10pt}  primary               & -6.7 & 10.0 & -5.5 & -30.5 & 16.5 &  25'189  &                                                 \\
 \hspace{10pt}  primary\_link               & -12.5 & 14.2 & -15.6 & -36.0 & 28.4 &  411  &                                                 \\
 \hspace{10pt}  secondary               & -4.0 & 9.5 & -3.5 & -27.6 & 20.4 &  11'275  &                                                 \\
 \hspace{10pt}  secondary\_link               & -12.1 & 10.5 & -11.0 & -35.3 & 10.6 &  110  &                                                 \\
 \hspace{10pt}  tertiary               & -3.5 & 9.6 & -3.5 & -26.7 & 23.8 &  23'872  &                                                 \\
 \hspace{10pt}  tertiary\_link               & -11.4 & 16.8 & -12.3 & -69.8 & 28.5 &  124  &                                                 \\
 \hspace{10pt}  unclassified               & -7.3 & 12.8 & -8.2 & -33.7 & 33.0 &  16'213  &                                                 \\
 \hspace{10pt}  residential               & -7.2 & 11.7 & -7.3 & -33.7 & 28.5 &  176'643  &                                                 \\
\bottomrule

        \caption[Key figures London ]{Key figures London for the generated data from 20 randomly sampled days.
        \textbf{num\_edges} number of edges in the street network graph;
        \textbf{num\_nodes} number of nodes in the street network graph;
        \textbf{num\_edges\_per\_cell} number of edges a cell (row,col,heading) has in its intersecting cells;
        \textbf{num\_intersecting\_cells} number of cells (row,col,heading) in an edge's intersecting cells;
        \textbf{node\_degree} number of (unique) neighbor nodes per node;
        \textbf{length\_meters} free flow speed derived from data;
        \textbf{speed\_kph} signalled speed;
        \textbf{free\_flow\_kph} free flow speed derived from data;
        \textbf{free\_flow\_kph-speed\_kph} difference
        }
    \label{tab:key_figures:/iarai/public/t4c/data_pipeline/release20221026_residential_unclassified/2022:London:}
    \end{longtable}
    \end{small}
    
\subsubsection{Segment density map  London (2022)}
\mbox{}
\nopagebreak{}
\begin{figure}[H]
\centering
\includegraphics[width=0.85\textwidth]{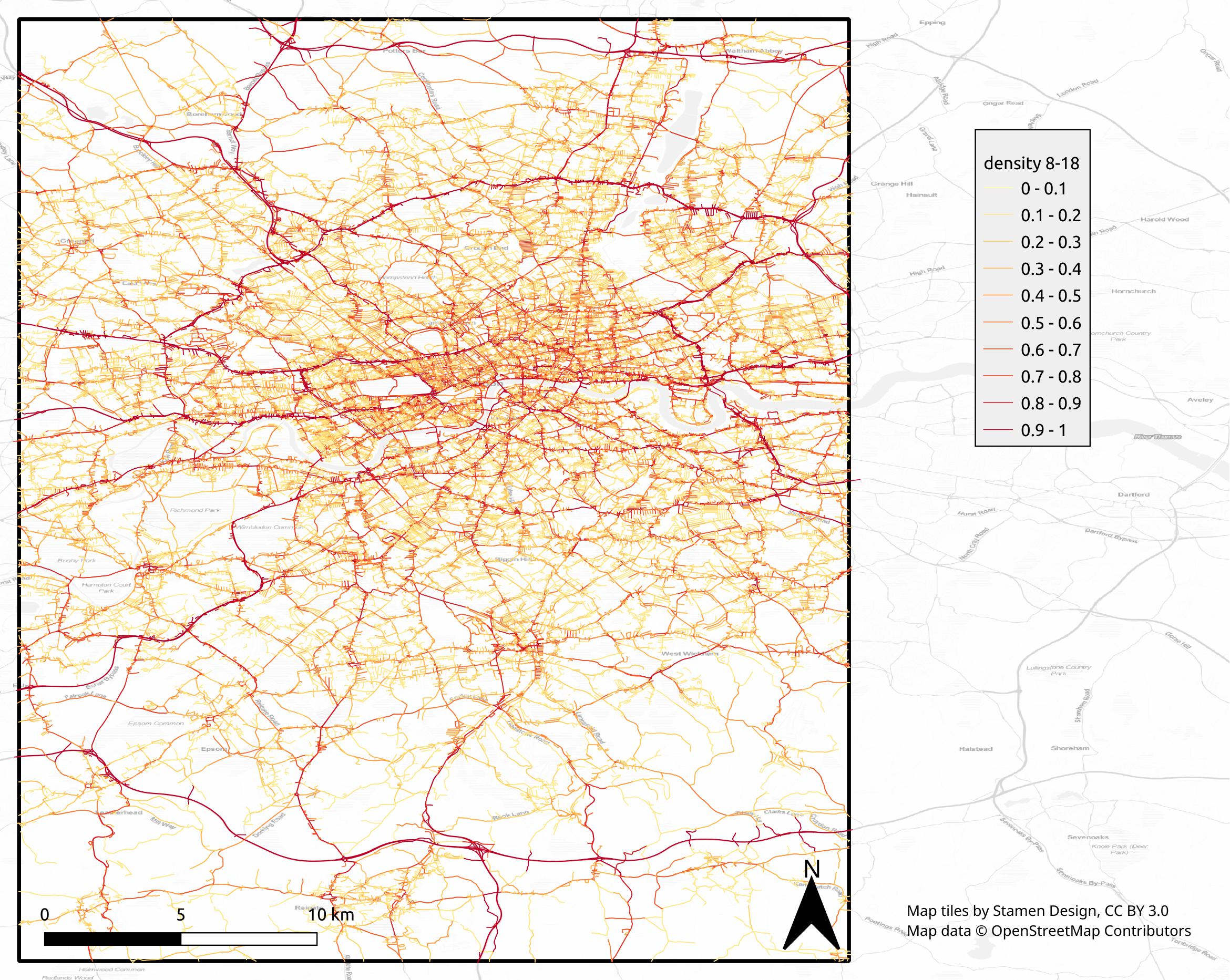}
\caption[Segment-wise density 8am--6pm London]{Segment-wise density 8am--6pm London from 20 randomly sampled days.}
\label{figures/speed_stats/density_8_18_london_2022.jpg}
\end{figure}
\clearpage
\subsubsection{Daily density profile  London  (2022) } 
\mbox{}
\nopagebreak{}
\begin{figure}[H]
\centering
\includegraphics[width=0.85\textwidth]{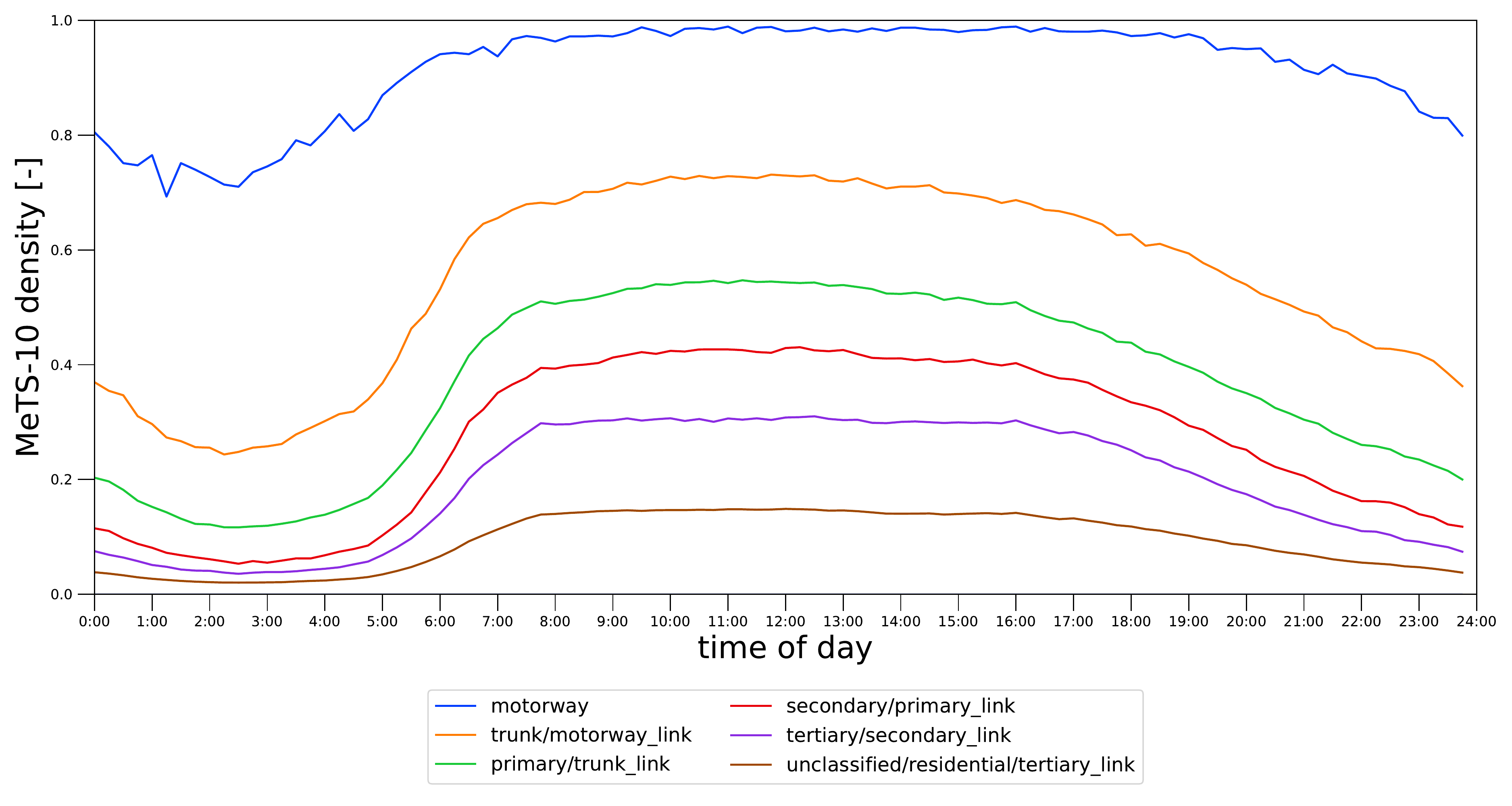}
\caption[Daily density profile London]{Daily density profile for different road types for London . Data from 20 randomly sampled days.}
\label{figures/speed_stats/speed_stats_coverage_london_2022_by_highway.pdf}
\end{figure}
\subsubsection{Daily speed profile  London  (2022) }
\mbox{}
\nopagebreak{}
\begin{figure}[H]
\centering
\includegraphics[width=0.85\textwidth]{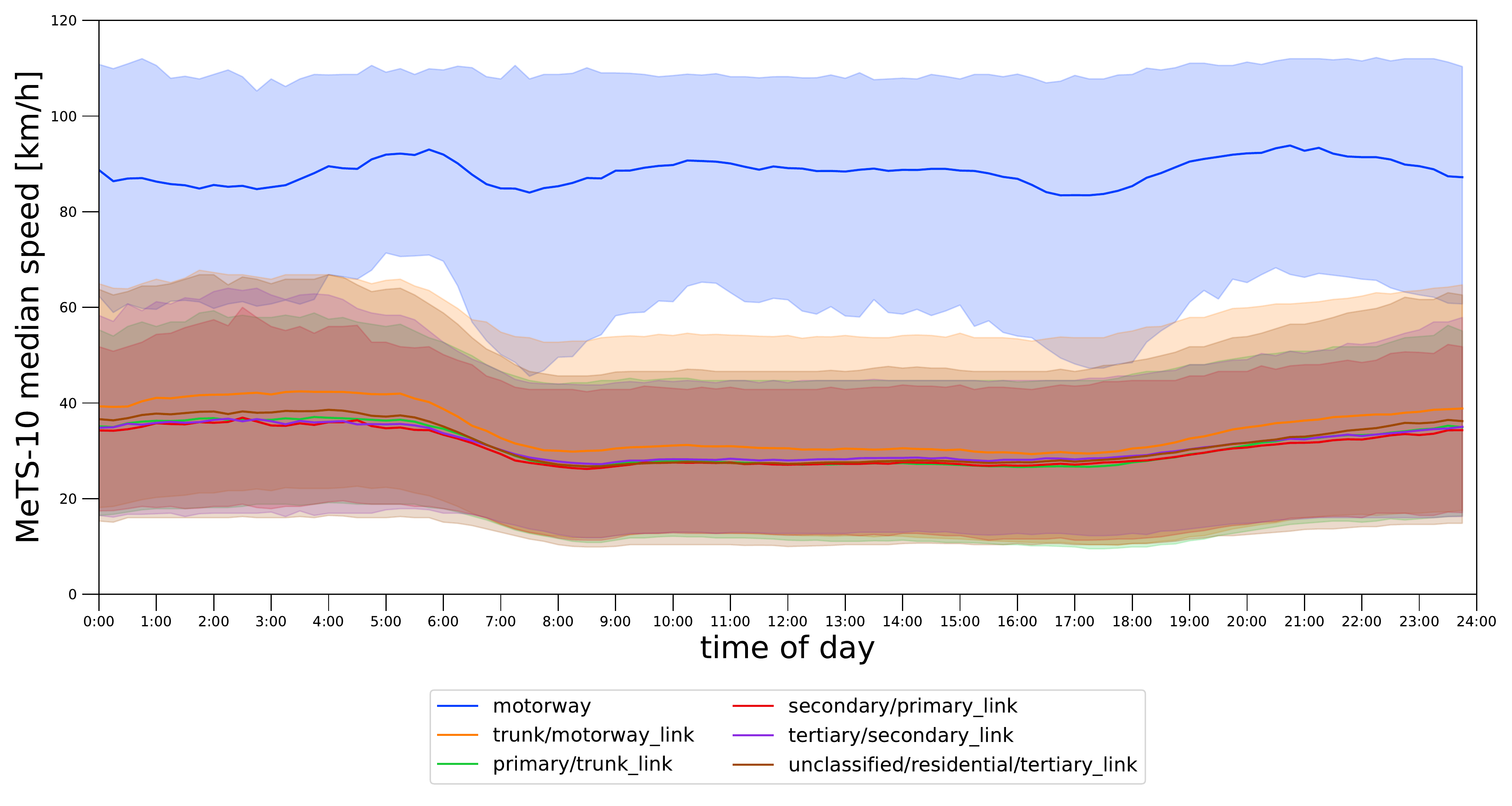}
\caption[Daily median 15 min speeds of all intersecting cells profile London]{Daily median 15 min speeds of all intersecting cells profile for different road types for London . The error hull is the 80\% data interval [10.0--90.0 percentiles] of daily means from 20 randomly sampled days.}
\label{figures/speed_stats/speed_stats_median_speed_kph_london_2022_by_highway.pdf}
\end{figure}
\clearpage

\subsection{Key Figures Madrid (2022)}
\subsubsection{Road graph map Madrid (2022)}
\mbox{}
\nopagebreak{}
\begin{figure}[H]
\centering
\includegraphics[width=0.85\textwidth]{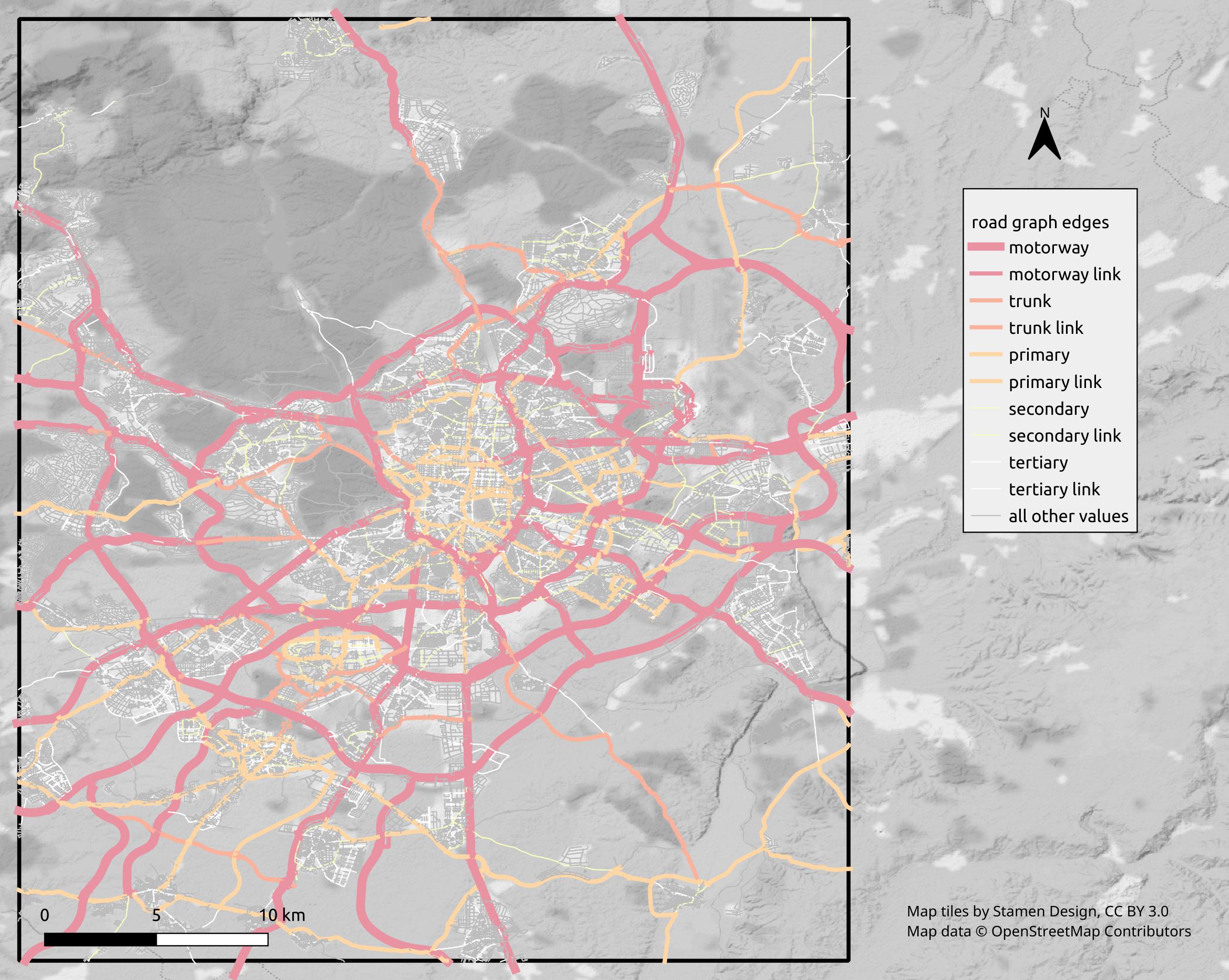}
\caption[Road graph Madrid]{Road graph Madrid, OSM color scheme (2022).}
\label{figures/speed_stats/road_graph_madrid_2022.jpg}
\end{figure}
\subsubsection{Static data  Madrid  (2022) }
\mbox{}\nopagebreak
\begin{small}
\begin{longtable}{p{4cm}rrrrrrrr}
\toprule
Attribute      & {mean} &{std} & {median}  & {q01} & {q99} & {data points} & {sum}  \\
\midrule
 bounding box                &  &  &  &  &  &  -3.927---3.491 / 40.177--40.672 &                                                \\
 num\_edges                &  &  &  &  &  &  143'402 &                                                \\
 \hspace{10pt}  motorway               &  &  &  &  &  &  1277 &                                                \\
 \hspace{10pt}  motorway\_link               &  &  &  &  &  &  2647 &                                                \\
 \hspace{10pt}  trunk               &  &  &  &  &  &  1017 &                                                \\
 \hspace{10pt}  trunk\_link               &  &  &  &  &  &  373 &                                                \\
 \hspace{10pt}  primary               &  &  &  &  &  &  6265 &                                                \\
 \hspace{10pt}  primary\_link               &  &  &  &  &  &  475 &                                                \\
 \hspace{10pt}  secondary               &  &  &  &  &  &  6855 &                                                \\
 \hspace{10pt}  secondary\_link               &  &  &  &  &  &  276 &                                                \\
 \hspace{10pt}  tertiary               &  &  &  &  &  &  16147 &                                                \\
 \hspace{10pt}  tertiary\_link               &  &  &  &  &  &  310 &                                                \\
 \hspace{10pt}  unclassified               &  &  &  &  &  &  5669 &                                                \\
 \hspace{10pt}  residential               &  &  &  &  &  &  102091 &                                                \\
 num\_nodes                &  &  &  &  &  &  71757 &                                                 \\
 num\_edges\_per\_cell                & 1.1 & 0.4 & 1.0 & 1.0 & 3.0 &  467'013 &                                                 \\
 num\_intersecting\_cells                & 3.5 & 3.7 & 3.0 & 1.0 & 16.0 &  143'402 &                                                 \\
 node\_degree                & 3.0 & 0.7 & 3.0 & 1.0 & 4.0 &  71'757 &                                                 \\
 length\_meters                & 110.2 & 193.2 & 66.0 & 4.7 & 776.8 &  143'402 & 1.6e+07                                                \\
 \hspace{10pt}  motorway               & 878.7 & 842.8 & 660.3 & 13.2 & 4'440.7 &  1'277  & 1.1e+06                                                \\
 \hspace{10pt}  motorway\_link               & 343.5 & 287.4 & 286.5 & 15.6 & 1'344.9 &  2'647  & 9.1e+05                                                \\
 \hspace{10pt}  trunk               & 254.4 & 483.3 & 60.4 & 4.9 & 2'285.9 &  1'017  & 2.6e+05                                                \\
 \hspace{10pt}  trunk\_link               & 227.4 & 235.6 & 175.3 & 12.4 & 923.6 &  373  & 8.5e+04                                                \\
 \hspace{10pt}  primary               & 130.1 & 335.0 & 44.6 & 3.3 & 1'346.9 &  6'265  & 8.1e+05                                                \\
 \hspace{10pt}  primary\_link               & 132.5 & 135.1 & 85.7 & 8.1 & 654.7 &  475  & 6.3e+04                                                \\
 \hspace{10pt}  secondary               & 94.2 & 247.8 & 41.0 & 3.2 & 686.8 &  6'855  & 6.5e+05                                                \\
 \hspace{10pt}  secondary\_link               & 73.9 & 94.8 & 38.2 & 6.0 & 420.4 &  276  & 2.0e+04                                                \\
 \hspace{10pt}  tertiary               & 88.6 & 147.2 & 45.6 & 3.2 & 602.5 &  16'147  & 1.4e+06                                                \\
 \hspace{10pt}  tertiary\_link               & 77.3 & 112.8 & 35.7 & 6.5 & 581.5 &  310  & 2.4e+04                                                \\
 \hspace{10pt}  unclassified               & 169.9 & 331.4 & 76.3 & 5.6 & 1'687.0 &  5'669  & 9.6e+05                                                \\
 \hspace{10pt}  residential               & 92.7 & 92.4 & 68.0 & 5.2 & 452.5 &  102'091  & 9.5e+06                                                \\
 speed\_kph                & 68.6 & 1'334.2 & 76.6 & 20.0 & 151.3 &  143'402 &                                                 \\
 \hspace{10pt}  motorway               & 98.1 & 15.5 & 100.0 & 50.0 & 120.0 &  1'277  &                                                 \\
 \hspace{10pt}  motorway\_link               & 60.4 & 13.0 & 59.2 & 40.0 & 100.0 &  2'647  &                                                 \\
 \hspace{10pt}  trunk               & 63.8 & 19.1 & 71.2 & 30.0 & 100.0 &  1'017  &                                                 \\
 \hspace{10pt}  trunk\_link               & 52.8 & 9.2 & 52.1 & 40.0 & 82.8 &  373  &                                                 \\
 \hspace{10pt}  primary               & 87.7 & 50.7 & 50.0 & 30.0 & 151.3 &  6'265  &                                                 \\
 \hspace{10pt}  primary\_link               & 48.9 & 6.8 & 49.0 & 40.0 & 90.0 &  475  &                                                 \\
 \hspace{10pt}  secondary               & 56.5 & 120.6 & 61.3 & 30.0 & 70.0 &  6'855  &                                                 \\
 \hspace{10pt}  secondary\_link               & 45.1 & 4.7 & 45.3 & 30.0 & 60.0 &  276  &                                                 \\
 \hspace{10pt}  tertiary               & 44.6 & 5.8 & 44.8 & 20.0 & 50.0 &  16'147  &                                                 \\
 \hspace{10pt}  tertiary\_link               & 41.7 & 5.7 & 41.8 & 20.0 & 60.0 &  310  &                                                 \\
 \hspace{10pt}  unclassified               & 40.6 & 4.2 & 40.5 & 20.0 & 60.0 &  5'669  &                                                 \\
 \hspace{10pt}  residential               & 73.8 & 1'580.9 & 76.6 & 20.0 & 76.6 &  102'091  &                                                 \\
 free\_flow\_kph                & 34.8 & 17.9 & 30.6 & 9.9 & 101.2 &  141'365 &                                                 \\
 \hspace{10pt}  motorway               & 93.3 & 15.4 & 95.2 & 51.2 & 120.0 &  1'277  &                                                 \\
 \hspace{10pt}  motorway\_link               & 86.7 & 18.0 & 90.4 & 32.1 & 120.0 &  2'645  &                                                 \\
 \hspace{10pt}  trunk               & 57.1 & 24.0 & 49.4 & 24.0 & 102.7 &  1'017  &                                                 \\
 \hspace{10pt}  trunk\_link               & 76.8 & 18.3 & 81.4 & 28.7 & 105.7 &  373  &                                                 \\
 \hspace{10pt}  primary               & 42.2 & 15.9 & 38.6 & 17.9 & 95.2 &  6'261  &                                                 \\
 \hspace{10pt}  primary\_link               & 61.4 & 24.2 & 61.6 & 10.7 & 101.0 &  474  &                                                 \\
 \hspace{10pt}  secondary               & 37.9 & 13.3 & 35.3 & 18.4 & 88.8 &  6'855  &                                                 \\
 \hspace{10pt}  secondary\_link               & 49.6 & 20.1 & 46.6 & 9.1 & 87.9 &  272  &                                                 \\
 \hspace{10pt}  tertiary               & 36.2 & 13.6 & 33.9 & 16.9 & 91.9 &  16'144  &                                                 \\
 \hspace{10pt}  tertiary\_link               & 47.3 & 22.0 & 41.9 & 11.7 & 101.6 &  307  &                                                 \\
 \hspace{10pt}  unclassified               & 43.9 & 25.0 & 34.8 & 8.9 & 107.0 &  5'448  &                                                 \\
 \hspace{10pt}  residential               & 30.8 & 13.5 & 28.7 & 8.9 & 89.9 &  100'292  &                                                 \\
 free\_flow\_kph-speed\_kph                & -33.7 & 1'343.9 & -36.6 & -108.0 & 46.4 &  141'365 &                                                 \\
 \hspace{10pt}  motorway               & -4.8 & 14.1 & -4.6 & -46.5 & 26.0 &  1'277  &                                                 \\
 \hspace{10pt}  motorway\_link               & 26.3 & 19.5 & 29.3 & -26.3 & 60.8 &  2'645  &                                                 \\
 \hspace{10pt}  trunk               & -6.7 & 20.4 & -3.3 & -44.4 & 40.5 &  1'017  &                                                 \\
 \hspace{10pt}  trunk\_link               & 24.1 & 19.3 & 27.2 & -30.5 & 60.7 &  373  &                                                 \\
 \hspace{10pt}  primary               & -45.5 & 50.0 & -18.0 & -129.7 & 23.6 &  6'261  &                                                 \\
 \hspace{10pt}  primary\_link               & 12.6 & 24.7 & 12.6 & -41.7 & 53.0 &  474  &                                                 \\
 \hspace{10pt}  secondary               & -18.6 & 120.7 & -17.5 & -41.9 & 28.3 &  6'855  &                                                 \\
 \hspace{10pt}  secondary\_link               & 4.5 & 20.0 & 2.3 & -36.8 & 42.4 &  272  &                                                 \\
 \hspace{10pt}  tertiary               & -8.3 & 14.5 & -10.9 & -31.2 & 48.1 &  16'144  &                                                 \\
 \hspace{10pt}  tertiary\_link               & 5.6 & 22.3 & -0.6 & -31.2 & 59.8 &  307  &                                                 \\
 \hspace{10pt}  unclassified               & 3.2 & 24.7 & -5.7 & -31.6 & 67.3 &  5'448  &                                                 \\
 \hspace{10pt}  residential               & -43.0 & 1'595.0 & -44.6 & -67.0 & 22.7 &  100'292  &                                                 \\
\bottomrule

        \caption[Key figures Madrid ]{Key figures Madrid for the generated data from 20 randomly sampled days.
        \textbf{num\_edges} number of edges in the street network graph;
        \textbf{num\_nodes} number of nodes in the street network graph;
        \textbf{num\_edges\_per\_cell} number of edges a cell (row,col,heading) has in its intersecting cells;
        \textbf{num\_intersecting\_cells} number of cells (row,col,heading) in an edge's intersecting cells;
        \textbf{node\_degree} number of (unique) neighbor nodes per node;
        \textbf{length\_meters} free flow speed derived from data;
        \textbf{speed\_kph} signalled speed;
        \textbf{free\_flow\_kph} free flow speed derived from data;
        \textbf{free\_flow\_kph-speed\_kph} difference
        }
    \label{tab:key_figures:/iarai/public/t4c/data_pipeline/release20221026_residential_unclassified/2022:Madrid:}
    \end{longtable}
    \end{small}
    
\subsubsection{Segment density map  Madrid (2022)}
\mbox{}
\nopagebreak{}
\begin{figure}[H]
\centering
\includegraphics[width=0.85\textwidth]{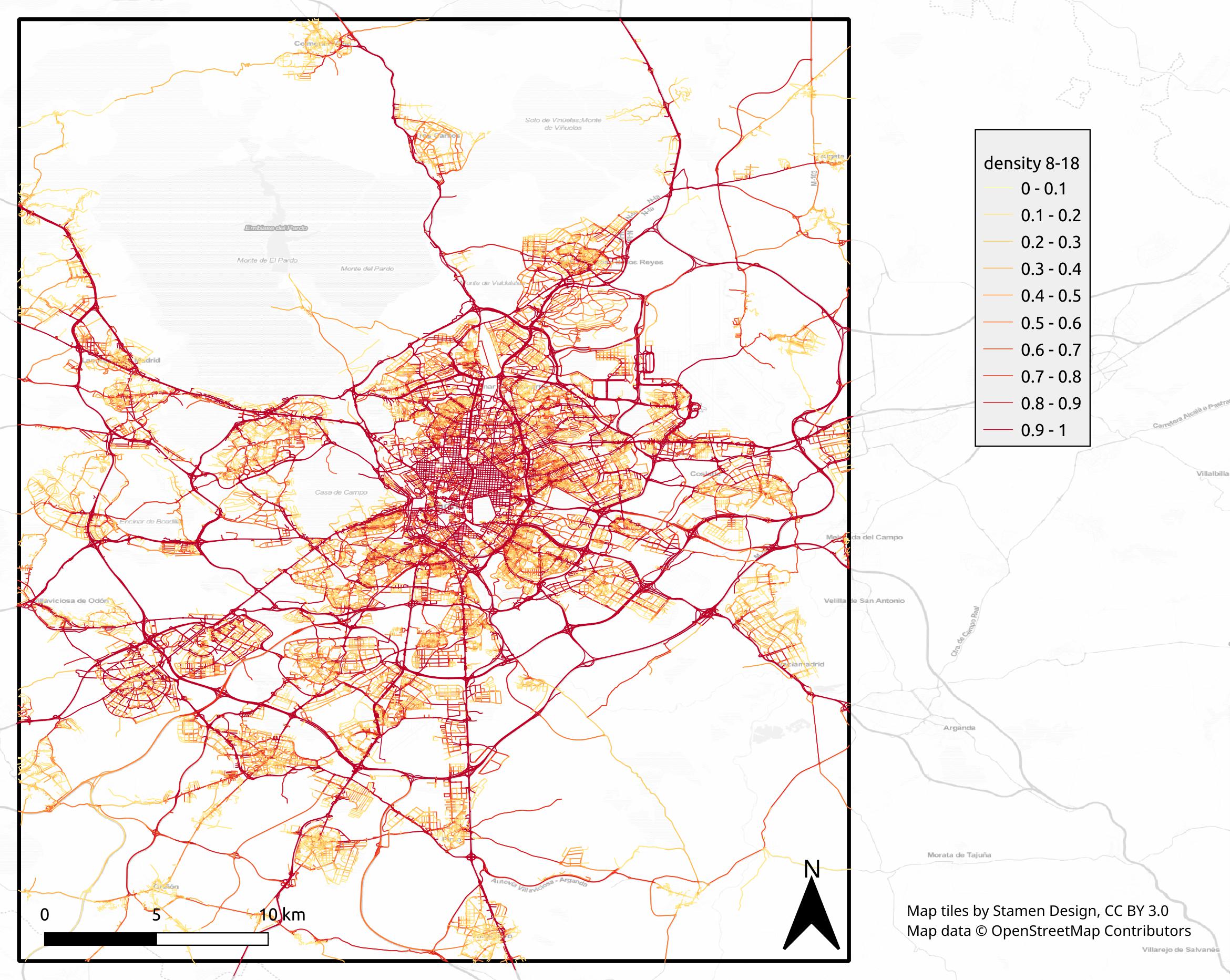}
\caption[Segment-wise density 8am--6pm Madrid]{Segment-wise density 8am--6pm Madrid from 20 randomly sampled days.}
\label{figures/speed_stats/density_8_18_madrid_2022.jpg}
\end{figure}
\clearpage
\subsubsection{Daily density profile  Madrid  (2022) } 
\mbox{}
\nopagebreak{}
\begin{figure}[H]
\centering
\includegraphics[width=0.85\textwidth]{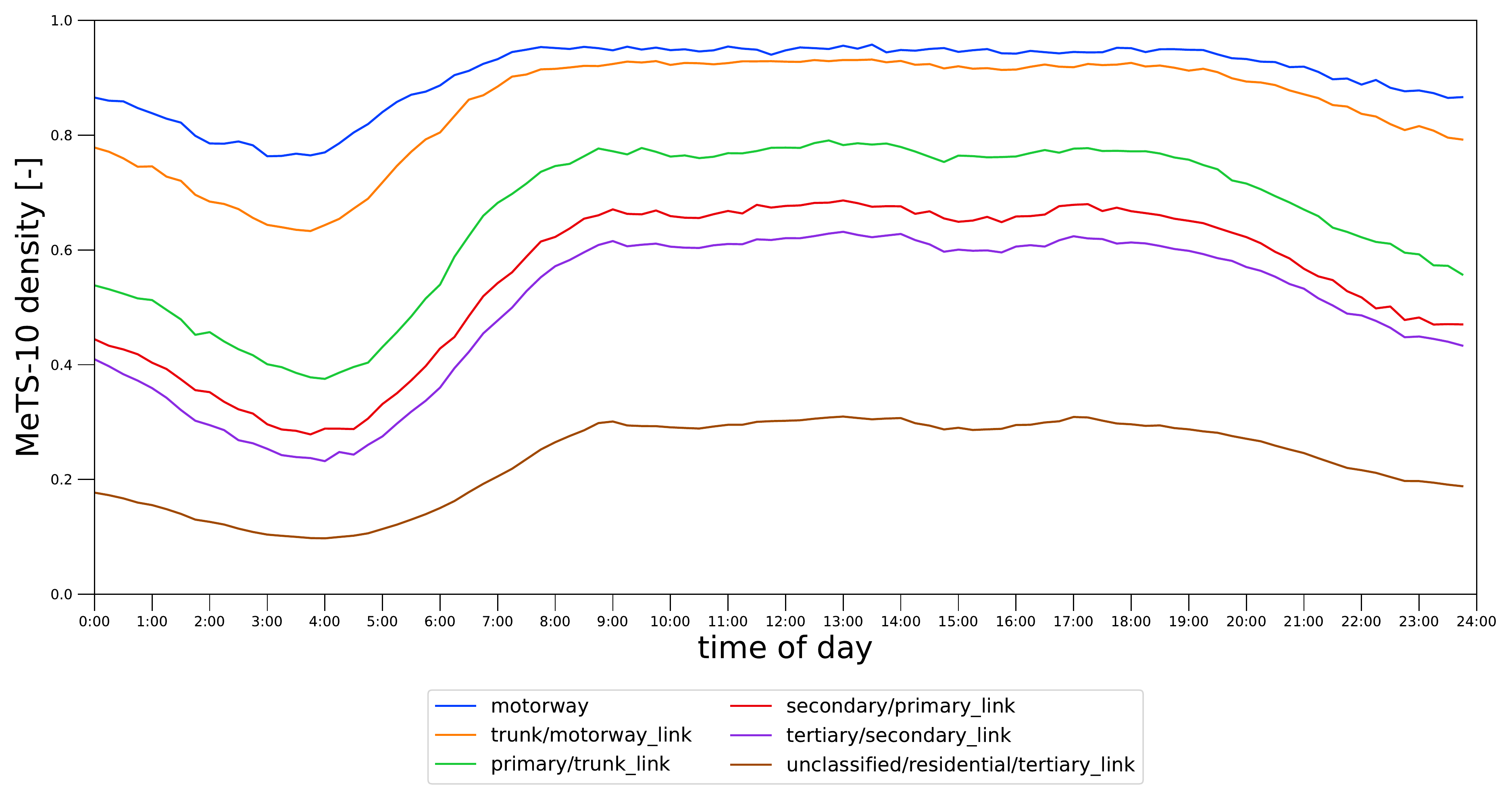}
\caption[Daily density profile Madrid]{Daily density profile for different road types for Madrid . Data from 20 randomly sampled days.}
\label{figures/speed_stats/speed_stats_coverage_madrid_2022_by_highway.pdf}
\end{figure}
\subsubsection{Daily speed profile  Madrid  (2022) }
\mbox{}
\nopagebreak{}
\begin{figure}[H]
\centering
\includegraphics[width=0.85\textwidth]{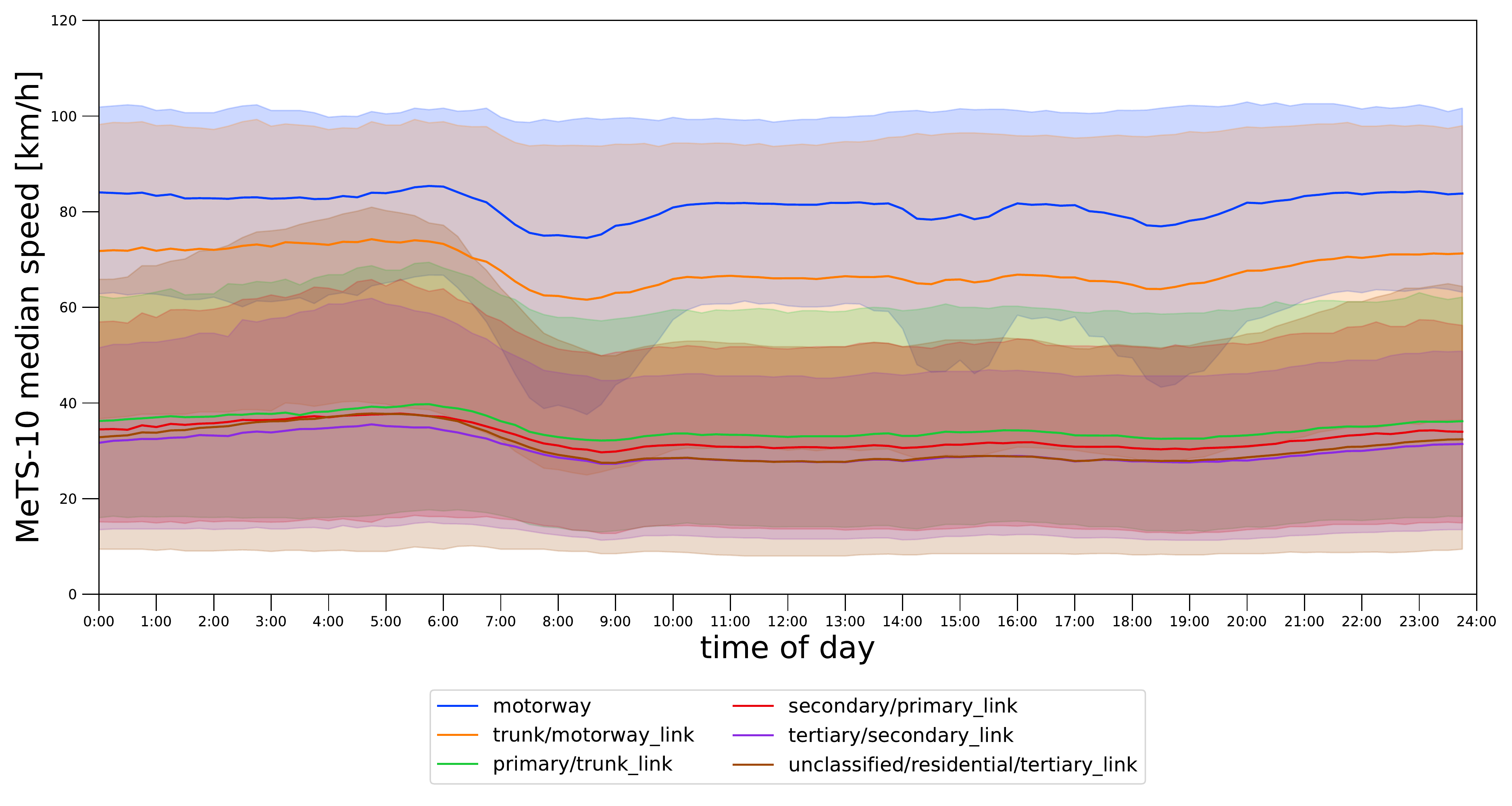}
\caption[Daily median 15 min speeds of all intersecting cells profile Madrid]{Daily median 15 min speeds of all intersecting cells profile for different road types for Madrid . The error hull is the 80\% data interval [10.0--90.0 percentiles] of daily means from 20 randomly sampled days.}
\label{figures/speed_stats/speed_stats_median_speed_kph_madrid_2022_by_highway.pdf}
\end{figure}
\clearpage

\subsection{Key Figures Melbourne (2022)}
\subsubsection{Road graph map Melbourne (2022)}
\mbox{}
\nopagebreak{}
\begin{figure}[H]
\centering
\includegraphics[width=0.85\textwidth]{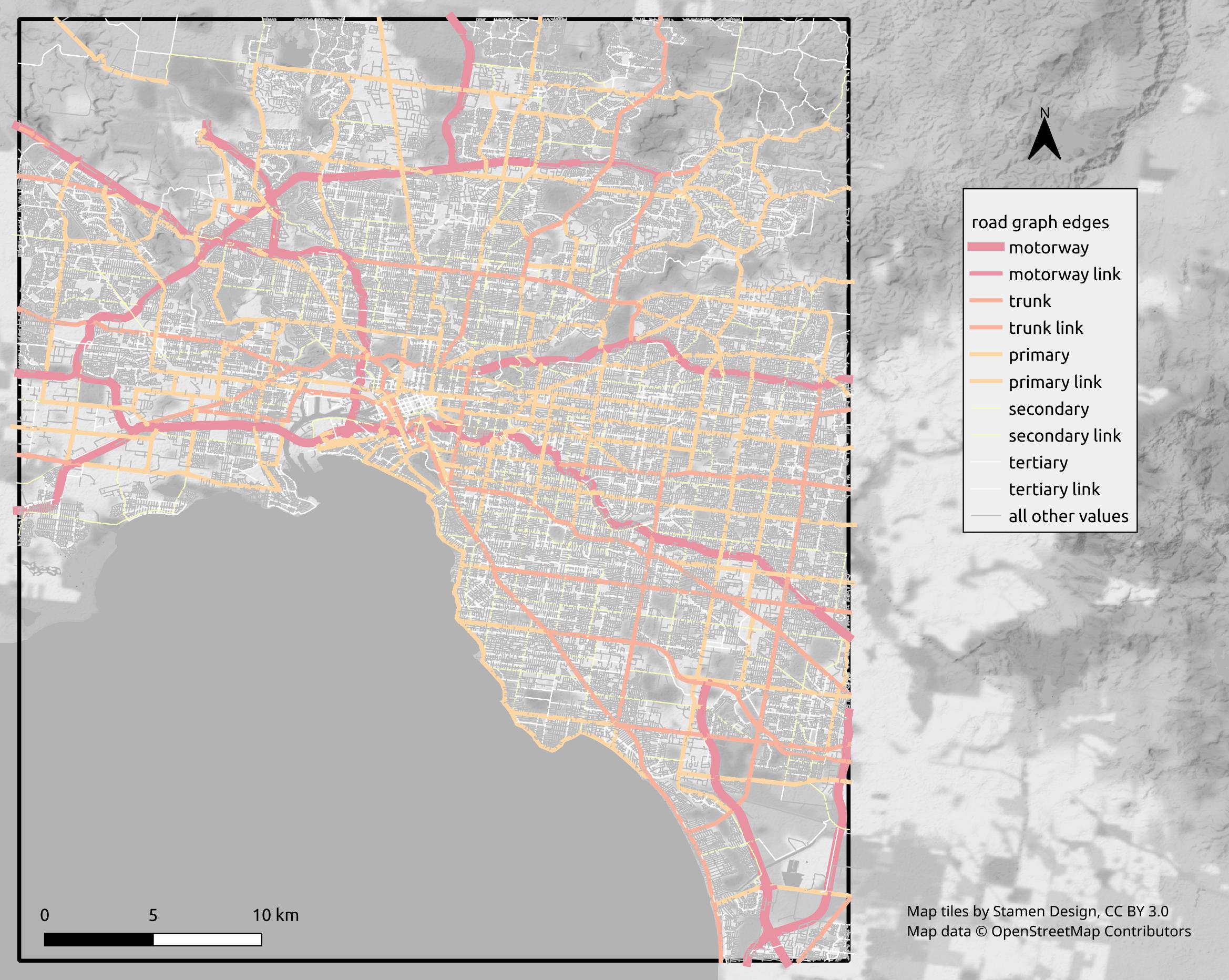}
\caption[Road graph Melbourne]{Road graph Melbourne, OSM color scheme (2022).}
\label{figures/speed_stats/road_graph_melbourne_2022.jpg}
\end{figure}
\subsubsection{Static data  Melbourne  (2022) }
\mbox{}\nopagebreak
\begin{small}
\begin{longtable}{p{4cm}rrrrrrrr}
\toprule
Attribute      & {mean} &{std} & {median}  & {q01} & {q99} & {data points} & {sum}  \\
\midrule
 bounding box                &  &  &  &  &  &  144.757--145.193 / -38.106---37.611 &                                                \\
 num\_edges                &  &  &  &  &  &  230'654 &                                                \\
 \hspace{10pt}  motorway               &  &  &  &  &  &  354 &                                                \\
 \hspace{10pt}  motorway\_link               &  &  &  &  &  &  896 &                                                \\
 \hspace{10pt}  trunk               &  &  &  &  &  &  5382 &                                                \\
 \hspace{10pt}  trunk\_link               &  &  &  &  &  &  762 &                                                \\
 \hspace{10pt}  primary               &  &  &  &  &  &  13913 &                                                \\
 \hspace{10pt}  primary\_link               &  &  &  &  &  &  1574 &                                                \\
 \hspace{10pt}  secondary               &  &  &  &  &  &  10342 &                                                \\
 \hspace{10pt}  secondary\_link               &  &  &  &  &  &  394 &                                                \\
 \hspace{10pt}  tertiary               &  &  &  &  &  &  30557 &                                                \\
 \hspace{10pt}  tertiary\_link               &  &  &  &  &  &  1001 &                                                \\
 \hspace{10pt}  unclassified               &  &  &  &  &  &  7381 &                                                \\
 \hspace{10pt}  residential               &  &  &  &  &  &  158098 &                                                \\
 num\_nodes                &  &  &  &  &  &  103062 &                                                 \\
 num\_edges\_per\_cell                & 1.1 & 0.4 & 1.0 & 1.0 & 3.0 &  891'475 &                                                 \\
 num\_intersecting\_cells                & 4.2 & 3.1 & 4.0 & 1.0 & 14.0 &  230'654 &                                                 \\
 node\_degree                & 2.8 & 0.8 & 3.0 & 1.0 & 4.0 &  103'062 &                                                 \\
 length\_meters                & 105.3 & 123.1 & 81.5 & 2.7 & 497.8 &  230'654 & 2.4e+07                                                \\
 \hspace{10pt}  motorway               & 1'102.6 & 774.5 & 976.5 & 26.1 & 4'090.9 &  354  & 3.9e+05                                                \\
 \hspace{10pt}  motorway\_link               & 260.7 & 318.6 & 102.0 & 10.2 & 1'524.8 &  896  & 2.3e+05                                                \\
 \hspace{10pt}  trunk               & 105.4 & 139.8 & 68.1 & 3.9 & 657.1 &  5'382  & 5.7e+05                                                \\
 \hspace{10pt}  trunk\_link               & 36.0 & 39.8 & 22.8 & 7.8 & 182.3 &  762  & 2.7e+04                                                \\
 \hspace{10pt}  primary               & 99.3 & 115.6 & 69.4 & 3.0 & 564.1 &  13'913  & 1.4e+06                                                \\
 \hspace{10pt}  primary\_link               & 30.5 & 34.3 & 17.0 & 6.2 & 146.5 &  1'574  & 4.8e+04                                                \\
 \hspace{10pt}  secondary               & 86.5 & 104.6 & 64.2 & 2.3 & 457.6 &  10'342  & 9.0e+05                                                \\
 \hspace{10pt}  secondary\_link               & 40.3 & 28.6 & 37.1 & 5.1 & 124.3 &  394  & 1.6e+04                                                \\
 \hspace{10pt}  tertiary               & 78.5 & 111.2 & 54.4 & 1.6 & 421.8 &  30'557  & 2.4e+06                                                \\
 \hspace{10pt}  tertiary\_link               & 22.9 & 24.5 & 12.7 & 2.0 & 105.6 &  1'001  & 2.3e+04                                                \\
 \hspace{10pt}  unclassified               & 155.5 & 235.7 & 91.0 & 3.7 & 1'068.5 &  7'381  & 1.1e+06                                                \\
 \hspace{10pt}  residential               & 108.5 & 99.3 & 87.6 & 3.1 & 443.9 &  158'098  & 1.7e+07                                                \\
 speed\_kph                & 51.1 & 6.8 & 48.7 & 40.0 & 80.0 &  230'654 &                                                 \\
 \hspace{10pt}  motorway               & 93.6 & 9.5 & 100.0 & 80.0 & 100.0 &  354  &                                                 \\
 \hspace{10pt}  motorway\_link               & 77.2 & 11.0 & 78.9 & 50.0 & 100.0 &  896  &                                                 \\
 \hspace{10pt}  trunk               & 68.6 & 9.5 & 70.0 & 40.0 & 80.0 &  5'382  &                                                 \\
 \hspace{10pt}  trunk\_link               & 65.2 & 5.0 & 65.4 & 50.0 & 80.0 &  762  &                                                 \\
 \hspace{10pt}  primary               & 62.5 & 8.3 & 60.0 & 40.0 & 80.0 &  13'913  &                                                 \\
 \hspace{10pt}  primary\_link               & 57.1 & 3.3 & 57.0 & 40.0 & 70.0 &  1'574  &                                                 \\
 \hspace{10pt}  secondary               & 58.8 & 5.9 & 60.0 & 40.0 & 80.0 &  10'342  &                                                 \\
 \hspace{10pt}  secondary\_link               & 59.4 & 2.5 & 59.6 & 50.0 & 70.0 &  394  &                                                 \\
 \hspace{10pt}  tertiary               & 51.4 & 5.8 & 51.4 & 40.0 & 70.0 &  30'557  &                                                 \\
 \hspace{10pt}  tertiary\_link               & 54.4 & 3.0 & 54.6 & 40.0 & 60.0 &  1'001  &                                                 \\
 \hspace{10pt}  unclassified               & 49.1 & 4.8 & 49.0 & 20.0 & 60.0 &  7'381  &                                                 \\
 \hspace{10pt}  residential               & 48.7 & 2.4 & 48.7 & 40.0 & 50.0 &  158'098  &                                                 \\
 free\_flow\_kph                & 37.7 & 18.3 & 37.2 & 0.0 & 82.8 &  190'471 &                                                 \\
 \hspace{10pt}  motorway               & 87.6 & 15.0 & 94.0 & 16.7 & 99.8 &  354  &                                                 \\
 \hspace{10pt}  motorway\_link               & 73.0 & 23.1 & 80.0 & 16.9 & 98.8 &  894  &                                                 \\
 \hspace{10pt}  trunk               & 54.9 & 13.1 & 56.0 & 25.3 & 77.9 &  5'381  &                                                 \\
 \hspace{10pt}  trunk\_link               & 44.8 & 20.3 & 46.2 & 0.0 & 80.5 &  748  &                                                 \\
 \hspace{10pt}  primary               & 49.8 & 12.9 & 51.8 & 24.0 & 80.0 &  13'906  &                                                 \\
 \hspace{10pt}  primary\_link               & 45.2 & 19.1 & 44.7 & 1.1 & 92.5 &  1'527  &                                                 \\
 \hspace{10pt}  secondary               & 47.6 & 12.6 & 49.9 & 19.6 & 85.6 &  10'326  &                                                 \\
 \hspace{10pt}  secondary\_link               & 41.7 & 14.5 & 40.9 & 12.3 & 82.6 &  385  &                                                 \\
 \hspace{10pt}  tertiary               & 39.5 & 12.2 & 39.1 & 10.1 & 73.9 &  29'809  &                                                 \\
 \hspace{10pt}  tertiary\_link               & 37.6 & 15.4 & 36.7 & 0.0 & 74.6 &  930  &                                                 \\
 \hspace{10pt}  unclassified               & 36.0 & 18.6 & 32.2 & 0.0 & 92.2 &  7'035  &                                                 \\
 \hspace{10pt}  residential               & 33.7 & 18.6 & 32.9 & 0.0 & 80.0 &  119'176  &                                                 \\
 free\_flow\_kph-speed\_kph                & -13.9 & 17.0 & -13.8 & -50.0 & 29.9 &  190'471 &                                                 \\
 \hspace{10pt}  motorway               & -6.0 & 14.9 & -2.1 & -80.1 & 17.1 &  354  &                                                 \\
 \hspace{10pt}  motorway\_link               & -4.1 & 23.3 & -1.2 & -62.0 & 37.9 &  894  &                                                 \\
 \hspace{10pt}  trunk               & -13.7 & 12.1 & -10.4 & -47.4 & 6.4 &  5'381  &                                                 \\
 \hspace{10pt}  trunk\_link               & -20.4 & 20.5 & -18.8 & -65.4 & 22.9 &  748  &                                                 \\
 \hspace{10pt}  primary               & -12.6 & 12.3 & -9.6 & -43.9 & 18.6 &  13'906  &                                                 \\
 \hspace{10pt}  primary\_link               & -11.9 & 19.1 & -11.5 & -55.9 & 31.3 &  1'527  &                                                 \\
 \hspace{10pt}  secondary               & -11.2 & 12.4 & -9.2 & -39.5 & 30.6 &  10'326  &                                                 \\
 \hspace{10pt}  secondary\_link               & -17.8 & 14.8 & -18.7 & -49.4 & 23.0 &  385  &                                                 \\
 \hspace{10pt}  tertiary               & -11.9 & 11.7 & -11.9 & -42.5 & 19.2 &  29'809  &                                                 \\
 \hspace{10pt}  tertiary\_link               & -16.8 & 15.3 & -17.5 & -54.6 & 20.2 &  930  &                                                 \\
 \hspace{10pt}  unclassified               & -13.1 & 18.9 & -16.5 & -49.0 & 42.7 &  7'035  &                                                 \\
 \hspace{10pt}  residential               & -14.8 & 18.7 & -15.5 & -50.0 & 31.9 &  119'176  &                                                 \\
\bottomrule

        \caption[Key figures Melbourne ]{Key figures Melbourne for the generated data from 20 randomly sampled days.
        \textbf{num\_edges} number of edges in the street network graph;
        \textbf{num\_nodes} number of nodes in the street network graph;
        \textbf{num\_edges\_per\_cell} number of edges a cell (row,col,heading) has in its intersecting cells;
        \textbf{num\_intersecting\_cells} number of cells (row,col,heading) in an edge's intersecting cells;
        \textbf{node\_degree} number of (unique) neighbor nodes per node;
        \textbf{length\_meters} free flow speed derived from data;
        \textbf{speed\_kph} signalled speed;
        \textbf{free\_flow\_kph} free flow speed derived from data;
        \textbf{free\_flow\_kph-speed\_kph} difference
        }
    \label{tab:key_figures:/iarai/public/t4c/data_pipeline/release20221026_residential_unclassified/2022:Melbourne:}
    \end{longtable}
    \end{small}
    
\subsubsection{Segment density map  Melbourne (2022)}
\mbox{}
\nopagebreak{}
\begin{figure}[H]
\centering
\includegraphics[width=0.85\textwidth]{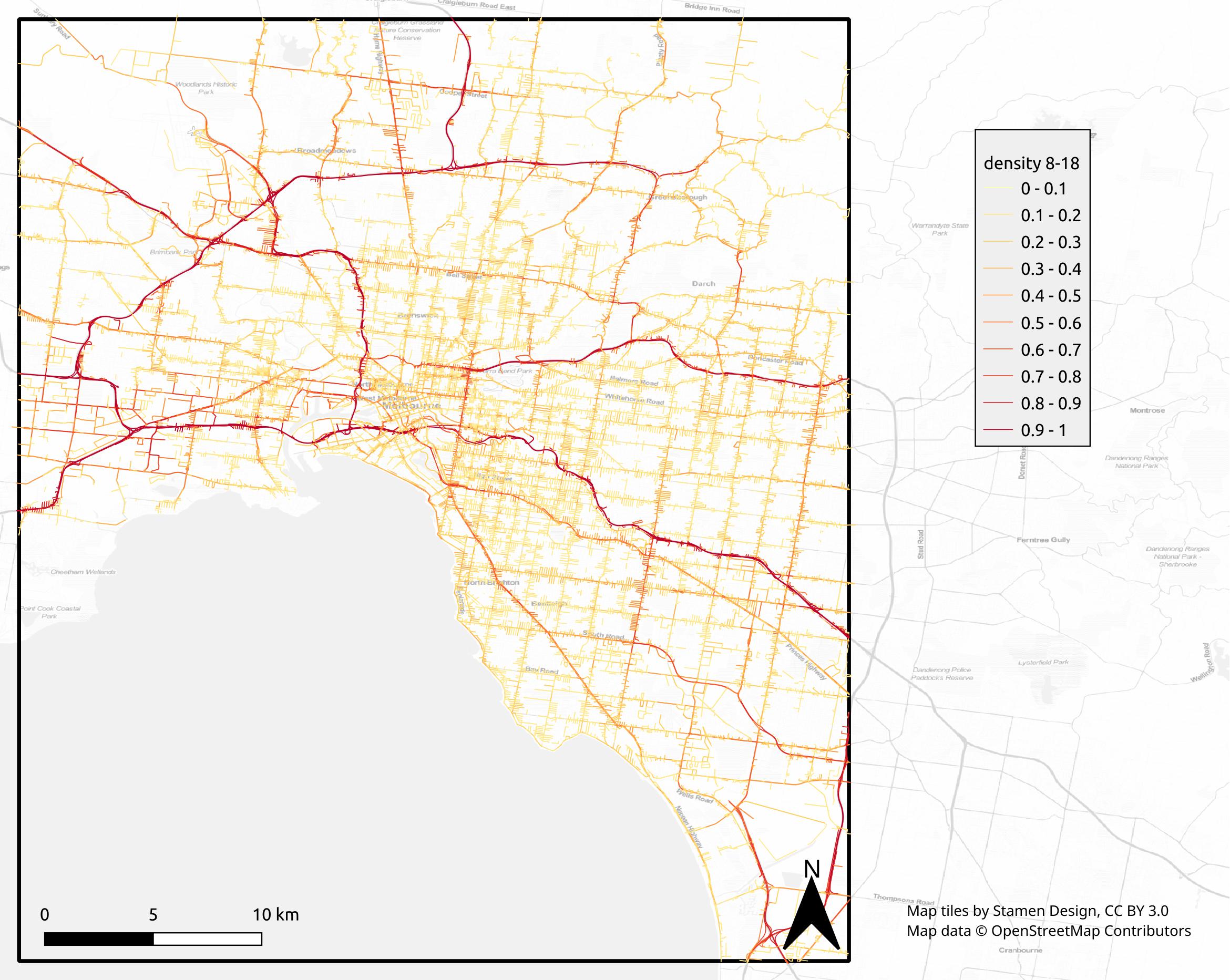}
\caption[Segment-wise density 8am--6pm Melbourne]{Segment-wise density 8am--6pm Melbourne from 20 randomly sampled days.}
\label{figures/speed_stats/density_8_18_melbourne_2022.jpg}
\end{figure}
\clearpage
\subsubsection{Daily density profile  Melbourne  (2022) } 
\mbox{}
\nopagebreak{}
\begin{figure}[H]
\centering
\includegraphics[width=0.85\textwidth]{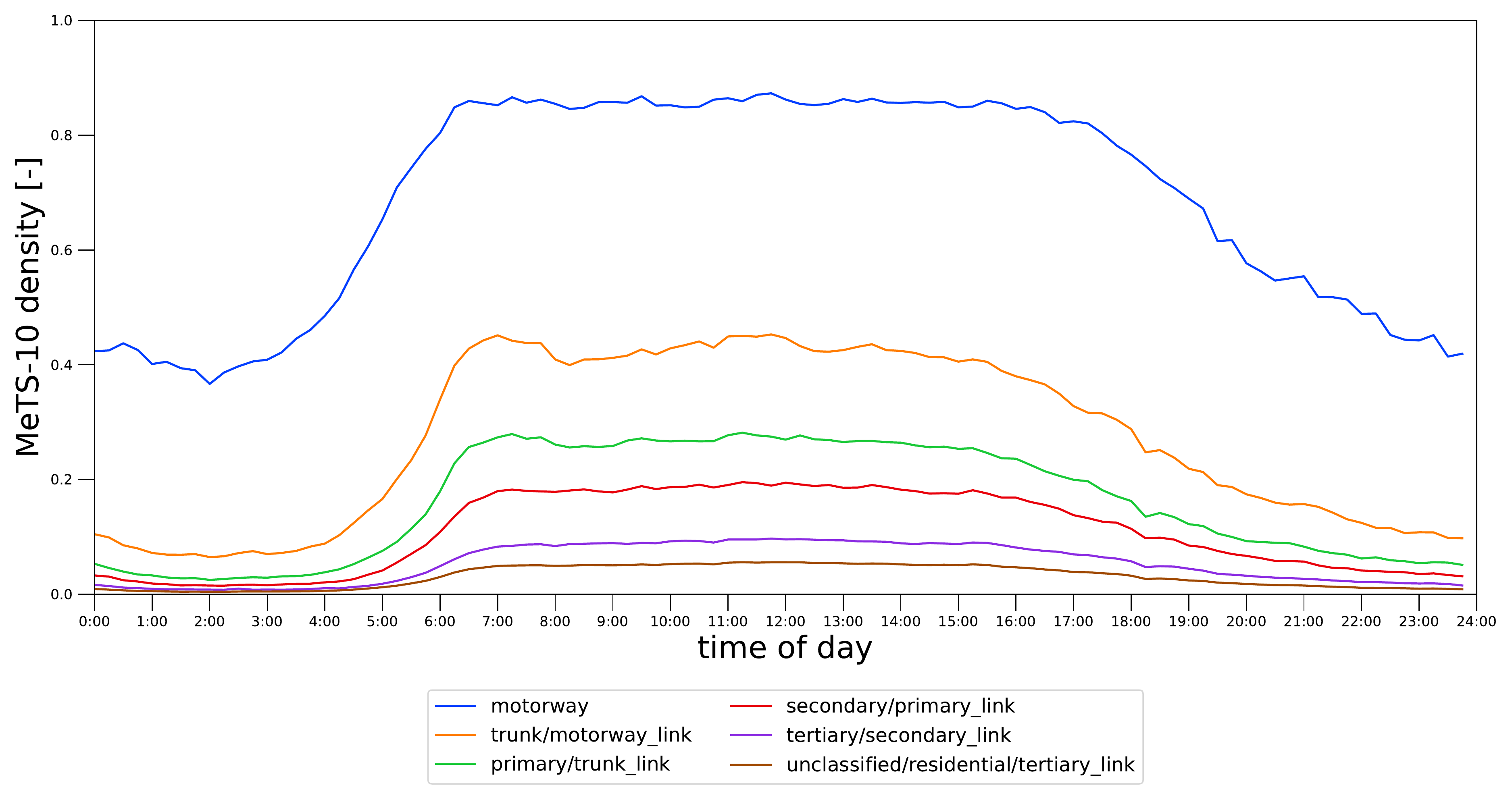}
\caption[Daily density profile Melbourne]{Daily density profile for different road types for Melbourne . Data from 20 randomly sampled days.}
\label{figures/speed_stats/speed_stats_coverage_melbourne_2022_by_highway.pdf}
\end{figure}
\subsubsection{Daily speed profile  Melbourne  (2022) }
\mbox{}
\nopagebreak{}
\begin{figure}[H]
\centering
\includegraphics[width=0.85\textwidth]{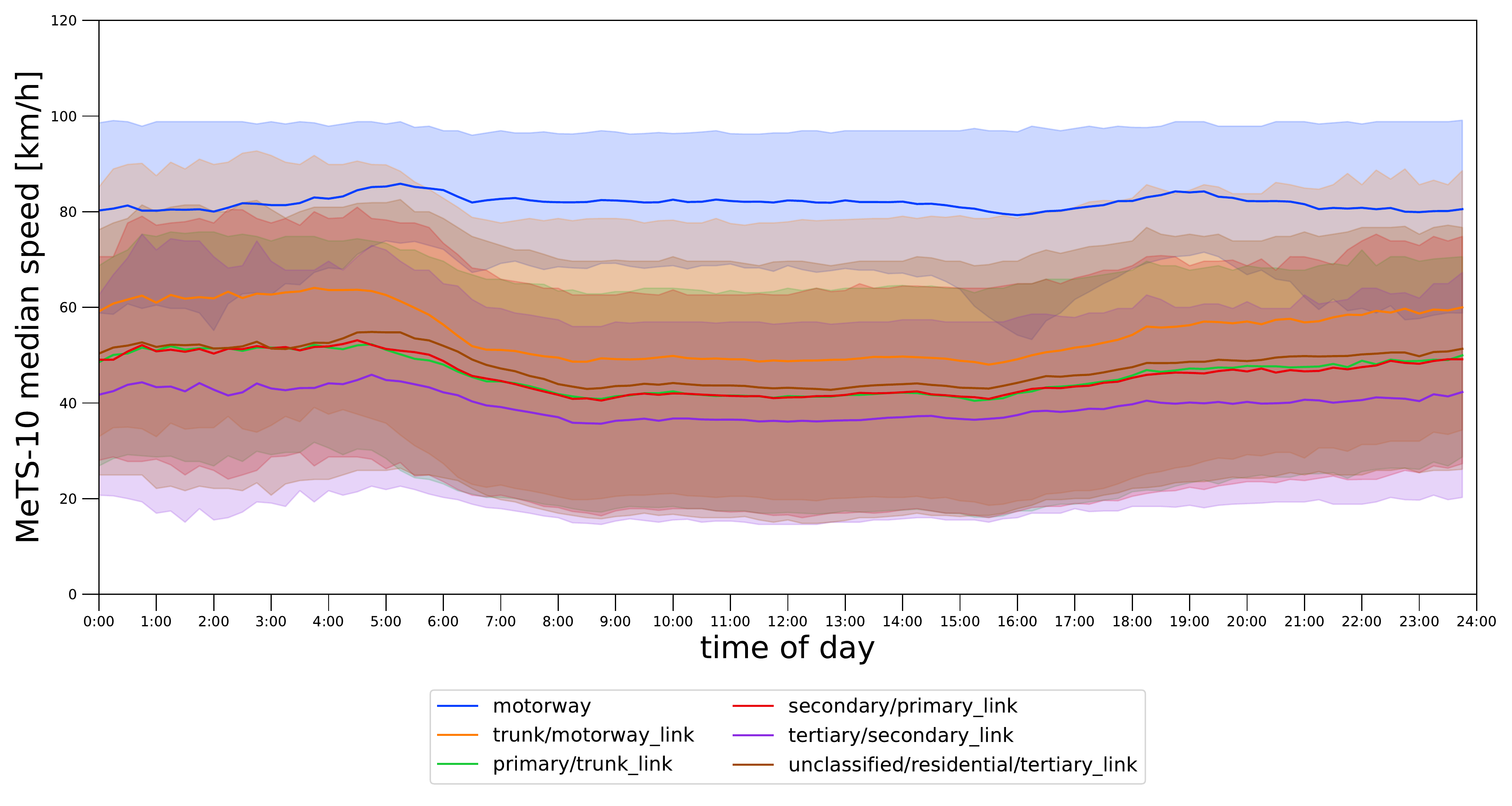}
\caption[Daily median 15 min speeds of all intersecting cells profile Melbourne]{Daily median 15 min speeds of all intersecting cells profile for different road types for Melbourne . The error hull is the 80\% data interval [10.0--90.0 percentiles] of daily means from 20 randomly sampled days.}
\label{figures/speed_stats/speed_stats_median_speed_kph_melbourne_2022_by_highway.pdf}
\end{figure}
\clearpage

\pagebreak

\section{Key Figures Uber Validation Historic Road Graph}\label{appendix:key_figures_uber}

\subsection{Key Figures London}
\subsubsection{Road graph map London}
\mbox{}
\nopagebreak{}
\begin{figure}[H]
\centering
\includegraphics[width=0.85\textwidth]{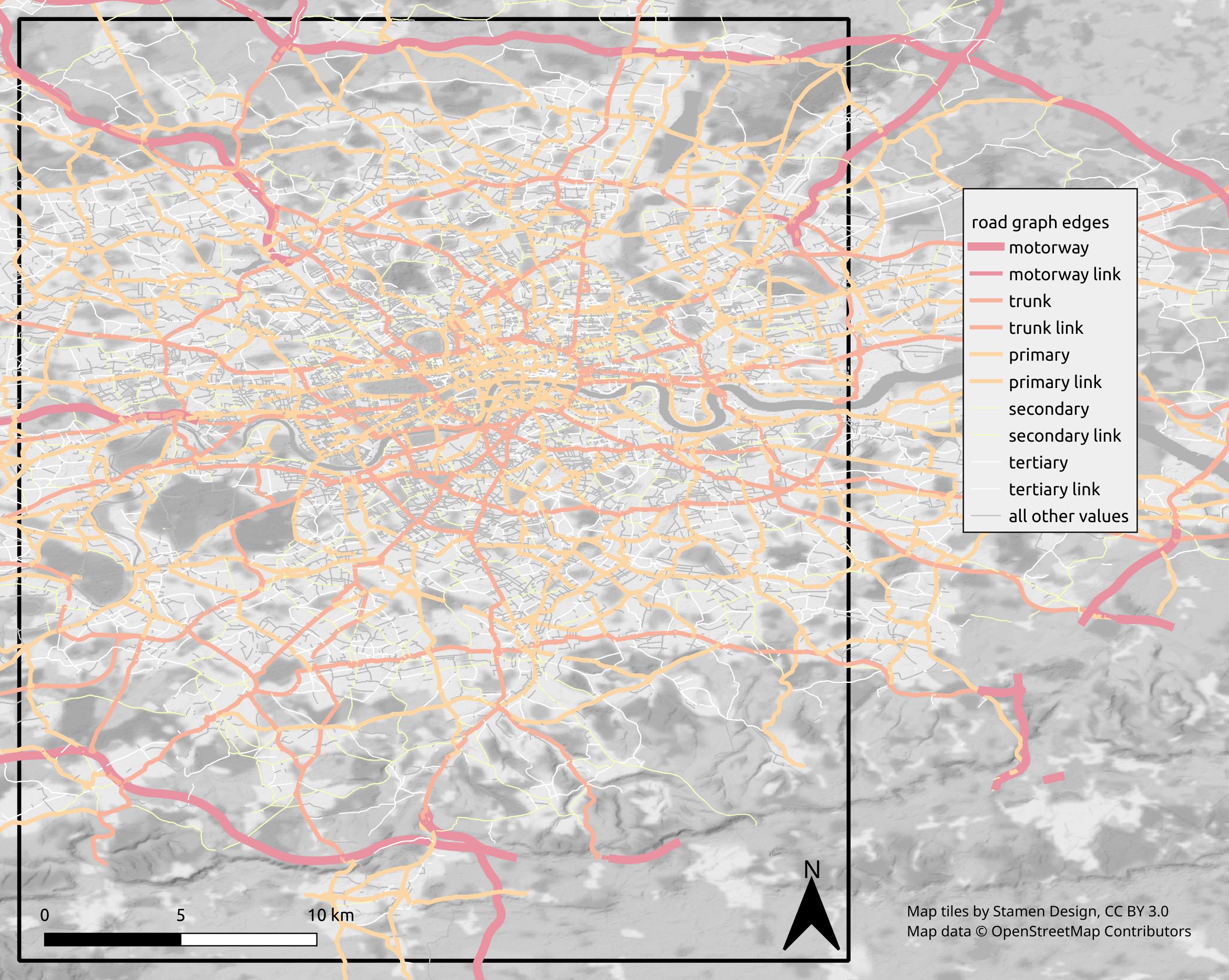}
\caption[Road graph London]{Road graph London, OSM color scheme.}
\label{figures/speed_stats_05_val01_uber/road_graph_london.jpg}
\end{figure}
\subsubsection{Static data  London   (full historic road graph)}
\mbox{}\nopagebreak
\begin{small}
\begin{longtable}{p{4cm}rrrrrrrr}
\toprule
Attribute      & {mean} &{std} & {median}  & {q01} & {q99} & {data points} & {sum}  \\
\midrule
 bounding box  (full historic road graph)               &  &  &  &  &  &  -0.369--0.067 / 51.205--51.7 &                                                \\
 num\_edges  (full historic road graph)               &  &  &  &  &  &  234'308 &                                                \\
 \hspace{10pt}  motorway               &  &  &  &  &  &  1520 &                                                \\
 \hspace{10pt}  motorway\_link               &  &  &  &  &  &  894 &                                                \\
 \hspace{10pt}  trunk               &  &  &  &  &  &  17347 &                                                \\
 \hspace{10pt}  trunk\_link               &  &  &  &  &  &  1600 &                                                \\
 \hspace{10pt}  primary               &  &  &  &  &  &  62005 &                                                \\
 \hspace{10pt}  primary\_link               &  &  &  &  &  &  907 &                                                \\
 \hspace{10pt}  secondary               &  &  &  &  &  &  28003 &                                                \\
 \hspace{10pt}  secondary\_link               &  &  &  &  &  &  135 &                                                \\
 \hspace{10pt}  tertiary               &  &  &  &  &  &  59977 &                                                \\
 \hspace{10pt}  tertiary\_link               &  &  &  &  &  &  257 &                                                \\
 \hspace{10pt}  unclassified               &  &  &  &  &  &  12447 &                                                \\
 \hspace{10pt}  residential               &  &  &  &  &  &  49110 &                                                \\
 \hspace{10pt}  living\_street               &  &  &  &  &  &  58 &                                                \\
 \hspace{10pt}  service               &  &  &  &  &  &  44 &                                                \\
 \hspace{10pt}  cycleway               &  &  &  &  &  &  1 &                                                \\
 \hspace{10pt}  road               &  &  &  &  &  &  2 &                                                \\
 \hspace{10pt}  construction               &  &  &  &  &  &  1 &                                                \\
 num\_nodes  (full historic road graph)               &  &  &  &  &  &  140412 &                                                 \\
 num\_edges\_per\_cell  (full historic road graph)               & 1.1 & 0.4 & 1.0 & 1.0 & 3.0 &  355'950 &                                                 \\
 num\_intersecting\_cells  (full historic road graph)               & 1.7 & 2.2 & 1.0 & 0.0 & 9.0 &  234'308 &                                                 \\
 node\_degree  (full historic road graph)               & 2.2 & 0.6 & 2.0 & 1.0 & 4.0 &  140'412 &                                                 \\
 length\_meters  (full historic road graph)               & 108.2 & 172.9 & 65.6 & 4.7 & 748.0 &  234'308 & 2.5e+07                                                \\
 \hspace{10pt}  motorway               & 691.1 & 1'030.4 & 286.2 & 19.1 & 4'846.2 &  1'520  & 1.1e+06                                                \\
 \hspace{10pt}  motorway\_link               & 293.3 & 308.7 & 182.6 & 14.3 & 1'437.8 &  894  & 2.6e+05                                                \\
 \hspace{10pt}  trunk               & 144.1 & 271.7 & 63.2 & 5.0 & 1'332.8 &  17'347  & 2.5e+06                                                \\
 \hspace{10pt}  trunk\_link               & 178.8 & 198.2 & 97.1 & 8.9 & 834.9 &  1'600  & 2.9e+05                                                \\
 \hspace{10pt}  primary               & 91.0 & 129.0 & 55.0 & 4.3 & 620.1 &  62'005  & 5.6e+06                                                \\
 \hspace{10pt}  primary\_link               & 84.0 & 104.8 & 48.8 & 5.2 & 572.8 &  907  & 7.6e+04                                                \\
 \hspace{10pt}  secondary               & 104.5 & 143.8 & 63.2 & 4.6 & 721.5 &  28'003  & 2.9e+06                                                \\
 \hspace{10pt}  secondary\_link               & 54.7 & 60.9 & 33.6 & 6.3 & 294.2 &  135  & 7.4e+03                                                \\
 \hspace{10pt}  tertiary               & 104.9 & 129.4 & 68.5 & 4.8 & 628.2 &  59'977  & 6.3e+06                                                \\
 \hspace{10pt}  tertiary\_link               & 77.3 & 91.4 & 47.6 & 4.1 & 426.4 &  257  & 2.0e+04                                                \\
 \hspace{10pt}  unclassified               & 104.1 & 152.4 & 60.9 & 4.4 & 776.5 &  12'447  & 1.3e+06                                                \\
 \hspace{10pt}  residential               & 101.6 & 94.8 & 76.6 & 5.0 & 469.3 &  49'110  & 5.0e+06                                                \\
 \hspace{10pt}  living\_street               & 68.2 & 52.7 & 49.7 & 4.8 & 207.9 &  58  & 4.0e+03                                                \\
 \hspace{10pt}  service               & 66.6 & 83.8 & 30.0 & 4.7 & 337.7 &  44  & 2.9e+03                                                \\
 \hspace{10pt}  cycleway               & 31.0 & nan & 31.0 & 31.0 & 31.0 &  1  & 3.1e+01                                                \\
 \hspace{10pt}  road               & 45.8 & 1.8 & 45.8 & 44.6 & 47.1 &  2  & 9.2e+01                                                \\
 \hspace{10pt}  construction               & 14.8 & nan & 14.8 & 14.8 & 14.8 &  1  & 1.5e+01                                                \\
 speed\_kph  (full historic road graph)               & 47.3 & 16.1 & 48.3 & 32.2 & 112.7 &  234'308 &                                                 \\
 \hspace{10pt}  motorway               & 110.6 & 8.3 & 112.7 & 64.4 & 112.7 &  1'520  &                                                 \\
 \hspace{10pt}  motorway\_link               & 106.0 & 14.2 & 112.7 & 64.4 & 112.7 &  894  &                                                 \\
 \hspace{10pt}  trunk               & 62.1 & 21.5 & 48.3 & 32.2 & 112.7 &  17'347  &                                                 \\
 \hspace{10pt}  trunk\_link               & 72.9 & 21.6 & 72.8 & 32.2 & 112.7 &  1'600  &                                                 \\
 \hspace{10pt}  primary               & 50.4 & 13.5 & 48.3 & 32.2 & 96.6 &  62'005  &                                                 \\
 \hspace{10pt}  primary\_link               & 54.2 & 13.9 & 48.3 & 32.2 & 96.6 &  907  &                                                 \\
 \hspace{10pt}  secondary               & 48.5 & 14.2 & 48.3 & 32.2 & 96.6 &  28'003  &                                                 \\
 \hspace{10pt}  secondary\_link               & 47.9 & 14.2 & 48.3 & 32.2 & 91.1 &  135  &                                                 \\
 \hspace{10pt}  tertiary               & 46.2 & 12.9 & 46.2 & 32.2 & 96.6 &  59'977  &                                                 \\
 \hspace{10pt}  tertiary\_link               & 52.2 & 12.7 & 52.0 & 32.2 & 96.6 &  257  &                                                 \\
 \hspace{10pt}  unclassified               & 40.4 & 10.3 & 40.4 & 32.2 & 96.6 &  12'447  &                                                 \\
 \hspace{10pt}  residential               & 36.3 & 6.1 & 32.2 & 32.2 & 48.3 &  49'110  &                                                 \\
 \hspace{10pt}  living\_street               & 33.1 & 3.6 & 32.2 & 32.2 & 48.3 &  58  &                                                 \\
 \hspace{10pt}  service               & 33.8 & 3.3 & 33.6 & 32.2 & 48.3 &  44  &                                                 \\
 \hspace{10pt}  cycleway               & 32.2 & nan & 32.2 & 32.2 & 32.2 &  1  &                                                 \\
 \hspace{10pt}  road               & 52.8 & 0.0 & 52.8 & 52.8 & 52.8 &  2  &                                                 \\
 \hspace{10pt}  construction               & 32.2 & nan & 32.2 & 32.2 & 32.2 &  1  &                                                 \\
 free\_flow\_kph  (full historic road graph)               & 34.0 & 11.8 & 32.0 & 15.3 & 77.2 &  136'149 &                                                 \\
 \hspace{10pt}  motorway               & 103.2 & 14.4 & 109.2 & 59.8 & 117.6 &  311  &                                                 \\
 \hspace{10pt}  motorway\_link               & 89.0 & 21.2 & 94.8 & 41.5 & 117.5 &  125  &                                                 \\
 \hspace{10pt}  trunk               & 41.2 & 14.2 & 38.1 & 20.2 & 85.6 &  11'928  &                                                 \\
 \hspace{10pt}  trunk\_link               & 54.7 & 19.7 & 57.9 & 18.7 & 102.0 &  779  &                                                 \\
 \hspace{10pt}  primary               & 35.8 & 10.0 & 32.9 & 18.4 & 67.4 &  33'921  &                                                 \\
 \hspace{10pt}  primary\_link               & 44.0 & 19.0 & 38.6 & 17.9 & 80.4 &  511  &                                                 \\
 \hspace{10pt}  secondary               & 35.1 & 10.0 & 32.9 & 17.9 & 67.8 &  14'742  &                                                 \\
 \hspace{10pt}  secondary\_link               & 32.2 & 14.7 & 28.7 & 17.8 & 78.1 &  88  &                                                 \\
 \hspace{10pt}  tertiary               & 34.6 & 10.2 & 32.5 & 17.9 & 67.8 &  28'749  &                                                 \\
 \hspace{10pt}  tertiary\_link               & 48.9 & 20.0 & 46.6 & 19.8 & 78.6 &  134  &                                                 \\
 \hspace{10pt}  unclassified               & 28.9 & 11.7 & 27.1 & 11.8 & 75.8 &  7'505  &                                                 \\
 \hspace{10pt}  residential               & 28.8 & 8.4 & 27.8 & 12.7 & 54.6 &  37'281  &                                                 \\
 \hspace{10pt}  living\_street               & 23.1 & 4.3 & 24.9 & 12.9 & 28.6 &  53  &                                                 \\
 \hspace{10pt}  service               & 28.0 & 18.2 & 19.8 & 13.4 & 76.8 &  19  &                                                 \\
 \hspace{10pt}  road               & 26.4 & 0.7 & 26.4 & 25.9 & 26.8 &  2  &                                                 \\
 \hspace{10pt}  construction               & 14.1 & nan & 14.1 & 14.1 & 14.1 &  1  &                                                 \\
 free\_flow\_kph-speed\_kph  (full historic road graph)               & -7.6 & 10.1 & -6.3 & -31.4 & 18.6 &  136'149 &                                                 \\
 \hspace{10pt}  motorway               & -4.4 & 12.0 & -2.6 & -46.4 & 23.9 &  311  &                                                 \\
 \hspace{10pt}  motorway\_link               & -12.9 & 19.1 & -7.8 & -64.9 & 12.7 &  125  &                                                 \\
 \hspace{10pt}  trunk               & -11.5 & 9.9 & -11.3 & -39.2 & 10.0 &  11'928  &                                                 \\
 \hspace{10pt}  trunk\_link               & -6.4 & 18.2 & -5.5 & -48.2 & 32.5 &  779  &                                                 \\
 \hspace{10pt}  primary               & -9.2 & 10.1 & -7.8 & -30.9 & 13.6 &  33'921  &                                                 \\
 \hspace{10pt}  primary\_link               & -5.8 & 16.7 & -5.8 & -37.6 & 29.8 &  511  &                                                 \\
 \hspace{10pt}  secondary               & -7.1 & 10.1 & -5.4 & -31.5 & 14.4 &  14'742  &                                                 \\
 \hspace{10pt}  secondary\_link               & -10.2 & 14.5 & -10.3 & -30.4 & 29.5 &  88  &                                                 \\
 \hspace{10pt}  tertiary               & -6.0 & 10.2 & -4.5 & -29.5 & 20.9 &  28'749  &                                                 \\
 \hspace{10pt}  tertiary\_link               & -0.3 & 19.6 & -3.5 & -31.1 & 36.8 &  134  &                                                 \\
 \hspace{10pt}  unclassified               & -7.9 & 11.5 & -8.2 & -32.8 & 31.5 &  7'505  &                                                 \\
 \hspace{10pt}  residential               & -6.3 & 8.8 & -5.8 & -29.7 & 17.5 &  37'281  &                                                 \\
 \hspace{10pt}  living\_street               & -10.1 & 5.2 & -8.0 & -21.9 & -3.6 &  53  &                                                 \\
 \hspace{10pt}  service               & -4.5 & 18.1 & -12.4 & -19.6 & 44.6 &  19  &                                                 \\
 \hspace{10pt}  road               & -26.4 & 0.7 & -26.4 & -26.9 & -26.0 &  2  &                                                 \\
 \hspace{10pt}  construction               & -18.1 & nan & -18.1 & -18.1 & -18.1 &  1  &                                                 \\
\bottomrule

        \caption[Key figures London  (full historic road graph)]{Key figures London for the generated data from 20 randomly sampled days (full historic road graph).
        \textbf{num\_edges} number of edges in the street network graph;
        \textbf{num\_nodes} number of nodes in the street network graph;
        \textbf{num\_edges\_per\_cell} number of edges a cell (row,col,heading) has in its intersecting cells;
        \textbf{num\_intersecting\_cells} number of cells (row,col,heading) in an edge's intersecting cells;
        \textbf{node\_degree} number of (unique) neighbor nodes per node;
        \textbf{length\_meters} free flow speed derived from data;
        \textbf{speed\_kph} signalled speed;
        \textbf{free\_flow\_kph} free flow speed derived from data;
        \textbf{free\_flow\_kph-speed\_kph} difference
        }
    \label{tab:key_figures:/iarai/public/t4c/data_pipeline/release20221028_historic_uber:London: (full historic road graph)}
    \end{longtable}
    \end{small}
    
\subsubsection{Static data  London   (MeTS-10 extent (bounding box))}
\mbox{}\nopagebreak
\begin{small}
\begin{longtable}{p{4cm}rrrrrrrr}
\toprule
Attribute      & {mean} &{std} & {median}  & {q01} & {q99} & {data points} & {sum}  \\
\midrule
 bounding box  (MeTS-10 extent (bounding box))               &  &  &  &  &  &  -0.369--0.067 / 51.205--51.7 &                                                \\
 num\_edges  (MeTS-10 extent (bounding box))               &  &  &  &  &  &  136'138 &                                                \\
 \hspace{10pt}  motorway               &  &  &  &  &  &  311 &                                                \\
 \hspace{10pt}  motorway\_link               &  &  &  &  &  &  125 &                                                \\
 \hspace{10pt}  trunk               &  &  &  &  &  &  11928 &                                                \\
 \hspace{10pt}  trunk\_link               &  &  &  &  &  &  779 &                                                \\
 \hspace{10pt}  primary               &  &  &  &  &  &  33917 &                                                \\
 \hspace{10pt}  primary\_link               &  &  &  &  &  &  511 &                                                \\
 \hspace{10pt}  secondary               &  &  &  &  &  &  14738 &                                                \\
 \hspace{10pt}  secondary\_link               &  &  &  &  &  &  88 &                                                \\
 \hspace{10pt}  tertiary               &  &  &  &  &  &  28749 &                                                \\
 \hspace{10pt}  tertiary\_link               &  &  &  &  &  &  134 &                                                \\
 \hspace{10pt}  unclassified               &  &  &  &  &  &  7506 &                                                \\
 \hspace{10pt}  residential               &  &  &  &  &  &  37277 &                                                \\
 \hspace{10pt}  living\_street               &  &  &  &  &  &  53 &                                                \\
 \hspace{10pt}  service               &  &  &  &  &  &  19 &                                                \\
 \hspace{10pt}  road               &  &  &  &  &  &  2 &                                                \\
 \hspace{10pt}  construction               &  &  &  &  &  &  1 &                                                \\
 num\_nodes  (MeTS-10 extent (bounding box))               &  &  &  &  &  &  77427 &                                                 \\
 num\_edges\_per\_cell  (MeTS-10 extent (bounding box))               & 1.1 & 0.4 & 1.0 & 1.0 & 3.0 &  355'931 &                                                 \\
 num\_intersecting\_cells  (MeTS-10 extent (bounding box))               & 2.9 & 2.2 & 2.0 & 1.0 & 10.0 &  136'138 &                                                 \\
 node\_degree  (MeTS-10 extent (bounding box))               & 2.2 & 0.6 & 2.0 & 1.0 & 4.0 &  77'427 &                                                 \\
 length\_meters  (MeTS-10 extent (bounding box))               & 92.4 & 124.6 & 62.4 & 4.6 & 521.7 &  136'138 & 1.3e+07                                                \\
 \hspace{10pt}  motorway               & 829.7 & 1'142.2 & 358.1 & 39.0 & 5'045.8 &  311  & 2.6e+05                                                \\
 \hspace{10pt}  motorway\_link               & 429.7 & 421.9 & 343.9 & 30.6 & 1'768.6 &  125  & 5.4e+04                                                \\
 \hspace{10pt}  trunk               & 102.9 & 162.1 & 57.2 & 4.8 & 764.9 &  11'928  & 1.2e+06                                                \\
 \hspace{10pt}  trunk\_link               & 112.9 & 132.8 & 57.8 & 6.9 & 628.9 &  779  & 8.8e+04                                                \\
 \hspace{10pt}  primary               & 74.2 & 88.6 & 50.7 & 4.2 & 435.2 &  33'917  & 2.5e+06                                                \\
 \hspace{10pt}  primary\_link               & 71.0 & 86.0 & 42.9 & 7.1 & 408.4 &  511  & 3.6e+04                                                \\
 \hspace{10pt}  secondary               & 86.7 & 104.9 & 58.8 & 4.5 & 473.7 &  14'738  & 1.3e+06                                                \\
 \hspace{10pt}  secondary\_link               & 44.3 & 49.9 & 29.6 & 5.9 & 244.1 &  88  & 3.9e+03                                                \\
 \hspace{10pt}  tertiary               & 95.4 & 105.3 & 66.8 & 5.1 & 515.2 &  28'749  & 2.7e+06                                                \\
 \hspace{10pt}  tertiary\_link               & 75.5 & 89.4 & 46.0 & 4.6 & 443.7 &  134  & 1.0e+04                                                \\
 \hspace{10pt}  unclassified               & 85.6 & 110.3 & 56.2 & 4.2 & 540.6 &  7'506  & 6.4e+05                                                \\
 \hspace{10pt}  residential               & 99.7 & 92.2 & 75.2 & 4.9 & 461.4 &  37'277  & 3.7e+06                                                \\
 \hspace{10pt}  living\_street               & 69.9 & 54.1 & 51.9 & 4.8 & 209.1 &  53  & 3.7e+03                                                \\
 \hspace{10pt}  service               & 46.9 & 62.2 & 19.6 & 5.1 & 210.4 &  19  & 8.9e+02                                                \\
 \hspace{10pt}  road               & 45.8 & 1.8 & 45.8 & 44.6 & 47.1 &  2  & 9.2e+01                                                \\
 \hspace{10pt}  construction               & 14.8 & nan & 14.8 & 14.8 & 14.8 &  1  & 1.5e+01                                                \\
 speed\_kph  (MeTS-10 extent (bounding box))               & 41.6 & 10.9 & 40.4 & 32.2 & 80.5 &  136'138 &                                                 \\
 \hspace{10pt}  motorway               & 107.6 & 13.2 & 112.7 & 64.4 & 112.7 &  311  &                                                 \\
 \hspace{10pt}  motorway\_link               & 101.9 & 18.0 & 112.7 & 48.3 & 112.7 &  125  &                                                 \\
 \hspace{10pt}  trunk               & 52.7 & 12.4 & 48.3 & 32.2 & 96.6 &  11'928  &                                                 \\
 \hspace{10pt}  trunk\_link               & 61.1 & 14.9 & 64.4 & 32.2 & 112.7 &  779  &                                                 \\
 \hspace{10pt}  primary               & 45.0 & 9.2 & 48.3 & 32.2 & 64.4 &  33'917  &                                                 \\
 \hspace{10pt}  primary\_link               & 49.8 & 11.4 & 48.3 & 32.2 & 80.5 &  511  &                                                 \\
 \hspace{10pt}  secondary               & 42.1 & 9.8 & 48.3 & 32.2 & 64.4 &  14'738  &                                                 \\
 \hspace{10pt}  secondary\_link               & 42.4 & 9.5 & 48.1 & 32.2 & 66.5 &  88  &                                                 \\
 \hspace{10pt}  tertiary               & 40.6 & 9.2 & 46.2 & 32.2 & 64.4 &  28'749  &                                                 \\
 \hspace{10pt}  tertiary\_link               & 49.1 & 11.5 & 48.3 & 32.2 & 80.5 &  134  &                                                 \\
 \hspace{10pt}  unclassified               & 36.8 & 7.3 & 32.2 & 24.1 & 48.3 &  7'506  &                                                 \\
 \hspace{10pt}  residential               & 35.2 & 5.3 & 32.2 & 32.2 & 48.3 &  37'277  &                                                 \\
 \hspace{10pt}  living\_street               & 33.2 & 3.7 & 32.2 & 32.2 & 48.3 &  53  &                                                 \\
 \hspace{10pt}  service               & 32.6 & 0.6 & 32.2 & 32.2 & 33.6 &  19  &                                                 \\
 \hspace{10pt}  road               & 52.8 & 0.0 & 52.8 & 52.8 & 52.8 &  2  &                                                 \\
 \hspace{10pt}  construction               & 32.2 & nan & 32.2 & 32.2 & 32.2 &  1  &                                                 \\
 free\_flow\_kph  (MeTS-10 extent (bounding box))               & 34.0 & 11.8 & 32.0 & 15.3 & 77.2 &  136'137 &                                                 \\
 \hspace{10pt}  motorway               & 103.2 & 14.4 & 109.2 & 59.8 & 117.6 &  311  &                                                 \\
 \hspace{10pt}  motorway\_link               & 89.0 & 21.2 & 94.8 & 41.5 & 117.5 &  125  &                                                 \\
 \hspace{10pt}  trunk               & 41.2 & 14.2 & 38.1 & 20.2 & 85.6 &  11'928  &                                                 \\
 \hspace{10pt}  trunk\_link               & 54.7 & 19.7 & 57.9 & 18.7 & 102.0 &  779  &                                                 \\
 \hspace{10pt}  primary               & 35.8 & 10.0 & 32.9 & 18.4 & 67.4 &  33'917  &                                                 \\
 \hspace{10pt}  primary\_link               & 44.0 & 19.0 & 38.6 & 17.9 & 80.4 &  511  &                                                 \\
 \hspace{10pt}  secondary               & 35.1 & 10.0 & 32.9 & 17.9 & 67.8 &  14'738  &                                                 \\
 \hspace{10pt}  secondary\_link               & 32.2 & 14.7 & 28.7 & 17.8 & 78.1 &  88  &                                                 \\
 \hspace{10pt}  tertiary               & 34.6 & 10.2 & 32.5 & 17.9 & 67.8 &  28'749  &                                                 \\
 \hspace{10pt}  tertiary\_link               & 48.9 & 20.0 & 46.6 & 19.8 & 78.6 &  134  &                                                 \\
 \hspace{10pt}  unclassified               & 28.9 & 11.7 & 27.1 & 11.8 & 75.8 &  7'505  &                                                 \\
 \hspace{10pt}  residential               & 28.8 & 8.4 & 27.8 & 12.7 & 54.6 &  37'277  &                                                 \\
 \hspace{10pt}  living\_street               & 23.1 & 4.3 & 24.9 & 12.9 & 28.6 &  53  &                                                 \\
 \hspace{10pt}  service               & 28.0 & 18.2 & 19.8 & 13.4 & 76.8 &  19  &                                                 \\
 \hspace{10pt}  road               & 26.4 & 0.7 & 26.4 & 25.9 & 26.8 &  2  &                                                 \\
 \hspace{10pt}  construction               & 14.1 & nan & 14.1 & 14.1 & 14.1 &  1  &                                                 \\
 free\_flow\_kph-speed\_kph  (MeTS-10 extent (bounding box))               & -7.6 & 10.1 & -6.3 & -31.4 & 18.6 &  136'137 &                                                 \\
 \hspace{10pt}  motorway               & -4.4 & 12.0 & -2.6 & -46.4 & 23.9 &  311  &                                                 \\
 \hspace{10pt}  motorway\_link               & -12.9 & 19.1 & -7.8 & -64.9 & 12.7 &  125  &                                                 \\
 \hspace{10pt}  trunk               & -11.5 & 9.9 & -11.3 & -39.2 & 10.0 &  11'928  &                                                 \\
 \hspace{10pt}  trunk\_link               & -6.4 & 18.2 & -5.5 & -48.2 & 32.5 &  779  &                                                 \\
 \hspace{10pt}  primary               & -9.2 & 10.1 & -7.8 & -30.9 & 13.6 &  33'917  &                                                 \\
 \hspace{10pt}  primary\_link               & -5.8 & 16.7 & -5.8 & -37.6 & 29.8 &  511  &                                                 \\
 \hspace{10pt}  secondary               & -7.1 & 10.1 & -5.4 & -31.4 & 14.4 &  14'738  &                                                 \\
 \hspace{10pt}  secondary\_link               & -10.2 & 14.5 & -10.3 & -30.4 & 29.5 &  88  &                                                 \\
 \hspace{10pt}  tertiary               & -6.0 & 10.2 & -4.5 & -29.5 & 20.9 &  28'749  &                                                 \\
 \hspace{10pt}  tertiary\_link               & -0.3 & 19.6 & -3.5 & -31.1 & 36.8 &  134  &                                                 \\
 \hspace{10pt}  unclassified               & -7.9 & 11.5 & -8.2 & -32.8 & 31.5 &  7'505  &                                                 \\
 \hspace{10pt}  residential               & -6.3 & 8.8 & -5.8 & -29.7 & 17.5 &  37'277  &                                                 \\
 \hspace{10pt}  living\_street               & -10.1 & 5.2 & -8.0 & -21.9 & -3.6 &  53  &                                                 \\
 \hspace{10pt}  service               & -4.5 & 18.1 & -12.4 & -19.6 & 44.6 &  19  &                                                 \\
 \hspace{10pt}  road               & -26.4 & 0.7 & -26.4 & -26.9 & -26.0 &  2  &                                                 \\
 \hspace{10pt}  construction               & -18.1 & nan & -18.1 & -18.1 & -18.1 &  1  &                                                 \\
\bottomrule

        \caption[Key figures London  (MeTS-10 extent (bounding box))]{Key figures London for the generated data from 20 randomly sampled days (MeTS-10 extent (bounding box)).
        \textbf{num\_edges} number of edges in the street network graph;
        \textbf{num\_nodes} number of nodes in the street network graph;
        \textbf{num\_edges\_per\_cell} number of edges a cell (row,col,heading) has in its intersecting cells;
        \textbf{num\_intersecting\_cells} number of cells (row,col,heading) in an edge's intersecting cells;
        \textbf{node\_degree} number of (unique) neighbor nodes per node;
        \textbf{length\_meters} free flow speed derived from data;
        \textbf{speed\_kph} signalled speed;
        \textbf{free\_flow\_kph} free flow speed derived from data;
        \textbf{free\_flow\_kph-speed\_kph} difference
        }
    \label{tab:key_figures:/iarai/public/t4c/data_pipeline/release20221028_historic_uber:London: (MeTS-10 extent (bounding box))}
    \end{longtable}
    \end{small}
    
\subsubsection{Segment density map  London}
\mbox{}
\nopagebreak{}
\begin{figure}[H]
\centering
\includegraphics[width=0.85\textwidth]{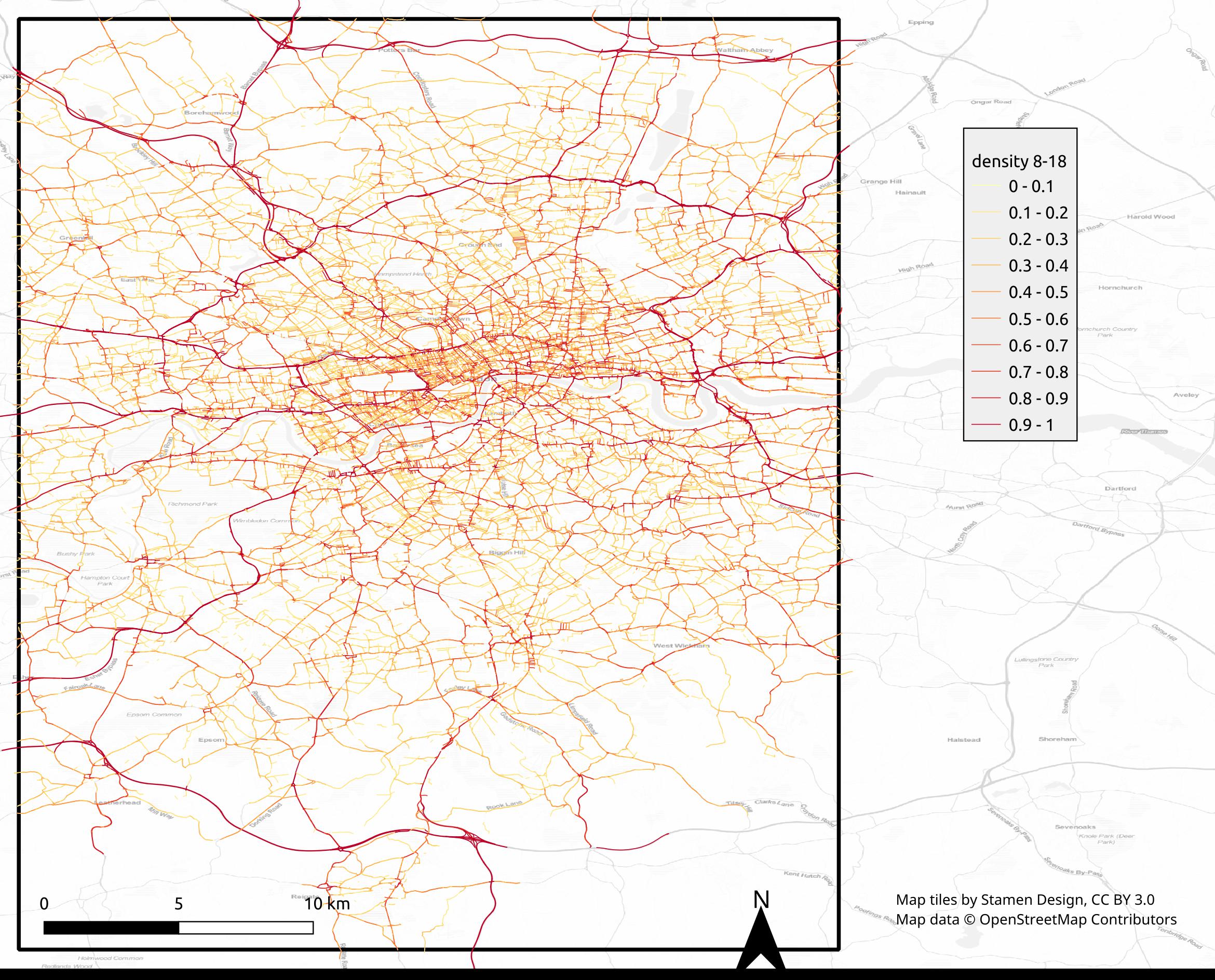}
\caption[Segment-wise density 8am--6pm London]{Segment-wise density 8am--6pm London from 20 randomly sampled days.}
\label{figures/speed_stats_05_val01_uber/density_8_18_london.jpg}
\end{figure}
\clearpage
\subsubsection{Daily density profile  London   (full historic road graph)} 
\mbox{}
\nopagebreak{}
\begin{figure}[H]
\centering
\includegraphics[width=0.85\textwidth]{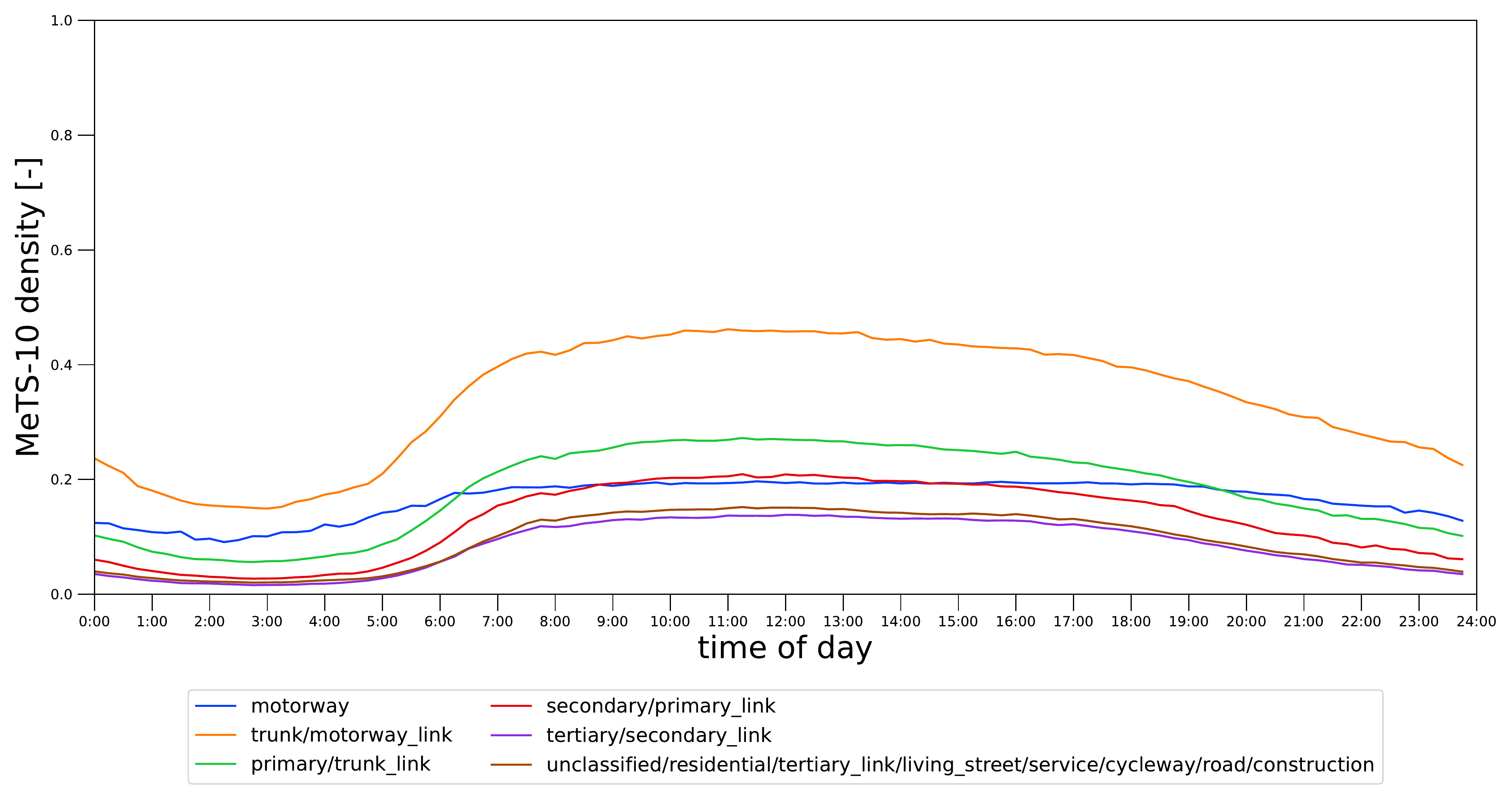}
\caption[Daily density profile London  (full historic road graph)]{Daily density profile for different road types for London  (full historic road graph). Data from 20 randomly sampled days.}
\label{figures/speed_stats_05_val01_uber/speed_stats_coverage_london_by_highway.pdf}
\end{figure}
\subsubsection{Daily speed profile  London   (full historic road graph)}
\mbox{}
\nopagebreak{}
\begin{figure}[H]
\centering
\includegraphics[width=0.85\textwidth]{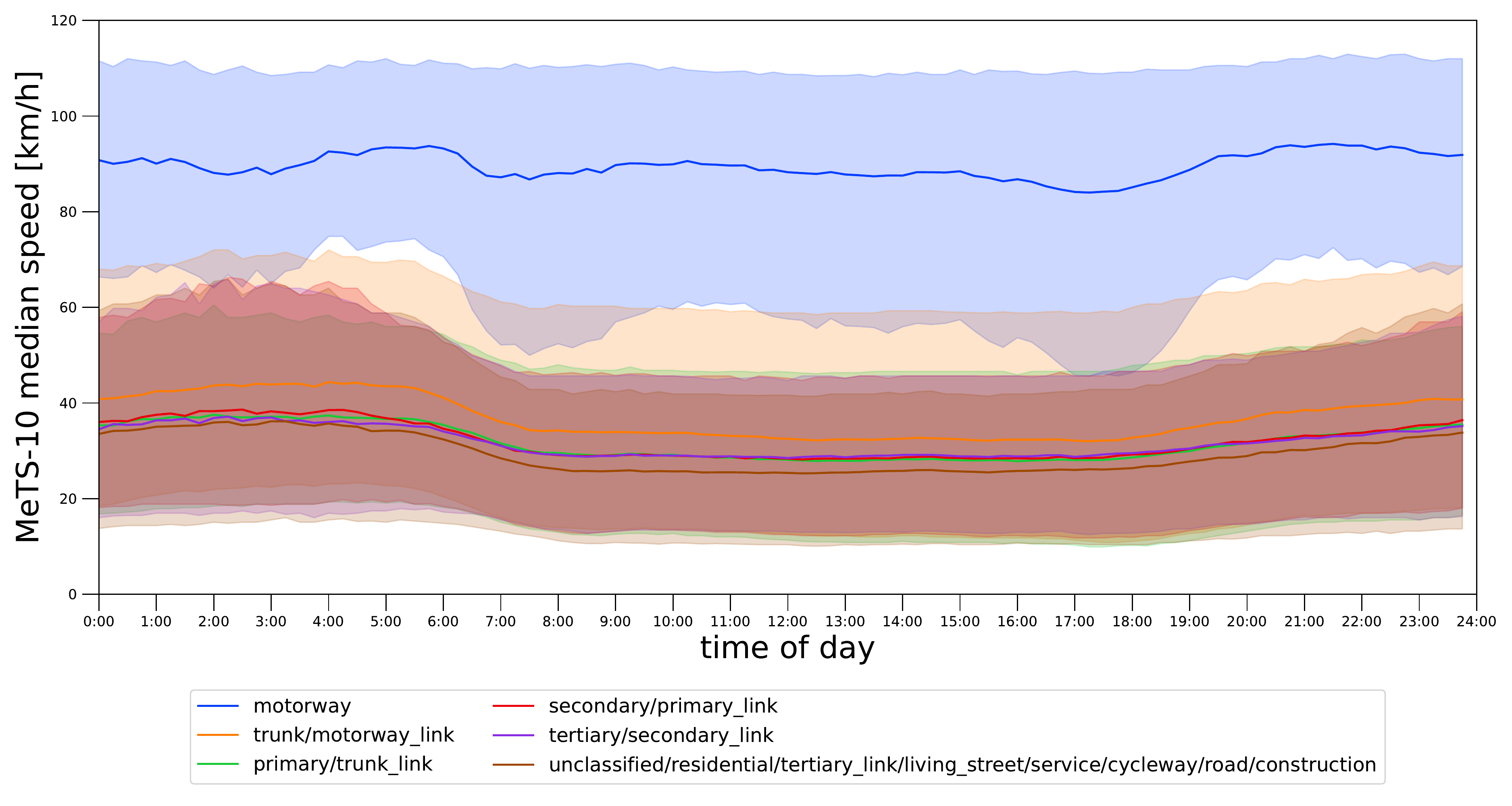}
\caption[Daily median 15 min speeds of all intersecting cells profile London  (full historic road graph)]{Daily median 15 min speeds of all intersecting cells profile for different road types for London  (full historic road graph). The error hull is the 80\% data interval [10.0--90.0 percentiles] of daily means from 20 randomly sampled days.}
\label{figures/speed_stats_05_val01_uber/speed_stats_median_speed_kph_london_by_highway.pdf}
\end{figure}
\subsubsection{Daily density profile  London   (MeTS-10 extent (bounding box))}
\mbox{}
\nopagebreak{}
\begin{figure}[H]
\centering
\includegraphics[width=0.85\textwidth]{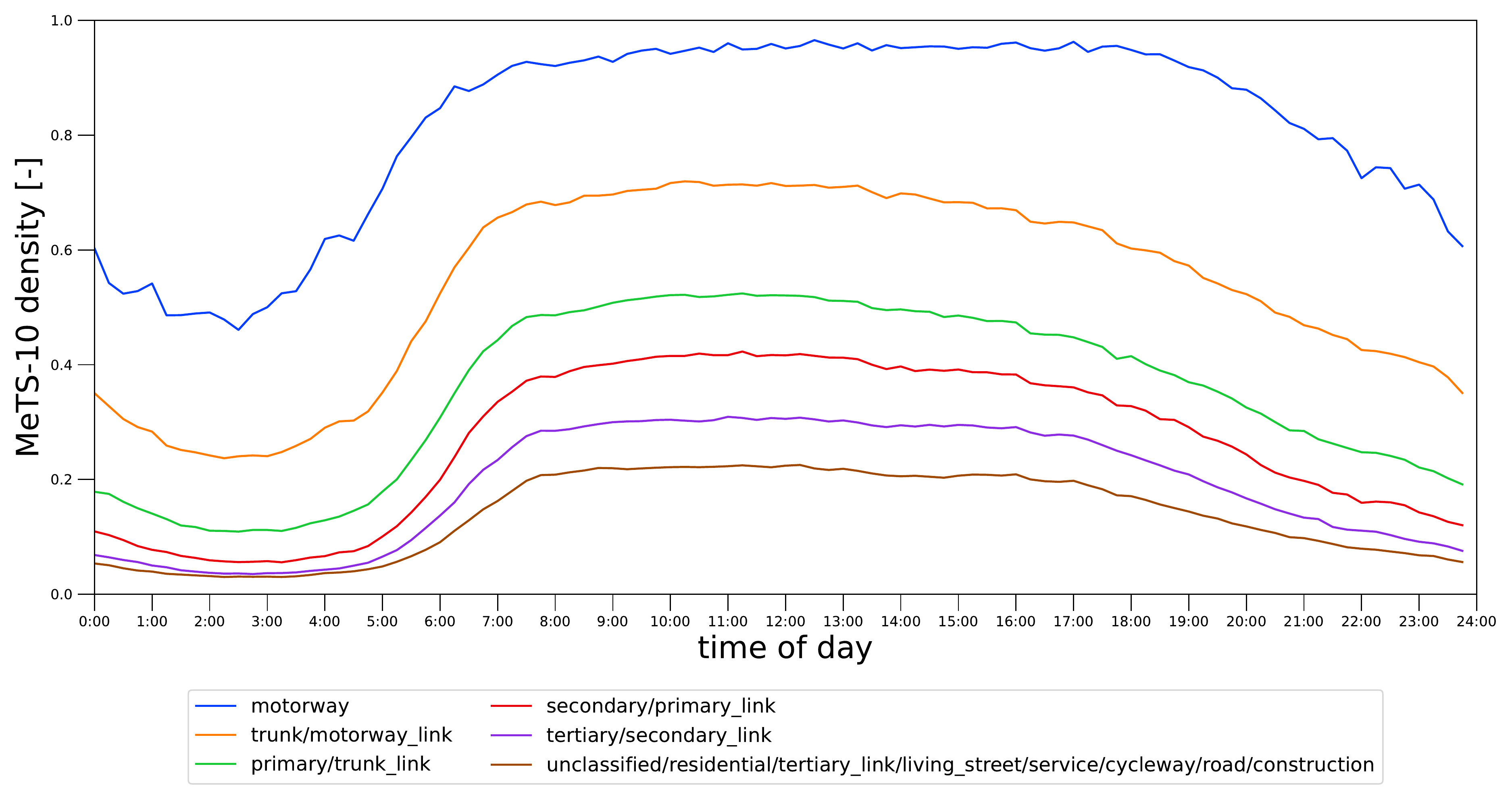}
\caption[Daily density profile London  (MeTS-10 extent (bounding box))]{Daily density profile for different road types for London  (MeTS-10 extent (bounding box)). Data from 20 randomly sampled days.}
\label{figures/speed_stats_05_val01_uber/speed_stats_coverage_london_by_highway_in_bb.pdf}
\end{figure}
\subsubsection{Daily speed profile  London   (MeTS-10 extent (bounding box))}
\mbox{}
\nopagebreak{}
\begin{figure}[H]
\centering
\includegraphics[width=0.85\textwidth]{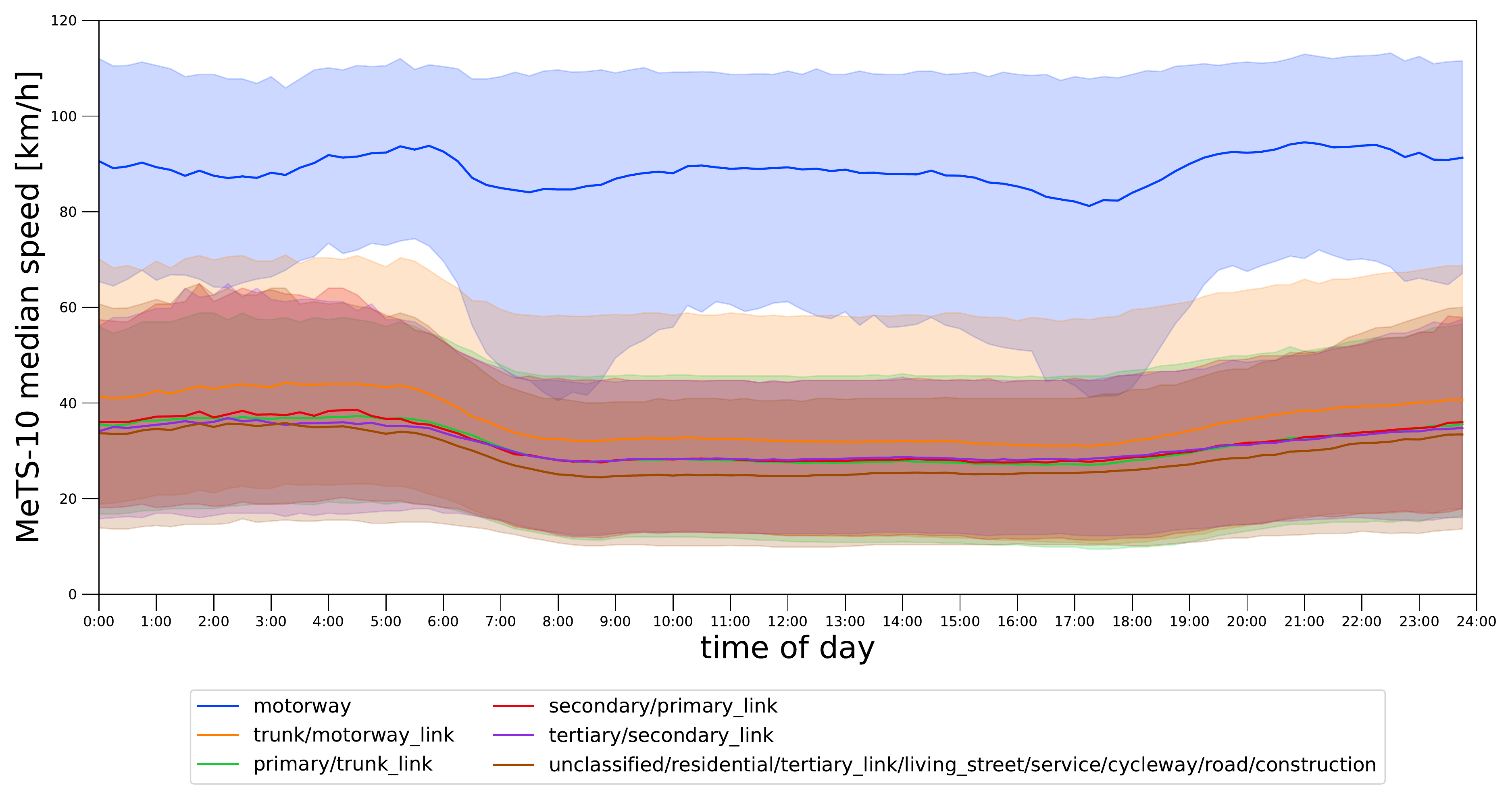}
\caption[Daily median 15 min speeds of all intersecting cells profile London  (MeTS-10 extent (bounding box))]{Daily median 15 min speeds of all intersecting cells profile for different road types for London  (MeTS-10 extent (bounding box)). The error hull is the 80\% data interval [10.0--90.0 percentiles] of daily means from 20 randomly sampled days.}
\label{figures/speed_stats_05_val01_uber/speed_stats_median_speed_kph_london_by_highway_in_bb.pdf}
\end{figure}
\clearpage

\subsection{Key Figures Berlin}
\subsubsection{Road graph map Berlin}
\mbox{}
\nopagebreak{}
\begin{figure}[H]
\centering
\includegraphics[width=0.85\textwidth]{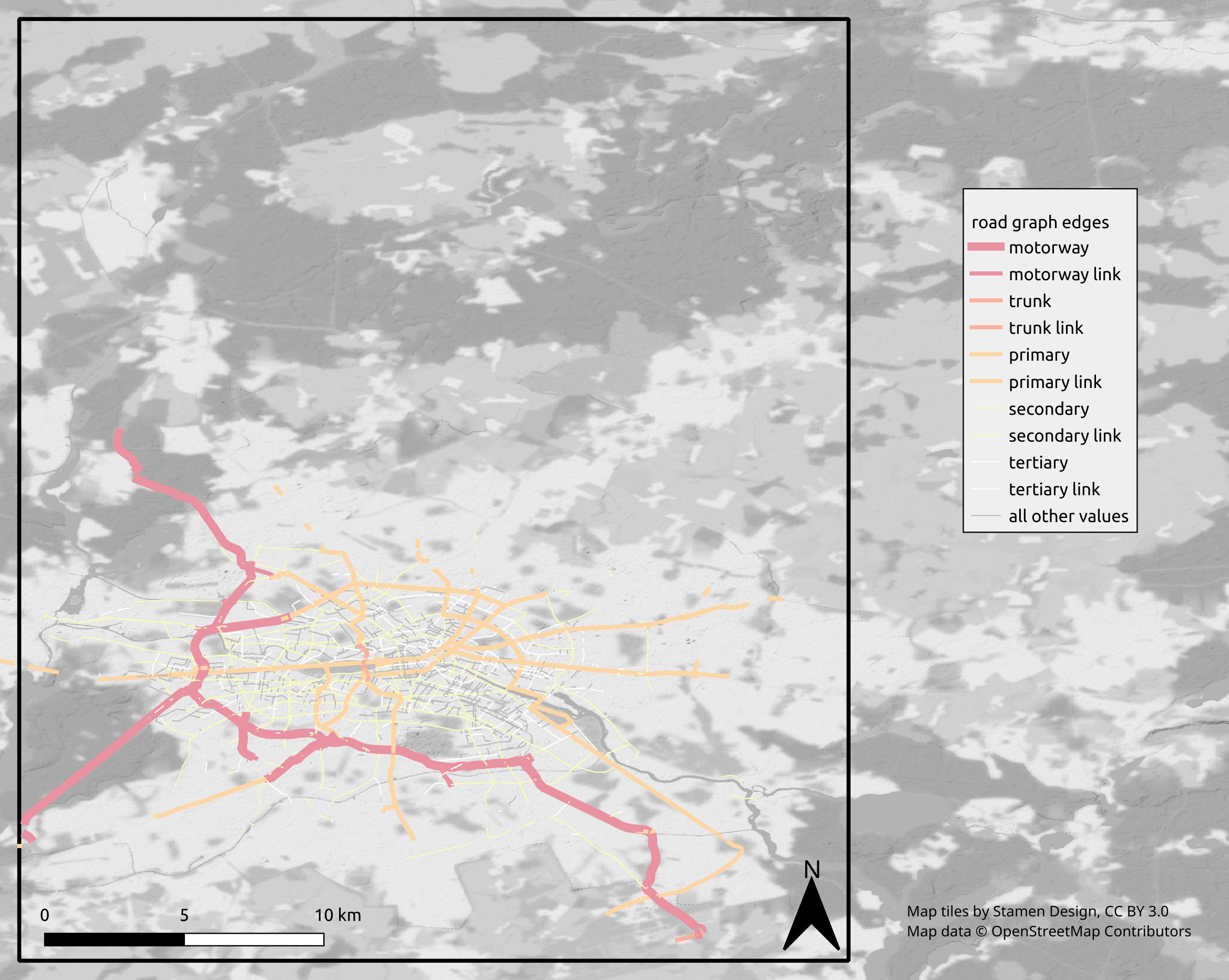}
\caption[Road graph Berlin]{Road graph Berlin, OSM color scheme.}
\label{figures/speed_stats_05_val01_uber/road_graph_berlin.jpg}
\end{figure}
\subsubsection{Static data  Berlin   (full historic road graph)}
\mbox{}\nopagebreak
\begin{small}
\begin{longtable}{p{4cm}rrrrrrrr}
\toprule
Attribute      & {mean} &{std} & {median}  & {q01} & {q99} & {data points} & {sum}  \\
\midrule
 bounding box  (full historic road graph)               &  &  &  &  &  &  13.189--13.625 / 52.359--52.854 &                                                \\
 num\_edges  (full historic road graph)               &  &  &  &  &  &  16'279 &                                                \\
 \hspace{10pt}  motorway               &  &  &  &  &  &  515 &                                                \\
 \hspace{10pt}  motorway\_link               &  &  &  &  &  &  384 &                                                \\
 \hspace{10pt}  trunk               &  &  &  &  &  &  72 &                                                \\
 \hspace{10pt}  trunk\_link               &  &  &  &  &  &  4 &                                                \\
 \hspace{10pt}  primary               &  &  &  &  &  &  2937 &                                                \\
 \hspace{10pt}  primary\_link               &  &  &  &  &  &  15 &                                                \\
 \hspace{10pt}  secondary               &  &  &  &  &  &  6464 &                                                \\
 \hspace{10pt}  secondary\_link               &  &  &  &  &  &  21 &                                                \\
 \hspace{10pt}  tertiary               &  &  &  &  &  &  2782 &                                                \\
 \hspace{10pt}  tertiary\_link               &  &  &  &  &  &  2 &                                                \\
 \hspace{10pt}  unclassified               &  &  &  &  &  &  56 &                                                \\
 \hspace{10pt}  residential               &  &  &  &  &  &  2934 &                                                \\
 \hspace{10pt}  living\_street               &  &  &  &  &  &  91 &                                                \\
 \hspace{10pt}  construction               &  &  &  &  &  &  2 &                                                \\
 num\_nodes  (full historic road graph)               &  &  &  &  &  &  12655 &                                                 \\
 num\_edges\_per\_cell  (full historic road graph)               & 1.1 & 0.3 & 1.0 & 1.0 & 3.0 &  46'509 &                                                 \\
 num\_intersecting\_cells  (full historic road graph)               & 3.1 & 2.2 & 2.0 & 1.0 & 12.0 &  16'279 &                                                 \\
 node\_degree  (full historic road graph)               & 2.2 & 0.7 & 2.0 & 1.0 & 4.0 &  12'655 &                                                 \\
 length\_meters  (full historic road graph)               & 103.4 & 120.9 & 71.8 & 5.2 & 536.7 &  16'279 & 1.7e+06                                                \\
 \hspace{10pt}  motorway               & 269.5 & 322.6 & 166.2 & 13.9 & 1'670.6 &  515  & 1.4e+05                                                \\
 \hspace{10pt}  motorway\_link               & 144.0 & 118.0 & 116.4 & 13.2 & 531.2 &  384  & 5.5e+04                                                \\
 \hspace{10pt}  trunk               & 337.4 & 572.4 & 162.9 & 3.7 & 2'542.3 &  72  & 2.4e+04                                                \\
 \hspace{10pt}  trunk\_link               & 87.3 & 97.4 & 54.2 & 12.1 & 224.5 &  4  & 3.5e+02                                                \\
 \hspace{10pt}  primary               & 105.3 & 116.2 & 72.0 & 5.6 & 603.9 &  2'937  & 3.1e+05                                                \\
 \hspace{10pt}  primary\_link               & 36.3 & 32.9 & 23.3 & 10.7 & 122.0 &  15  & 5.4e+02                                                \\
 \hspace{10pt}  secondary               & 91.4 & 89.8 & 66.2 & 5.5 & 413.8 &  6'464  & 5.9e+05                                                \\
 \hspace{10pt}  secondary\_link               & 50.7 & 47.3 & 27.3 & 15.4 & 148.3 &  21  & 1.1e+03                                                \\
 \hspace{10pt}  tertiary               & 90.2 & 85.8 & 65.8 & 4.9 & 378.2 &  2'782  & 2.5e+05                                                \\
 \hspace{10pt}  tertiary\_link               & 19.2 & 0.0 & 19.2 & 19.2 & 19.2 &  2  & 3.8e+01                                                \\
 \hspace{10pt}  unclassified               & 75.9 & 80.8 & 46.1 & 4.3 & 351.9 &  56  & 4.2e+03                                                \\
 \hspace{10pt}  residential               & 101.2 & 86.4 & 77.1 & 4.2 & 373.4 &  2'934  & 3.0e+05                                                \\
 \hspace{10pt}  living\_street               & 111.4 & 96.5 & 88.3 & 6.9 & 369.8 &  91  & 1.0e+04                                                \\
 \hspace{10pt}  construction               & 36.7 & 0.0 & 36.7 & 36.7 & 36.7 &  2  & 7.3e+01                                                \\
 speed\_kph  (full historic road graph)               & 46.6 & 11.2 & 50.0 & 30.0 & 80.0 &  16'279 &                                                 \\
 \hspace{10pt}  motorway               & 77.3 & 11.8 & 80.0 & 40.0 & 120.0 &  515  &                                                 \\
 \hspace{10pt}  motorway\_link               & 60.1 & 10.0 & 60.0 & 40.0 & 80.0 &  384  &                                                 \\
 \hspace{10pt}  trunk               & 70.6 & 22.0 & 65.0 & 50.0 & 100.0 &  72  &                                                 \\
 \hspace{10pt}  trunk\_link               & 30.0 & 0.0 & 30.0 & 30.0 & 30.0 &  4  &                                                 \\
 \hspace{10pt}  primary               & 48.8 & 6.0 & 50.0 & 30.0 & 60.0 &  2'937  &                                                 \\
 \hspace{10pt}  primary\_link               & 48.0 & 7.7 & 50.0 & 30.0 & 58.6 &  15  &                                                 \\
 \hspace{10pt}  secondary               & 48.8 & 5.3 & 50.0 & 30.0 & 60.0 &  6'464  &                                                 \\
 \hspace{10pt}  secondary\_link               & 50.0 & 0.0 & 50.0 & 50.0 & 50.0 &  21  &                                                 \\
 \hspace{10pt}  tertiary               & 45.7 & 8.1 & 50.0 & 30.0 & 50.0 &  2'782  &                                                 \\
 \hspace{10pt}  tertiary\_link               & 50.0 & 0.0 & 50.0 & 50.0 & 50.0 &  2  &                                                 \\
 \hspace{10pt}  unclassified               & 37.7 & 10.3 & 30.0 & 20.0 & 50.0 &  56  &                                                 \\
 \hspace{10pt}  residential               & 32.5 & 7.7 & 30.0 & 10.0 & 50.0 &  2'934  &                                                 \\
 \hspace{10pt}  living\_street               & 50.0 & 0.0 & 50.0 & 50.0 & 50.0 &  91  &                                                 \\
 \hspace{10pt}  construction               & 50.0 & 0.0 & 50.0 & 50.0 & 50.0 &  2  &                                                 \\
 free\_flow\_kph  (full historic road graph)               & 45.1 & 13.5 & 43.5 & 22.1 & 89.9 &  16'229 &                                                 \\
 \hspace{10pt}  motorway               & 84.3 & 10.5 & 85.2 & 56.8 & 115.1 &  515  &                                                 \\
 \hspace{10pt}  motorway\_link               & 77.6 & 14.2 & 81.5 & 36.7 & 114.7 &  384  &                                                 \\
 \hspace{10pt}  trunk               & 59.5 & 19.6 & 52.9 & 33.2 & 96.0 &  46  &                                                 \\
 \hspace{10pt}  primary               & 47.1 & 8.8 & 47.1 & 28.7 & 70.8 &  2'919  &                                                 \\
 \hspace{10pt}  primary\_link               & 42.9 & 11.7 & 41.1 & 25.9 & 64.8 &  15  &                                                 \\
 \hspace{10pt}  secondary               & 45.9 & 9.2 & 45.6 & 26.5 & 80.3 &  6'464  &                                                 \\
 \hspace{10pt}  secondary\_link               & 41.6 & 14.0 & 39.8 & 20.2 & 77.2 &  21  &                                                 \\
 \hspace{10pt}  tertiary               & 41.4 & 8.8 & 40.9 & 23.4 & 75.6 &  2'780  &                                                 \\
 \hspace{10pt}  tertiary\_link               & 38.8 & 0.0 & 38.8 & 38.8 & 38.8 &  2  &                                                 \\
 \hspace{10pt}  unclassified               & 44.8 & 7.8 & 46.1 & 21.8 & 54.3 &  56  &                                                 \\
 \hspace{10pt}  residential               & 34.2 & 7.7 & 33.9 & 18.8 & 55.9 &  2'934  &                                                 \\
 \hspace{10pt}  living\_street               & 26.0 & 6.1 & 24.9 & 16.5 & 37.9 &  91  &                                                 \\
 \hspace{10pt}  construction               & 42.4 & 0.0 & 42.4 & 42.4 & 42.4 &  2  &                                                 \\
 free\_flow\_kph-speed\_kph  (full historic road graph)               & -1.4 & 10.8 & -1.3 & -26.0 & 30.2 &  16'229 &                                                 \\
 \hspace{10pt}  motorway               & 7.0 & 7.5 & 6.9 & -9.6 & 29.0 &  515  &                                                 \\
 \hspace{10pt}  motorway\_link               & 17.5 & 13.1 & 20.6 & -15.5 & 38.7 &  384  &                                                 \\
 \hspace{10pt}  trunk               & 2.9 & 10.1 & 2.9 & -16.8 & 26.0 &  46  &                                                 \\
 \hspace{10pt}  primary               & -1.6 & 8.2 & -1.1 & -20.5 & 19.2 &  2'919  &                                                 \\
 \hspace{10pt}  primary\_link               & -5.1 & 16.7 & -8.9 & -24.1 & 34.2 &  15  &                                                 \\
 \hspace{10pt}  secondary               & -2.9 & 9.9 & -3.4 & -23.4 & 31.5 &  6'464  &                                                 \\
 \hspace{10pt}  secondary\_link               & -8.4 & 14.0 & -10.2 & -29.8 & 27.2 &  21  &                                                 \\
 \hspace{10pt}  tertiary               & -4.4 & 10.8 & -5.3 & -26.0 & 25.6 &  2'780  &                                                 \\
 \hspace{10pt}  tertiary\_link               & -11.2 & 0.0 & -11.2 & -11.2 & -11.2 &  2  &                                                 \\
 \hspace{10pt}  unclassified               & 7.1 & 13.6 & 12.4 & -28.2 & 23.3 &  56  &                                                 \\
 \hspace{10pt}  residential               & 1.6 & 10.6 & 2.5 & -29.6 & 26.2 &  2'934  &                                                 \\
 \hspace{10pt}  living\_street               & -24.0 & 6.1 & -25.1 & -33.5 & -12.1 &  91  &                                                 \\
 \hspace{10pt}  construction               & -7.6 & 0.0 & -7.6 & -7.6 & -7.6 &  2  &                                                 \\
\bottomrule

        \caption[Key figures Berlin  (full historic road graph)]{Key figures Berlin for the generated data from 20 randomly sampled days (full historic road graph).
        \textbf{num\_edges} number of edges in the street network graph;
        \textbf{num\_nodes} number of nodes in the street network graph;
        \textbf{num\_edges\_per\_cell} number of edges a cell (row,col,heading) has in its intersecting cells;
        \textbf{num\_intersecting\_cells} number of cells (row,col,heading) in an edge's intersecting cells;
        \textbf{node\_degree} number of (unique) neighbor nodes per node;
        \textbf{length\_meters} free flow speed derived from data;
        \textbf{speed\_kph} signalled speed;
        \textbf{free\_flow\_kph} free flow speed derived from data;
        \textbf{free\_flow\_kph-speed\_kph} difference
        }
    \label{tab:key_figures:/iarai/public/t4c/data_pipeline/release20221028_historic_uber:Berlin: (full historic road graph)}
    \end{longtable}
    \end{small}
    
\subsubsection{Static data  Berlin   (MeTS-10 extent (bounding box))}
\mbox{}\nopagebreak
\begin{small}
\begin{longtable}{p{4cm}rrrrrrrr}
\toprule
Attribute      & {mean} &{std} & {median}  & {q01} & {q99} & {data points} & {sum}  \\
\midrule
 bounding box  (MeTS-10 extent (bounding box))               &  &  &  &  &  &  13.189--13.625 / 52.359--52.854 &                                                \\
 num\_edges  (MeTS-10 extent (bounding box))               &  &  &  &  &  &  16'229 &                                                \\
 \hspace{10pt}  motorway               &  &  &  &  &  &  515 &                                                \\
 \hspace{10pt}  motorway\_link               &  &  &  &  &  &  384 &                                                \\
 \hspace{10pt}  trunk               &  &  &  &  &  &  46 &                                                \\
 \hspace{10pt}  primary               &  &  &  &  &  &  2919 &                                                \\
 \hspace{10pt}  primary\_link               &  &  &  &  &  &  15 &                                                \\
 \hspace{10pt}  secondary               &  &  &  &  &  &  6464 &                                                \\
 \hspace{10pt}  secondary\_link               &  &  &  &  &  &  21 &                                                \\
 \hspace{10pt}  tertiary               &  &  &  &  &  &  2780 &                                                \\
 \hspace{10pt}  tertiary\_link               &  &  &  &  &  &  2 &                                                \\
 \hspace{10pt}  unclassified               &  &  &  &  &  &  56 &                                                \\
 \hspace{10pt}  residential               &  &  &  &  &  &  2934 &                                                \\
 \hspace{10pt}  living\_street               &  &  &  &  &  &  91 &                                                \\
 \hspace{10pt}  construction               &  &  &  &  &  &  2 &                                                \\
 num\_nodes  (MeTS-10 extent (bounding box))               &  &  &  &  &  &  12603 &                                                 \\
 num\_edges\_per\_cell  (MeTS-10 extent (bounding box))               & 1.1 & 0.3 & 1.0 & 1.0 & 3.0 &  46'509 &                                                 \\
 num\_intersecting\_cells  (MeTS-10 extent (bounding box))               & 3.1 & 2.2 & 2.0 & 1.0 & 12.0 &  16'229 &                                                 \\
 node\_degree  (MeTS-10 extent (bounding box))               & 2.2 & 0.7 & 2.0 & 1.0 & 4.0 &  12'603 &                                                 \\
 length\_meters  (MeTS-10 extent (bounding box))               & 102.6 & 114.6 & 71.7 & 5.2 & 531.4 &  16'229 & 1.7e+06                                                \\
 \hspace{10pt}  motorway               & 269.5 & 322.6 & 166.2 & 13.9 & 1'670.6 &  515  & 1.4e+05                                                \\
 \hspace{10pt}  motorway\_link               & 144.0 & 118.0 & 116.4 & 13.2 & 531.2 &  384  & 5.5e+04                                                \\
 \hspace{10pt}  trunk               & 191.8 & 209.7 & 114.0 & 3.6 & 835.0 &  46  & 8.8e+03                                                \\
 \hspace{10pt}  primary               & 105.4 & 116.5 & 72.0 & 5.6 & 605.4 &  2'919  & 3.1e+05                                                \\
 \hspace{10pt}  primary\_link               & 36.3 & 32.9 & 23.3 & 10.7 & 122.0 &  15  & 5.4e+02                                                \\
 \hspace{10pt}  secondary               & 91.4 & 89.8 & 66.2 & 5.5 & 413.8 &  6'464  & 5.9e+05                                                \\
 \hspace{10pt}  secondary\_link               & 50.7 & 47.3 & 27.3 & 15.4 & 148.3 &  21  & 1.1e+03                                                \\
 \hspace{10pt}  tertiary               & 90.2 & 85.8 & 65.8 & 4.9 & 378.3 &  2'780  & 2.5e+05                                                \\
 \hspace{10pt}  tertiary\_link               & 19.2 & 0.0 & 19.2 & 19.2 & 19.2 &  2  & 3.8e+01                                                \\
 \hspace{10pt}  unclassified               & 75.9 & 80.8 & 46.1 & 4.3 & 351.9 &  56  & 4.2e+03                                                \\
 \hspace{10pt}  residential               & 101.2 & 86.4 & 77.1 & 4.2 & 373.4 &  2'934  & 3.0e+05                                                \\
 \hspace{10pt}  living\_street               & 111.4 & 96.5 & 88.3 & 6.9 & 369.8 &  91  & 1.0e+04                                                \\
 \hspace{10pt}  construction               & 36.7 & 0.0 & 36.7 & 36.7 & 36.7 &  2  & 7.3e+01                                                \\
 speed\_kph  (MeTS-10 extent (bounding box))               & 46.5 & 11.0 & 50.0 & 30.0 & 80.0 &  16'229 &                                                 \\
 \hspace{10pt}  motorway               & 77.3 & 11.8 & 80.0 & 40.0 & 120.0 &  515  &                                                 \\
 \hspace{10pt}  motorway\_link               & 60.1 & 10.0 & 60.0 & 40.0 & 80.0 &  384  &                                                 \\
 \hspace{10pt}  trunk               & 56.5 & 11.4 & 50.0 & 50.0 & 80.0 &  46  &                                                 \\
 \hspace{10pt}  primary               & 48.8 & 6.0 & 50.0 & 30.0 & 60.0 &  2'919  &                                                 \\
 \hspace{10pt}  primary\_link               & 48.0 & 7.7 & 50.0 & 30.0 & 58.6 &  15  &                                                 \\
 \hspace{10pt}  secondary               & 48.8 & 5.3 & 50.0 & 30.0 & 60.0 &  6'464  &                                                 \\
 \hspace{10pt}  secondary\_link               & 50.0 & 0.0 & 50.0 & 50.0 & 50.0 &  21  &                                                 \\
 \hspace{10pt}  tertiary               & 45.7 & 8.1 & 50.0 & 30.0 & 50.0 &  2'780  &                                                 \\
 \hspace{10pt}  tertiary\_link               & 50.0 & 0.0 & 50.0 & 50.0 & 50.0 &  2  &                                                 \\
 \hspace{10pt}  unclassified               & 37.7 & 10.3 & 30.0 & 20.0 & 50.0 &  56  &                                                 \\
 \hspace{10pt}  residential               & 32.5 & 7.7 & 30.0 & 10.0 & 50.0 &  2'934  &                                                 \\
 \hspace{10pt}  living\_street               & 50.0 & 0.0 & 50.0 & 50.0 & 50.0 &  91  &                                                 \\
 \hspace{10pt}  construction               & 50.0 & 0.0 & 50.0 & 50.0 & 50.0 &  2  &                                                 \\
 free\_flow\_kph  (MeTS-10 extent (bounding box))               & 45.1 & 13.5 & 43.5 & 22.1 & 89.9 &  16'229 &                                                 \\
 \hspace{10pt}  motorway               & 84.3 & 10.5 & 85.2 & 56.8 & 115.1 &  515  &                                                 \\
 \hspace{10pt}  motorway\_link               & 77.6 & 14.2 & 81.5 & 36.7 & 114.7 &  384  &                                                 \\
 \hspace{10pt}  trunk               & 59.5 & 19.6 & 52.9 & 33.2 & 96.0 &  46  &                                                 \\
 \hspace{10pt}  primary               & 47.1 & 8.8 & 47.1 & 28.7 & 70.8 &  2'919  &                                                 \\
 \hspace{10pt}  primary\_link               & 42.9 & 11.7 & 41.1 & 25.9 & 64.8 &  15  &                                                 \\
 \hspace{10pt}  secondary               & 45.9 & 9.2 & 45.6 & 26.5 & 80.3 &  6'464  &                                                 \\
 \hspace{10pt}  secondary\_link               & 41.6 & 14.0 & 39.8 & 20.2 & 77.2 &  21  &                                                 \\
 \hspace{10pt}  tertiary               & 41.4 & 8.8 & 40.9 & 23.4 & 75.6 &  2'780  &                                                 \\
 \hspace{10pt}  tertiary\_link               & 38.8 & 0.0 & 38.8 & 38.8 & 38.8 &  2  &                                                 \\
 \hspace{10pt}  unclassified               & 44.8 & 7.8 & 46.1 & 21.8 & 54.3 &  56  &                                                 \\
 \hspace{10pt}  residential               & 34.2 & 7.7 & 33.9 & 18.8 & 55.9 &  2'934  &                                                 \\
 \hspace{10pt}  living\_street               & 26.0 & 6.1 & 24.9 & 16.5 & 37.9 &  91  &                                                 \\
 \hspace{10pt}  construction               & 42.4 & 0.0 & 42.4 & 42.4 & 42.4 &  2  &                                                 \\
 free\_flow\_kph-speed\_kph  (MeTS-10 extent (bounding box))               & -1.4 & 10.8 & -1.3 & -26.0 & 30.2 &  16'229 &                                                 \\
 \hspace{10pt}  motorway               & 7.0 & 7.5 & 6.9 & -9.6 & 29.0 &  515  &                                                 \\
 \hspace{10pt}  motorway\_link               & 17.5 & 13.1 & 20.6 & -15.5 & 38.7 &  384  &                                                 \\
 \hspace{10pt}  trunk               & 2.9 & 10.1 & 2.9 & -16.8 & 26.0 &  46  &                                                 \\
 \hspace{10pt}  primary               & -1.6 & 8.2 & -1.1 & -20.5 & 19.2 &  2'919  &                                                 \\
 \hspace{10pt}  primary\_link               & -5.1 & 16.7 & -8.9 & -24.1 & 34.2 &  15  &                                                 \\
 \hspace{10pt}  secondary               & -2.9 & 9.9 & -3.4 & -23.4 & 31.5 &  6'464  &                                                 \\
 \hspace{10pt}  secondary\_link               & -8.4 & 14.0 & -10.2 & -29.8 & 27.2 &  21  &                                                 \\
 \hspace{10pt}  tertiary               & -4.4 & 10.8 & -5.3 & -26.0 & 25.6 &  2'780  &                                                 \\
 \hspace{10pt}  tertiary\_link               & -11.2 & 0.0 & -11.2 & -11.2 & -11.2 &  2  &                                                 \\
 \hspace{10pt}  unclassified               & 7.1 & 13.6 & 12.4 & -28.2 & 23.3 &  56  &                                                 \\
 \hspace{10pt}  residential               & 1.6 & 10.6 & 2.5 & -29.6 & 26.2 &  2'934  &                                                 \\
 \hspace{10pt}  living\_street               & -24.0 & 6.1 & -25.1 & -33.5 & -12.1 &  91  &                                                 \\
 \hspace{10pt}  construction               & -7.6 & 0.0 & -7.6 & -7.6 & -7.6 &  2  &                                                 \\
\bottomrule

        \caption[Key figures Berlin  (MeTS-10 extent (bounding box))]{Key figures Berlin for the generated data from 20 randomly sampled days (MeTS-10 extent (bounding box)).
        \textbf{num\_edges} number of edges in the street network graph;
        \textbf{num\_nodes} number of nodes in the street network graph;
        \textbf{num\_edges\_per\_cell} number of edges a cell (row,col,heading) has in its intersecting cells;
        \textbf{num\_intersecting\_cells} number of cells (row,col,heading) in an edge's intersecting cells;
        \textbf{node\_degree} number of (unique) neighbor nodes per node;
        \textbf{length\_meters} free flow speed derived from data;
        \textbf{speed\_kph} signalled speed;
        \textbf{free\_flow\_kph} free flow speed derived from data;
        \textbf{free\_flow\_kph-speed\_kph} difference
        }
    \label{tab:key_figures:/iarai/public/t4c/data_pipeline/release20221028_historic_uber:Berlin: (MeTS-10 extent (bounding box))}
    \end{longtable}
    \end{small}
    
\subsubsection{Segment density map  Berlin}
\mbox{}
\nopagebreak{}
\begin{figure}[H]
\centering
\includegraphics[width=0.85\textwidth]{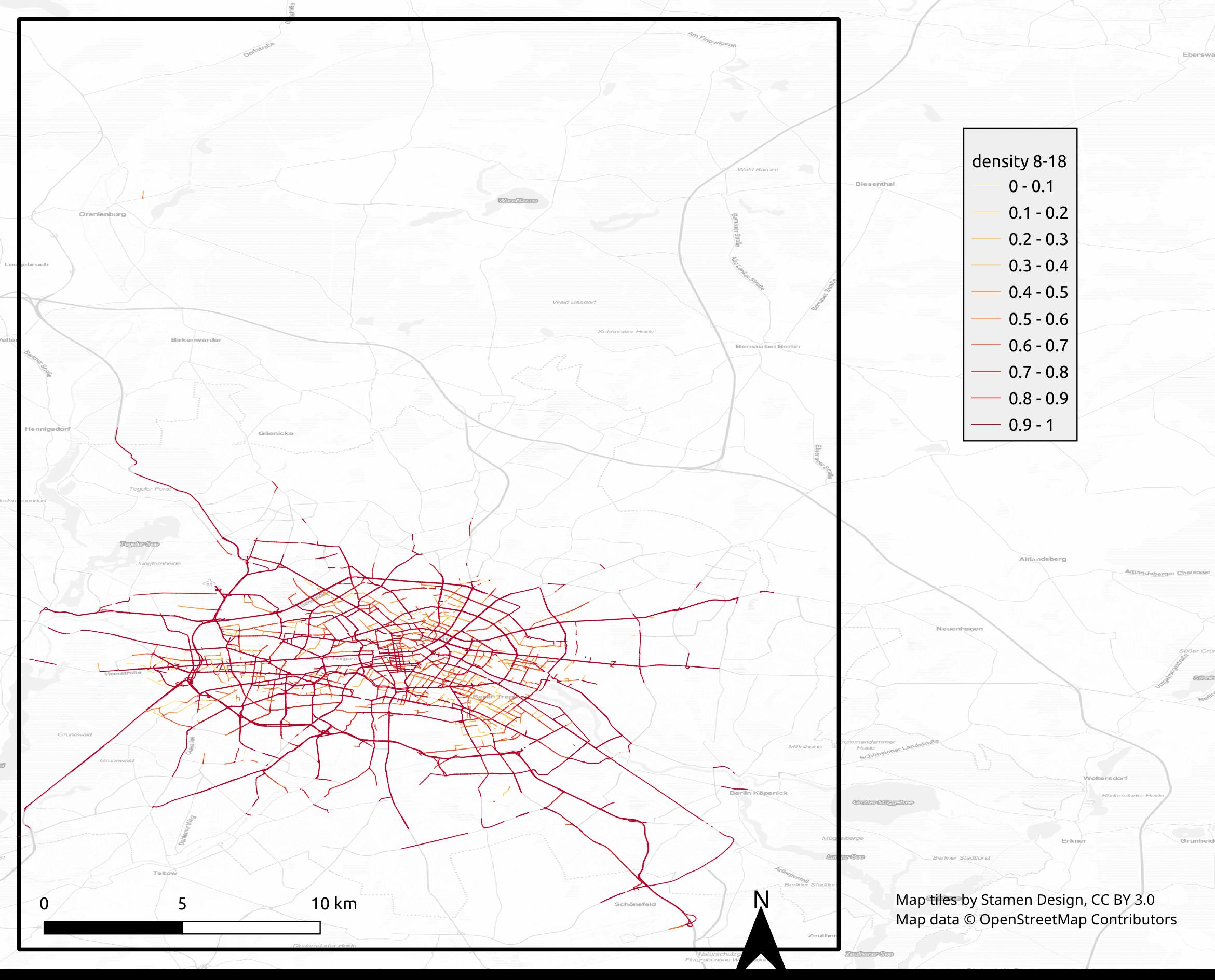}
\caption[Segment-wise density 8am--6pm Berlin]{Segment-wise density 8am--6pm Berlin from 20 randomly sampled days.}
\label{figures/speed_stats_05_val01_uber/density_8_18_berlin.jpg}
\end{figure}
\clearpage
\subsubsection{Daily density profile  Berlin   (full historic road graph)} 
\mbox{}
\nopagebreak{}
\begin{figure}[H]
\centering
\includegraphics[width=0.85\textwidth]{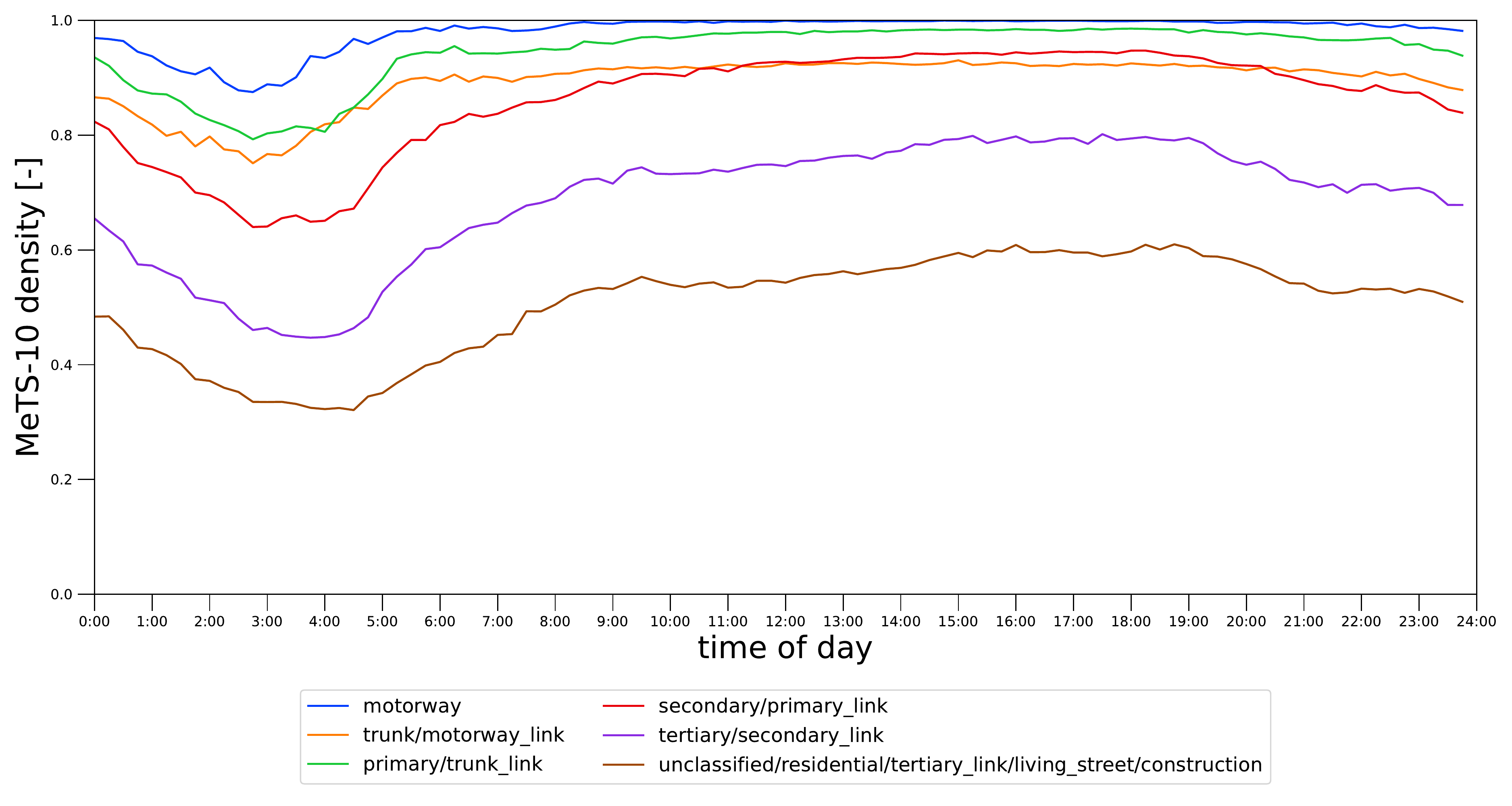}
\caption[Daily density profile Berlin  (full historic road graph)]{Daily density profile for different road types for Berlin  (full historic road graph). Data from 20 randomly sampled days.}
\label{figures/speed_stats_05_val01_uber/speed_stats_coverage_berlin_by_highway.pdf}
\end{figure}
\subsubsection{Daily speed profile  Berlin   (full historic road graph)}
\mbox{}
\nopagebreak{}
\begin{figure}[H]
\centering
\includegraphics[width=0.85\textwidth]{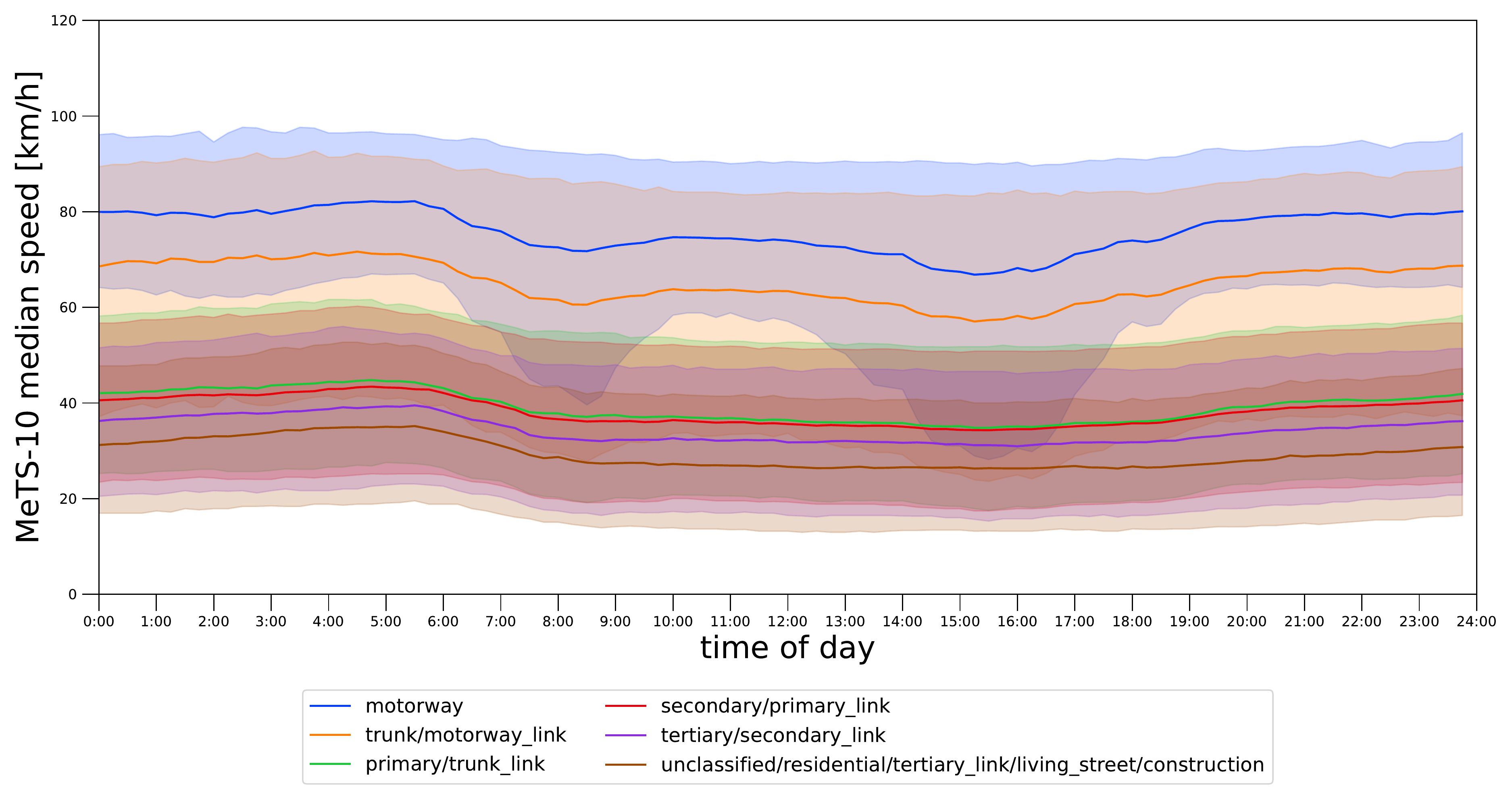}
\caption[Daily median 15 min speeds of all intersecting cells profile Berlin  (full historic road graph)]{Daily median 15 min speeds of all intersecting cells profile for different road types for Berlin  (full historic road graph). The error hull is the 80\% data interval [10.0--90.0 percentiles] of daily means from 20 randomly sampled days.}
\label{figures/speed_stats_05_val01_uber/speed_stats_median_speed_kph_berlin_by_highway.pdf}
\end{figure}
\subsubsection{Daily density profile  Berlin   (MeTS-10 extent (bounding box))}
\mbox{}
\nopagebreak{}
\begin{figure}[H]
\centering
\includegraphics[width=0.85\textwidth]{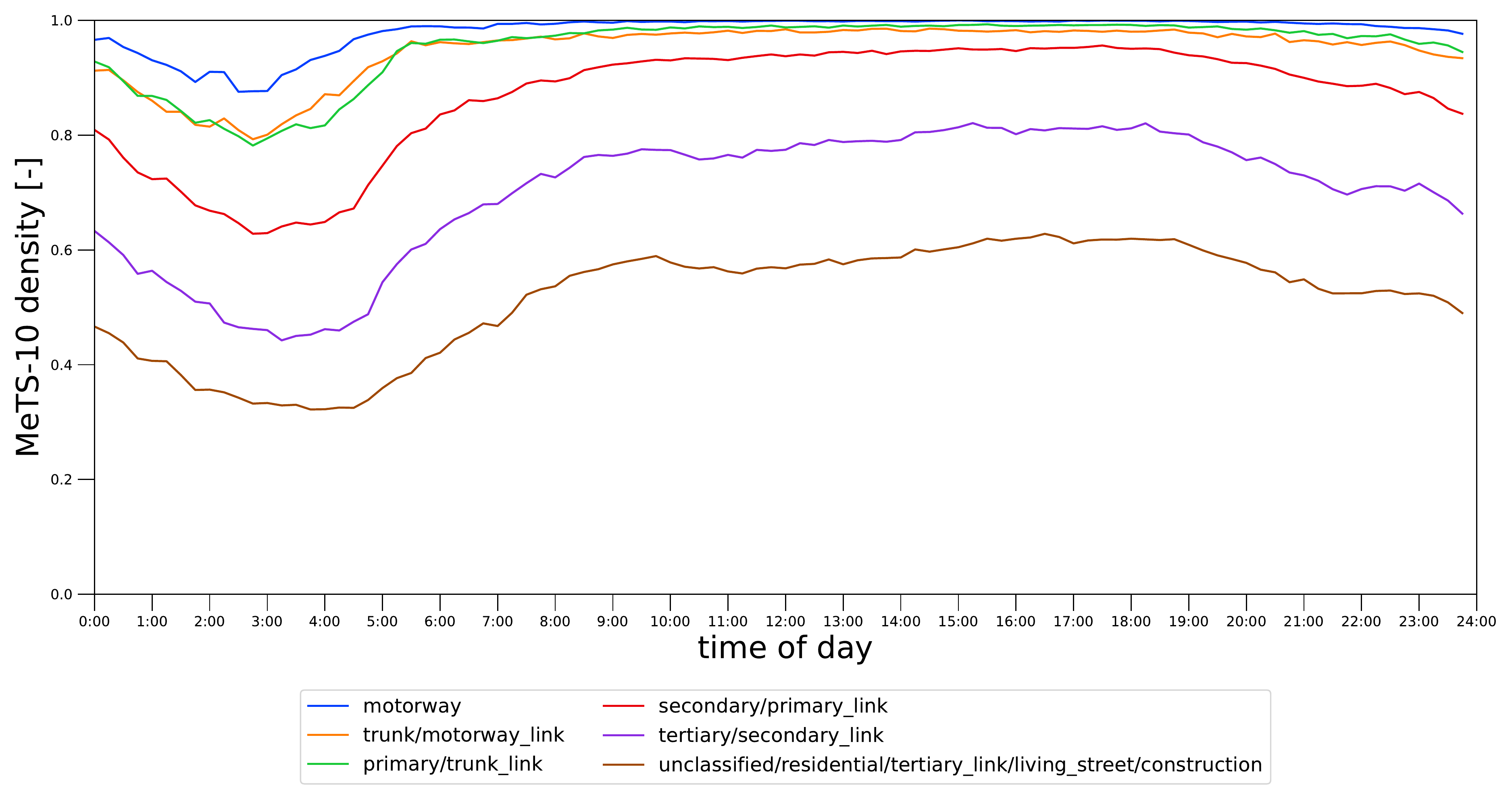}
\caption[Daily density profile Berlin  (MeTS-10 extent (bounding box))]{Daily density profile for different road types for Berlin  (MeTS-10 extent (bounding box)). Data from 20 randomly sampled days.}
\label{figures/speed_stats_05_val01_uber/speed_stats_coverage_berlin_by_highway_in_bb.pdf}
\end{figure}
\subsubsection{Daily speed profile  Berlin   (MeTS-10 extent (bounding box))}
\mbox{}
\nopagebreak{}
\begin{figure}[H]
\centering
\includegraphics[width=0.85\textwidth]{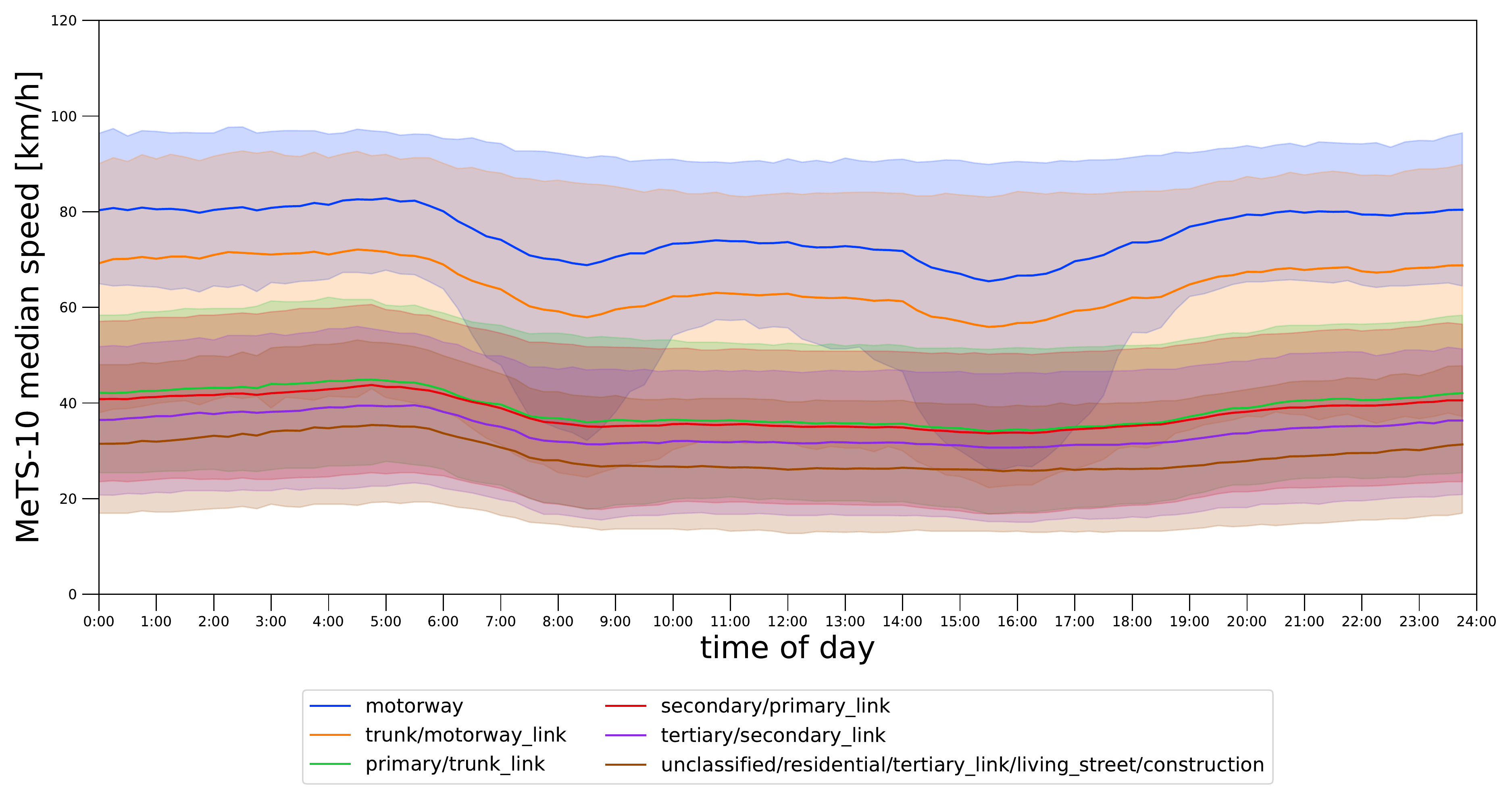}
\caption[Daily median 15 min speeds of all intersecting cells profile Berlin  (MeTS-10 extent (bounding box))]{Daily median 15 min speeds of all intersecting cells profile for different road types for Berlin  (MeTS-10 extent (bounding box)). The error hull is the 80\% data interval [10.0--90.0 percentiles] of daily means from 20 randomly sampled days.}
\label{figures/speed_stats_05_val01_uber/speed_stats_median_speed_kph_berlin_by_highway_in_bb.pdf}
\end{figure}
\clearpage

\subsection{Key Figures Barcelona}
\subsubsection{Road graph map Barcelona}
\mbox{}
\nopagebreak{}
\begin{figure}[H]
\centering
\includegraphics[width=0.85\textwidth]{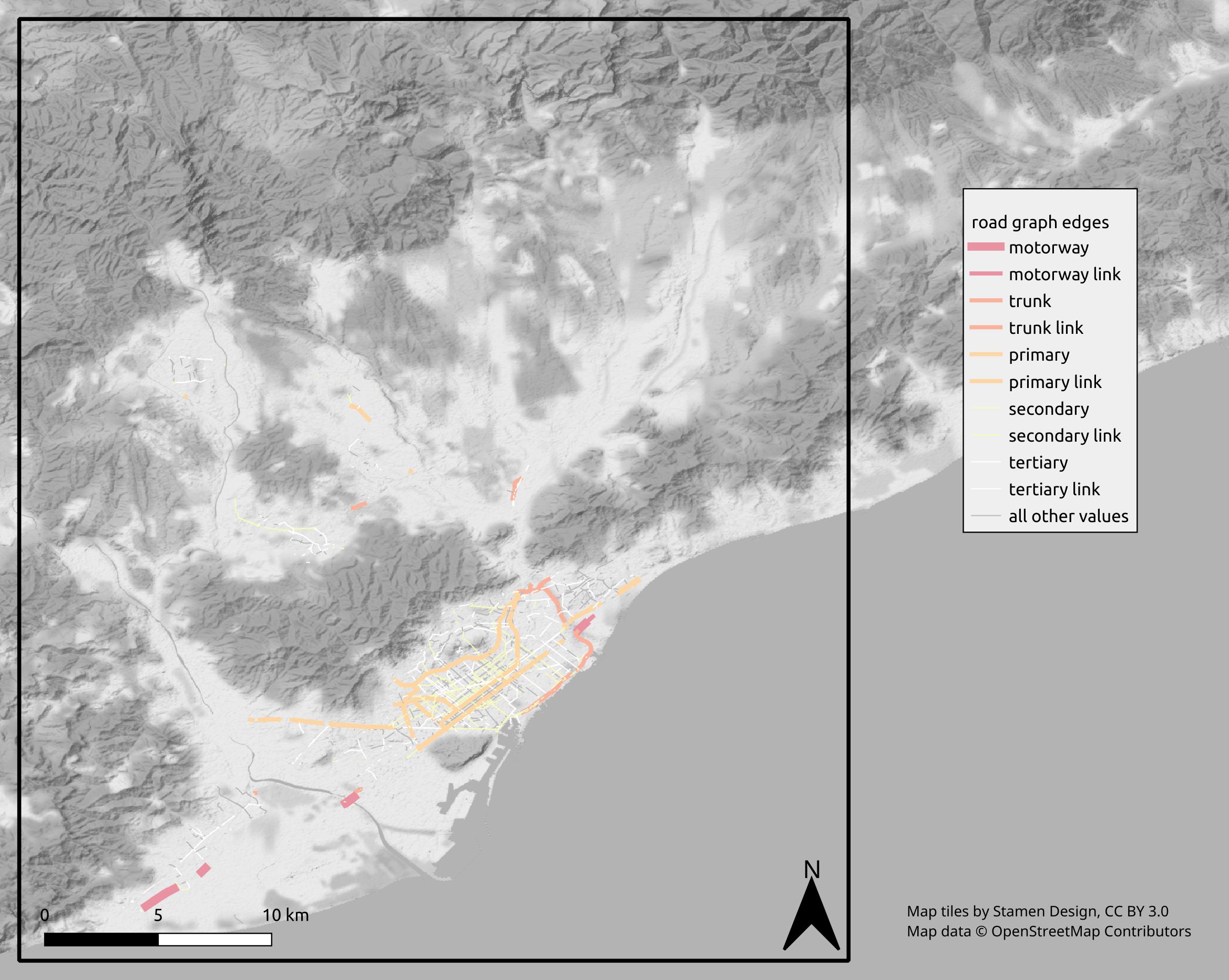}
\caption[Road graph Barcelona]{Road graph Barcelona, OSM color scheme.}
\label{figures/speed_stats_05_val01_uber/road_graph_barcelona.jpg}
\end{figure}
\subsubsection{Static data  Barcelona   (full historic road graph)}
\mbox{}\nopagebreak
\begin{small}
\begin{longtable}{p{4cm}rrrrrrrr}
\toprule
Attribute      & {mean} &{std} & {median}  & {q01} & {q99} & {data points} & {sum}  \\
\midrule
 bounding box  (full historic road graph)               &  &  &  &  &  &  1.925--2.361 / 41.253--41.748 &                                                \\
 num\_edges  (full historic road graph)               &  &  &  &  &  &  5'943 &                                                \\
 \hspace{10pt}  motorway               &  &  &  &  &  &  16 &                                                \\
 \hspace{10pt}  motorway\_link               &  &  &  &  &  &  3 &                                                \\
 \hspace{10pt}  trunk               &  &  &  &  &  &  56 &                                                \\
 \hspace{10pt}  trunk\_link               &  &  &  &  &  &  44 &                                                \\
 \hspace{10pt}  primary               &  &  &  &  &  &  709 &                                                \\
 \hspace{10pt}  primary\_link               &  &  &  &  &  &  216 &                                                \\
 \hspace{10pt}  secondary               &  &  &  &  &  &  1278 &                                                \\
 \hspace{10pt}  secondary\_link               &  &  &  &  &  &  92 &                                                \\
 \hspace{10pt}  tertiary               &  &  &  &  &  &  2014 &                                                \\
 \hspace{10pt}  tertiary\_link               &  &  &  &  &  &  130 &                                                \\
 \hspace{10pt}  unclassified               &  &  &  &  &  &  20 &                                                \\
 \hspace{10pt}  residential               &  &  &  &  &  &  1341 &                                                \\
 \hspace{10pt}  living\_street               &  &  &  &  &  &  24 &                                                \\
 num\_nodes  (full historic road graph)               &  &  &  &  &  &  5530 &                                                 \\
 num\_edges\_per\_cell  (full historic road graph)               & 1.1 & 0.4 & 1.0 & 1.0 & 3.0 &  13'823 &                                                 \\
 num\_intersecting\_cells  (full historic road graph)               & 2.5 & 1.5 & 2.0 & 1.0 & 8.0 &  5'943 &                                                 \\
 node\_degree  (full historic road graph)               & 2.0 & 0.7 & 2.0 & 1.0 & 4.0 &  5'530 &                                                 \\
 length\_meters  (full historic road graph)               & 80.0 & 74.0 & 61.9 & 3.9 & 327.3 &  5'943 & 4.8e+05                                                \\
 \hspace{10pt}  motorway               & 225.8 & 223.5 & 133.2 & 25.4 & 651.4 &  16  & 3.6e+03                                                \\
 \hspace{10pt}  motorway\_link               & 202.8 & 130.0 & 196.9 & 78.3 & 333.0 &  3  & 6.1e+02                                                \\
 \hspace{10pt}  trunk               & 212.8 & 202.4 & 130.6 & 7.9 & 880.2 &  56  & 1.2e+04                                                \\
 \hspace{10pt}  trunk\_link               & 122.5 & 102.9 & 90.1 & 18.7 & 439.1 &  44  & 5.4e+03                                                \\
 \hspace{10pt}  primary               & 83.5 & 78.4 & 69.0 & 4.3 & 350.4 &  709  & 5.9e+04                                                \\
 \hspace{10pt}  primary\_link               & 71.6 & 65.1 & 44.5 & 4.6 & 274.5 &  216  & 1.5e+04                                                \\
 \hspace{10pt}  secondary               & 86.8 & 80.8 & 68.0 & 4.4 & 365.5 &  1'278  & 1.1e+05                                                \\
 \hspace{10pt}  secondary\_link               & 52.3 & 70.4 & 32.1 & 2.7 & 279.8 &  92  & 4.8e+03                                                \\
 \hspace{10pt}  tertiary               & 73.1 & 62.2 & 56.0 & 3.6 & 275.8 &  2'014  & 1.5e+05                                                \\
 \hspace{10pt}  tertiary\_link               & 67.6 & 70.3 & 33.4 & 3.6 & 258.8 &  130  & 8.8e+03                                                \\
 \hspace{10pt}  unclassified               & 89.9 & 99.5 & 48.1 & 7.1 & 366.2 &  20  & 1.8e+03                                                \\
 \hspace{10pt}  residential               & 77.8 & 58.6 & 65.5 & 4.0 & 266.4 &  1'341  & 1.0e+05                                                \\
 \hspace{10pt}  living\_street               & 68.6 & 42.6 & 63.2 & 6.2 & 153.9 &  24  & 1.6e+03                                                \\
 speed\_kph  (full historic road graph)               & 47.3 & 7.3 & 49.7 & 30.0 & 80.0 &  5'943 &                                                 \\
 \hspace{10pt}  motorway               & 90.0 & 11.0 & 90.0 & 80.0 & 117.0 &  16  &                                                 \\
 \hspace{10pt}  motorway\_link               & 56.7 & 5.8 & 60.0 & 50.2 & 60.0 &  3  &                                                 \\
 \hspace{10pt}  trunk               & 78.5 & 6.4 & 80.0 & 60.0 & 89.0 &  56  &                                                 \\
 \hspace{10pt}  trunk\_link               & 57.3 & 9.9 & 57.4 & 40.0 & 80.0 &  44  &                                                 \\
 \hspace{10pt}  primary               & 49.7 & 2.5 & 50.0 & 30.8 & 50.0 &  709  &                                                 \\
 \hspace{10pt}  primary\_link               & 47.2 & 7.3 & 50.0 & 20.0 & 58.5 &  216  &                                                 \\
 \hspace{10pt}  secondary               & 49.7 & 1.5 & 50.0 & 40.0 & 50.0 &  1'278  &                                                 \\
 \hspace{10pt}  secondary\_link               & 49.1 & 3.2 & 50.0 & 30.0 & 50.0 &  92  &                                                 \\
 \hspace{10pt}  tertiary               & 49.5 & 2.6 & 50.0 & 31.3 & 50.0 &  2'014  &                                                 \\
 \hspace{10pt}  tertiary\_link               & 49.3 & 3.0 & 50.0 & 30.0 & 50.0 &  130  &                                                 \\
 \hspace{10pt}  unclassified               & 41.4 & 6.0 & 41.4 & 30.0 & 50.0 &  20  &                                                 \\
 \hspace{10pt}  residential               & 38.6 & 5.4 & 38.6 & 30.0 & 50.0 &  1'341  &                                                 \\
 \hspace{10pt}  living\_street               & 16.6 & 7.8 & 16.3 & 10.0 & 30.0 &  24  &                                                 \\
 free\_flow\_kph  (full historic road graph)               & 37.1 & 14.8 & 35.3 & 14.4 & 91.4 &  5'943 &                                                 \\
 \hspace{10pt}  motorway               & 103.7 & 13.4 & 93.6 & 90.8 & 119.8 &  16  &                                                 \\
 \hspace{10pt}  motorway\_link               & 85.4 & 17.2 & 94.7 & 66.2 & 96.0 &  3  &                                                 \\
 \hspace{10pt}  trunk               & 80.8 & 8.6 & 82.1 & 63.5 & 99.8 &  56  &                                                 \\
 \hspace{10pt}  trunk\_link               & 74.9 & 12.9 & 76.7 & 39.9 & 92.7 &  44  &                                                 \\
 \hspace{10pt}  primary               & 41.6 & 11.4 & 42.1 & 20.9 & 90.2 &  709  &                                                 \\
 \hspace{10pt}  primary\_link               & 46.2 & 13.2 & 47.5 & 18.9 & 81.7 &  216  &                                                 \\
 \hspace{10pt}  secondary               & 38.8 & 11.5 & 37.6 & 17.8 & 80.9 &  1'278  &                                                 \\
 \hspace{10pt}  secondary\_link               & 39.0 & 14.8 & 35.3 & 18.4 & 85.0 &  92  &                                                 \\
 \hspace{10pt}  tertiary               & 35.7 & 13.6 & 32.9 & 15.1 & 88.5 &  2'014  &                                                 \\
 \hspace{10pt}  tertiary\_link               & 35.9 & 11.5 & 34.1 & 14.3 & 78.8 &  130  &                                                 \\
 \hspace{10pt}  unclassified               & 47.4 & 29.2 & 36.7 & 19.2 & 117.1 &  20  &                                                 \\
 \hspace{10pt}  residential               & 30.0 & 12.2 & 28.7 & 11.5 & 83.0 &  1'341  &                                                 \\
 \hspace{10pt}  living\_street               & 25.0 & 10.4 & 20.5 & 9.1 & 43.8 &  24  &                                                 \\
 free\_flow\_kph-speed\_kph  (full historic road graph)               & -10.2 & 13.7 & -11.9 & -33.5 & 38.5 &  5'943 &                                                 \\
 \hspace{10pt}  motorway               & 13.7 & 13.5 & 11.9 & -6.4 & 29.8 &  16  &                                                 \\
 \hspace{10pt}  motorway\_link               & 28.8 & 11.4 & 34.7 & 16.0 & 36.0 &  3  &                                                 \\
 \hspace{10pt}  trunk               & 2.3 & 7.6 & 3.2 & -13.3 & 19.8 &  56  &                                                 \\
 \hspace{10pt}  trunk\_link               & 17.6 & 15.1 & 16.7 & -14.2 & 52.7 &  44  &                                                 \\
 \hspace{10pt}  primary               & -8.1 & 11.4 & -7.3 & -29.1 & 40.2 &  709  &                                                 \\
 \hspace{10pt}  primary\_link               & -1.0 & 15.8 & -1.1 & -31.0 & 50.0 &  216  &                                                 \\
 \hspace{10pt}  secondary               & -10.9 & 11.8 & -12.1 & -32.0 & 31.2 &  1'278  &                                                 \\
 \hspace{10pt}  secondary\_link               & -10.1 & 15.5 & -14.7 & -29.0 & 45.0 &  92  &                                                 \\
 \hspace{10pt}  tertiary               & -13.9 & 13.5 & -16.1 & -34.9 & 36.3 &  2'014  &                                                 \\
 \hspace{10pt}  tertiary\_link               & -13.4 & 12.2 & -15.2 & -35.0 & 29.7 &  130  &                                                 \\
 \hspace{10pt}  unclassified               & 6.0 & 29.9 & -3.8 & -22.2 & 75.7 &  20  &                                                 \\
 \hspace{10pt}  residential               & -8.6 & 12.9 & -9.4 & -33.1 & 44.4 &  1'341  &                                                 \\
 \hspace{10pt}  living\_street               & 8.4 & 17.0 & 5.0 & -20.9 & 33.8 &  24  &                                                 \\
\bottomrule

        \caption[Key figures Barcelona  (full historic road graph)]{Key figures Barcelona for the generated data from 20 randomly sampled days (full historic road graph).
        \textbf{num\_edges} number of edges in the street network graph;
        \textbf{num\_nodes} number of nodes in the street network graph;
        \textbf{num\_edges\_per\_cell} number of edges a cell (row,col,heading) has in its intersecting cells;
        \textbf{num\_intersecting\_cells} number of cells (row,col,heading) in an edge's intersecting cells;
        \textbf{node\_degree} number of (unique) neighbor nodes per node;
        \textbf{length\_meters} free flow speed derived from data;
        \textbf{speed\_kph} signalled speed;
        \textbf{free\_flow\_kph} free flow speed derived from data;
        \textbf{free\_flow\_kph-speed\_kph} difference
        }
    \label{tab:key_figures:/iarai/public/t4c/data_pipeline/release20221028_historic_uber:Barcelona: (full historic road graph)}
    \end{longtable}
    \end{small}
    
\subsubsection{Static data  Barcelona   (MeTS-10 extent (bounding box))}
\mbox{}\nopagebreak
\begin{small}
\begin{longtable}{p{4cm}rrrrrrrr}
\toprule
Attribute      & {mean} &{std} & {median}  & {q01} & {q99} & {data points} & {sum}  \\
\midrule
 bounding box  (MeTS-10 extent (bounding box))               &  &  &  &  &  &  1.925--2.361 / 41.253--41.748 &                                                \\
 num\_edges  (MeTS-10 extent (bounding box))               &  &  &  &  &  &  5'943 &                                                \\
 \hspace{10pt}  motorway               &  &  &  &  &  &  16 &                                                \\
 \hspace{10pt}  motorway\_link               &  &  &  &  &  &  3 &                                                \\
 \hspace{10pt}  trunk               &  &  &  &  &  &  56 &                                                \\
 \hspace{10pt}  trunk\_link               &  &  &  &  &  &  44 &                                                \\
 \hspace{10pt}  primary               &  &  &  &  &  &  709 &                                                \\
 \hspace{10pt}  primary\_link               &  &  &  &  &  &  216 &                                                \\
 \hspace{10pt}  secondary               &  &  &  &  &  &  1278 &                                                \\
 \hspace{10pt}  secondary\_link               &  &  &  &  &  &  92 &                                                \\
 \hspace{10pt}  tertiary               &  &  &  &  &  &  2014 &                                                \\
 \hspace{10pt}  tertiary\_link               &  &  &  &  &  &  130 &                                                \\
 \hspace{10pt}  unclassified               &  &  &  &  &  &  20 &                                                \\
 \hspace{10pt}  residential               &  &  &  &  &  &  1341 &                                                \\
 \hspace{10pt}  living\_street               &  &  &  &  &  &  24 &                                                \\
 num\_nodes  (MeTS-10 extent (bounding box))               &  &  &  &  &  &  5530 &                                                 \\
 num\_edges\_per\_cell  (MeTS-10 extent (bounding box))               & 1.1 & 0.4 & 1.0 & 1.0 & 3.0 &  13'823 &                                                 \\
 num\_intersecting\_cells  (MeTS-10 extent (bounding box))               & 2.5 & 1.5 & 2.0 & 1.0 & 8.0 &  5'943 &                                                 \\
 node\_degree  (MeTS-10 extent (bounding box))               & 2.0 & 0.7 & 2.0 & 1.0 & 4.0 &  5'530 &                                                 \\
 length\_meters  (MeTS-10 extent (bounding box))               & 80.0 & 74.0 & 61.9 & 3.9 & 327.3 &  5'943 & 4.8e+05                                                \\
 \hspace{10pt}  motorway               & 225.8 & 223.5 & 133.2 & 25.4 & 651.4 &  16  & 3.6e+03                                                \\
 \hspace{10pt}  motorway\_link               & 202.8 & 130.0 & 196.9 & 78.3 & 333.0 &  3  & 6.1e+02                                                \\
 \hspace{10pt}  trunk               & 212.8 & 202.4 & 130.6 & 7.9 & 880.2 &  56  & 1.2e+04                                                \\
 \hspace{10pt}  trunk\_link               & 122.5 & 102.9 & 90.1 & 18.7 & 439.1 &  44  & 5.4e+03                                                \\
 \hspace{10pt}  primary               & 83.5 & 78.4 & 69.0 & 4.3 & 350.4 &  709  & 5.9e+04                                                \\
 \hspace{10pt}  primary\_link               & 71.6 & 65.1 & 44.5 & 4.6 & 274.5 &  216  & 1.5e+04                                                \\
 \hspace{10pt}  secondary               & 86.8 & 80.8 & 68.0 & 4.4 & 365.5 &  1'278  & 1.1e+05                                                \\
 \hspace{10pt}  secondary\_link               & 52.3 & 70.4 & 32.1 & 2.7 & 279.8 &  92  & 4.8e+03                                                \\
 \hspace{10pt}  tertiary               & 73.1 & 62.2 & 56.0 & 3.6 & 275.8 &  2'014  & 1.5e+05                                                \\
 \hspace{10pt}  tertiary\_link               & 67.6 & 70.3 & 33.4 & 3.6 & 258.8 &  130  & 8.8e+03                                                \\
 \hspace{10pt}  unclassified               & 89.9 & 99.5 & 48.1 & 7.1 & 366.2 &  20  & 1.8e+03                                                \\
 \hspace{10pt}  residential               & 77.8 & 58.6 & 65.5 & 4.0 & 266.4 &  1'341  & 1.0e+05                                                \\
 \hspace{10pt}  living\_street               & 68.6 & 42.6 & 63.2 & 6.2 & 153.9 &  24  & 1.6e+03                                                \\
 speed\_kph  (MeTS-10 extent (bounding box))               & 47.3 & 7.3 & 49.7 & 30.0 & 80.0 &  5'943 &                                                 \\
 \hspace{10pt}  motorway               & 90.0 & 11.0 & 90.0 & 80.0 & 117.0 &  16  &                                                 \\
 \hspace{10pt}  motorway\_link               & 56.7 & 5.8 & 60.0 & 50.2 & 60.0 &  3  &                                                 \\
 \hspace{10pt}  trunk               & 78.5 & 6.4 & 80.0 & 60.0 & 89.0 &  56  &                                                 \\
 \hspace{10pt}  trunk\_link               & 57.3 & 9.9 & 57.4 & 40.0 & 80.0 &  44  &                                                 \\
 \hspace{10pt}  primary               & 49.7 & 2.5 & 50.0 & 30.8 & 50.0 &  709  &                                                 \\
 \hspace{10pt}  primary\_link               & 47.2 & 7.3 & 50.0 & 20.0 & 58.5 &  216  &                                                 \\
 \hspace{10pt}  secondary               & 49.7 & 1.5 & 50.0 & 40.0 & 50.0 &  1'278  &                                                 \\
 \hspace{10pt}  secondary\_link               & 49.1 & 3.2 & 50.0 & 30.0 & 50.0 &  92  &                                                 \\
 \hspace{10pt}  tertiary               & 49.5 & 2.6 & 50.0 & 31.3 & 50.0 &  2'014  &                                                 \\
 \hspace{10pt}  tertiary\_link               & 49.3 & 3.0 & 50.0 & 30.0 & 50.0 &  130  &                                                 \\
 \hspace{10pt}  unclassified               & 41.4 & 6.0 & 41.4 & 30.0 & 50.0 &  20  &                                                 \\
 \hspace{10pt}  residential               & 38.6 & 5.4 & 38.6 & 30.0 & 50.0 &  1'341  &                                                 \\
 \hspace{10pt}  living\_street               & 16.6 & 7.8 & 16.3 & 10.0 & 30.0 &  24  &                                                 \\
 free\_flow\_kph  (MeTS-10 extent (bounding box))               & 37.1 & 14.8 & 35.3 & 14.4 & 91.4 &  5'943 &                                                 \\
 \hspace{10pt}  motorway               & 103.7 & 13.4 & 93.6 & 90.8 & 119.8 &  16  &                                                 \\
 \hspace{10pt}  motorway\_link               & 85.4 & 17.2 & 94.7 & 66.2 & 96.0 &  3  &                                                 \\
 \hspace{10pt}  trunk               & 80.8 & 8.6 & 82.1 & 63.5 & 99.8 &  56  &                                                 \\
 \hspace{10pt}  trunk\_link               & 74.9 & 12.9 & 76.7 & 39.9 & 92.7 &  44  &                                                 \\
 \hspace{10pt}  primary               & 41.6 & 11.4 & 42.1 & 20.9 & 90.2 &  709  &                                                 \\
 \hspace{10pt}  primary\_link               & 46.2 & 13.2 & 47.5 & 18.9 & 81.7 &  216  &                                                 \\
 \hspace{10pt}  secondary               & 38.8 & 11.5 & 37.6 & 17.8 & 80.9 &  1'278  &                                                 \\
 \hspace{10pt}  secondary\_link               & 39.0 & 14.8 & 35.3 & 18.4 & 85.0 &  92  &                                                 \\
 \hspace{10pt}  tertiary               & 35.7 & 13.6 & 32.9 & 15.1 & 88.5 &  2'014  &                                                 \\
 \hspace{10pt}  tertiary\_link               & 35.9 & 11.5 & 34.1 & 14.3 & 78.8 &  130  &                                                 \\
 \hspace{10pt}  unclassified               & 47.4 & 29.2 & 36.7 & 19.2 & 117.1 &  20  &                                                 \\
 \hspace{10pt}  residential               & 30.0 & 12.2 & 28.7 & 11.5 & 83.0 &  1'341  &                                                 \\
 \hspace{10pt}  living\_street               & 25.0 & 10.4 & 20.5 & 9.1 & 43.8 &  24  &                                                 \\
 free\_flow\_kph-speed\_kph  (MeTS-10 extent (bounding box))               & -10.2 & 13.7 & -11.9 & -33.5 & 38.5 &  5'943 &                                                 \\
 \hspace{10pt}  motorway               & 13.7 & 13.5 & 11.9 & -6.4 & 29.8 &  16  &                                                 \\
 \hspace{10pt}  motorway\_link               & 28.8 & 11.4 & 34.7 & 16.0 & 36.0 &  3  &                                                 \\
 \hspace{10pt}  trunk               & 2.3 & 7.6 & 3.2 & -13.3 & 19.8 &  56  &                                                 \\
 \hspace{10pt}  trunk\_link               & 17.6 & 15.1 & 16.7 & -14.2 & 52.7 &  44  &                                                 \\
 \hspace{10pt}  primary               & -8.1 & 11.4 & -7.3 & -29.1 & 40.2 &  709  &                                                 \\
 \hspace{10pt}  primary\_link               & -1.0 & 15.8 & -1.1 & -31.0 & 50.0 &  216  &                                                 \\
 \hspace{10pt}  secondary               & -10.9 & 11.8 & -12.1 & -32.0 & 31.2 &  1'278  &                                                 \\
 \hspace{10pt}  secondary\_link               & -10.1 & 15.5 & -14.7 & -29.0 & 45.0 &  92  &                                                 \\
 \hspace{10pt}  tertiary               & -13.9 & 13.5 & -16.1 & -34.9 & 36.3 &  2'014  &                                                 \\
 \hspace{10pt}  tertiary\_link               & -13.4 & 12.2 & -15.2 & -35.0 & 29.7 &  130  &                                                 \\
 \hspace{10pt}  unclassified               & 6.0 & 29.9 & -3.8 & -22.2 & 75.7 &  20  &                                                 \\
 \hspace{10pt}  residential               & -8.6 & 12.9 & -9.4 & -33.1 & 44.4 &  1'341  &                                                 \\
 \hspace{10pt}  living\_street               & 8.4 & 17.0 & 5.0 & -20.9 & 33.8 &  24  &                                                 \\
\bottomrule

        \caption[Key figures Barcelona  (MeTS-10 extent (bounding box))]{Key figures Barcelona for the generated data from 20 randomly sampled days (MeTS-10 extent (bounding box)).
        \textbf{num\_edges} number of edges in the street network graph;
        \textbf{num\_nodes} number of nodes in the street network graph;
        \textbf{num\_edges\_per\_cell} number of edges a cell (row,col,heading) has in its intersecting cells;
        \textbf{num\_intersecting\_cells} number of cells (row,col,heading) in an edge's intersecting cells;
        \textbf{node\_degree} number of (unique) neighbor nodes per node;
        \textbf{length\_meters} free flow speed derived from data;
        \textbf{speed\_kph} signalled speed;
        \textbf{free\_flow\_kph} free flow speed derived from data;
        \textbf{free\_flow\_kph-speed\_kph} difference
        }
    \label{tab:key_figures:/iarai/public/t4c/data_pipeline/release20221028_historic_uber:Barcelona: (MeTS-10 extent (bounding box))}
    \end{longtable}
    \end{small}
    
\subsubsection{Segment density map  Barcelona}
\mbox{}
\nopagebreak{}
\begin{figure}[H]
\centering
\includegraphics[width=0.85\textwidth]{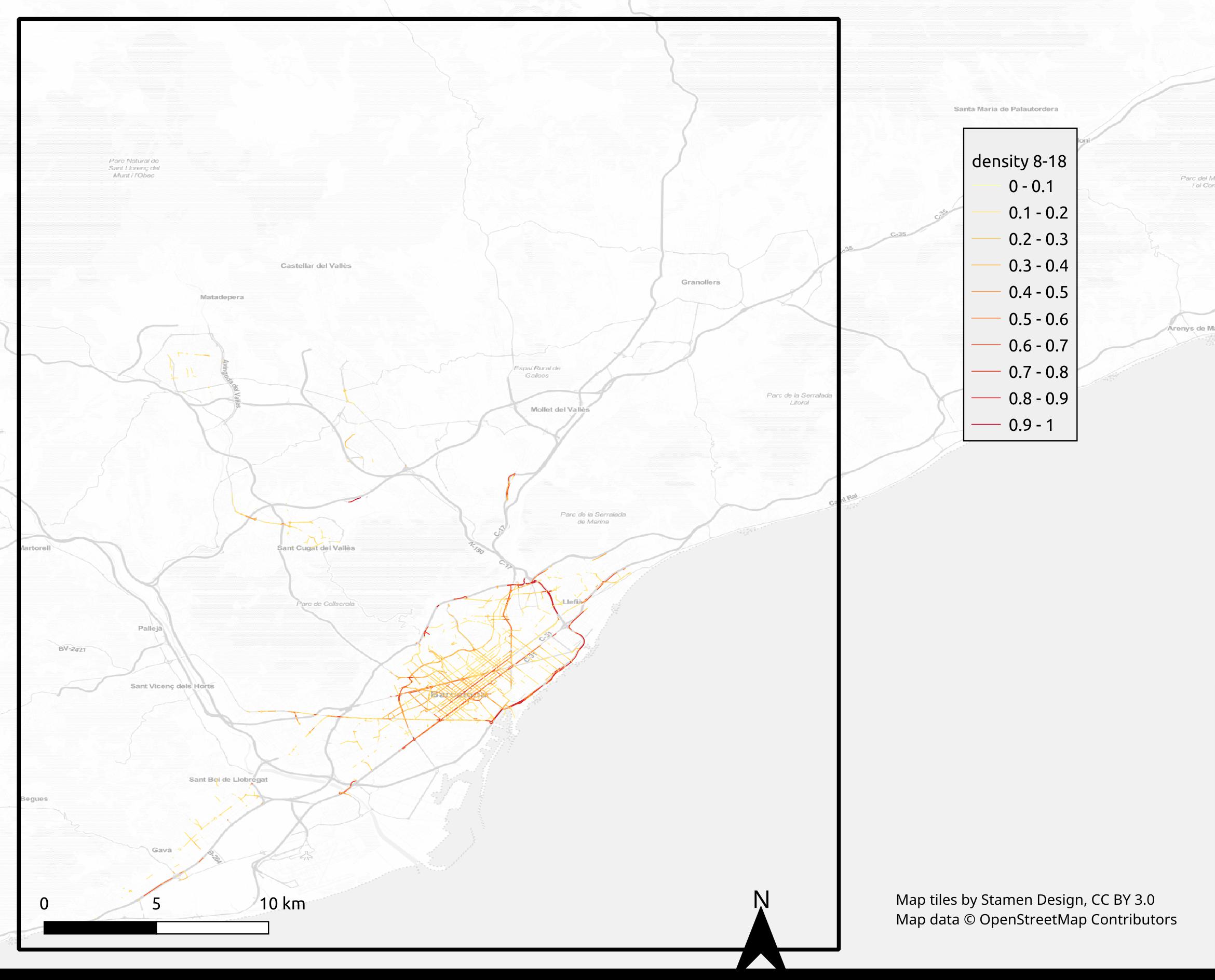}
\caption[Segment-wise density 8am--6pm Barcelona]{Segment-wise density 8am--6pm Barcelona from 20 randomly sampled days.}
\label{figures/speed_stats_05_val01_uber/density_8_18_barcelona.jpg}
\end{figure}
\clearpage
\subsubsection{Daily density profile  Barcelona   (full historic road graph)} 
\mbox{}
\nopagebreak{}
\begin{figure}[H]
\centering
\includegraphics[width=0.85\textwidth]{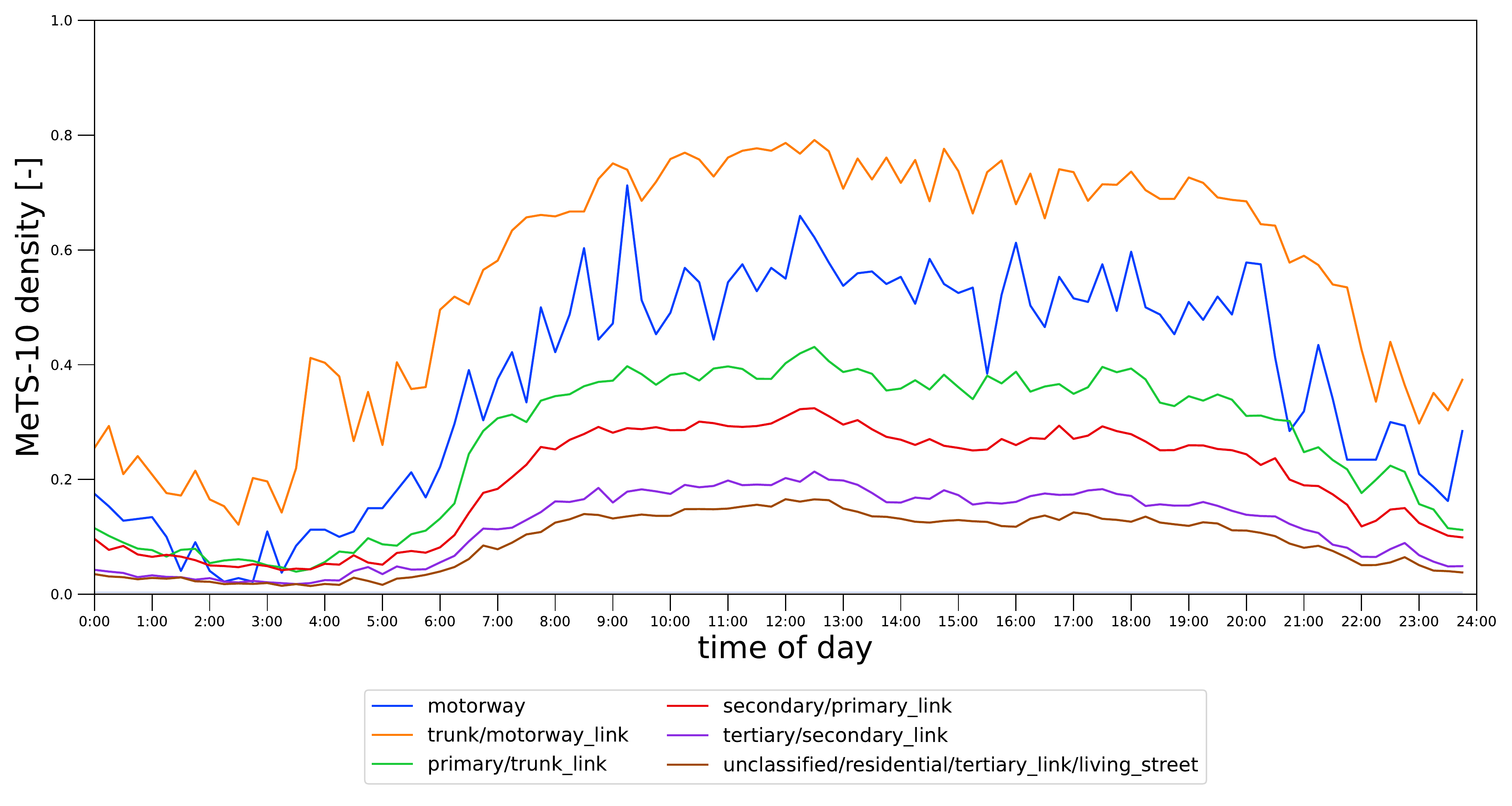}
\caption[Daily density profile Barcelona  (full historic road graph)]{Daily density profile for different road types for Barcelona  (full historic road graph). Data from 20 randomly sampled days.}
\label{figures/speed_stats_05_val01_uber/speed_stats_coverage_barcelona_by_highway.pdf}
\end{figure}
\subsubsection{Daily speed profile  Barcelona   (full historic road graph)}
\mbox{}
\nopagebreak{}
\begin{figure}[H]
\centering
\includegraphics[width=0.85\textwidth]{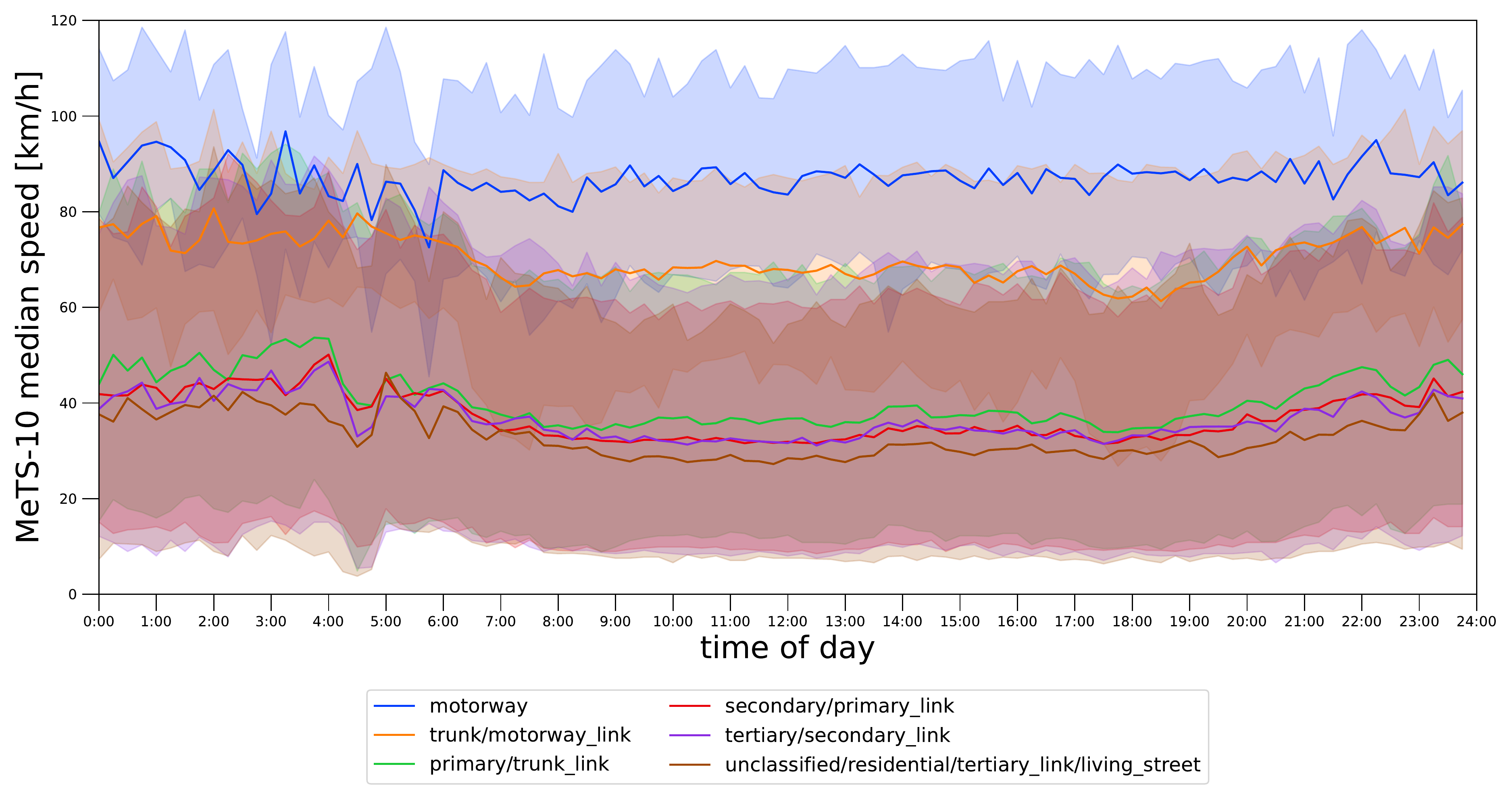}
\caption[Daily median 15 min speeds of all intersecting cells profile Barcelona  (full historic road graph)]{Daily median 15 min speeds of all intersecting cells profile for different road types for Barcelona  (full historic road graph). The error hull is the 80\% data interval [10.0--90.0 percentiles] of daily means from 20 randomly sampled days.}
\label{figures/speed_stats_05_val01_uber/speed_stats_median_speed_kph_barcelona_by_highway.pdf}
\end{figure}
\subsubsection{Daily density profile  Barcelona   (MeTS-10 extent (bounding box))}
\mbox{}
\nopagebreak{}
\begin{figure}[H]
\centering
\includegraphics[width=0.85\textwidth]{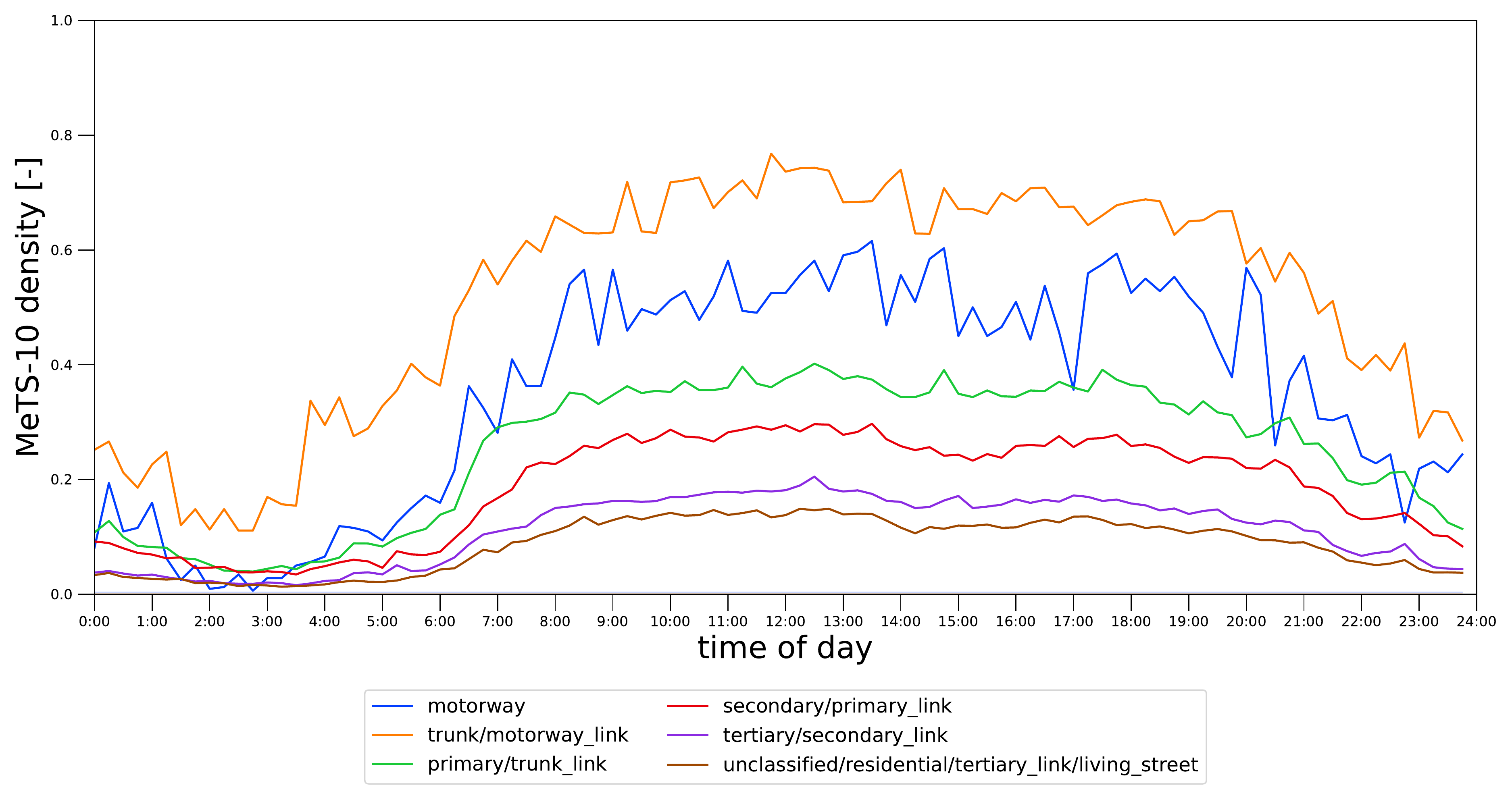}
\caption[Daily density profile Barcelona  (MeTS-10 extent (bounding box))]{Daily density profile for different road types for Barcelona  (MeTS-10 extent (bounding box)). Data from 20 randomly sampled days.}
\label{figures/speed_stats_05_val01_uber/speed_stats_coverage_barcelona_by_highway_in_bb.pdf}
\end{figure}
\subsubsection{Daily speed profile  Barcelona   (MeTS-10 extent (bounding box))}
\mbox{}
\nopagebreak{}
\begin{figure}[H]
\centering
\includegraphics[width=0.85\textwidth]{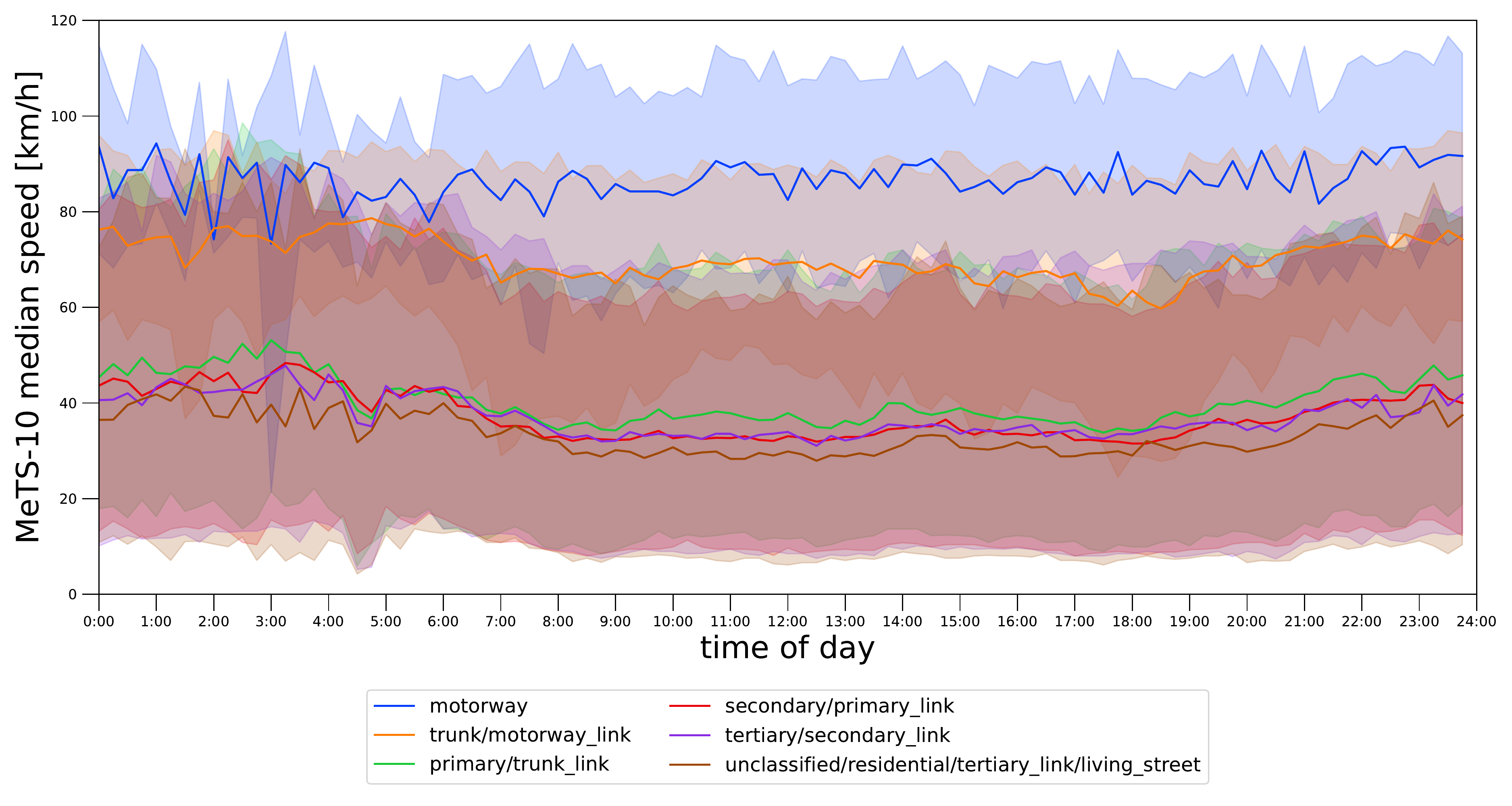}
\caption[Daily median 15 min speeds of all intersecting cells profile Barcelona  (MeTS-10 extent (bounding box))]{Daily median 15 min speeds of all intersecting cells profile for different road types for Barcelona  (MeTS-10 extent (bounding box)). The error hull is the 80\% data interval [10.0--90.0 percentiles] of daily means from 20 randomly sampled days.}
\label{figures/speed_stats_05_val01_uber/speed_stats_median_speed_kph_barcelona_by_highway_in_bb.pdf}
\end{figure}
\clearpage

\pagebreak

\section{Key Figures Uber Comparison}\label{appendix:uber_comparision}

\subsection{Temporal coverage all 3 cities}
\mbox{}
\nopagebreak
\begin{figure}[h]
  \centering
  \begin{subfigure}[b]{0.9\textwidth}
  \includegraphics[width=0.95\textwidth]{figures/05_val01_uber/uber03_spatial_coverage_city_comparison_bb_density_u_density_t.pdf}
  \caption{Uber segments in \t4c bounding box only}
  \end{subfigure}
  \\
  \begin{subfigure}[b]{0.9\textwidth}
  \includegraphics[width=0.95\textwidth]{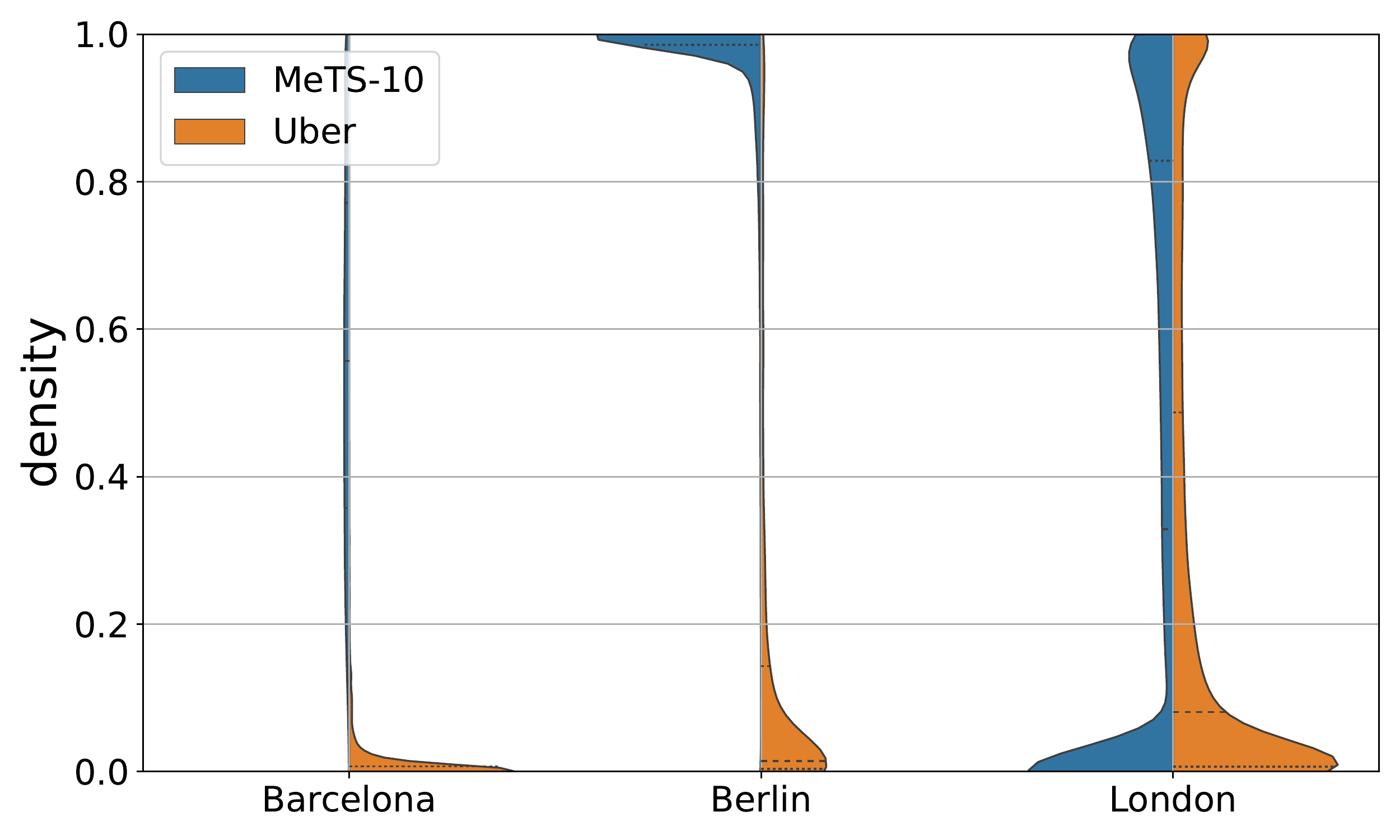}
  \caption{full historic road graph (all Uber segments)}
  \end{subfigure}
  \caption{Violin plot of segment-wise temporal coverage for \mcswts{} and Uber on the historic road graph for the 3 cities Barcelona, Berlin and London. 
  }
  \label{fig:05val02-uber03_spatial_coverage_city_comparison_bb_density_u_density_t}
\end{figure}

\clearpage


\subsection{Barcelona}
\mbox{}
\nopagebreak{}
\begin{figure}[H]
\centering
\includegraphics[width=0.85\textwidth]{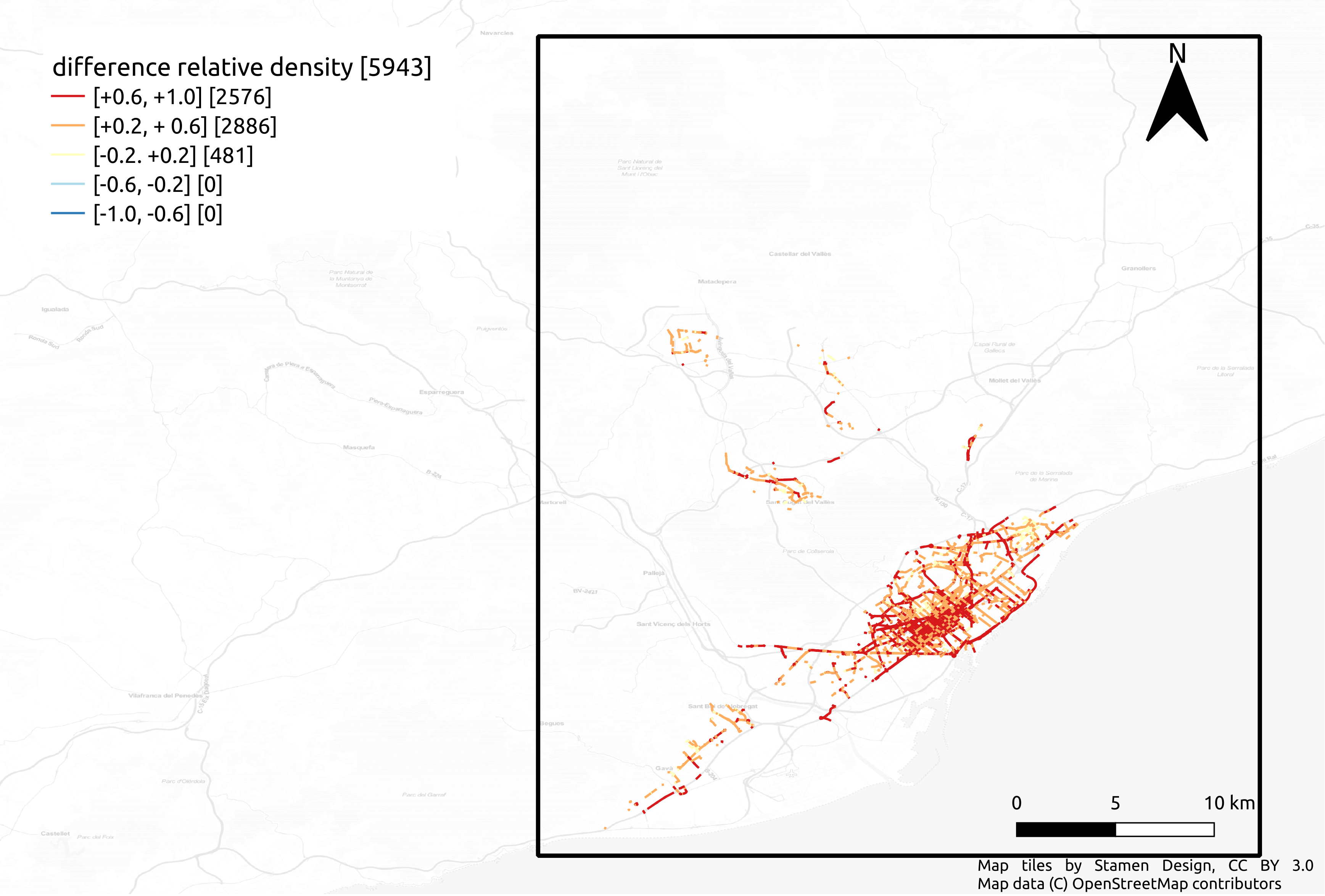}
\caption{Segment density differences of Uber and \mcswts{} on the historic road graph Barcelona (8am--6pm). The color encoding shows the edge density difference, negative means higher temporal coverage of \mcswts{} and positive values mean higher temporal coverage..}
\end{figure}

\begin{figure}[H]
\centering
\includegraphics[width=0.85\textwidth]{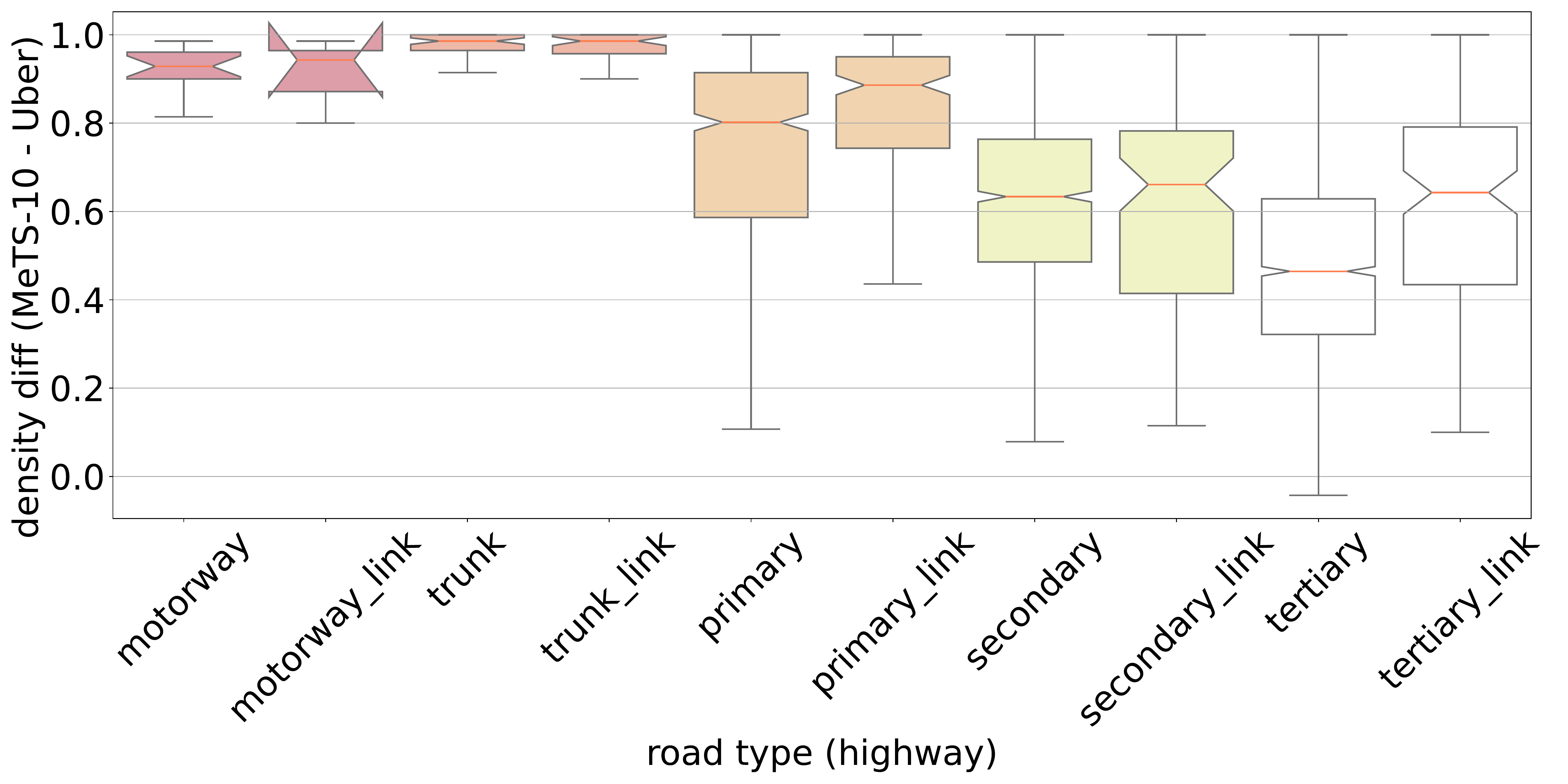}
\caption{Segment density differences Uber and \mcswts{} on the historic road graph Barcelona daytime (8am--6pm, segments within 	4c bounding box only). Mean density difference by road type (\ie{} OSM highway attribute); positive density difference means higher temporal coverage of \mcswts{} and negative mean higher temporal coverage.}
\end{figure}

\begin{figure}[H]
\centering
\begin{subfigure}[b]{0.45\textwidth}
\includegraphics[width=\textwidth]{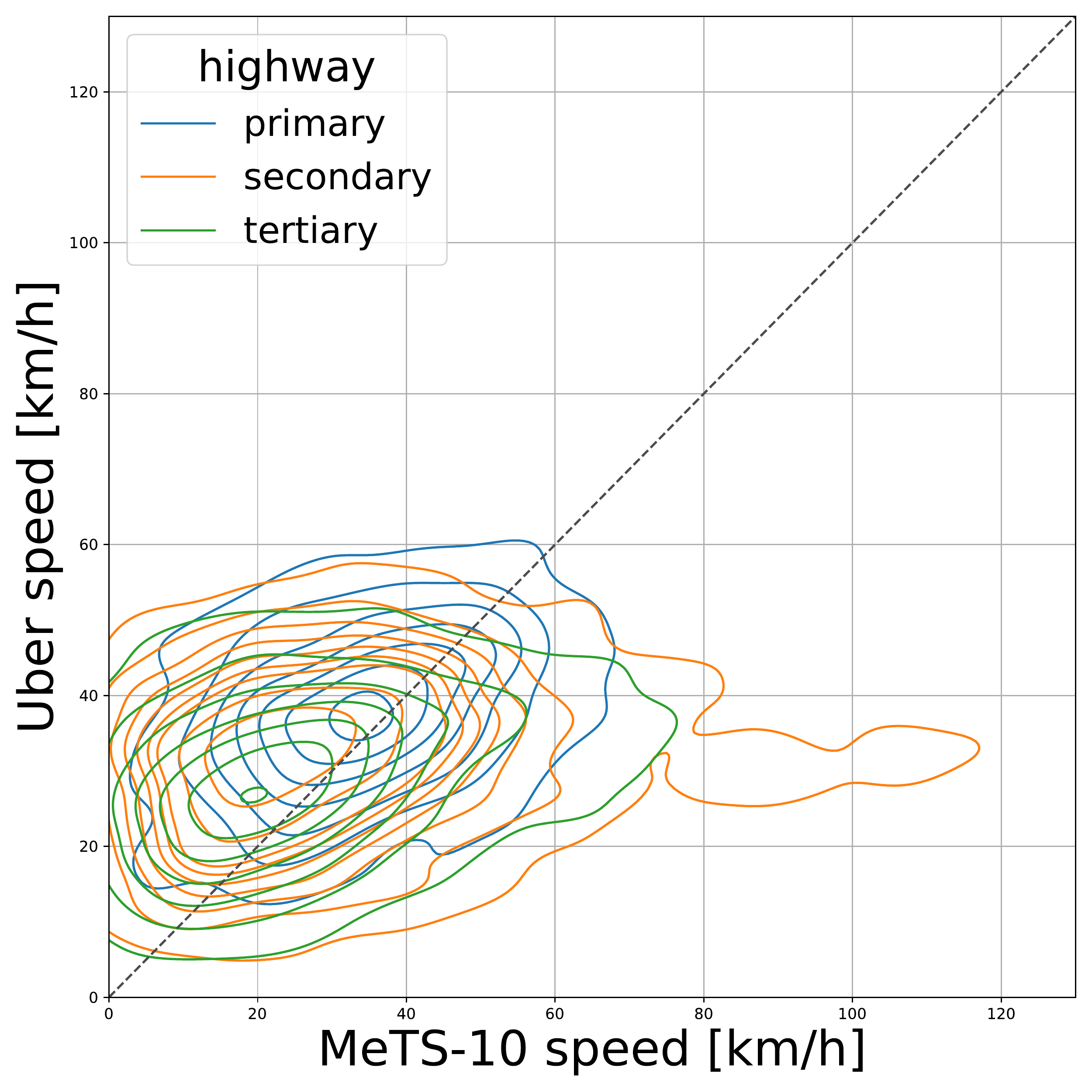}
\caption{KDE non-link road types.}
\end{subfigure}
\begin{subfigure}[b]{0.45\textwidth}
\includegraphics[width=\textwidth]{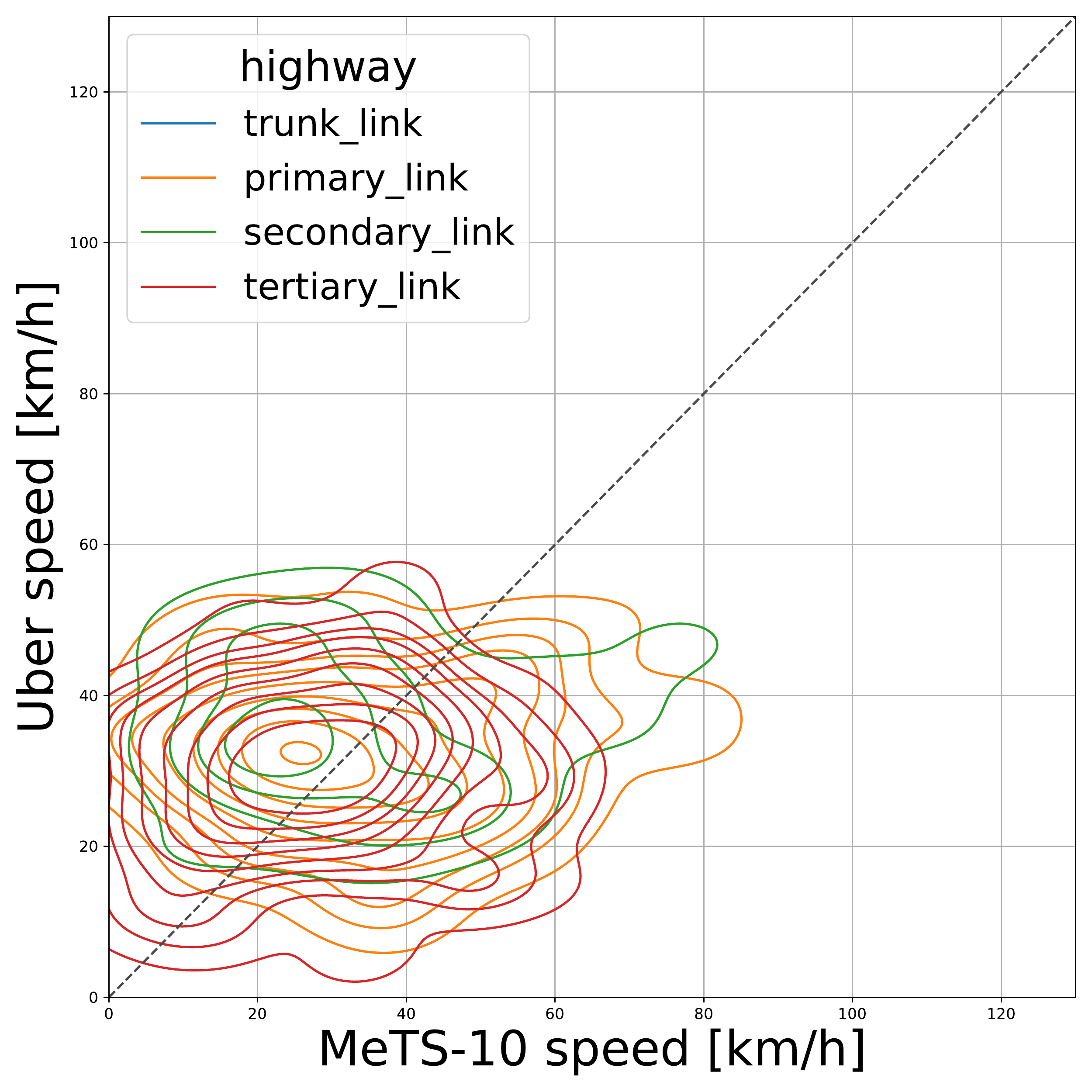}
\caption{KDE link road types.}
\end{subfigure}
\begin{subfigure}[b]{0.9\textwidth}
\includegraphics[width=\textwidth]{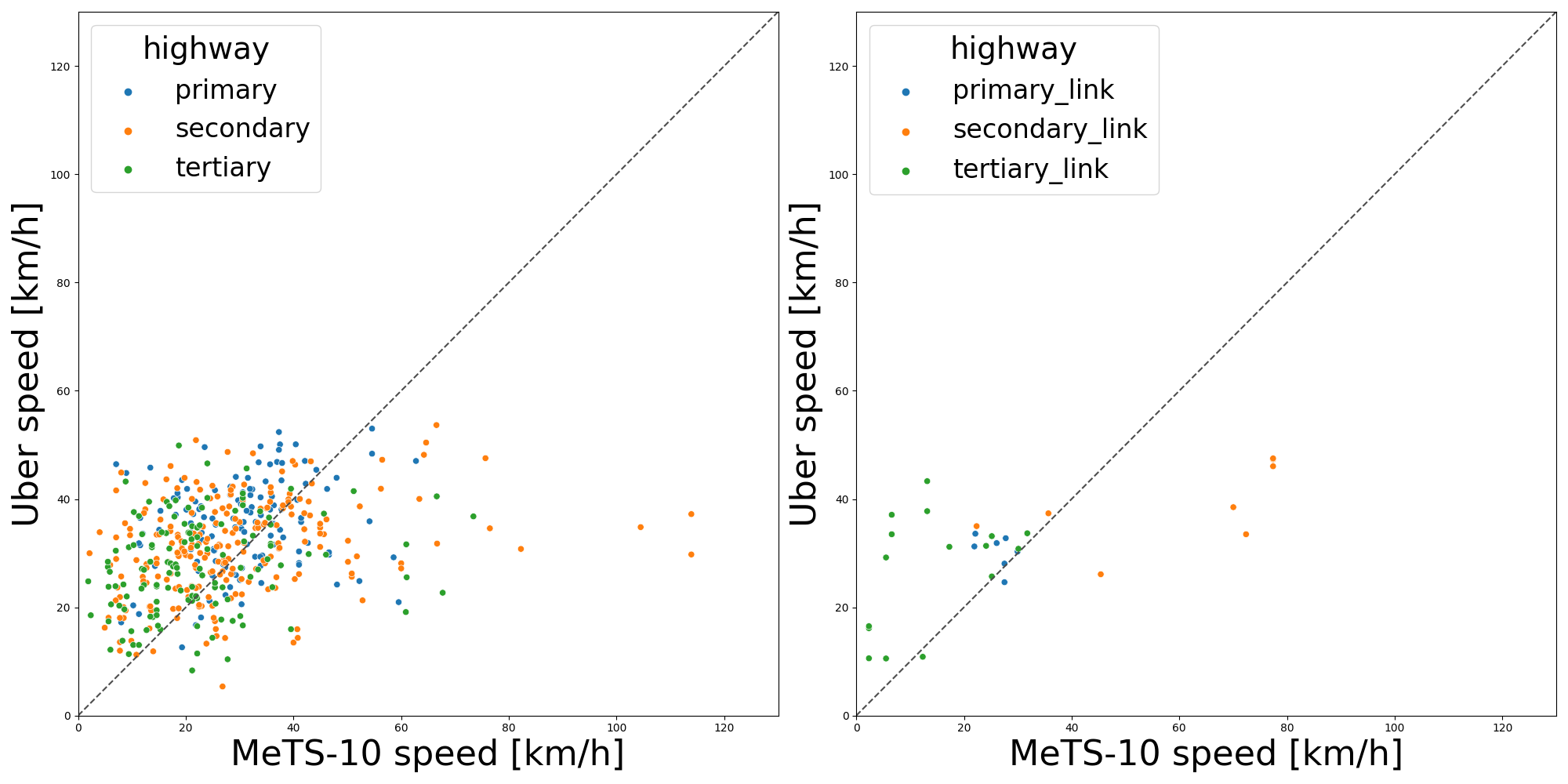}
\caption{Scatter non-link (left) and link (right) road types.}
\end{subfigure}
\caption{Kernel Distribution Estimation and Scatter Plots of speeds of \mcswts{} (x-axis, \texttt{median\_speed\_kph}) and Uber (y-axis, \texttt{speed\_kph\_mean}) on the historic road graph Barcelona daytime (8am--6pm) on the matching data, \ie{} within \mcswts{} bounding box only and where data is available at the same time and segment, for the most important road types.}
\end{figure}

\begin{figure}[H]
\centering
\includegraphics[width=\textwidth]{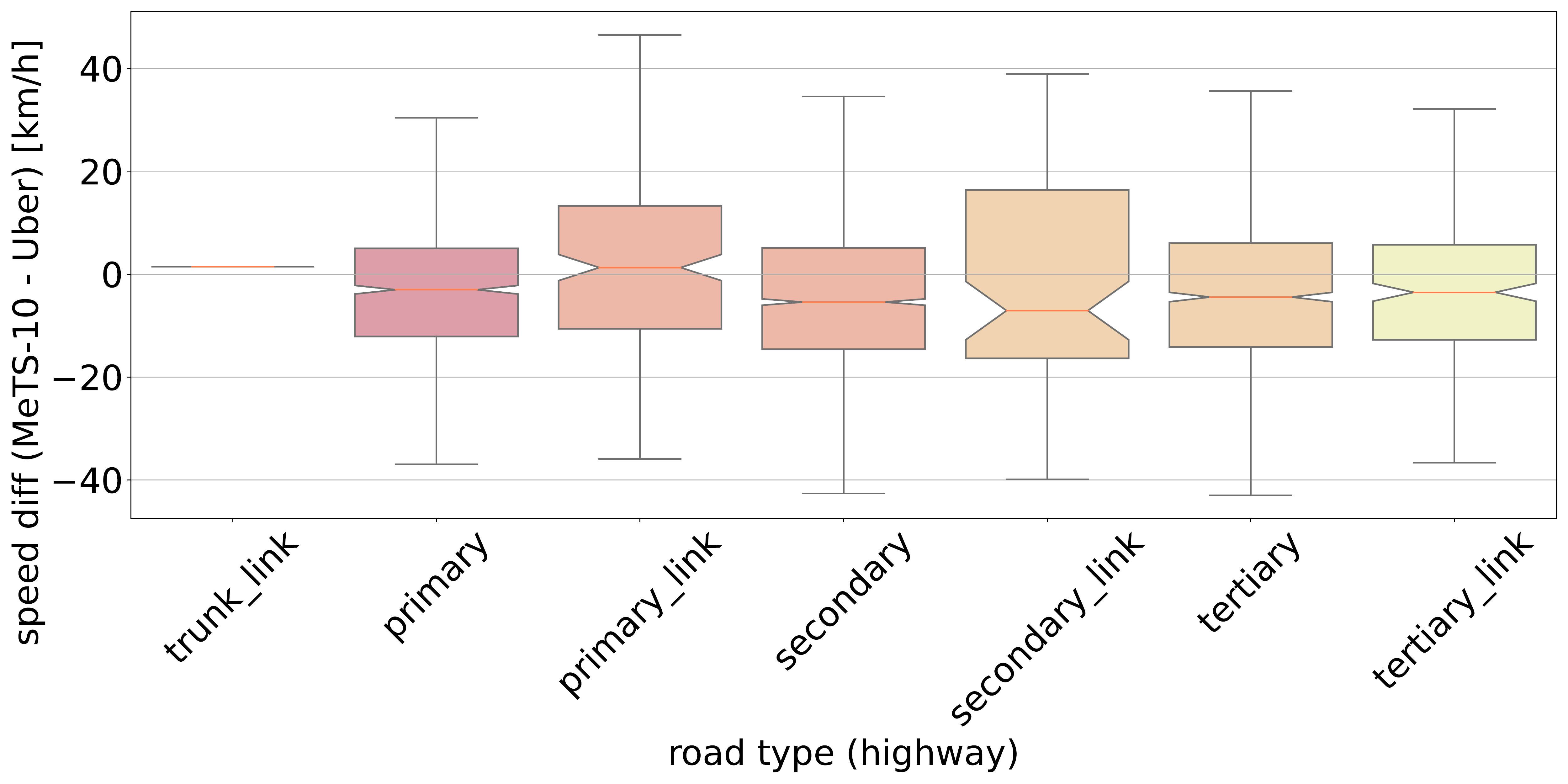}
\caption{Speed differences Uber and \mcswts{} on the historic road graph Barcelona daytime (8am--6pm) on the matching data, \ie{} within \mcswts{} bounding box only and where data is available at the same time and segment. Mean difference by road class (OSM highway attribute). Positive speed difference means higher values in \mcswts.}
\end{figure}
\begin{figure}[H]
\centering
\includegraphics[width=\textwidth]{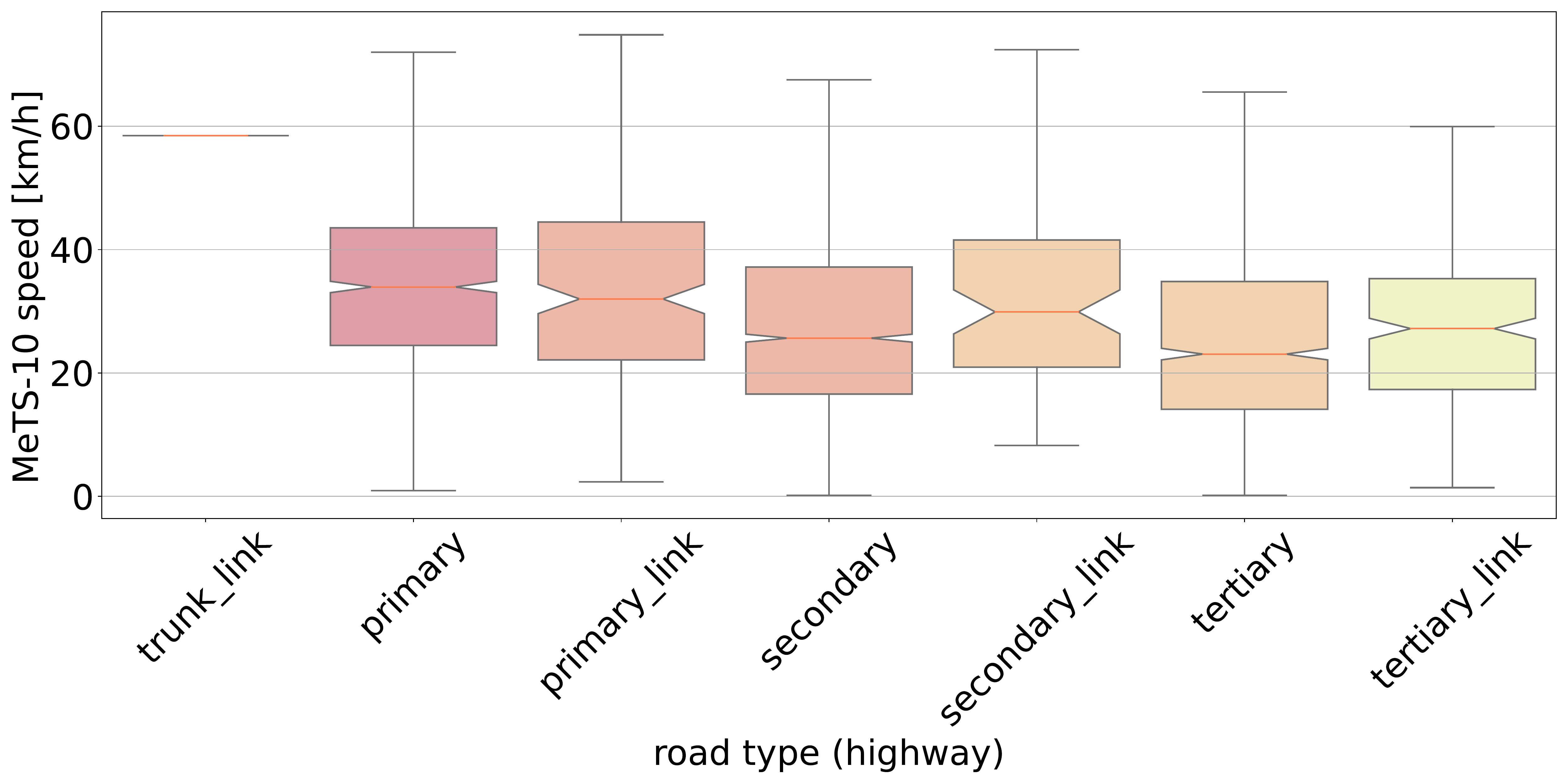}
\caption{\mcswts{} speeds on the historic road graph Barcelona daytime (8am--6pm) on the matching data, \ie{} within \mcswts{} bounding box only and where data is available at the same time and segment. By road class (OSM highway attribute).}
\end{figure}

\begin{figure}[ht]
  \centering
  \includegraphics[width=0.48\textwidth]{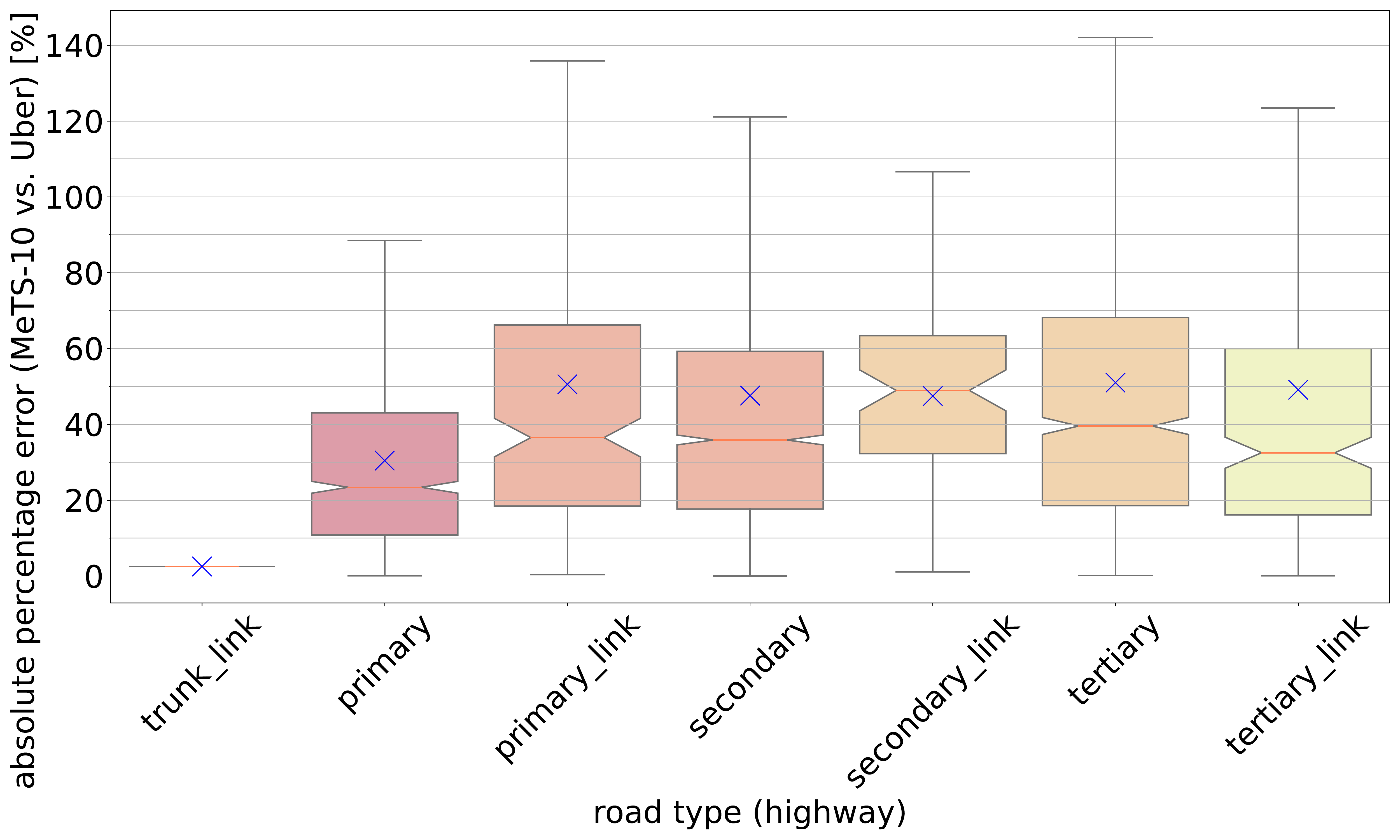}
  \caption{Barcelona absolute percentage error \mcswts{} vs. Uber by road type. Blue crosses indicate the mean per road type.}
  \label{fig:Barcelona_Uber_ape}
\end{figure}

\begin{figure}[ht!]
  \centering
  \includegraphics[width=0.48\textwidth]{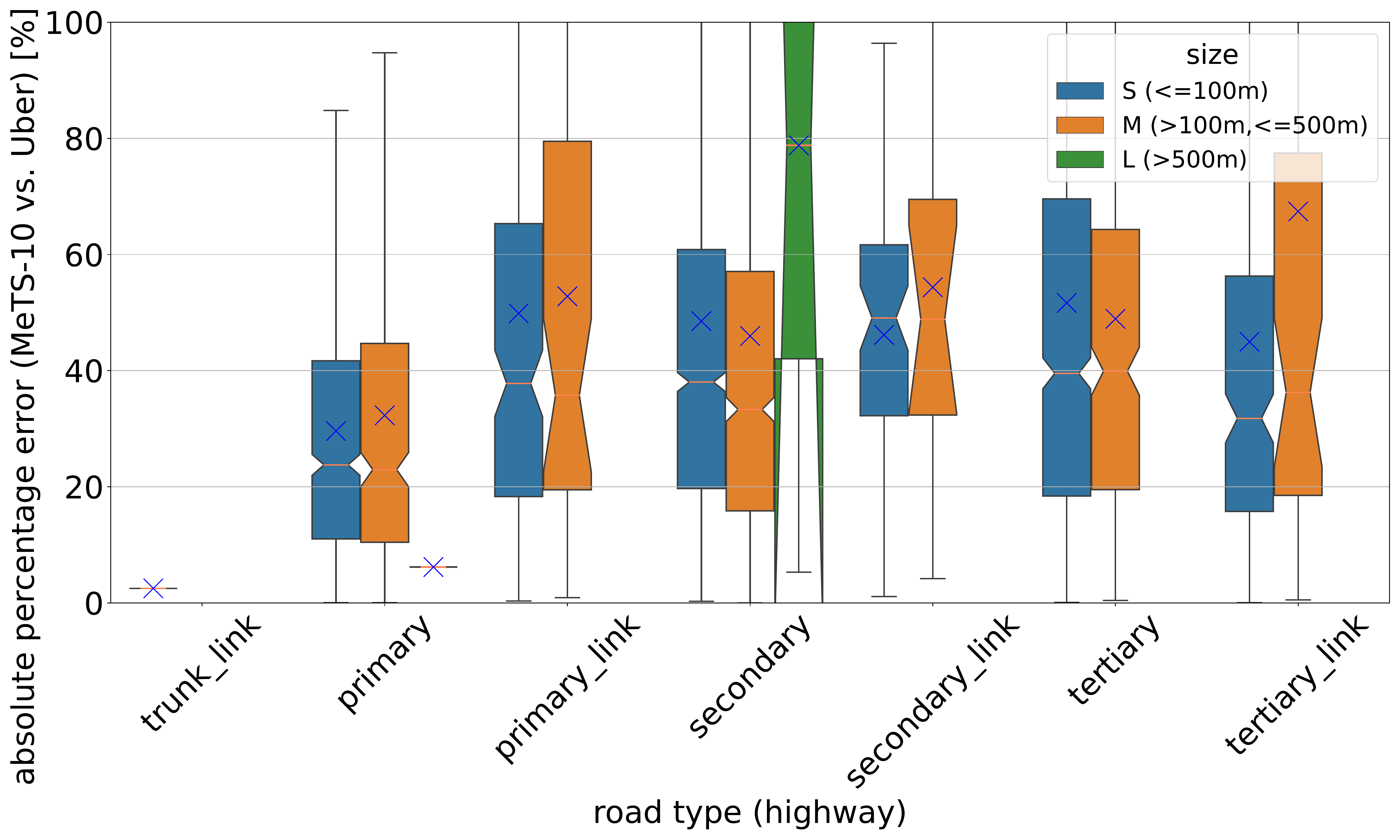}
  \caption{Barcelona absolute percentage error \mcswts{} vs. Uber by road type and segment length. Blue crosses indicate the mean per road type.}
  \label{fig:Barcelona_Uber_ape_size}
\end{figure}

\begin{figure}[ht!]
  \centering
  \includegraphics[width=0.48\textwidth]{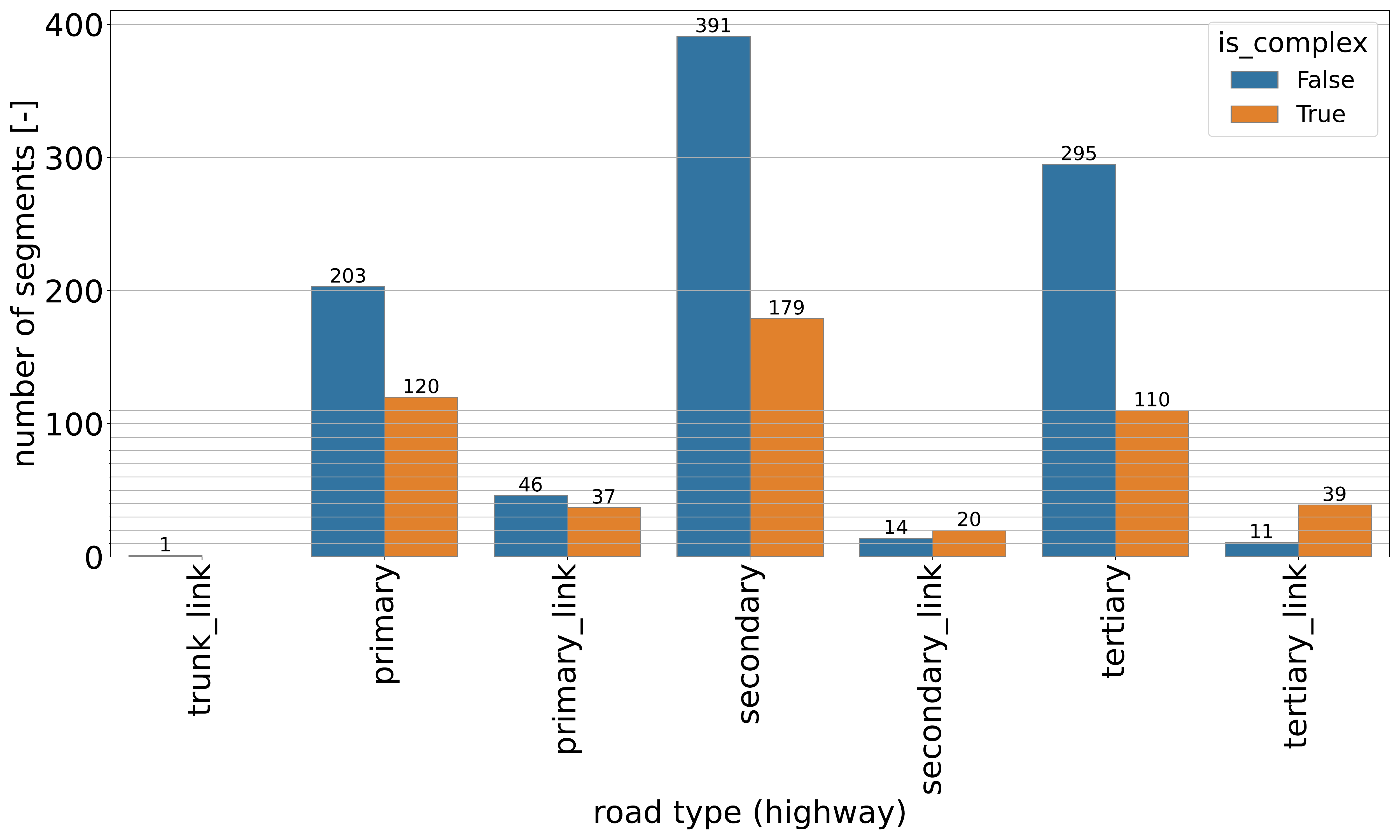}
  \caption{Segment counts \mcswts{} -- Uber matched data.}
  \label{fig:Barcelona_Uber_segment_counts_complex_non_complex}
\end{figure}

\begin{figure}[ht!]
  \centering
  \begin{subfigure}[b]{0.45\textwidth}
  \includegraphics[width=1.0\textwidth]{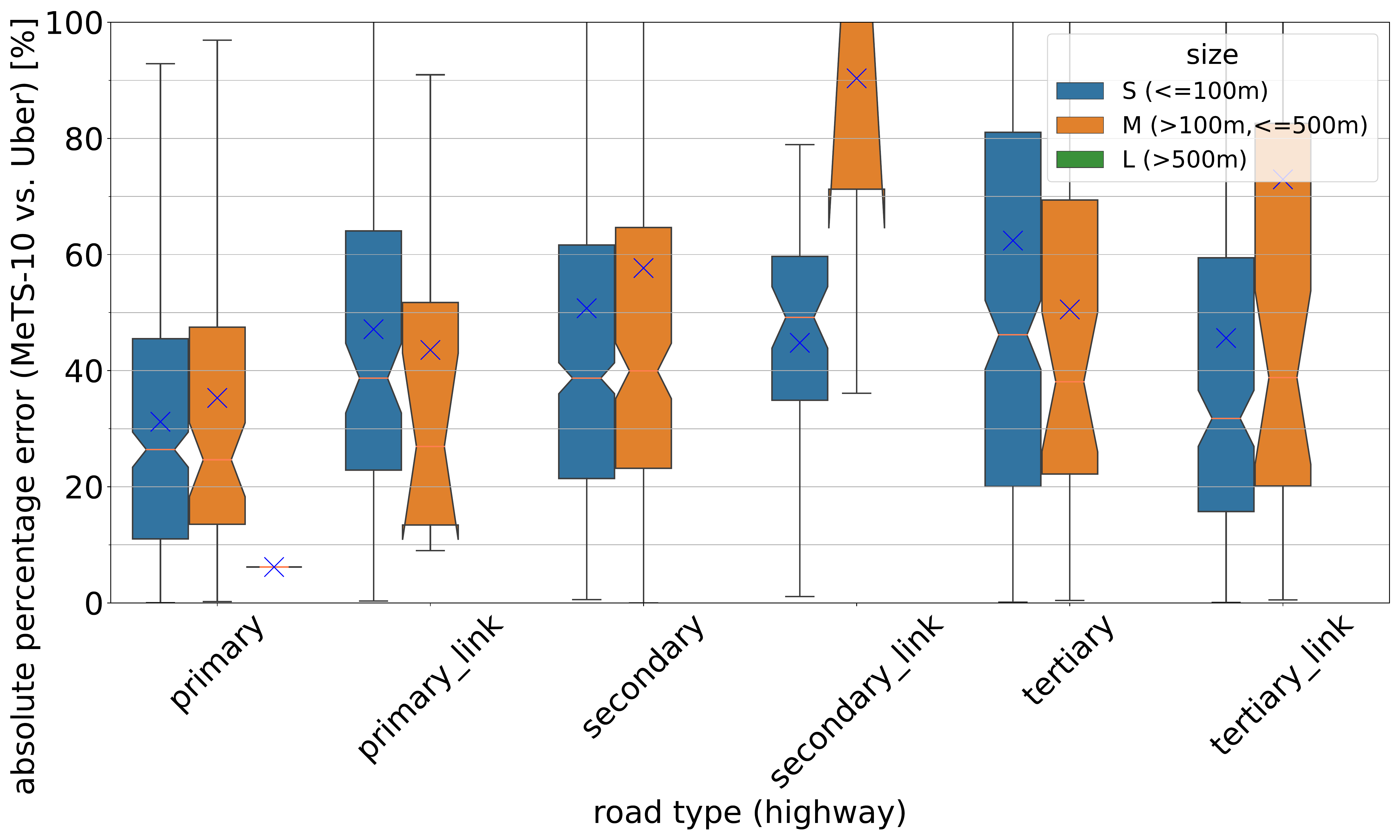}
  \caption{complex road segments}
  \label{fig:Barcelona_Uber_ape_complex}
  \end{subfigure}
  \begin{subfigure}[b]{0.45\textwidth}
  \includegraphics[width=1.0\textwidth]{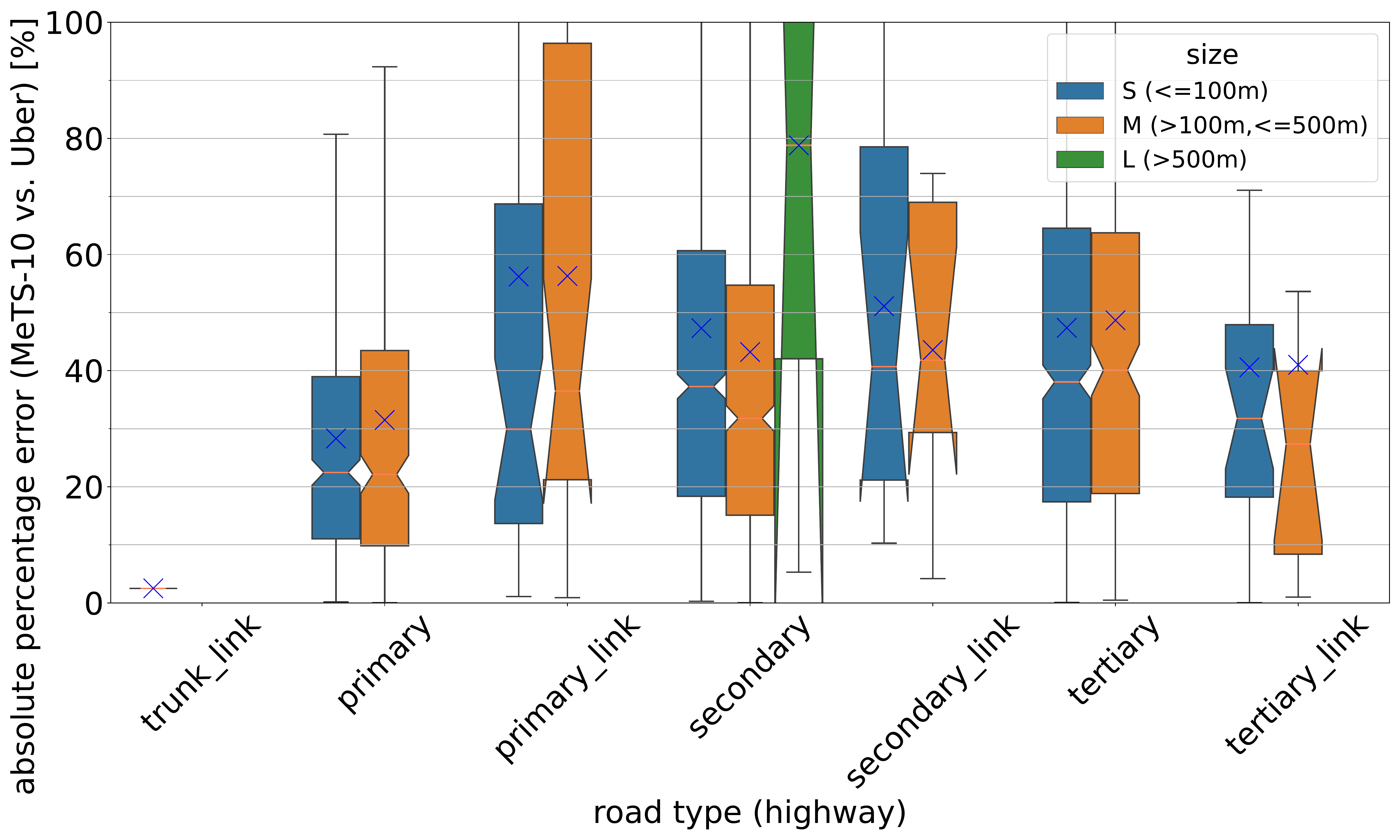}
  \label{fig:Barcelona_Uber_ape_non_complex}
  \caption{non-complex road segments}
  \end{subfigure}
  \caption{Barcelona absolute percentage error \mcswts{} vs. Uber by road type and segment length. Blue crosses indicate the mean per road type.}
  \label{fig:Barcelona_Uber_ape_complex_non_complex}
\end{figure}

\subsection{Berlin}
\mbox{}
\nopagebreak{}
\begin{figure}[H]
\centering
\includegraphics[width=0.85\textwidth]{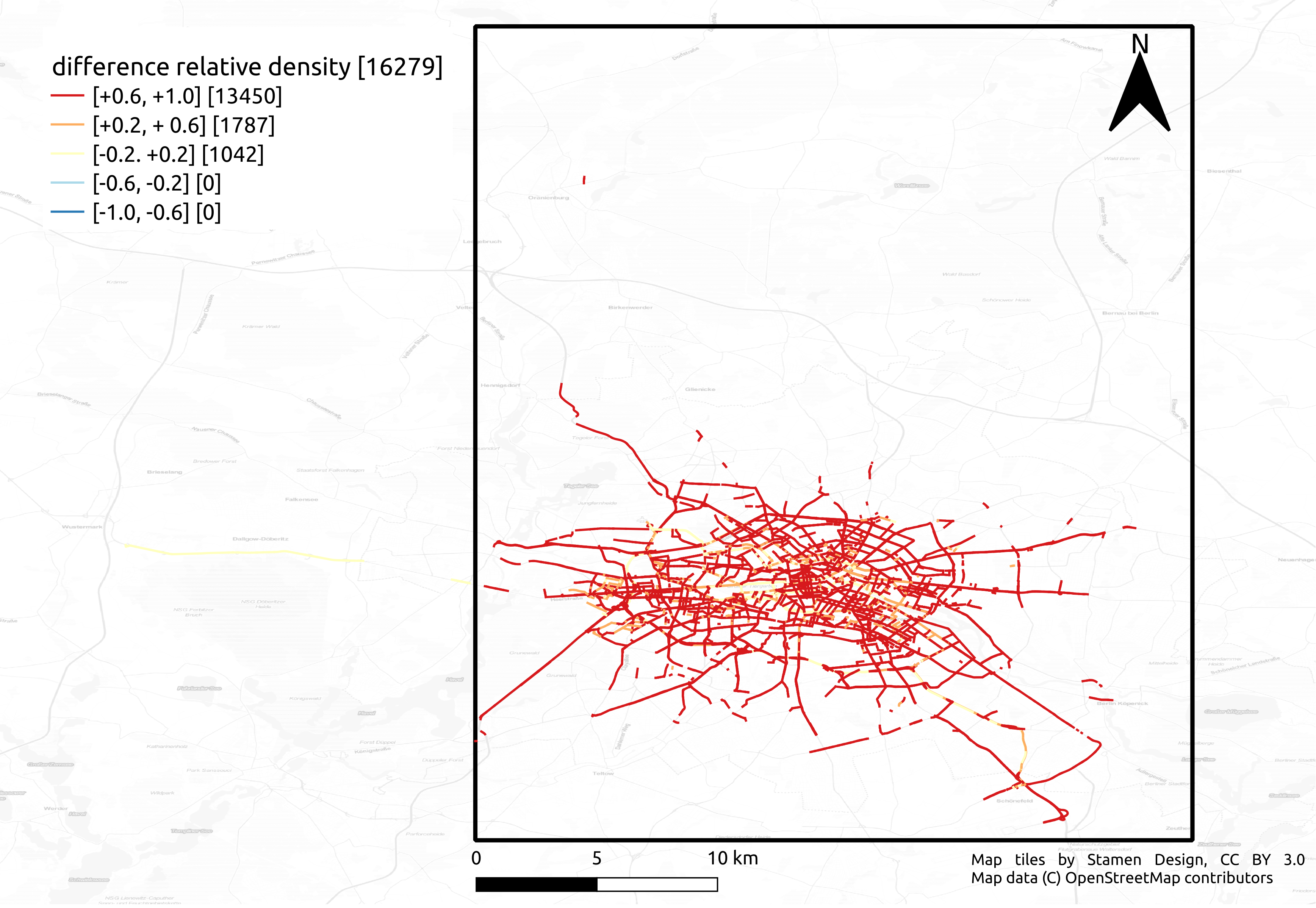}
\caption{Segment density differences of Uber and \mcswts{} on the historic road graph Berlin (8am--6pm). The color encoding shows the edge density difference, negative means higher temporal coverage of \mcswts{} and positive values mean higher temporal coverage..}
\end{figure}

\begin{figure}[H]
\centering
\includegraphics[width=0.85\textwidth]{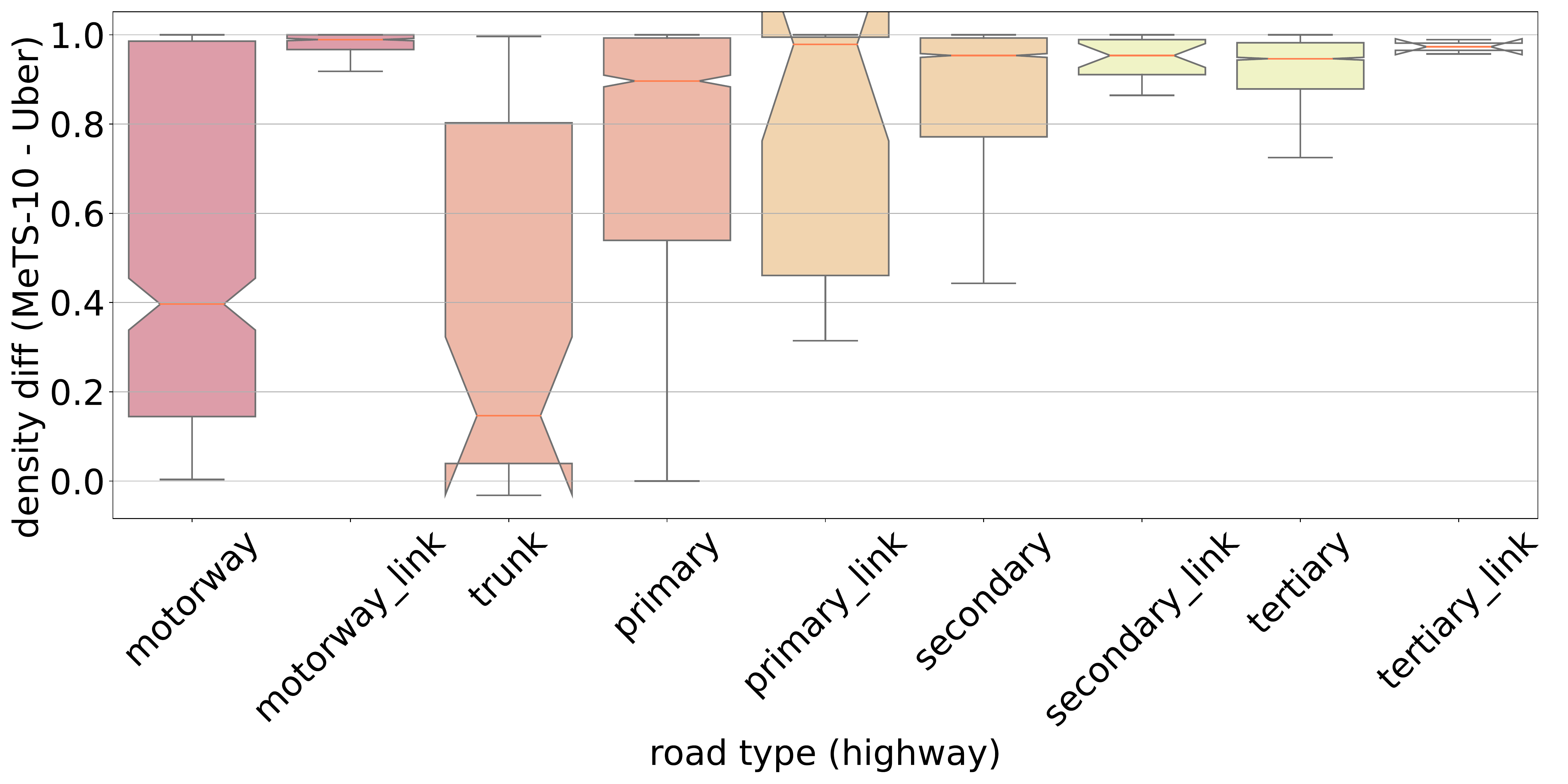}
\caption{Segment density differences Uber and \mcswts{} on the historic road graph Berlin daytime (8am--6pm, segments within 	4c bounding box only). Mean density difference by road class (\ie{} OSM highway attribute); positive density difference means higher temporal coverage of \mcswts{} and negative mean higher temporal coverage.}
\end{figure}

\begin{figure}[H]
\centering
\begin{subfigure}[b]{0.45\textwidth}
\includegraphics[width=\textwidth]{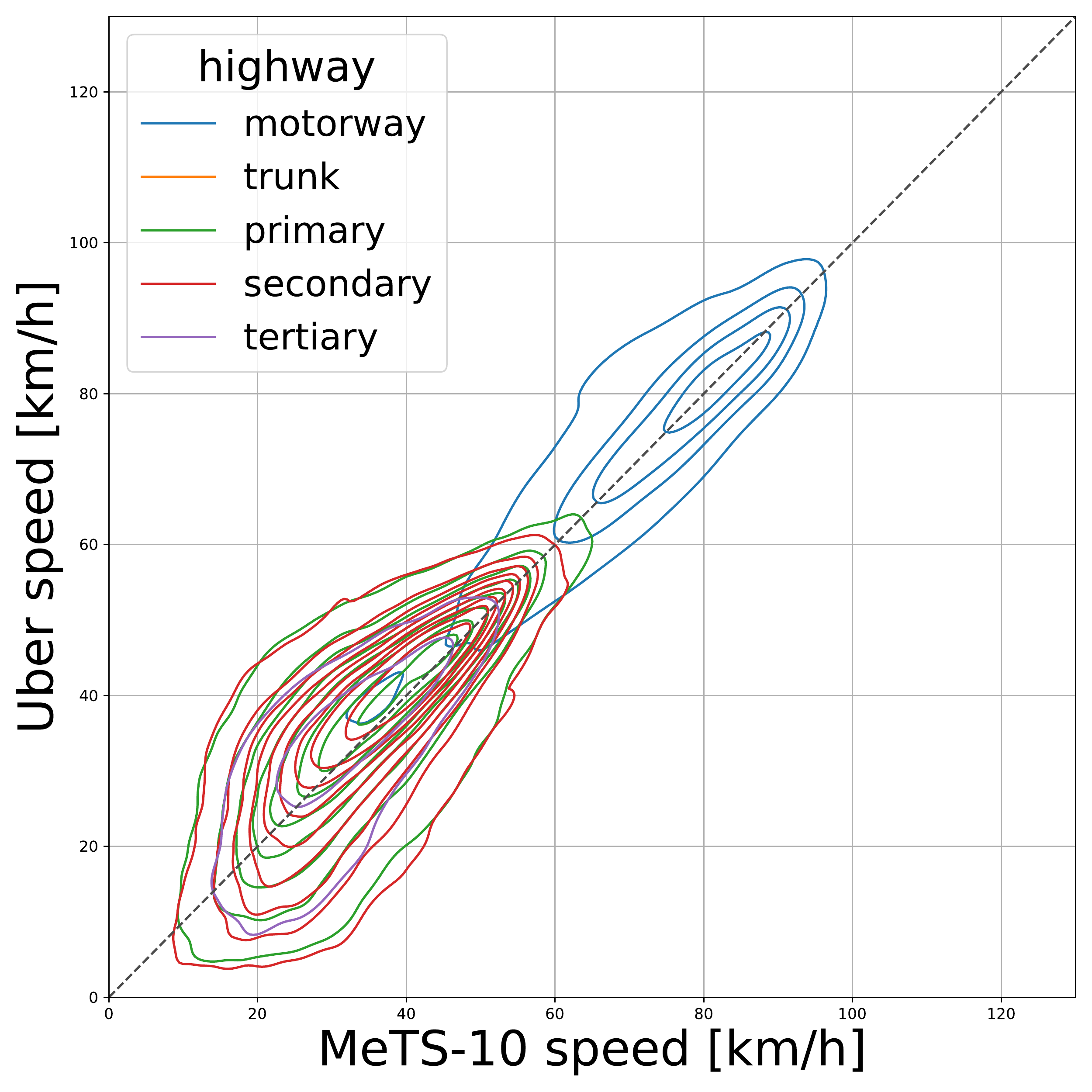}
\caption{KDE non-link road types.}
\end{subfigure}
\begin{subfigure}[b]{0.45\textwidth}
\includegraphics[width=\textwidth]{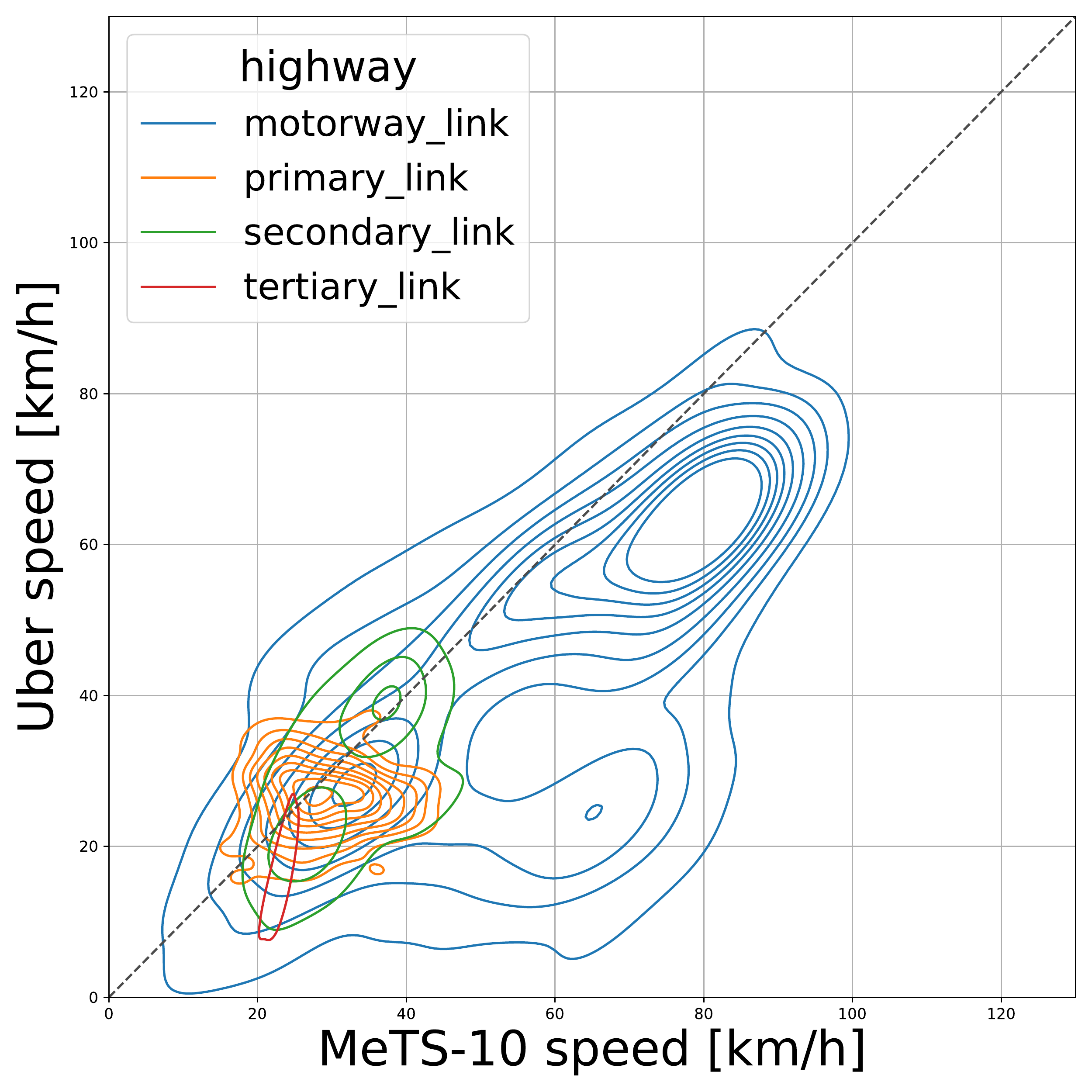}
\caption{KDE link road types.}
\end{subfigure}
\begin{subfigure}[b]{0.9\textwidth}
\includegraphics[width=\textwidth]{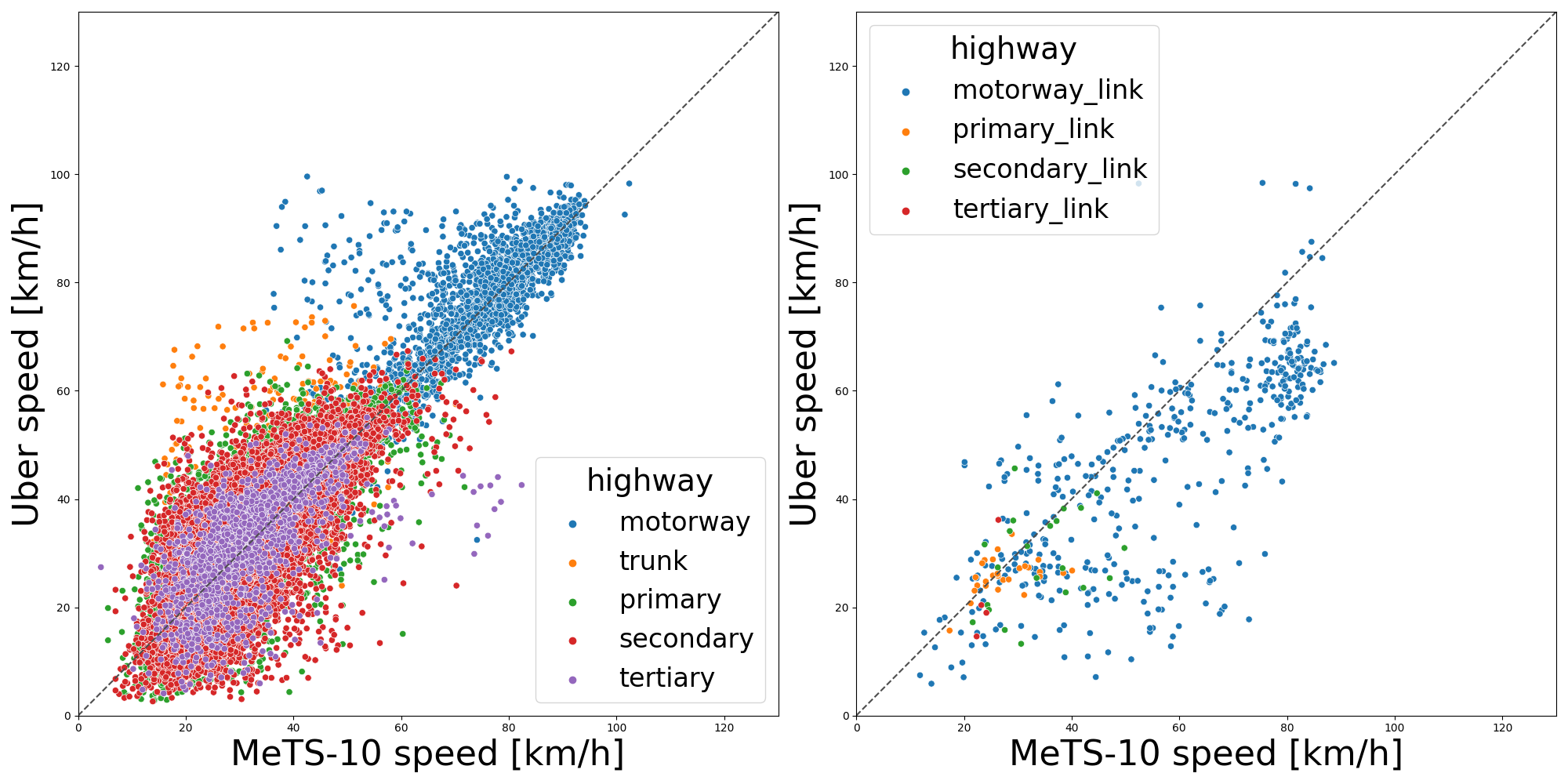}
\caption{Scatter non-link (left) and link (right) road types.}
\end{subfigure}
\caption{Kernel Distribution Estimation and Scatter Plots of speeds of \mcswts{} (x-axis, \texttt{median\_speed\_kph}) and Uber (y-axis, \texttt{speed\_kph\_mean}) on the historic road graph Berlin daytime (8am--6pm) on the matching data, \ie{} within \mcswts{} bounding box only and where data is available at the same time and segment, for the most important road types.}
\end{figure}

\begin{figure}[H]
\centering
\includegraphics[width=\textwidth]{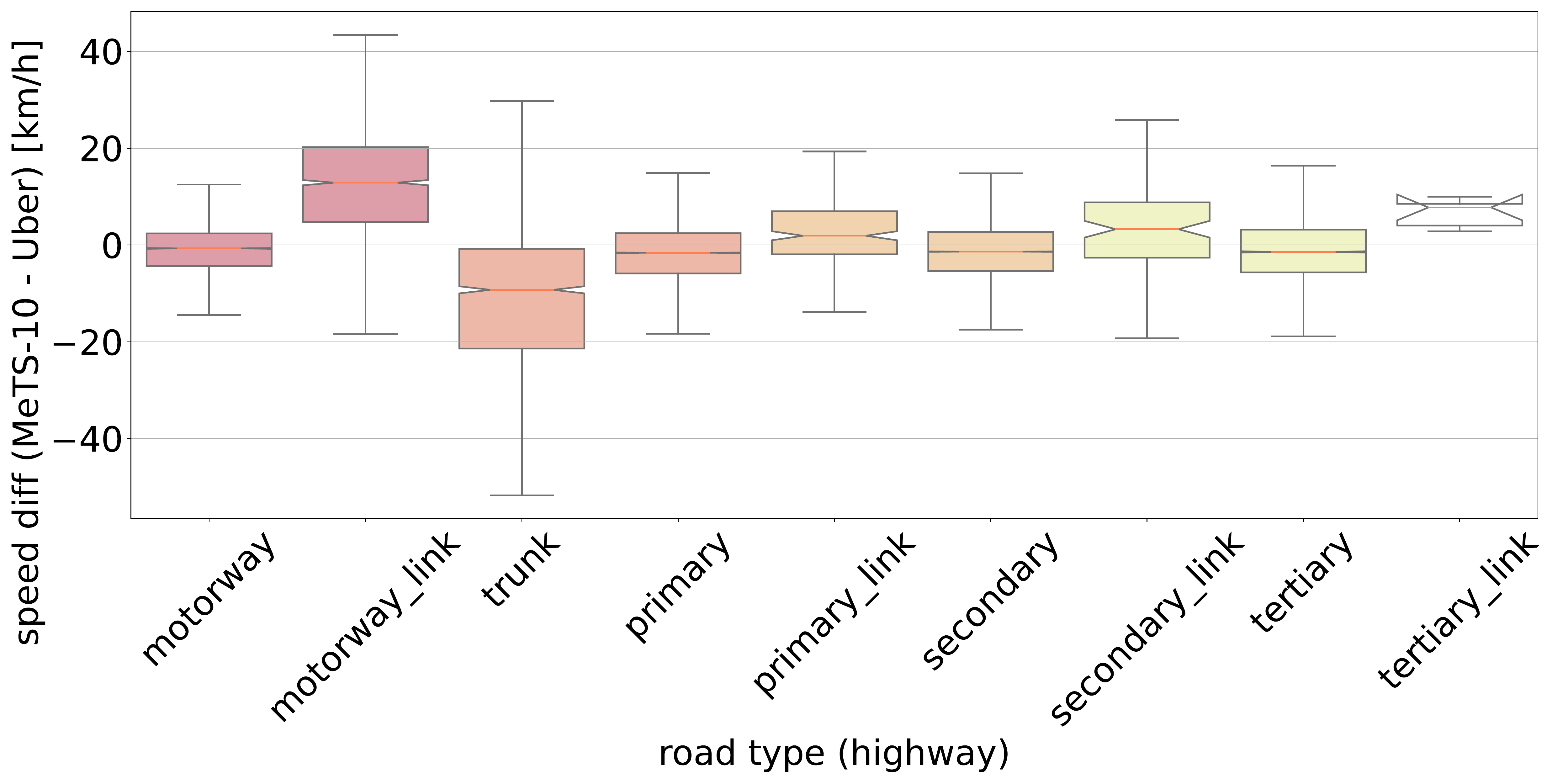}
\caption{Speed differences Uber and \mcswts{} on the historic road graph Berlin daytime (8am--6pm) on the matching data, \ie{} within \mcswts{} bounding box only and where data is available at the same time and segment. Mean difference by road class (OSM highway attribute). Positive speed difference means higher values in \mcswts.}
\end{figure}
\begin{figure}[H]
\centering
\includegraphics[width=\textwidth]{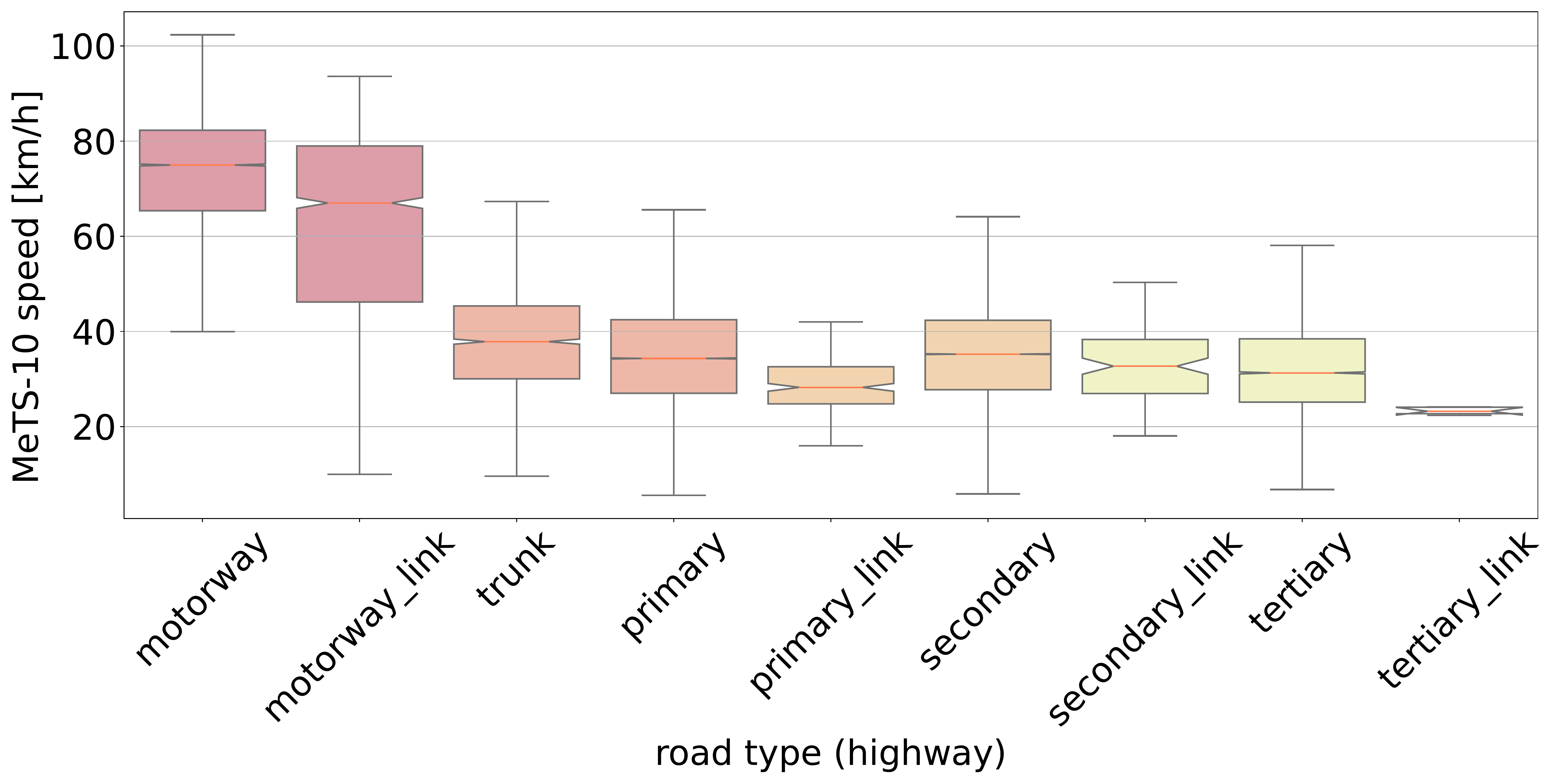}
\caption{\mcswts{} speeds on the historic road graph Berlin daytime (8am--6pm) on the matching data, \ie{} within \mcswts{} bounding box only and where data is available at the same time and segment. By road class (OSM highway attribute).}
\end{figure}

\begin{figure}[ht]
  \centering
  \includegraphics[width=0.48\textwidth]{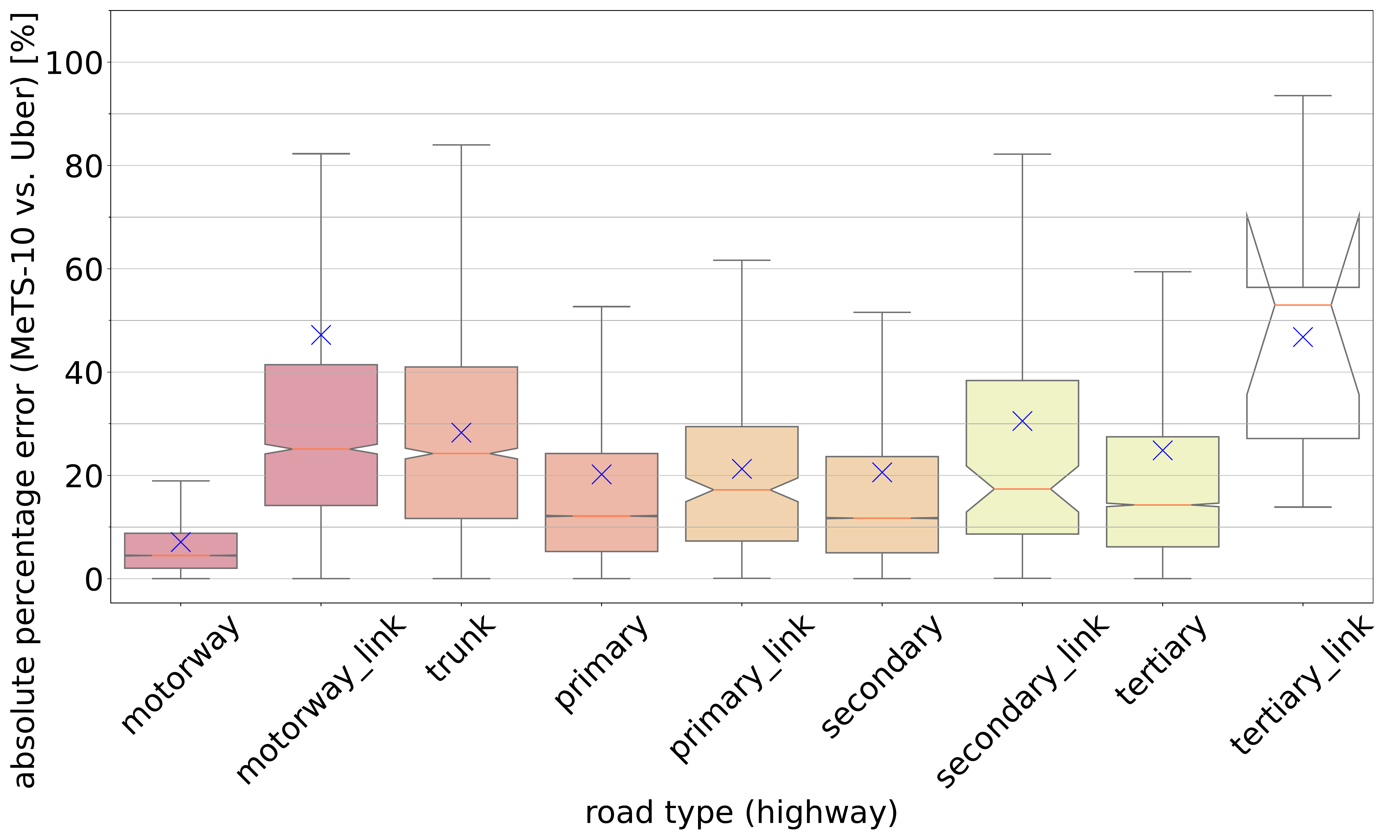}
  \caption{Berlin absolute percentage error \mcswts{} vs. Uber by road type. Blue crosses indicate the mean per road type.}
  \label{fig:Berlin_Uber_ape}
\end{figure}

\begin{figure}[ht!]
  \centering
  \includegraphics[width=0.48\textwidth]{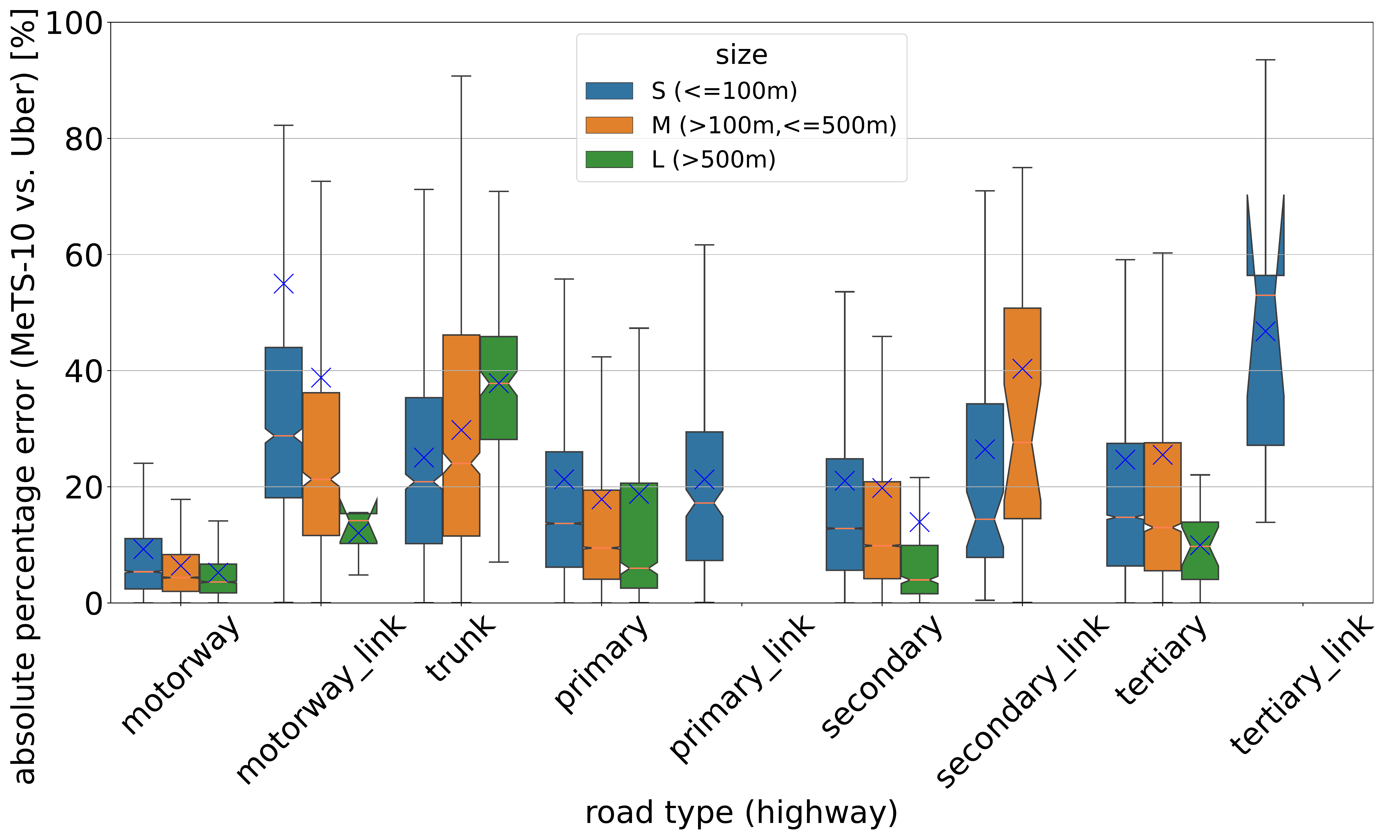}
  \caption{Berlin absolute percentage error \mcswts{} vs. Uber by road type and segment length. Blue crosses indicate the mean per road type.}
  \label{fig:Berlin_Uber_ape_size}
\end{figure}

\begin{figure}[ht!]
  \centering
  \includegraphics[width=0.48\textwidth]{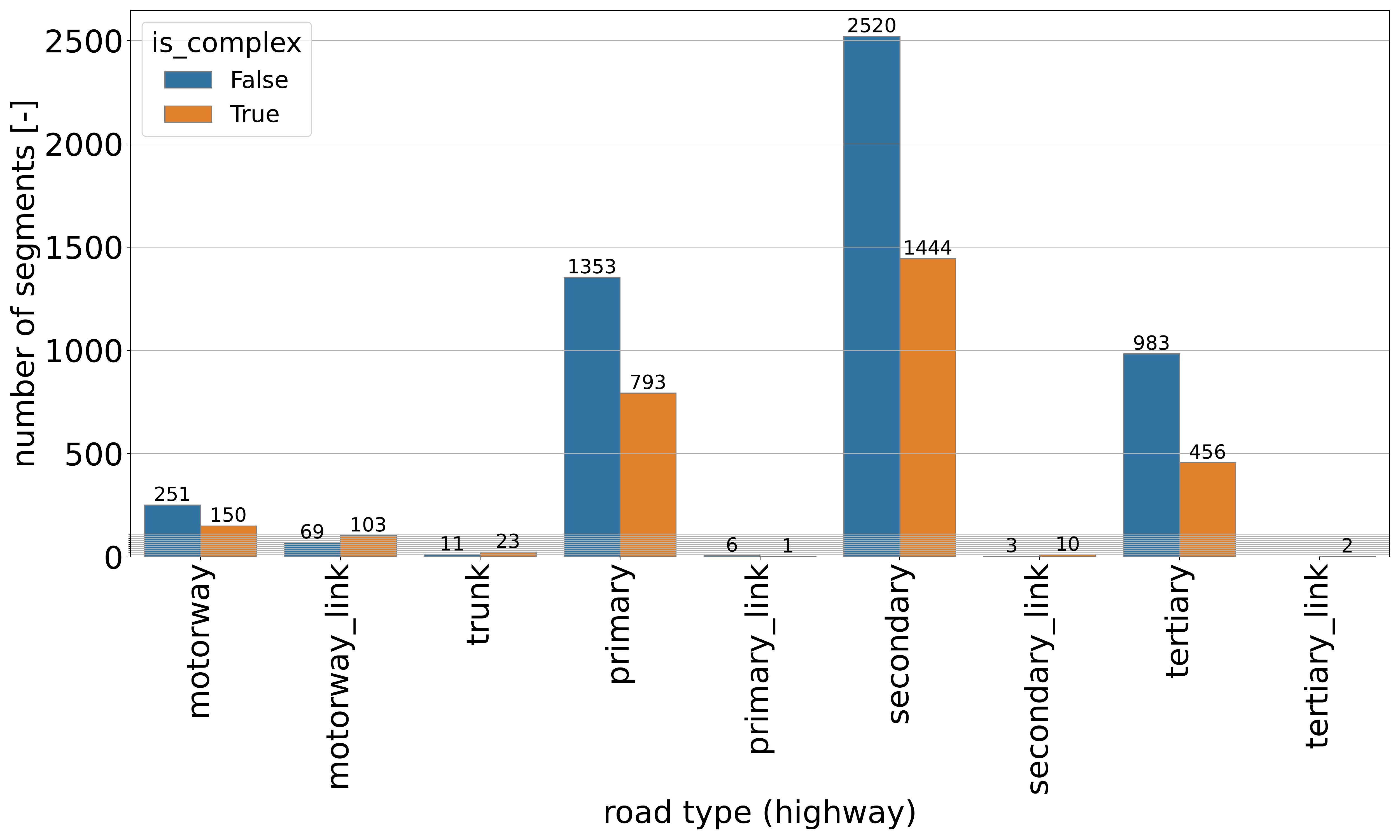}
  \caption{Segment counts \mcswts{} -- Uber matched data.}
  \label{fig:Berlin_Uber_segment_counts_complex_non_complex}
\end{figure}

\begin{figure}[ht!]
  \centering
  \begin{subfigure}[b]{0.45\textwidth}
  \includegraphics[width=1.0\textwidth]{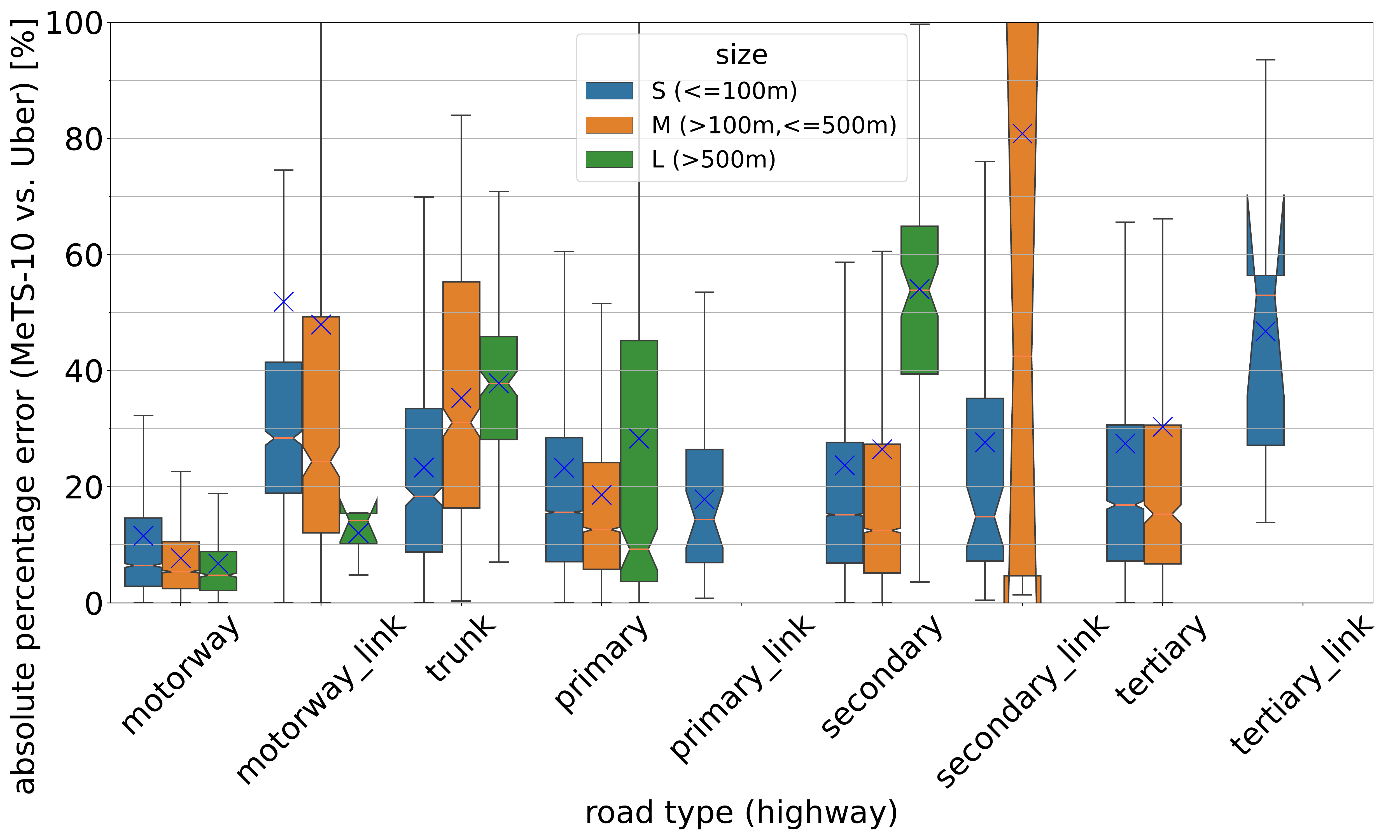}
  \caption{complex road segments}
  \label{fig:Berlin_Uber_ape_complex}
  \end{subfigure}
  \begin{subfigure}[b]{0.45\textwidth}
  \includegraphics[width=1.0\textwidth]{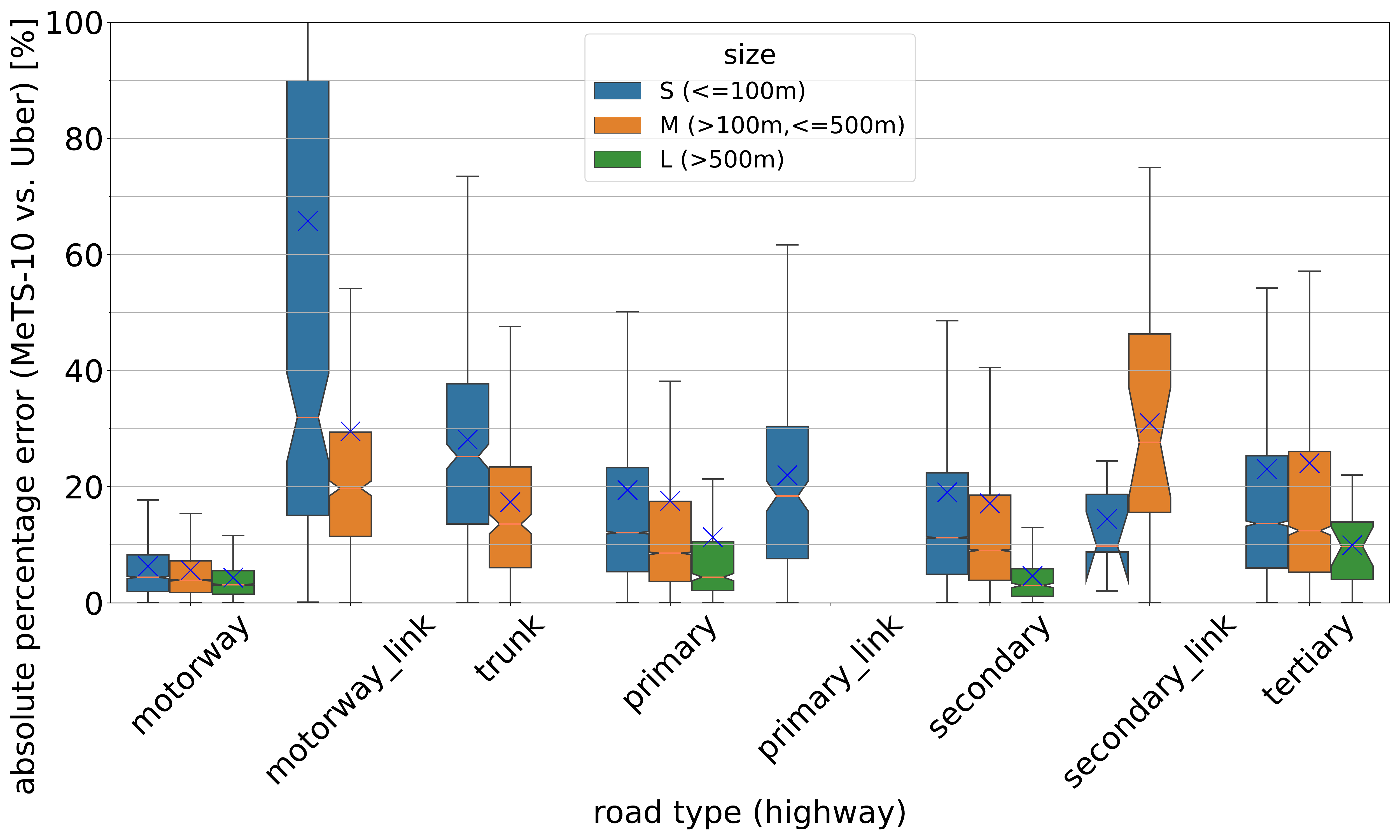}
  \label{fig:Berlin_Uber_ape_non_complex}
  \caption{non-complex road segments}
  \end{subfigure}
  \caption{Berlin absolute percentage error \mcswts{} vs. Uber by road type and segment length. Blue crosses indicate the mean per road type.}
  \label{fig:Berlin_Uber_ape_complex_non_complex}
\end{figure}

\subsection{London}
\mbox{}
\nopagebreak{}
\begin{figure}[H]
\centering
\includegraphics[width=0.85\textwidth]{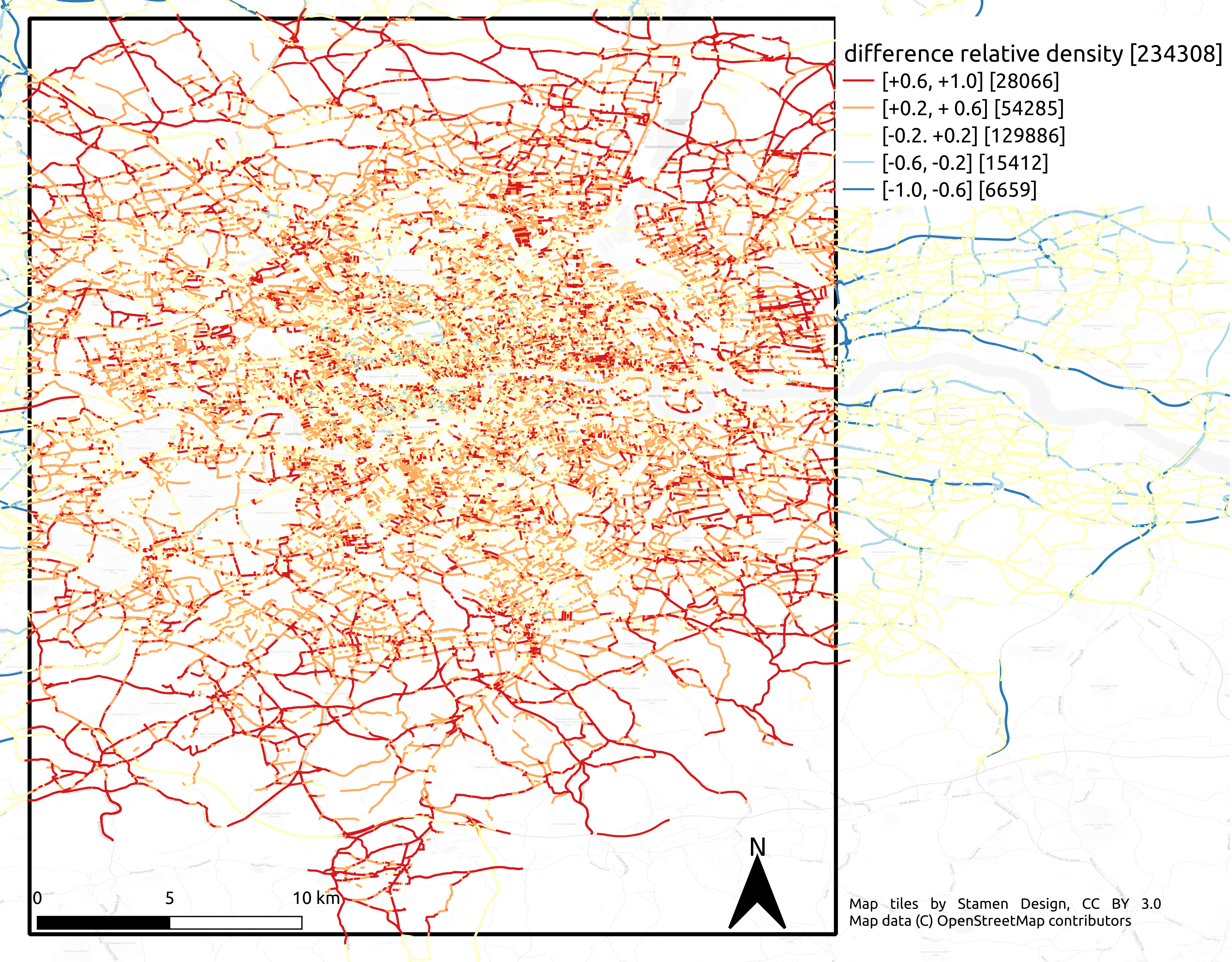}
\caption{Segment density differences of Uber and \mcswts{} on the historic road graph London (8am--6pm). The color encoding shows the edge density difference, negative means higher temporal coverage of \mcswts{} and positive values mean higher temporal coverage..}
\end{figure}

\begin{figure}[H]
\centering
\includegraphics[width=0.85\textwidth]{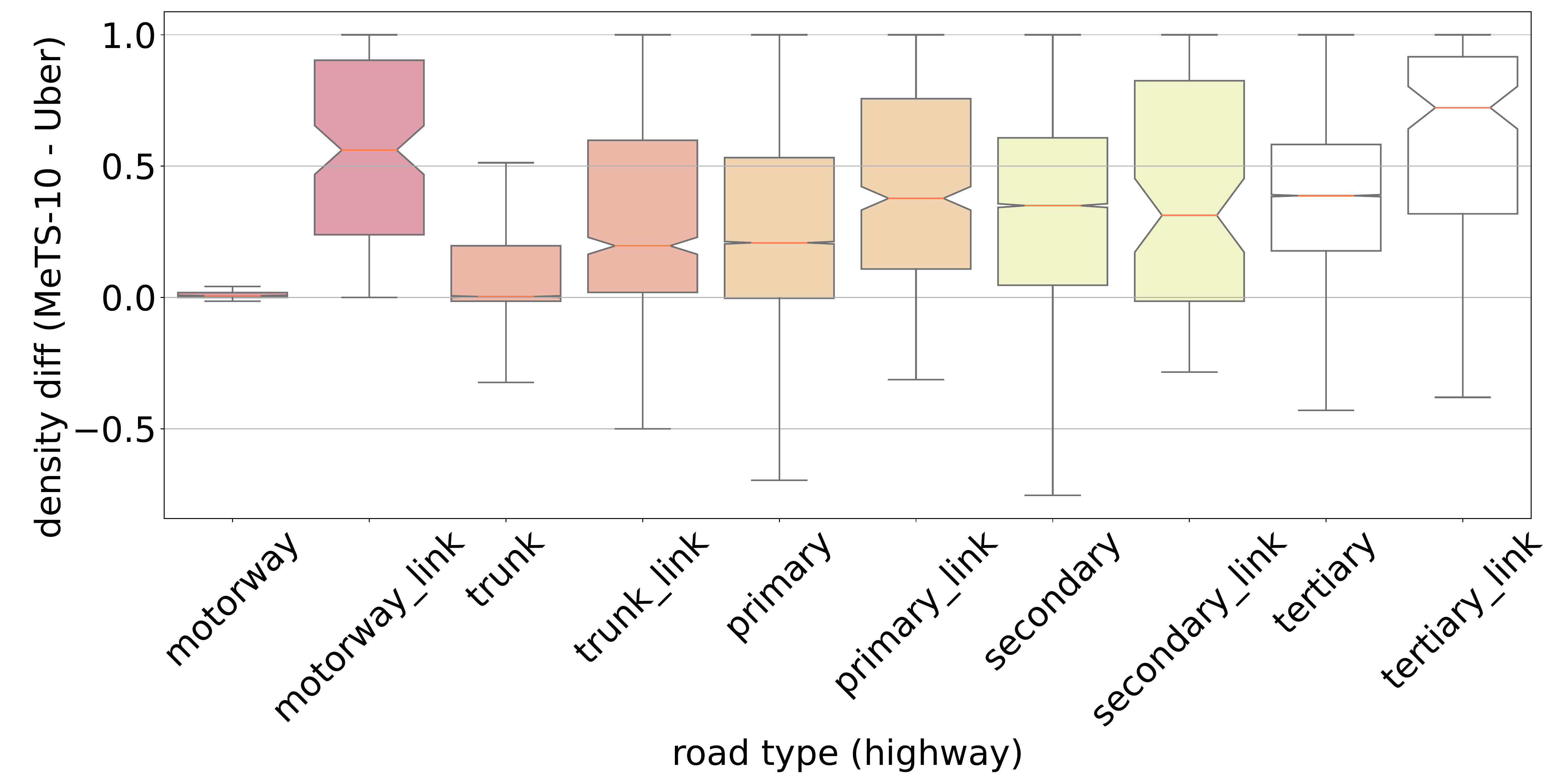}
\caption{Segment density differences Uber and \mcswts{} on the historic road graph London daytime (8am--6pm, segments within 	4c bounding box only). Mean density difference by road class (\ie{} OSM highway attribute); positive density difference means higher temporal coverage of \mcswts{} and negative mean higher temporal coverage.}
\end{figure}

\begin{figure}[H]
\centering
\begin{subfigure}[b]{0.45\textwidth}
\includegraphics[width=\textwidth]{figures/05_val01_uber/London_Uber_kde_highway_non_links.pdf}
\caption{KDE non-link road types.}
\end{subfigure}
\begin{subfigure}[b]{0.45\textwidth}
\includegraphics[width=\textwidth]{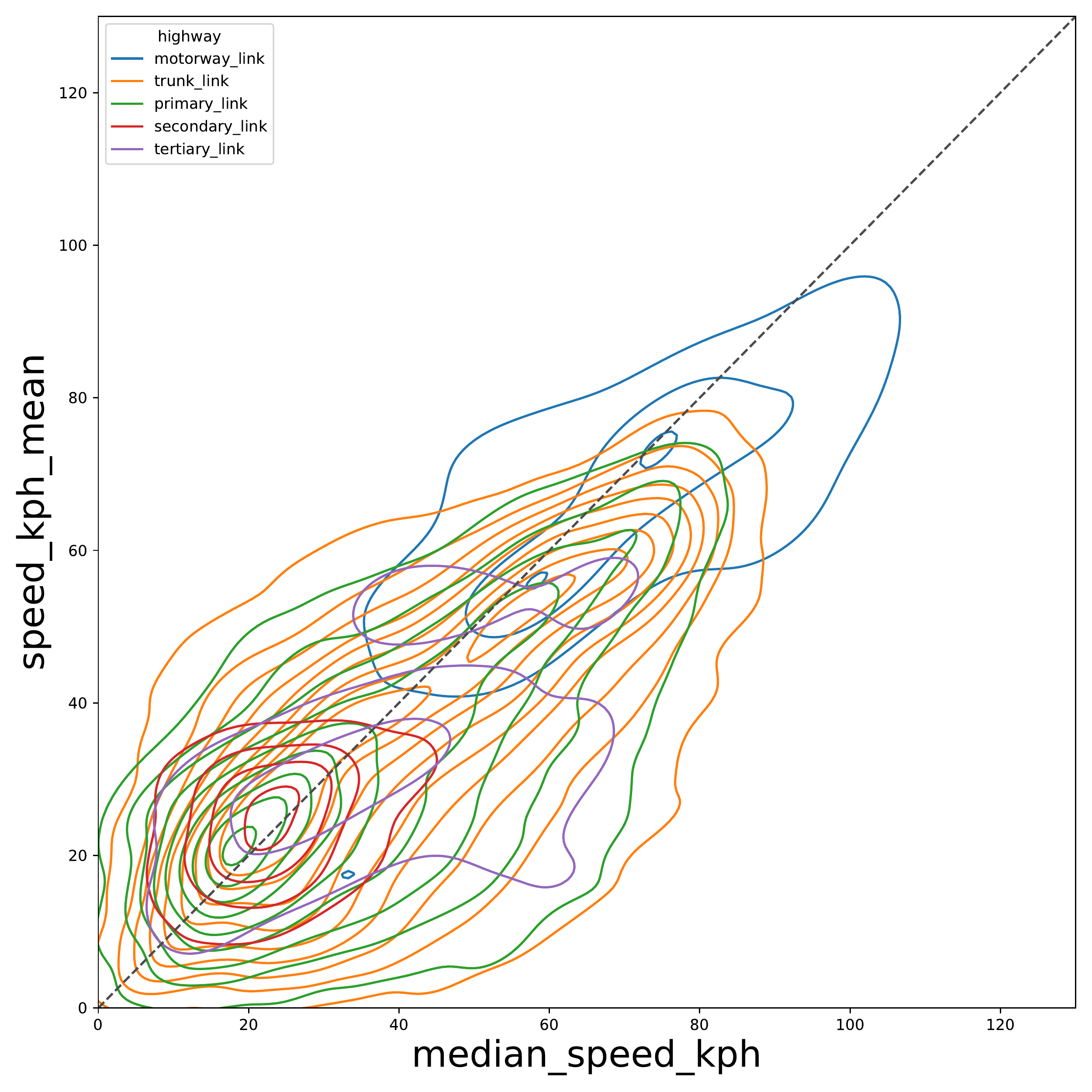}
\caption{KDE link road types.}
\end{subfigure}
\begin{subfigure}[b]{0.9\textwidth}
\includegraphics[width=\textwidth]{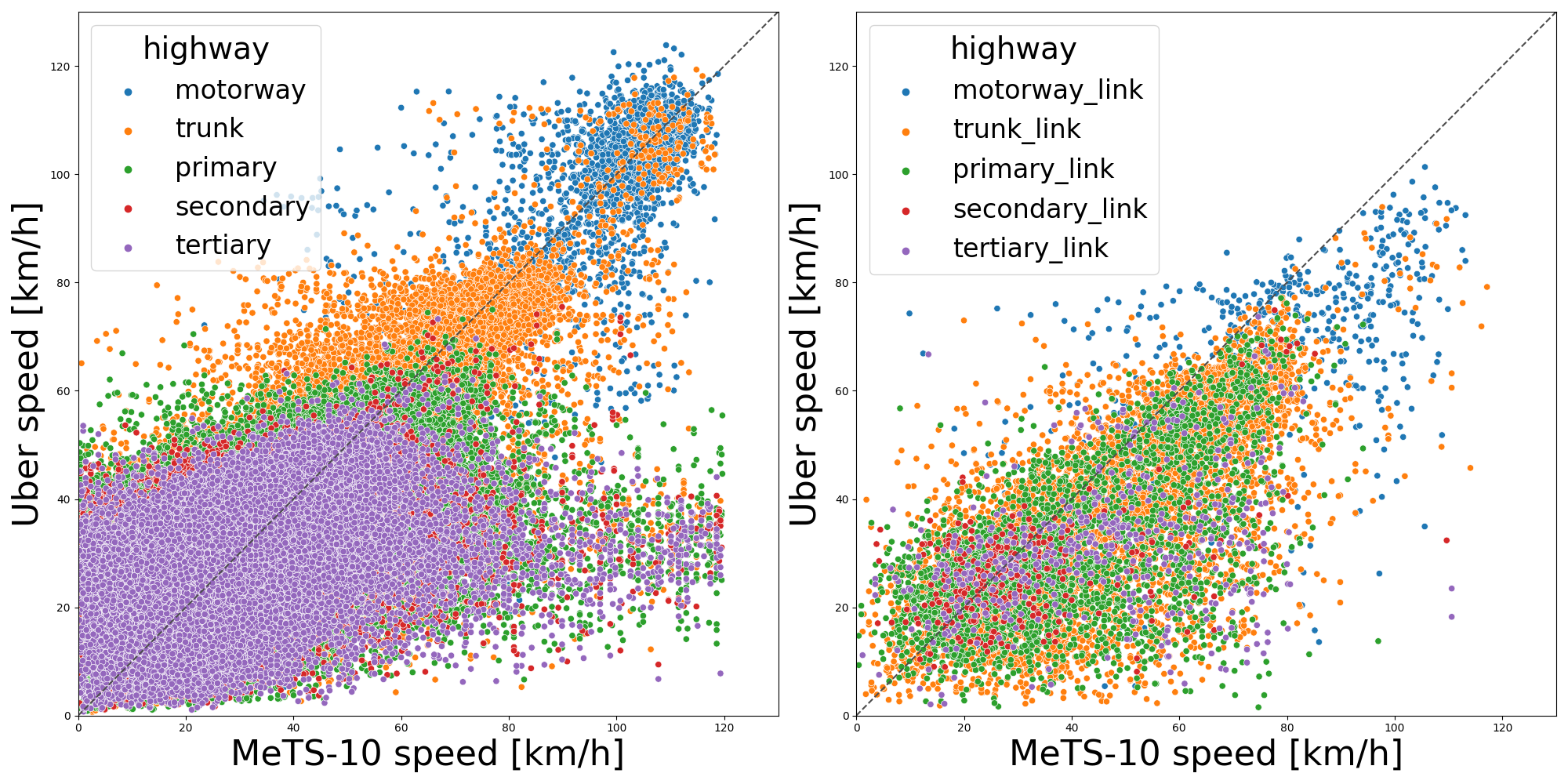}
\caption{Scatter non-link (left) and link (right) road types.}
\end{subfigure}
\caption{Kernel Distribution Estimation and Scatter Plots of speeds of \mcswts{} (x-axis, \texttt{median\_speed\_kph}) and Uber (y-axis, \texttt{speed\_kph\_mean}) on the historic road graph London daytime (8am--6pm) on the matching data, \ie{} within \mcswts{} bounding box only and where data is available at the same time and segment, for the most important road types.}
\end{figure}

\begin{figure}[H]
\centering
\includegraphics[width=\textwidth]{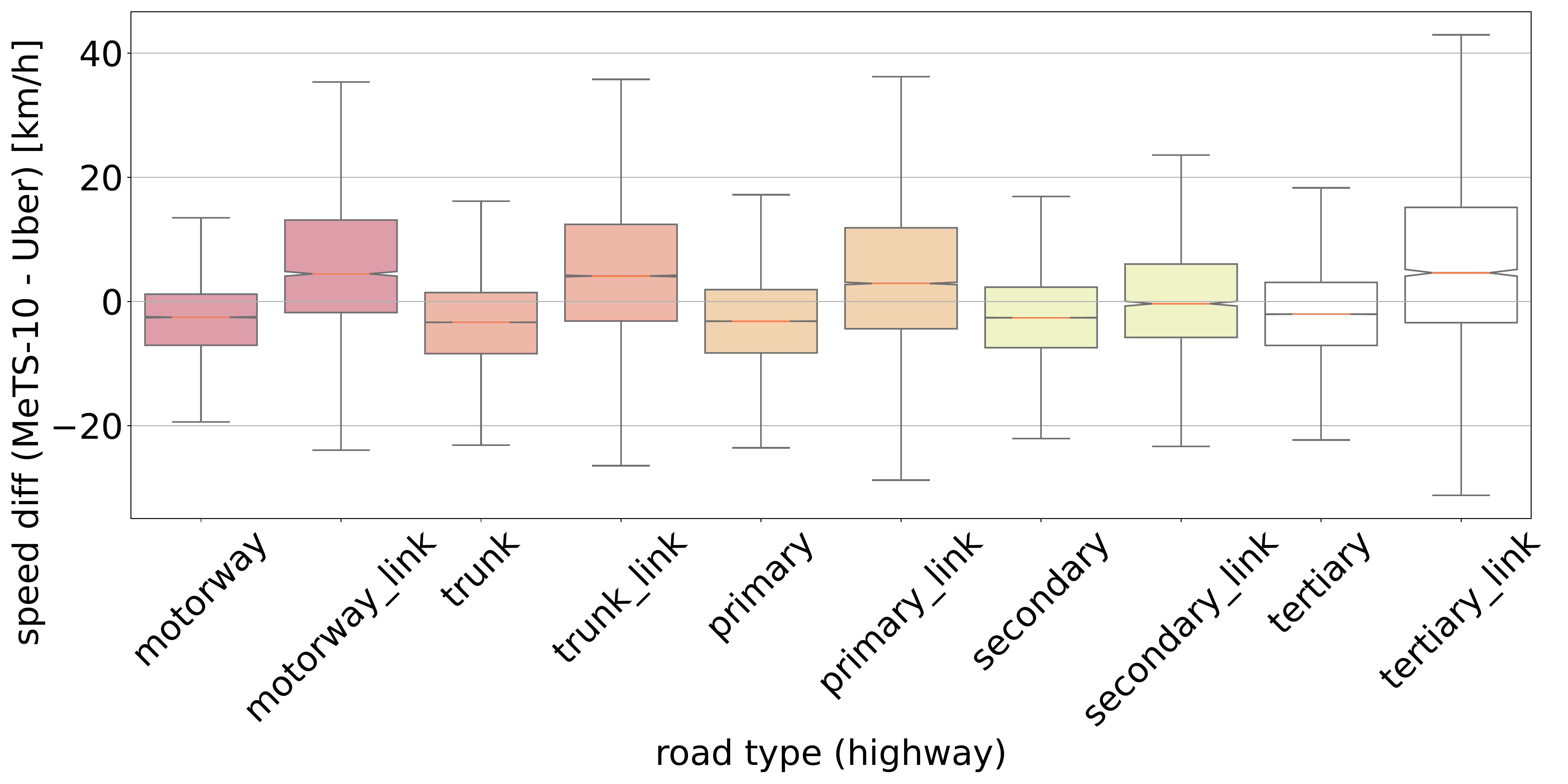}
\caption{Speed differences Uber and \mcswts{} on the historic road graph London daytime (8am--6pm) on the matching data, \ie{} within \mcswts{} bounding box only and where data is available at the same time and segment. Mean difference by road class (OSM highway attribute). Positive speed difference means higher values in \mcswts.}
\end{figure}
\begin{figure}[H]
\centering
\includegraphics[width=\textwidth]{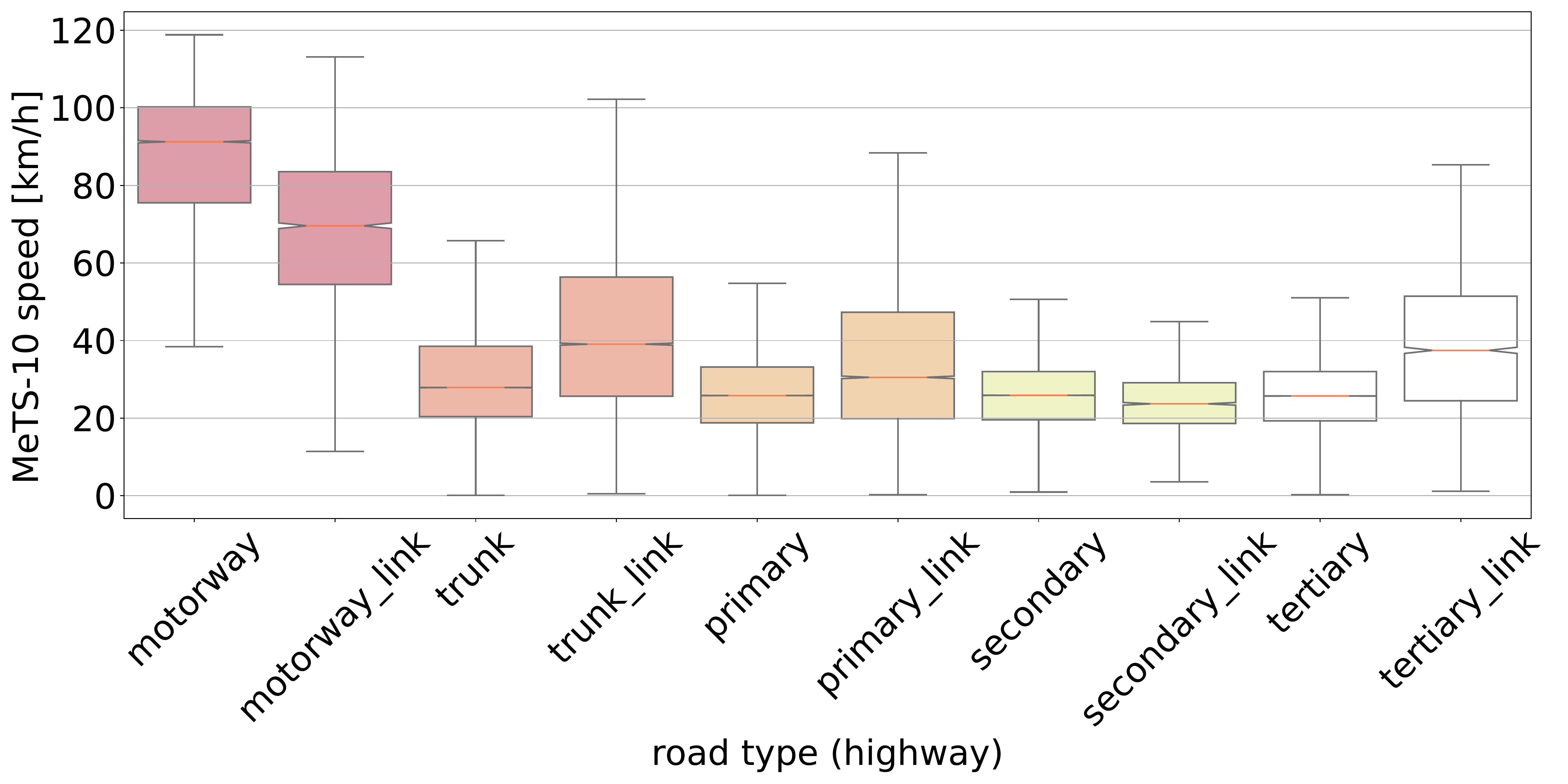}
\caption{\mcswts{} speeds on the historic road graph London daytime (8am--6pm) on the matching data, \ie{} within \mcswts{} bounding box only and where data is available at the same time and segment. By road class (OSM highway attribute).}
\end{figure}

\begin{figure}[ht]
  \centering
  \includegraphics[width=0.48\textwidth]{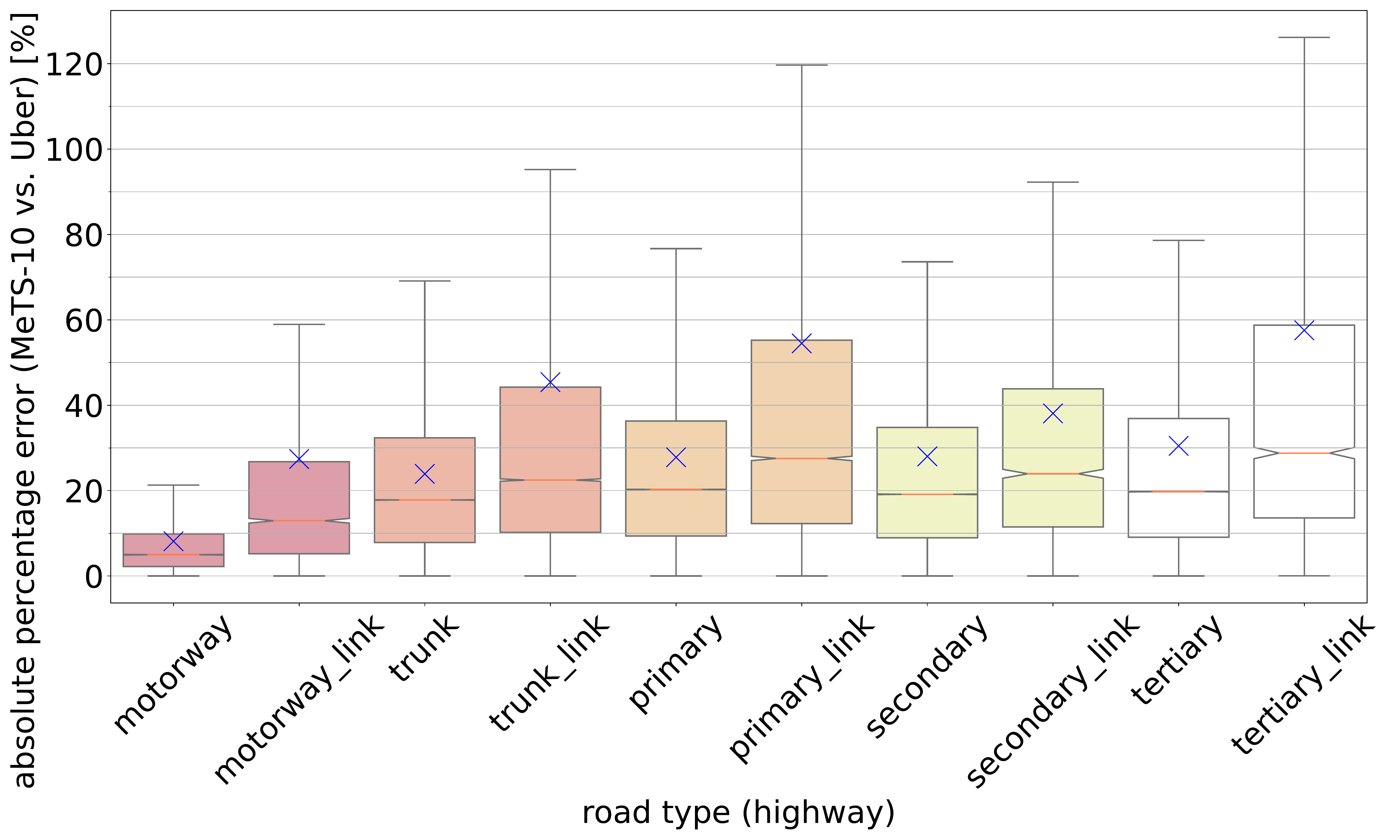}
  \caption{London absolute percentage error \mcswts{} vs. Uber by road type. Blue crosses indicate the mean per road type.}
\end{figure}

\begin{figure}[ht!]
  \centering
  \includegraphics[width=0.48\textwidth]{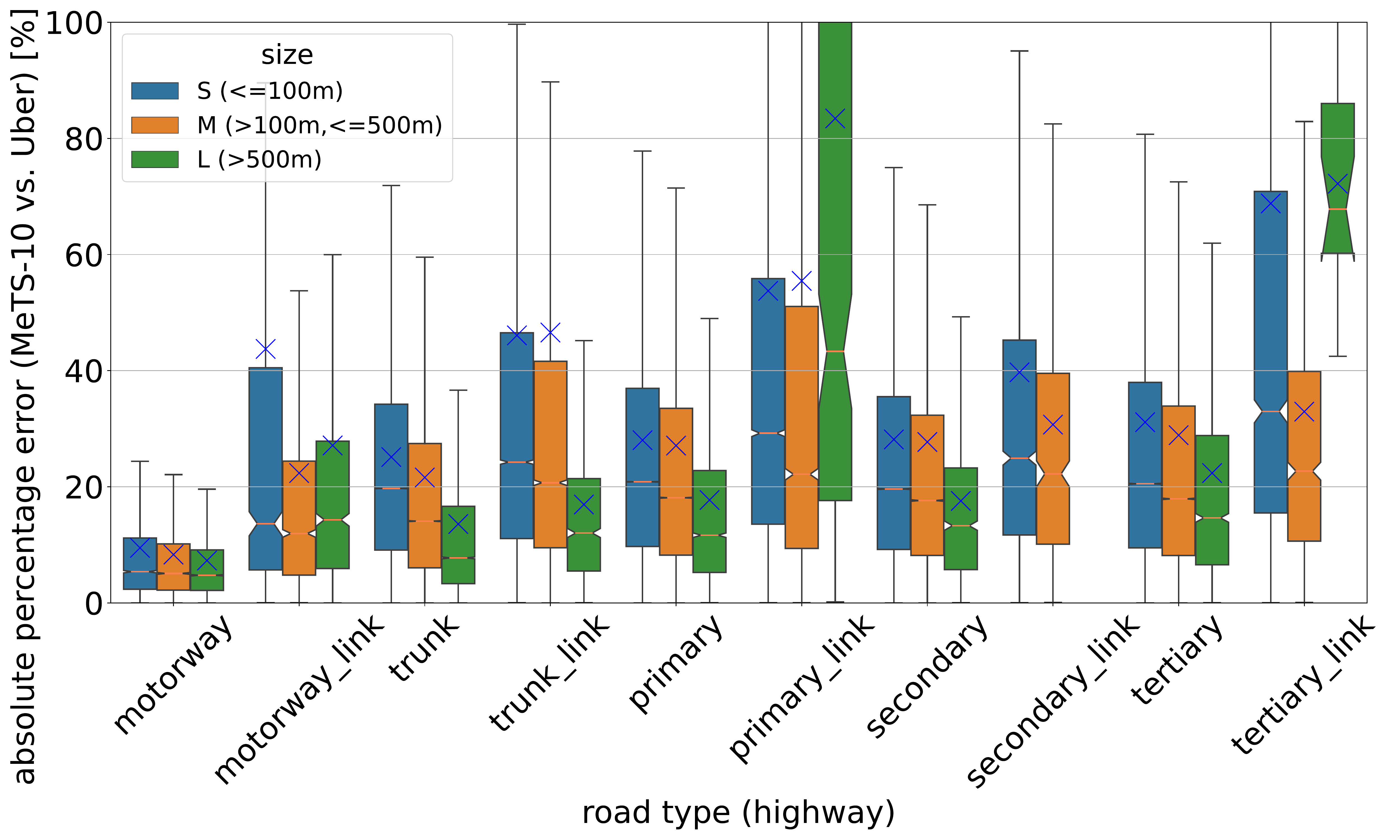}
  \caption{London absolute percentage error \mcswts{} vs. Uber by road type and segment length. Blue crosses indicate the mean per road type.}
\end{figure}

\begin{figure}[ht!]
  \centering
  \includegraphics[width=0.48\textwidth]{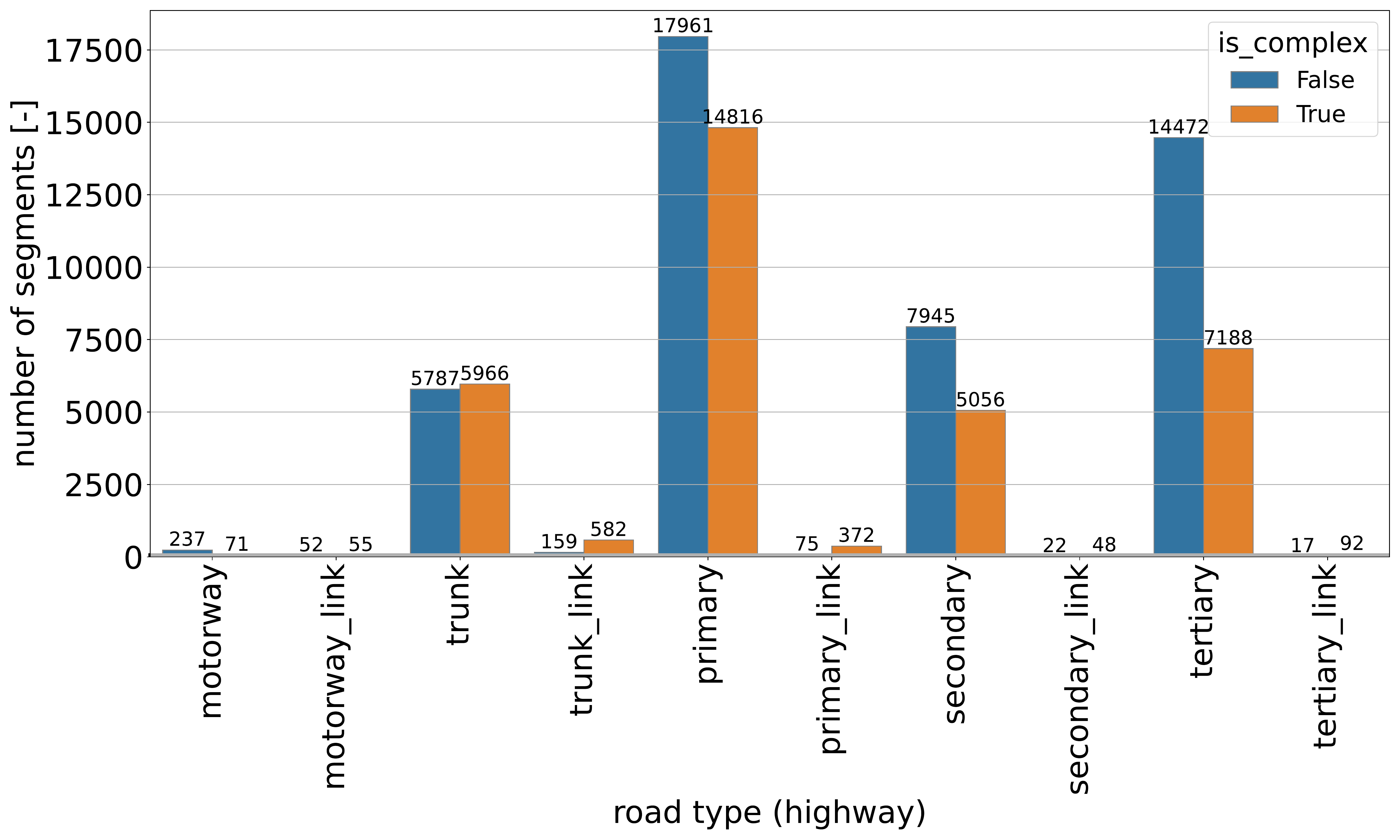}
  \caption{Segment counts \mcswts{} -- Uber matched data.}
\end{figure}

\begin{figure}[ht!]
  \centering
  \begin{subfigure}[b]{0.45\textwidth}
  \includegraphics[width=1.0\textwidth]{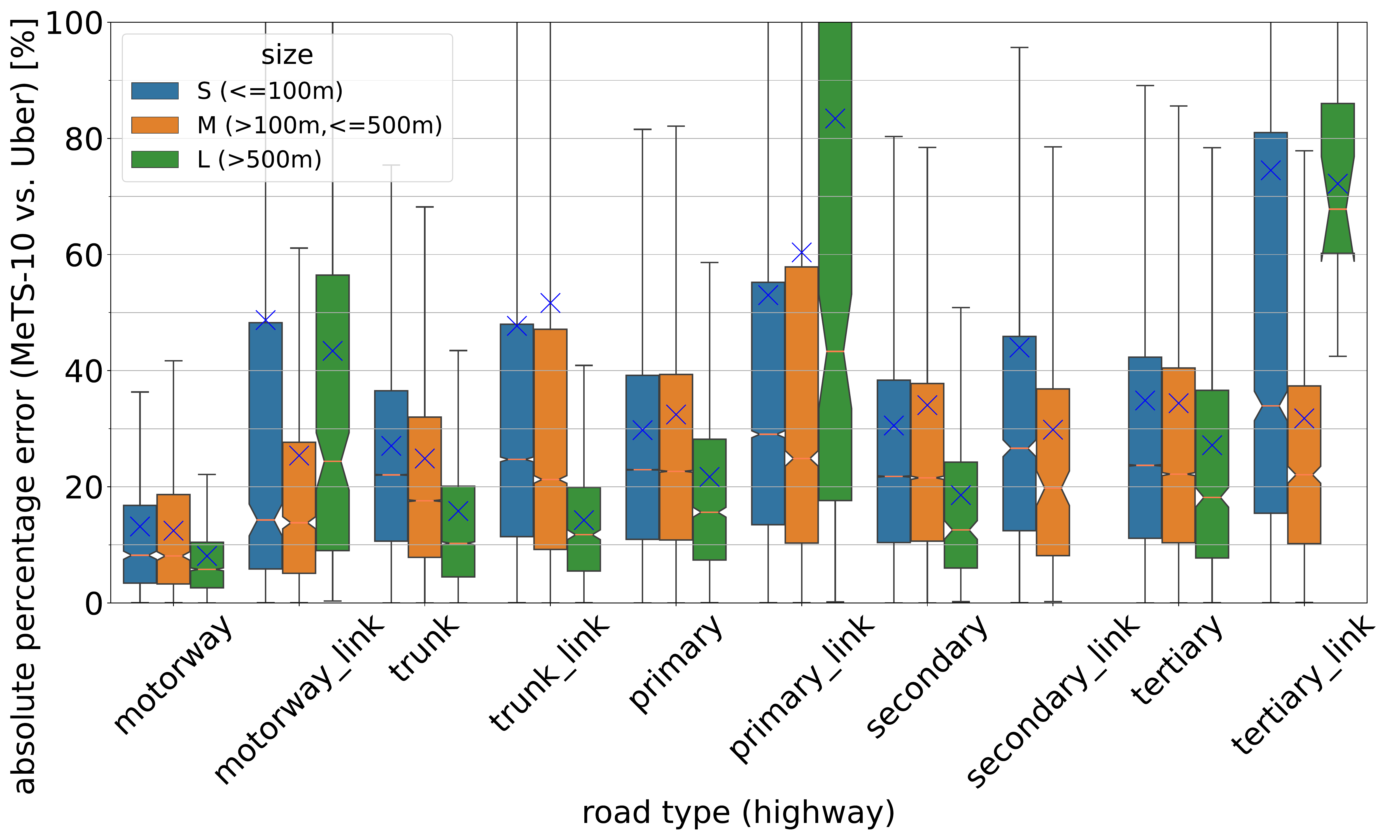}
  \caption{complex road segments}
  \end{subfigure}
  \begin{subfigure}[b]{0.45\textwidth}
  \includegraphics[width=1.0\textwidth]{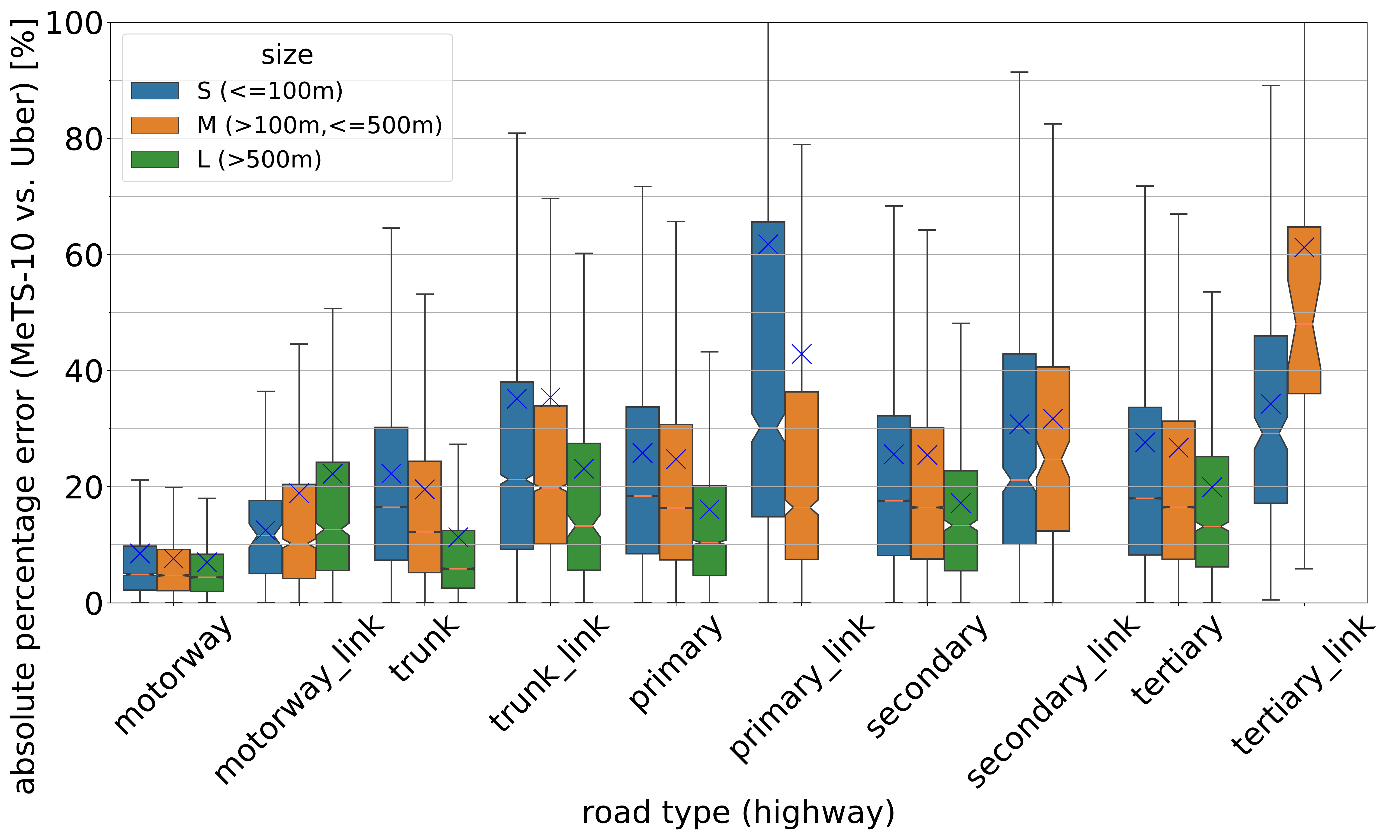}
  \caption{non-complex road segments}
  \end{subfigure}
  \caption{London absolute percentage error \mcswts{} vs. Uber by road type and segment length. Blue crosses indicate the mean per road type.}
\end{figure}

\pagebreak
\clearpage
\twocolumn
\tableofcontents

\end{document}